%% file: paper.tex
\documentclass[11pt,a4paper]{article}

\usepackage[utf8]{inputenc}
\usepackage[T1]{fontenc}
\usepackage{lmodern}
\usepackage[margin=2.5cm]{geometry}
\usepackage{microtype}
\DisableLigatures[-]{family=tt*}
\usepackage{booktabs}
\usepackage{amsmath,amssymb}
\usepackage{xcolor}
\usepackage{array}
\usepackage{longtable}
\usepackage{placeins}
\usepackage{enumitem}
\setlist{itemsep=1pt, parsep=0pt, topsep=3pt}
\usepackage{graphicx}
\input{figures/style}
\input{generated/evidence}
\usepackage{caption}
\usepackage[hidelinks]{hyperref}
\usepackage[capitalise,nameinlink]{cleveref}
\newcommand{\inputrows}[1]{\csname @@input\endcsname #1 }

\newcommand{\MaterialsRepo}{https://github.com/swedishembedded/continuous-learning-materials}
\newcommand{\MaterialsName}{swedishembedded/continuous-learning-materials}
\newcommand{\MaterialsCommit}{b6ef00a64cce5dca67bedac3e2db78d6323d5e3e}
\makeatletter
\def\materials@nine#1#2#3#4#5#6#7#8#9{#1#2#3#4#5#6#7#8#9\materials@three}
\def\materials@three#1#2#3#4\relax{#1#2#3}
\edef\MaterialsCommitShort{\expandafter\materials@nine\MaterialsCommit\relax}
\makeatother

\title{Continuous Learning from Executable Environments\\[0.3em]
  \large Autonomous Skill Acquisition, Retention, and Transfer in Language Models}
\author{Martin Schr\"oder\\Swedish Embedded AB\\\texttt{info@swedishembedded.com}}
\date{}

\begin{document}
\maketitle

\input{sections/00-abstract}
\input{sections/01-introduction}
\input{sections/02-related}
\input{sections/03-problem}
\input{sections/04-framework}
\input{sections/05-design}
\input{sections/06-results}
\input{sections/07-discussion}
\input{sections/08-limitations}
\input{sections/09-reproducibility}
\input{sections/10-conclusion}

\bibliographystyle{plain}
\bibliography{references}

\appendix
\crefalias{section}{appendix}
\crefalias{subsection}{appendix}
\input{sections/90-appendix-artifacts}
\input{sections/91-appendix-claims}
\input{sections/92-appendix-results}
\input{sections/93-appendix-records}
\input{appendix-reproduction}
\end{document}

%% file: figures/style.tex
\usepackage[T1]{fontenc}
\usepackage{lmodern}
\usepackage{tikz}
\usetikzlibrary{positioning,arrows.meta,calc,fit,backgrounds}
\usepackage{pgfplots}
\usepgfplotslibrary{groupplots}
\pgfplotsset{compat=1.17}

\definecolor{okBlue}{HTML}{0072B2}
\definecolor{okOrange}{HTML}{E69F00}
\definecolor{okGreen}{HTML}{009E73}
\definecolor{okVermillion}{HTML}{D55E00}
\definecolor{okSky}{HTML}{56B4E9}
\definecolor{okPurple}{HTML}{CC79A7}
\definecolor{okGrey}{HTML}{6E6E6E}

\tikzset{
  fig/.style={font=\footnotesize, >={Stealth[length=2mm]}},
  box/.style={draw=black!75, align=center, inner sep=3pt, minimum height=10mm, text width=22mm, line width=0.5pt,
    font=\footnotesize},
  proc/.style={box, rounded corners=2.5pt, fill=#1},
  proc/.default=white,
  data/.style={box, fill=white, path picture={\draw[line width=3.2pt,black!75] (path picture bounding box.north west) --
    (path picture bounding box.south west);}},
  flow/.style={->, line width=0.7pt, black!80, rounded corners=2pt},
  measure/.style={->, line width=0.7pt, dashed, black!65, rounded corners=2pt},
  note/.style={font=\scriptsize, text=black!75, align=center, inner sep=1.5pt},
  lane/.style={draw=black!25, rounded corners=4pt, fill=#1, inner sep=4pt},
  listing/.style={draw=black!45, fill=#1, align=left, inner sep=3pt, line width=0.4pt, font=\scriptsize\ttfamily},
  listing/.default=white,
  caption/.style={font=\scriptsize\bfseries, text=black!80, inner sep=1pt, anchor=south west},
}
\pgfplotsset{
  paper/.style={tick label style={font=\footnotesize}, label style={font=\footnotesize},
    title style={font=\footnotesize\bfseries}, legend style={font=\scriptsize, draw=black!35, fill=white},
    legend cell align=left, ymajorgrids, grid style={black!12}, axis line style={black!70}, tick align=outside,
    xtick pos=bottom, ytick pos=left, yticklabel style={/pgf/number format/fixed}, scaled y ticks=false,
    /pgf/number format/1000 sep={}},
  discard if not/.style 2 args={x filter/.append code={\edef\tempa{\thisrow{#1}}\edef\tempb{#2}%
    \ifx\tempa\tempb\else\fi}},
  table/col sep=comma,
  reference/.style={only marks, mark=square, mark size=1.8pt, okGrey},
  candidate/.style={only marks, mark=*, mark size=1.8pt, okBlue},
  pooled/.style={only marks, mark=diamond*, mark size=2.6pt, okVermillion},
}

%% file: generated/evidence.tex
\newcommand{\ev}[1]{\ifcsname ev@#1\endcsname\csname ev@#1\endcsname\else\errmessage{No evidence key #1 (see generated/evidence-keys.csv)}\fi}

\expandafter\def\csname ev@A/4B/g1/exam/base-counted\endcsname{393}
\expandafter\def\csname ev@A/4B/g1/exam/base-passes\endcsname{0}
\expandafter\def\csname ev@A/4B/g1/exam/base-pct\endcsname{0.00}
\expandafter\def\csname ev@A/4B/g1/exam/base-scheduled\endcsname{400}
\expandafter\def\csname ev@A/4B/g1/exam/base-uncounted\endcsname{7}
\expandafter\def\csname ev@A/4B/g1/exam/cand-counted\endcsname{400}
\expandafter\def\csname ev@A/4B/g1/exam/cand-passes\endcsname{17}
\expandafter\def\csname ev@A/4B/g1/exam/cand-pct\endcsname{4.25}
\expandafter\def\csname ev@A/4B/g1/exam/cand-scheduled\endcsname{400}
\expandafter\def\csname ev@A/4B/g1/exam/cand-uncounted\endcsname{0}
\expandafter\def\csname ev@A/4B/g1/exam/ci-hi-pp\endcsname{+7.11}
\expandafter\def\csname ev@A/4B/g1/exam/ci-lo-pp\endcsname{+2.03}
\expandafter\def\csname ev@A/4B/g1/exam/diff-pp\endcsname{+4.33}
\expandafter\def\csname ev@A/4B/g1/exam/exams-solved\endcsname{12}
\expandafter\def\csname ev@A/4B/g1/exam/losses\endcsname{0}
\expandafter\def\csname ev@A/4B/g1/exam/outcome\endcsname{improvement supported}
\expandafter\def\csname ev@A/4B/g1/exam/p-recorded\endcsname{0.00024}
\expandafter\def\csname ev@A/4B/g1/exam/protocol\endcsname{A}
\expandafter\def\csname ev@A/4B/g1/exam/task-p\endcsname{0.00024}
\expandafter\def\csname ev@A/4B/g1/exam/tasks\endcsname{100}
\expandafter\def\csname ev@A/4B/g1/exam/ties\endcsname{88}
\expandafter\def\csname ev@A/4B/g1/exam/uncounted\endcsname{7}
\expandafter\def\csname ev@A/4B/g1/exam/uncounted-pct\endcsname{1.75}
\expandafter\def\csname ev@A/4B/g1/exam/unit\endcsname{task}
\expandafter\def\csname ev@A/4B/g1/exam/wins\endcsname{12}
\expandafter\def\csname ev@A/4B/g1/target/base-counted\endcsname{204}
\expandafter\def\csname ev@A/4B/g1/target/base-passes\endcsname{6}
\expandafter\def\csname ev@A/4B/g1/target/base-pct\endcsname{2.94}
\expandafter\def\csname ev@A/4B/g1/target/base-scheduled\endcsname{204}
\expandafter\def\csname ev@A/4B/g1/target/base-uncounted\endcsname{0}
\expandafter\def\csname ev@A/4B/g1/target/cand-counted\endcsname{204}
\expandafter\def\csname ev@A/4B/g1/target/cand-passes\endcsname{42}
\expandafter\def\csname ev@A/4B/g1/target/cand-pct\endcsname{20.59}
\expandafter\def\csname ev@A/4B/g1/target/cand-scheduled\endcsname{204}
\expandafter\def\csname ev@A/4B/g1/target/cand-uncounted\endcsname{0}
\expandafter\def\csname ev@A/4B/g1/target/diff-pp\endcsname{+17.65}
\expandafter\def\csname ev@A/4B/g1/target/losses\endcsname{0}
\expandafter\def\csname ev@A/4B/g1/target/outcome\endcsname{improvement supported}
\expandafter\def\csname ev@A/4B/g1/target/p-recorded\endcsname{\ensuremath{9.5\times 10^{-7}}}
\expandafter\def\csname ev@A/4B/g1/target/protocol\endcsname{A}
\expandafter\def\csname ev@A/4B/g1/target/task-p\endcsname{\ensuremath{9.5\times 10^{-7}}}
\expandafter\def\csname ev@A/4B/g1/target/tasks\endcsname{51}
\expandafter\def\csname ev@A/4B/g1/target/ties\endcsname{31}
\expandafter\def\csname ev@A/4B/g1/target/uncounted\endcsname{0}
\expandafter\def\csname ev@A/4B/g1/target/uncounted-pct\endcsname{0.00}
\expandafter\def\csname ev@A/4B/g1/target/unit\endcsname{task}
\expandafter\def\csname ev@A/4B/g1/target/wins\endcsname{20}
\expandafter\def\csname ev@A/4B/g2/exam/base-counted\endcsname{400}
\expandafter\def\csname ev@A/4B/g2/exam/base-passes\endcsname{11}
\expandafter\def\csname ev@A/4B/g2/exam/base-pct\endcsname{2.75}
\expandafter\def\csname ev@A/4B/g2/exam/base-scheduled\endcsname{400}
\expandafter\def\csname ev@A/4B/g2/exam/base-uncounted\endcsname{0}
\expandafter\def\csname ev@A/4B/g2/exam/cand-counted\endcsname{400}
\expandafter\def\csname ev@A/4B/g2/exam/cand-passes\endcsname{13}
\expandafter\def\csname ev@A/4B/g2/exam/cand-pct\endcsname{3.25}
\expandafter\def\csname ev@A/4B/g2/exam/cand-scheduled\endcsname{400}
\expandafter\def\csname ev@A/4B/g2/exam/cand-uncounted\endcsname{0}
\expandafter\def\csname ev@A/4B/g2/exam/ci-hi-pp\endcsname{+2.75}
\expandafter\def\csname ev@A/4B/g2/exam/ci-lo-pp\endcsname{-1.75}
\expandafter\def\csname ev@A/4B/g2/exam/diff-pp\endcsname{+0.50}
\expandafter\def\csname ev@A/4B/g2/exam/losses\endcsname{5}
\expandafter\def\csname ev@A/4B/g2/exam/outcome\endcsname{inconclusive}
\expandafter\def\csname ev@A/4B/g2/exam/p-recorded\endcsname{0.291}
\expandafter\def\csname ev@A/4B/g2/exam/protocol\endcsname{A}
\expandafter\def\csname ev@A/4B/g2/exam/task-p\endcsname{0.291}
\expandafter\def\csname ev@A/4B/g2/exam/tasks\endcsname{100}
\expandafter\def\csname ev@A/4B/g2/exam/ties\endcsname{87}
\expandafter\def\csname ev@A/4B/g2/exam/uncounted\endcsname{0}
\expandafter\def\csname ev@A/4B/g2/exam/uncounted-pct\endcsname{0.00}
\expandafter\def\csname ev@A/4B/g2/exam/unit\endcsname{task}
\expandafter\def\csname ev@A/4B/g2/exam/wins\endcsname{8}
\expandafter\def\csname ev@A/4B/g2/retention-g1/base-counted\endcsname{204}
\expandafter\def\csname ev@A/4B/g2/retention-g1/base-passes\endcsname{39}
\expandafter\def\csname ev@A/4B/g2/retention-g1/base-pct\endcsname{19.12}
\expandafter\def\csname ev@A/4B/g2/retention-g1/base-scheduled\endcsname{204}
\expandafter\def\csname ev@A/4B/g2/retention-g1/base-uncounted\endcsname{0}
\expandafter\def\csname ev@A/4B/g2/retention-g1/cand-counted\endcsname{204}
\expandafter\def\csname ev@A/4B/g2/retention-g1/cand-passes\endcsname{49}
\expandafter\def\csname ev@A/4B/g2/retention-g1/cand-pct\endcsname{24.02}
\expandafter\def\csname ev@A/4B/g2/retention-g1/cand-scheduled\endcsname{204}
\expandafter\def\csname ev@A/4B/g2/retention-g1/cand-uncounted\endcsname{0}
\expandafter\def\csname ev@A/4B/g2/retention-g1/ci-hi-pp\endcsname{+12.25}
\expandafter\def\csname ev@A/4B/g2/retention-g1/ci-lo-pp\endcsname{-1.47}
\expandafter\def\csname ev@A/4B/g2/retention-g1/diff-pp\endcsname{+4.90}
\expandafter\def\csname ev@A/4B/g2/retention-g1/losses\endcsname{8}
\expandafter\def\csname ev@A/4B/g2/retention-g1/outcome\endcsname{retention supported}
\expandafter\def\csname ev@A/4B/g2/retention-g1/p-recorded\endcsname{0.192}
\expandafter\def\csname ev@A/4B/g2/retention-g1/protocol\endcsname{A}
\expandafter\def\csname ev@A/4B/g2/retention-g1/task-p\endcsname{0.192}
\expandafter\def\csname ev@A/4B/g2/retention-g1/tasks\endcsname{51}
\expandafter\def\csname ev@A/4B/g2/retention-g1/ties\endcsname{30}
\expandafter\def\csname ev@A/4B/g2/retention-g1/uncounted\endcsname{0}
\expandafter\def\csname ev@A/4B/g2/retention-g1/uncounted-pct\endcsname{0.00}
\expandafter\def\csname ev@A/4B/g2/retention-g1/unit\endcsname{task}
\expandafter\def\csname ev@A/4B/g2/retention-g1/wins\endcsname{13}
\expandafter\def\csname ev@A/4B/g2/target/base-counted\endcsname{264}
\expandafter\def\csname ev@A/4B/g2/target/base-passes\endcsname{57}
\expandafter\def\csname ev@A/4B/g2/target/base-pct\endcsname{21.59}
\expandafter\def\csname ev@A/4B/g2/target/base-scheduled\endcsname{264}
\expandafter\def\csname ev@A/4B/g2/target/base-uncounted\endcsname{0}
\expandafter\def\csname ev@A/4B/g2/target/cand-counted\endcsname{264}
\expandafter\def\csname ev@A/4B/g2/target/cand-passes\endcsname{66}
\expandafter\def\csname ev@A/4B/g2/target/cand-pct\endcsname{25.00}
\expandafter\def\csname ev@A/4B/g2/target/cand-scheduled\endcsname{264}
\expandafter\def\csname ev@A/4B/g2/target/cand-uncounted\endcsname{0}
\expandafter\def\csname ev@A/4B/g2/target/ci-hi-pp\endcsname{+9.85}
\expandafter\def\csname ev@A/4B/g2/target/ci-lo-pp\endcsname{-3.03}
\expandafter\def\csname ev@A/4B/g2/target/diff-pp\endcsname{+3.41}
\expandafter\def\csname ev@A/4B/g2/target/losses\endcsname{11}
\expandafter\def\csname ev@A/4B/g2/target/outcome\endcsname{inconclusive}
\expandafter\def\csname ev@A/4B/g2/target/p-recorded\endcsname{0.221}
\expandafter\def\csname ev@A/4B/g2/target/promoted\endcsname{no}
\expandafter\def\csname ev@A/4B/g2/target/protocol\endcsname{A}
\expandafter\def\csname ev@A/4B/g2/target/task-p\endcsname{0.221}
\expandafter\def\csname ev@A/4B/g2/target/tasks\endcsname{66}
\expandafter\def\csname ev@A/4B/g2/target/ties\endcsname{39}
\expandafter\def\csname ev@A/4B/g2/target/uncounted\endcsname{0}
\expandafter\def\csname ev@A/4B/g2/target/uncounted-pct\endcsname{0.00}
\expandafter\def\csname ev@A/4B/g2/target/unit\endcsname{task}
\expandafter\def\csname ev@A/4B/g2/target/wins\endcsname{16}
\expandafter\def\csname ev@A/8B/g1.seed1/exam/base-counted\endcsname{379}
\expandafter\def\csname ev@A/8B/g1.seed1/exam/base-passes\endcsname{0}
\expandafter\def\csname ev@A/8B/g1.seed1/exam/base-pct\endcsname{0.00}
\expandafter\def\csname ev@A/8B/g1.seed1/exam/base-scheduled\endcsname{400}
\expandafter\def\csname ev@A/8B/g1.seed1/exam/base-uncounted\endcsname{21}
\expandafter\def\csname ev@A/8B/g1.seed1/exam/cand-counted\endcsname{400}
\expandafter\def\csname ev@A/8B/g1.seed1/exam/cand-passes\endcsname{28}
\expandafter\def\csname ev@A/8B/g1.seed1/exam/cand-pct\endcsname{7.00}
\expandafter\def\csname ev@A/8B/g1.seed1/exam/cand-scheduled\endcsname{400}
\expandafter\def\csname ev@A/8B/g1.seed1/exam/cand-uncounted\endcsname{0}
\expandafter\def\csname ev@A/8B/g1.seed1/exam/ci-hi-pp\endcsname{+10.36}
\expandafter\def\csname ev@A/8B/g1.seed1/exam/ci-lo-pp\endcsname{+4.19}
\expandafter\def\csname ev@A/8B/g1.seed1/exam/diff-pp\endcsname{+7.12}
\expandafter\def\csname ev@A/8B/g1.seed1/exam/exams-solved\endcsname{21}
\expandafter\def\csname ev@A/8B/g1.seed1/exam/losses\endcsname{0}
\expandafter\def\csname ev@A/8B/g1.seed1/exam/outcome\endcsname{excluded}
\expandafter\def\csname ev@A/8B/g1.seed1/exam/p-recorded\endcsname{\ensuremath{9.5\times 10^{-7}}}
\expandafter\def\csname ev@A/8B/g1.seed1/exam/protocol\endcsname{A}
\expandafter\def\csname ev@A/8B/g1.seed1/exam/task-p\endcsname{\ensuremath{9.5\times 10^{-7}}}
\expandafter\def\csname ev@A/8B/g1.seed1/exam/tasks\endcsname{100}
\expandafter\def\csname ev@A/8B/g1.seed1/exam/ties\endcsname{80}
\expandafter\def\csname ev@A/8B/g1.seed1/exam/uncounted\endcsname{21}
\expandafter\def\csname ev@A/8B/g1.seed1/exam/uncounted-pct\endcsname{5.25}
\expandafter\def\csname ev@A/8B/g1.seed1/exam/unit\endcsname{task}
\expandafter\def\csname ev@A/8B/g1.seed1/exam/wins\endcsname{20}
\expandafter\def\csname ev@A/8B/g1.seed1/target/base-counted\endcsname{203}
\expandafter\def\csname ev@A/8B/g1.seed1/target/base-passes\endcsname{0}
\expandafter\def\csname ev@A/8B/g1.seed1/target/base-pct\endcsname{0.00}
\expandafter\def\csname ev@A/8B/g1.seed1/target/base-scheduled\endcsname{204}
\expandafter\def\csname ev@A/8B/g1.seed1/target/base-uncounted\endcsname{1}
\expandafter\def\csname ev@A/8B/g1.seed1/target/cand-counted\endcsname{204}
\expandafter\def\csname ev@A/8B/g1.seed1/target/cand-passes\endcsname{53}
\expandafter\def\csname ev@A/8B/g1.seed1/target/cand-pct\endcsname{25.98}
\expandafter\def\csname ev@A/8B/g1.seed1/target/cand-scheduled\endcsname{204}
\expandafter\def\csname ev@A/8B/g1.seed1/target/cand-uncounted\endcsname{0}
\expandafter\def\csname ev@A/8B/g1.seed1/target/diff-pp\endcsname{+25.98}
\expandafter\def\csname ev@A/8B/g1.seed1/target/losses\endcsname{0}
\expandafter\def\csname ev@A/8B/g1.seed1/target/outcome\endcsname{improvement supported}
\expandafter\def\csname ev@A/8B/g1.seed1/target/p-recorded\endcsname{\ensuremath{1.5\times 10^{-8}}}
\expandafter\def\csname ev@A/8B/g1.seed1/target/protocol\endcsname{A}
\expandafter\def\csname ev@A/8B/g1.seed1/target/task-p\endcsname{\ensuremath{1.5\times 10^{-8}}}
\expandafter\def\csname ev@A/8B/g1.seed1/target/tasks\endcsname{51}
\expandafter\def\csname ev@A/8B/g1.seed1/target/ties\endcsname{25}
\expandafter\def\csname ev@A/8B/g1.seed1/target/uncounted\endcsname{1}
\expandafter\def\csname ev@A/8B/g1.seed1/target/uncounted-pct\endcsname{0.49}
\expandafter\def\csname ev@A/8B/g1.seed1/target/unit\endcsname{task}
\expandafter\def\csname ev@A/8B/g1.seed1/target/wins\endcsname{26}
\expandafter\def\csname ev@A/8B/g1.seed2/exam/base-counted\endcsname{385}
\expandafter\def\csname ev@A/8B/g1.seed2/exam/base-passes\endcsname{0}
\expandafter\def\csname ev@A/8B/g1.seed2/exam/base-pct\endcsname{0.00}
\expandafter\def\csname ev@A/8B/g1.seed2/exam/base-scheduled\endcsname{400}
\expandafter\def\csname ev@A/8B/g1.seed2/exam/base-uncounted\endcsname{15}
\expandafter\def\csname ev@A/8B/g1.seed2/exam/cand-counted\endcsname{400}
\expandafter\def\csname ev@A/8B/g1.seed2/exam/cand-passes\endcsname{24}
\expandafter\def\csname ev@A/8B/g1.seed2/exam/cand-pct\endcsname{6.00}
\expandafter\def\csname ev@A/8B/g1.seed2/exam/cand-scheduled\endcsname{400}
\expandafter\def\csname ev@A/8B/g1.seed2/exam/cand-uncounted\endcsname{0}
\expandafter\def\csname ev@A/8B/g1.seed2/exam/ci-hi-pp\endcsname{+9.00}
\expandafter\def\csname ev@A/8B/g1.seed2/exam/ci-lo-pp\endcsname{+2.89}
\expandafter\def\csname ev@A/8B/g1.seed2/exam/diff-pp\endcsname{+5.71}
\expandafter\def\csname ev@A/8B/g1.seed2/exam/exams-solved\endcsname{18}
\expandafter\def\csname ev@A/8B/g1.seed2/exam/losses\endcsname{0}
\expandafter\def\csname ev@A/8B/g1.seed2/exam/outcome\endcsname{excluded}
\expandafter\def\csname ev@A/8B/g1.seed2/exam/p-recorded\endcsname{\ensuremath{1.5\times 10^{-5}}}
\expandafter\def\csname ev@A/8B/g1.seed2/exam/protocol\endcsname{A}
\expandafter\def\csname ev@A/8B/g1.seed2/exam/task-p\endcsname{\ensuremath{1.5\times 10^{-5}}}
\expandafter\def\csname ev@A/8B/g1.seed2/exam/tasks\endcsname{100}
\expandafter\def\csname ev@A/8B/g1.seed2/exam/ties\endcsname{84}
\expandafter\def\csname ev@A/8B/g1.seed2/exam/uncounted\endcsname{15}
\expandafter\def\csname ev@A/8B/g1.seed2/exam/uncounted-pct\endcsname{3.75}
\expandafter\def\csname ev@A/8B/g1.seed2/exam/unit\endcsname{task}
\expandafter\def\csname ev@A/8B/g1.seed2/exam/wins\endcsname{16}
\expandafter\def\csname ev@A/8B/g1.seed2/target/base-counted\endcsname{200}
\expandafter\def\csname ev@A/8B/g1.seed2/target/base-passes\endcsname{0}
\expandafter\def\csname ev@A/8B/g1.seed2/target/base-pct\endcsname{0.00}
\expandafter\def\csname ev@A/8B/g1.seed2/target/base-scheduled\endcsname{204}
\expandafter\def\csname ev@A/8B/g1.seed2/target/base-uncounted\endcsname{4}
\expandafter\def\csname ev@A/8B/g1.seed2/target/cand-counted\endcsname{204}
\expandafter\def\csname ev@A/8B/g1.seed2/target/cand-passes\endcsname{54}
\expandafter\def\csname ev@A/8B/g1.seed2/target/cand-pct\endcsname{26.47}
\expandafter\def\csname ev@A/8B/g1.seed2/target/cand-scheduled\endcsname{204}
\expandafter\def\csname ev@A/8B/g1.seed2/target/cand-uncounted\endcsname{0}
\expandafter\def\csname ev@A/8B/g1.seed2/target/diff-pp\endcsname{+26.47}
\expandafter\def\csname ev@A/8B/g1.seed2/target/losses\endcsname{0}
\expandafter\def\csname ev@A/8B/g1.seed2/target/outcome\endcsname{improvement supported}
\expandafter\def\csname ev@A/8B/g1.seed2/target/p-recorded\endcsname{\ensuremath{9.3\times 10^{-10}}}
\expandafter\def\csname ev@A/8B/g1.seed2/target/protocol\endcsname{A}
\expandafter\def\csname ev@A/8B/g1.seed2/target/task-p\endcsname{\ensuremath{9.3\times 10^{-10}}}
\expandafter\def\csname ev@A/8B/g1.seed2/target/tasks\endcsname{51}
\expandafter\def\csname ev@A/8B/g1.seed2/target/ties\endcsname{21}
\expandafter\def\csname ev@A/8B/g1.seed2/target/uncounted\endcsname{4}
\expandafter\def\csname ev@A/8B/g1.seed2/target/uncounted-pct\endcsname{1.96}
\expandafter\def\csname ev@A/8B/g1.seed2/target/unit\endcsname{task}
\expandafter\def\csname ev@A/8B/g1.seed2/target/wins\endcsname{30}
\expandafter\def\csname ev@A/8B/g1/exam/base-counted\endcsname{381}
\expandafter\def\csname ev@A/8B/g1/exam/base-passes\endcsname{0}
\expandafter\def\csname ev@A/8B/g1/exam/base-pct\endcsname{0.00}
\expandafter\def\csname ev@A/8B/g1/exam/base-scheduled\endcsname{400}
\expandafter\def\csname ev@A/8B/g1/exam/base-uncounted\endcsname{19}
\expandafter\def\csname ev@A/8B/g1/exam/cand-counted\endcsname{400}
\expandafter\def\csname ev@A/8B/g1/exam/cand-passes\endcsname{22}
\expandafter\def\csname ev@A/8B/g1/exam/cand-pct\endcsname{5.50}
\expandafter\def\csname ev@A/8B/g1/exam/cand-scheduled\endcsname{400}
\expandafter\def\csname ev@A/8B/g1/exam/cand-uncounted\endcsname{0}
\expandafter\def\csname ev@A/8B/g1/exam/ci-hi-pp\endcsname{+8.03}
\expandafter\def\csname ev@A/8B/g1/exam/ci-lo-pp\endcsname{+2.86}
\expandafter\def\csname ev@A/8B/g1/exam/diff-pp\endcsname{+5.25}
\expandafter\def\csname ev@A/8B/g1/exam/exams-solved\endcsname{16}
\expandafter\def\csname ev@A/8B/g1/exam/losses\endcsname{0}
\expandafter\def\csname ev@A/8B/g1/exam/outcome\endcsname{excluded}
\expandafter\def\csname ev@A/8B/g1/exam/p-recorded\endcsname{\ensuremath{1.5\times 10^{-5}}}
\expandafter\def\csname ev@A/8B/g1/exam/protocol\endcsname{A}
\expandafter\def\csname ev@A/8B/g1/exam/task-p\endcsname{\ensuremath{1.5\times 10^{-5}}}
\expandafter\def\csname ev@A/8B/g1/exam/tasks\endcsname{100}
\expandafter\def\csname ev@A/8B/g1/exam/ties\endcsname{84}
\expandafter\def\csname ev@A/8B/g1/exam/uncounted\endcsname{19}
\expandafter\def\csname ev@A/8B/g1/exam/uncounted-pct\endcsname{4.75}
\expandafter\def\csname ev@A/8B/g1/exam/unit\endcsname{task}
\expandafter\def\csname ev@A/8B/g1/exam/wins\endcsname{16}
\expandafter\def\csname ev@A/8B/g2/exam/base-counted\endcsname{400}
\expandafter\def\csname ev@A/8B/g2/exam/base-passes\endcsname{23}
\expandafter\def\csname ev@A/8B/g2/exam/base-pct\endcsname{5.75}
\expandafter\def\csname ev@A/8B/g2/exam/base-scheduled\endcsname{400}
\expandafter\def\csname ev@A/8B/g2/exam/base-uncounted\endcsname{0}
\expandafter\def\csname ev@A/8B/g2/exam/cand-counted\endcsname{400}
\expandafter\def\csname ev@A/8B/g2/exam/cand-passes\endcsname{30}
\expandafter\def\csname ev@A/8B/g2/exam/cand-pct\endcsname{7.50}
\expandafter\def\csname ev@A/8B/g2/exam/cand-scheduled\endcsname{400}
\expandafter\def\csname ev@A/8B/g2/exam/cand-uncounted\endcsname{0}
\expandafter\def\csname ev@A/8B/g2/exam/ci-hi-pp\endcsname{+5.25}
\expandafter\def\csname ev@A/8B/g2/exam/ci-lo-pp\endcsname{-1.75}
\expandafter\def\csname ev@A/8B/g2/exam/diff-pp\endcsname{+1.75}
\expandafter\def\csname ev@A/8B/g2/exam/losses\endcsname{7}
\expandafter\def\csname ev@A/8B/g2/exam/outcome\endcsname{inconclusive}
\expandafter\def\csname ev@A/8B/g2/exam/p-recorded\endcsname{0.18}
\expandafter\def\csname ev@A/8B/g2/exam/protocol\endcsname{A}
\expandafter\def\csname ev@A/8B/g2/exam/task-p\endcsname{0.18}
\expandafter\def\csname ev@A/8B/g2/exam/tasks\endcsname{100}
\expandafter\def\csname ev@A/8B/g2/exam/ties\endcsname{81}
\expandafter\def\csname ev@A/8B/g2/exam/uncounted\endcsname{0}
\expandafter\def\csname ev@A/8B/g2/exam/uncounted-pct\endcsname{0.00}
\expandafter\def\csname ev@A/8B/g2/exam/unit\endcsname{task}
\expandafter\def\csname ev@A/8B/g2/exam/wins\endcsname{12}
\expandafter\def\csname ev@A/8B/g2/retention-g1/base-counted\endcsname{204}
\expandafter\def\csname ev@A/8B/g2/retention-g1/base-passes\endcsname{46}
\expandafter\def\csname ev@A/8B/g2/retention-g1/base-pct\endcsname{22.55}
\expandafter\def\csname ev@A/8B/g2/retention-g1/base-scheduled\endcsname{204}
\expandafter\def\csname ev@A/8B/g2/retention-g1/base-uncounted\endcsname{0}
\expandafter\def\csname ev@A/8B/g2/retention-g1/cand-counted\endcsname{204}
\expandafter\def\csname ev@A/8B/g2/retention-g1/cand-passes\endcsname{60}
\expandafter\def\csname ev@A/8B/g2/retention-g1/cand-pct\endcsname{29.41}
\expandafter\def\csname ev@A/8B/g2/retention-g1/cand-scheduled\endcsname{204}
\expandafter\def\csname ev@A/8B/g2/retention-g1/cand-uncounted\endcsname{0}
\expandafter\def\csname ev@A/8B/g2/retention-g1/ci-hi-pp\endcsname{+16.18}
\expandafter\def\csname ev@A/8B/g2/retention-g1/ci-lo-pp\endcsname{-2.45}
\expandafter\def\csname ev@A/8B/g2/retention-g1/diff-pp\endcsname{+6.86}
\expandafter\def\csname ev@A/8B/g2/retention-g1/losses\endcsname{10}
\expandafter\def\csname ev@A/8B/g2/retention-g1/outcome\endcsname{retention supported}
\expandafter\def\csname ev@A/8B/g2/retention-g1/p-recorded\endcsname{0.163}
\expandafter\def\csname ev@A/8B/g2/retention-g1/protocol\endcsname{A}
\expandafter\def\csname ev@A/8B/g2/retention-g1/task-p\endcsname{0.163}
\expandafter\def\csname ev@A/8B/g2/retention-g1/tasks\endcsname{51}
\expandafter\def\csname ev@A/8B/g2/retention-g1/ties\endcsname{25}
\expandafter\def\csname ev@A/8B/g2/retention-g1/uncounted\endcsname{0}
\expandafter\def\csname ev@A/8B/g2/retention-g1/uncounted-pct\endcsname{0.00}
\expandafter\def\csname ev@A/8B/g2/retention-g1/unit\endcsname{task}
\expandafter\def\csname ev@A/8B/g2/retention-g1/wins\endcsname{16}
\expandafter\def\csname ev@A/8B/g2/target/base-counted\endcsname{264}
\expandafter\def\csname ev@A/8B/g2/target/base-passes\endcsname{78}
\expandafter\def\csname ev@A/8B/g2/target/base-pct\endcsname{29.55}
\expandafter\def\csname ev@A/8B/g2/target/base-scheduled\endcsname{264}
\expandafter\def\csname ev@A/8B/g2/target/base-uncounted\endcsname{0}
\expandafter\def\csname ev@A/8B/g2/target/cand-counted\endcsname{264}
\expandafter\def\csname ev@A/8B/g2/target/cand-passes\endcsname{81}
\expandafter\def\csname ev@A/8B/g2/target/cand-pct\endcsname{30.68}
\expandafter\def\csname ev@A/8B/g2/target/cand-scheduled\endcsname{264}
\expandafter\def\csname ev@A/8B/g2/target/cand-uncounted\endcsname{0}
\expandafter\def\csname ev@A/8B/g2/target/ci-hi-pp\endcsname{+8.33}
\expandafter\def\csname ev@A/8B/g2/target/ci-lo-pp\endcsname{-6.44}
\expandafter\def\csname ev@A/8B/g2/target/diff-pp\endcsname{+1.14}
\expandafter\def\csname ev@A/8B/g2/target/losses\endcsname{10}
\expandafter\def\csname ev@A/8B/g2/target/outcome\endcsname{inconclusive}
\expandafter\def\csname ev@A/8B/g2/target/p-recorded\endcsname{0.212}
\expandafter\def\csname ev@A/8B/g2/target/promoted\endcsname{no}
\expandafter\def\csname ev@A/8B/g2/target/protocol\endcsname{A}
\expandafter\def\csname ev@A/8B/g2/target/task-p\endcsname{0.212}
\expandafter\def\csname ev@A/8B/g2/target/tasks\endcsname{66}
\expandafter\def\csname ev@A/8B/g2/target/ties\endcsname{41}
\expandafter\def\csname ev@A/8B/g2/target/uncounted\endcsname{0}
\expandafter\def\csname ev@A/8B/g2/target/uncounted-pct\endcsname{0.00}
\expandafter\def\csname ev@A/8B/g2/target/unit\endcsname{task}
\expandafter\def\csname ev@A/8B/g2/target/wins\endcsname{15}
\expandafter\def\csname ev@B-R/4B/g2/pooled/exam/base-counted\endcsname{400}
\expandafter\def\csname ev@B-R/4B/g2/pooled/exam/base-passes\endcsname{12}
\expandafter\def\csname ev@B-R/4B/g2/pooled/exam/base-pct\endcsname{3.00}
\expandafter\def\csname ev@B-R/4B/g2/pooled/exam/base-scheduled\endcsname{400}
\expandafter\def\csname ev@B-R/4B/g2/pooled/exam/base-uncounted\endcsname{0}
\expandafter\def\csname ev@B-R/4B/g2/pooled/exam/cand-counted\endcsname{2000}
\expandafter\def\csname ev@B-R/4B/g2/pooled/exam/cand-passes\endcsname{72}
\expandafter\def\csname ev@B-R/4B/g2/pooled/exam/cand-pct\endcsname{3.60}
\expandafter\def\csname ev@B-R/4B/g2/pooled/exam/cand-scheduled\endcsname{2000}
\expandafter\def\csname ev@B-R/4B/g2/pooled/exam/cand-uncounted\endcsname{0}
\expandafter\def\csname ev@B-R/4B/g2/pooled/exam/diff-pp\endcsname{+0.60}
\expandafter\def\csname ev@B-R/4B/g2/pooled/exam/losses\endcsname{46}
\expandafter\def\csname ev@B-R/4B/g2/pooled/exam/outcome\endcsname{inconclusive}
\expandafter\def\csname ev@B-R/4B/g2/pooled/exam/p-recorded\endcsname{0.14}
\expandafter\def\csname ev@B-R/4B/g2/pooled/exam/protocol\endcsname{B-R}
\expandafter\def\csname ev@B-R/4B/g2/pooled/exam/seeds\endcsname{5}
\expandafter\def\csname ev@B-R/4B/g2/pooled/exam/seeds-negative\endcsname{1}
\expandafter\def\csname ev@B-R/4B/g2/pooled/exam/seeds-positive\endcsname{4}
\expandafter\def\csname ev@B-R/4B/g2/pooled/exam/tasks\endcsname{100}
\expandafter\def\csname ev@B-R/4B/g2/pooled/exam/uncounted\endcsname{0}
\expandafter\def\csname ev@B-R/4B/g2/pooled/exam/uncounted-pct\endcsname{0.00}
\expandafter\def\csname ev@B-R/4B/g2/pooled/exam/unit\endcsname{attempt slot, summed over training seeds}
\expandafter\def\csname ev@B-R/4B/g2/pooled/exam/wins\endcsname{58}
\expandafter\def\csname ev@B-R/4B/g2/pooled/retention-g1/base-counted\endcsname{224}
\expandafter\def\csname ev@B-R/4B/g2/pooled/retention-g1/base-passes\endcsname{68}
\expandafter\def\csname ev@B-R/4B/g2/pooled/retention-g1/base-pct\endcsname{30.36}
\expandafter\def\csname ev@B-R/4B/g2/pooled/retention-g1/base-scheduled\endcsname{224}
\expandafter\def\csname ev@B-R/4B/g2/pooled/retention-g1/base-uncounted\endcsname{0}
\expandafter\def\csname ev@B-R/4B/g2/pooled/retention-g1/cand-counted\endcsname{1120}
\expandafter\def\csname ev@B-R/4B/g2/pooled/retention-g1/cand-passes\endcsname{410}
\expandafter\def\csname ev@B-R/4B/g2/pooled/retention-g1/cand-pct\endcsname{36.61}
\expandafter\def\csname ev@B-R/4B/g2/pooled/retention-g1/cand-scheduled\endcsname{1120}
\expandafter\def\csname ev@B-R/4B/g2/pooled/retention-g1/cand-uncounted\endcsname{0}
\expandafter\def\csname ev@B-R/4B/g2/pooled/retention-g1/diff-pp\endcsname{+6.25}
\expandafter\def\csname ev@B-R/4B/g2/pooled/retention-g1/losses\endcsname{84}
\expandafter\def\csname ev@B-R/4B/g2/pooled/retention-g1/outcome\endcsname{inconclusive}
\expandafter\def\csname ev@B-R/4B/g2/pooled/retention-g1/p-recorded\endcsname{\ensuremath{3.3\times 10^{-6}}}
\expandafter\def\csname ev@B-R/4B/g2/pooled/retention-g1/protocol\endcsname{B-R}
\expandafter\def\csname ev@B-R/4B/g2/pooled/retention-g1/seeds\endcsname{5}
\expandafter\def\csname ev@B-R/4B/g2/pooled/retention-g1/seeds-negative\endcsname{0}
\expandafter\def\csname ev@B-R/4B/g2/pooled/retention-g1/seeds-positive\endcsname{5}
\expandafter\def\csname ev@B-R/4B/g2/pooled/retention-g1/tasks\endcsname{56}
\expandafter\def\csname ev@B-R/4B/g2/pooled/retention-g1/uncounted\endcsname{0}
\expandafter\def\csname ev@B-R/4B/g2/pooled/retention-g1/uncounted-pct\endcsname{0.00}
\expandafter\def\csname ev@B-R/4B/g2/pooled/retention-g1/unit\endcsname{attempt slot, summed over training seeds}
\expandafter\def\csname ev@B-R/4B/g2/pooled/retention-g1/wins\endcsname{154}
\expandafter\def\csname ev@B-R/4B/g2/pooled/target/base-counted\endcsname{188}
\expandafter\def\csname ev@B-R/4B/g2/pooled/target/base-passes\endcsname{31}
\expandafter\def\csname ev@B-R/4B/g2/pooled/target/base-pct\endcsname{16.49}
\expandafter\def\csname ev@B-R/4B/g2/pooled/target/base-scheduled\endcsname{188}
\expandafter\def\csname ev@B-R/4B/g2/pooled/target/base-uncounted\endcsname{0}
\expandafter\def\csname ev@B-R/4B/g2/pooled/target/cand-counted\endcsname{940}
\expandafter\def\csname ev@B-R/4B/g2/pooled/target/cand-passes\endcsname{166}
\expandafter\def\csname ev@B-R/4B/g2/pooled/target/cand-pct\endcsname{17.66}
\expandafter\def\csname ev@B-R/4B/g2/pooled/target/cand-scheduled\endcsname{940}
\expandafter\def\csname ev@B-R/4B/g2/pooled/target/cand-uncounted\endcsname{0}
\expandafter\def\csname ev@B-R/4B/g2/pooled/target/diff-pp\endcsname{+1.17}
\expandafter\def\csname ev@B-R/4B/g2/pooled/target/losses\endcsname{65}
\expandafter\def\csname ev@B-R/4B/g2/pooled/target/outcome\endcsname{inconclusive}
\expandafter\def\csname ev@B-R/4B/g2/pooled/target/p-recorded\endcsname{0.2}
\expandafter\def\csname ev@B-R/4B/g2/pooled/target/protocol\endcsname{B-R}
\expandafter\def\csname ev@B-R/4B/g2/pooled/target/seeds\endcsname{5}
\expandafter\def\csname ev@B-R/4B/g2/pooled/target/seeds-negative\endcsname{1}
\expandafter\def\csname ev@B-R/4B/g2/pooled/target/seeds-positive\endcsname{4}
\expandafter\def\csname ev@B-R/4B/g2/pooled/target/tasks\endcsname{47}
\expandafter\def\csname ev@B-R/4B/g2/pooled/target/uncounted\endcsname{0}
\expandafter\def\csname ev@B-R/4B/g2/pooled/target/uncounted-pct\endcsname{0.00}
\expandafter\def\csname ev@B-R/4B/g2/pooled/target/unit\endcsname{attempt slot, summed over training seeds}
\expandafter\def\csname ev@B-R/4B/g2/pooled/target/wins\endcsname{76}
\expandafter\def\csname ev@B-R/4B/g2/s1/exam/attempt-ci-hi-pp\endcsname{+2.21}
\expandafter\def\csname ev@B-R/4B/g2/s1/exam/attempt-ci-lo-pp\endcsname{-2.73}
\expandafter\def\csname ev@B-R/4B/g2/s1/exam/base-counted\endcsname{400}
\expandafter\def\csname ev@B-R/4B/g2/s1/exam/base-passes\endcsname{12}
\expandafter\def\csname ev@B-R/4B/g2/s1/exam/base-pct\endcsname{3.00}
\expandafter\def\csname ev@B-R/4B/g2/s1/exam/base-scheduled\endcsname{400}
\expandafter\def\csname ev@B-R/4B/g2/s1/exam/base-uncounted\endcsname{0}
\expandafter\def\csname ev@B-R/4B/g2/s1/exam/cand-counted\endcsname{400}
\expandafter\def\csname ev@B-R/4B/g2/s1/exam/cand-passes\endcsname{11}
\expandafter\def\csname ev@B-R/4B/g2/s1/exam/cand-pct\endcsname{2.75}
\expandafter\def\csname ev@B-R/4B/g2/s1/exam/cand-scheduled\endcsname{400}
\expandafter\def\csname ev@B-R/4B/g2/s1/exam/cand-uncounted\endcsname{0}
\expandafter\def\csname ev@B-R/4B/g2/s1/exam/diff-pp\endcsname{-0.25}
\expandafter\def\csname ev@B-R/4B/g2/s1/exam/losses\endcsname{11}
\expandafter\def\csname ev@B-R/4B/g2/s1/exam/outcome\endcsname{inconclusive}
\expandafter\def\csname ev@B-R/4B/g2/s1/exam/p-recorded\endcsname{0.668}
\expandafter\def\csname ev@B-R/4B/g2/s1/exam/protocol\endcsname{B-R}
\expandafter\def\csname ev@B-R/4B/g2/s1/exam/task-losses-high\endcsname{11}
\expandafter\def\csname ev@B-R/4B/g2/s1/exam/task-losses-low\endcsname{1}
\expandafter\def\csname ev@B-R/4B/g2/s1/exam/task-p-high\endcsname{1}
\expandafter\def\csname ev@B-R/4B/g2/s1/exam/task-p-low\endcsname{0.046}
\expandafter\def\csname ev@B-R/4B/g2/s1/exam/task-wins-high\endcsname{10}
\expandafter\def\csname ev@B-R/4B/g2/s1/exam/task-wins-low\endcsname{0}
\expandafter\def\csname ev@B-R/4B/g2/s1/exam/tasks\endcsname{100}
\expandafter\def\csname ev@B-R/4B/g2/s1/exam/uncounted\endcsname{0}
\expandafter\def\csname ev@B-R/4B/g2/s1/exam/uncounted-pct\endcsname{0.00}
\expandafter\def\csname ev@B-R/4B/g2/s1/exam/unit\endcsname{attempt slot}
\expandafter\def\csname ev@B-R/4B/g2/s1/exam/wins\endcsname{10}
\expandafter\def\csname ev@B-R/4B/g2/s1/retention-g1/attempt-ci-hi-pp\endcsname{+12.13}
\expandafter\def\csname ev@B-R/4B/g2/s1/retention-g1/attempt-ci-lo-pp\endcsname{-5.06}
\expandafter\def\csname ev@B-R/4B/g2/s1/retention-g1/base-counted\endcsname{224}
\expandafter\def\csname ev@B-R/4B/g2/s1/retention-g1/base-passes\endcsname{68}
\expandafter\def\csname ev@B-R/4B/g2/s1/retention-g1/base-pct\endcsname{30.36}
\expandafter\def\csname ev@B-R/4B/g2/s1/retention-g1/base-scheduled\endcsname{224}
\expandafter\def\csname ev@B-R/4B/g2/s1/retention-g1/base-uncounted\endcsname{0}
\expandafter\def\csname ev@B-R/4B/g2/s1/retention-g1/cand-counted\endcsname{224}
\expandafter\def\csname ev@B-R/4B/g2/s1/retention-g1/cand-passes\endcsname{76}
\expandafter\def\csname ev@B-R/4B/g2/s1/retention-g1/cand-pct\endcsname{33.93}
\expandafter\def\csname ev@B-R/4B/g2/s1/retention-g1/cand-scheduled\endcsname{224}
\expandafter\def\csname ev@B-R/4B/g2/s1/retention-g1/cand-uncounted\endcsname{0}
\expandafter\def\csname ev@B-R/4B/g2/s1/retention-g1/diff-pp\endcsname{+3.57}
\expandafter\def\csname ev@B-R/4B/g2/s1/retention-g1/losses\endcsname{19}
\expandafter\def\csname ev@B-R/4B/g2/s1/retention-g1/outcome\endcsname{inconclusive}
\expandafter\def\csname ev@B-R/4B/g2/s1/retention-g1/p-recorded\endcsname{0.151}
\expandafter\def\csname ev@B-R/4B/g2/s1/retention-g1/protocol\endcsname{B-R}
\expandafter\def\csname ev@B-R/4B/g2/s1/retention-g1/task-losses-high\endcsname{19}
\expandafter\def\csname ev@B-R/4B/g2/s1/retention-g1/task-losses-low\endcsname{0}
\expandafter\def\csname ev@B-R/4B/g2/s1/retention-g1/task-p-high\endcsname{0.995}
\expandafter\def\csname ev@B-R/4B/g2/s1/retention-g1/task-p-low\endcsname{\ensuremath{5.7\times 10^{-5}}}
\expandafter\def\csname ev@B-R/4B/g2/s1/retention-g1/task-wins-high\endcsname{27}
\expandafter\def\csname ev@B-R/4B/g2/s1/retention-g1/task-wins-low\endcsname{2}
\expandafter\def\csname ev@B-R/4B/g2/s1/retention-g1/tasks\endcsname{56}
\expandafter\def\csname ev@B-R/4B/g2/s1/retention-g1/uncounted\endcsname{0}
\expandafter\def\csname ev@B-R/4B/g2/s1/retention-g1/uncounted-pct\endcsname{0.00}
\expandafter\def\csname ev@B-R/4B/g2/s1/retention-g1/unit\endcsname{attempt slot}
\expandafter\def\csname ev@B-R/4B/g2/s1/retention-g1/wins\endcsname{27}
\expandafter\def\csname ev@B-R/4B/g2/s1/target/attempt-ci-hi-pp\endcsname{+9.84}
\expandafter\def\csname ev@B-R/4B/g2/s1/target/attempt-ci-lo-pp\endcsname{-5.60}
\expandafter\def\csname ev@B-R/4B/g2/s1/target/base-counted\endcsname{188}
\expandafter\def\csname ev@B-R/4B/g2/s1/target/base-passes\endcsname{31}
\expandafter\def\csname ev@B-R/4B/g2/s1/target/base-pct\endcsname{16.49}
\expandafter\def\csname ev@B-R/4B/g2/s1/target/base-scheduled\endcsname{188}
\expandafter\def\csname ev@B-R/4B/g2/s1/target/base-uncounted\endcsname{0}
\expandafter\def\csname ev@B-R/4B/g2/s1/target/cand-counted\endcsname{188}
\expandafter\def\csname ev@B-R/4B/g2/s1/target/cand-passes\endcsname{35}
\expandafter\def\csname ev@B-R/4B/g2/s1/target/cand-pct\endcsname{18.62}
\expandafter\def\csname ev@B-R/4B/g2/s1/target/cand-scheduled\endcsname{188}
\expandafter\def\csname ev@B-R/4B/g2/s1/target/cand-uncounted\endcsname{0}
\expandafter\def\csname ev@B-R/4B/g2/s1/target/diff-pp\endcsname{+2.13}
\expandafter\def\csname ev@B-R/4B/g2/s1/target/losses\endcsname{13}
\expandafter\def\csname ev@B-R/4B/g2/s1/target/outcome\endcsname{inconclusive}
\expandafter\def\csname ev@B-R/4B/g2/s1/target/p-recorded\endcsname{0.292}
\expandafter\def\csname ev@B-R/4B/g2/s1/target/protocol\endcsname{B-R}
\expandafter\def\csname ev@B-R/4B/g2/s1/target/task-losses-high\endcsname{13}
\expandafter\def\csname ev@B-R/4B/g2/s1/target/task-losses-low\endcsname{0}
\expandafter\def\csname ev@B-R/4B/g2/s1/target/task-p-high\endcsname{0.989}
\expandafter\def\csname ev@B-R/4B/g2/s1/target/task-p-low\endcsname{0.0022}
\expandafter\def\csname ev@B-R/4B/g2/s1/target/task-wins-high\endcsname{17}
\expandafter\def\csname ev@B-R/4B/g2/s1/target/task-wins-low\endcsname{1}
\expandafter\def\csname ev@B-R/4B/g2/s1/target/tasks\endcsname{47}
\expandafter\def\csname ev@B-R/4B/g2/s1/target/uncounted\endcsname{0}
\expandafter\def\csname ev@B-R/4B/g2/s1/target/uncounted-pct\endcsname{0.00}
\expandafter\def\csname ev@B-R/4B/g2/s1/target/unit\endcsname{attempt slot}
\expandafter\def\csname ev@B-R/4B/g2/s1/target/wins\endcsname{17}
\expandafter\def\csname ev@B-R/4B/g2/s2/exam/attempt-ci-hi-pp\endcsname{+2.82}
\expandafter\def\csname ev@B-R/4B/g2/s2/exam/attempt-ci-lo-pp\endcsname{-2.30}
\expandafter\def\csname ev@B-R/4B/g2/s2/exam/base-counted\endcsname{400}
\expandafter\def\csname ev@B-R/4B/g2/s2/exam/base-passes\endcsname{12}
\expandafter\def\csname ev@B-R/4B/g2/s2/exam/base-pct\endcsname{3.00}
\expandafter\def\csname ev@B-R/4B/g2/s2/exam/base-scheduled\endcsname{400}
\expandafter\def\csname ev@B-R/4B/g2/s2/exam/base-uncounted\endcsname{0}
\expandafter\def\csname ev@B-R/4B/g2/s2/exam/cand-counted\endcsname{400}
\expandafter\def\csname ev@B-R/4B/g2/s2/exam/cand-passes\endcsname{13}
\expandafter\def\csname ev@B-R/4B/g2/s2/exam/cand-pct\endcsname{3.25}
\expandafter\def\csname ev@B-R/4B/g2/s2/exam/cand-scheduled\endcsname{400}
\expandafter\def\csname ev@B-R/4B/g2/s2/exam/cand-uncounted\endcsname{0}
\expandafter\def\csname ev@B-R/4B/g2/s2/exam/diff-pp\endcsname{+0.25}
\expandafter\def\csname ev@B-R/4B/g2/s2/exam/losses\endcsname{11}
\expandafter\def\csname ev@B-R/4B/g2/s2/exam/outcome\endcsname{inconclusive}
\expandafter\def\csname ev@B-R/4B/g2/s2/exam/p-recorded\endcsname{0.5}
\expandafter\def\csname ev@B-R/4B/g2/s2/exam/protocol\endcsname{B-R}
\expandafter\def\csname ev@B-R/4B/g2/s2/exam/task-losses-high\endcsname{11}
\expandafter\def\csname ev@B-R/4B/g2/s2/exam/task-losses-low\endcsname{0}
\expandafter\def\csname ev@B-R/4B/g2/s2/exam/task-p-high\endcsname{0.994}
\expandafter\def\csname ev@B-R/4B/g2/s2/exam/task-p-low\endcsname{0.018}
\expandafter\def\csname ev@B-R/4B/g2/s2/exam/task-wins-high\endcsname{12}
\expandafter\def\csname ev@B-R/4B/g2/s2/exam/task-wins-low\endcsname{1}
\expandafter\def\csname ev@B-R/4B/g2/s2/exam/tasks\endcsname{100}
\expandafter\def\csname ev@B-R/4B/g2/s2/exam/uncounted\endcsname{0}
\expandafter\def\csname ev@B-R/4B/g2/s2/exam/uncounted-pct\endcsname{0.00}
\expandafter\def\csname ev@B-R/4B/g2/s2/exam/unit\endcsname{attempt slot}
\expandafter\def\csname ev@B-R/4B/g2/s2/exam/wins\endcsname{12}
\expandafter\def\csname ev@B-R/4B/g2/s2/retention-g1/attempt-ci-hi-pp\endcsname{+15.30}
\expandafter\def\csname ev@B-R/4B/g2/s2/retention-g1/attempt-ci-lo-pp\endcsname{-2.05}
\expandafter\def\csname ev@B-R/4B/g2/s2/retention-g1/base-counted\endcsname{224}
\expandafter\def\csname ev@B-R/4B/g2/s2/retention-g1/base-passes\endcsname{68}
\expandafter\def\csname ev@B-R/4B/g2/s2/retention-g1/base-pct\endcsname{30.36}
\expandafter\def\csname ev@B-R/4B/g2/s2/retention-g1/base-scheduled\endcsname{224}
\expandafter\def\csname ev@B-R/4B/g2/s2/retention-g1/base-uncounted\endcsname{0}
\expandafter\def\csname ev@B-R/4B/g2/s2/retention-g1/cand-counted\endcsname{224}
\expandafter\def\csname ev@B-R/4B/g2/s2/retention-g1/cand-passes\endcsname{83}
\expandafter\def\csname ev@B-R/4B/g2/s2/retention-g1/cand-pct\endcsname{37.05}
\expandafter\def\csname ev@B-R/4B/g2/s2/retention-g1/cand-scheduled\endcsname{224}
\expandafter\def\csname ev@B-R/4B/g2/s2/retention-g1/cand-uncounted\endcsname{0}
\expandafter\def\csname ev@B-R/4B/g2/s2/retention-g1/diff-pp\endcsname{+6.70}
\expandafter\def\csname ev@B-R/4B/g2/s2/retention-g1/losses\endcsname{12}
\expandafter\def\csname ev@B-R/4B/g2/s2/retention-g1/outcome\endcsname{inconclusive}
\expandafter\def\csname ev@B-R/4B/g2/s2/retention-g1/p-recorded\endcsname{0.012}
\expandafter\def\csname ev@B-R/4B/g2/s2/retention-g1/protocol\endcsname{B-R}
\expandafter\def\csname ev@B-R/4B/g2/s2/retention-g1/task-losses-high\endcsname{12}
\expandafter\def\csname ev@B-R/4B/g2/s2/retention-g1/task-losses-low\endcsname{0}
\expandafter\def\csname ev@B-R/4B/g2/s2/retention-g1/task-p-high\endcsname{0.916}
\expandafter\def\csname ev@B-R/4B/g2/s2/retention-g1/task-p-low\endcsname{\ensuremath{4.2\times 10^{-6}}}
\expandafter\def\csname ev@B-R/4B/g2/s2/retention-g1/task-wins-high\endcsname{27}
\expandafter\def\csname ev@B-R/4B/g2/s2/retention-g1/task-wins-low\endcsname{4}
\expandafter\def\csname ev@B-R/4B/g2/s2/retention-g1/tasks\endcsname{56}
\expandafter\def\csname ev@B-R/4B/g2/s2/retention-g1/uncounted\endcsname{0}
\expandafter\def\csname ev@B-R/4B/g2/s2/retention-g1/uncounted-pct\endcsname{0.00}
\expandafter\def\csname ev@B-R/4B/g2/s2/retention-g1/unit\endcsname{attempt slot}
\expandafter\def\csname ev@B-R/4B/g2/s2/retention-g1/wins\endcsname{27}
\expandafter\def\csname ev@B-R/4B/g2/s2/target/attempt-ci-hi-pp\endcsname{+8.70}
\expandafter\def\csname ev@B-R/4B/g2/s2/target/attempt-ci-lo-pp\endcsname{-6.58}
\expandafter\def\csname ev@B-R/4B/g2/s2/target/base-counted\endcsname{188}
\expandafter\def\csname ev@B-R/4B/g2/s2/target/base-passes\endcsname{31}
\expandafter\def\csname ev@B-R/4B/g2/s2/target/base-pct\endcsname{16.49}
\expandafter\def\csname ev@B-R/4B/g2/s2/target/base-scheduled\endcsname{188}
\expandafter\def\csname ev@B-R/4B/g2/s2/target/base-uncounted\endcsname{0}
\expandafter\def\csname ev@B-R/4B/g2/s2/target/cand-counted\endcsname{188}
\expandafter\def\csname ev@B-R/4B/g2/s2/target/cand-passes\endcsname{33}
\expandafter\def\csname ev@B-R/4B/g2/s2/target/cand-pct\endcsname{17.55}
\expandafter\def\csname ev@B-R/4B/g2/s2/target/cand-scheduled\endcsname{188}
\expandafter\def\csname ev@B-R/4B/g2/s2/target/cand-uncounted\endcsname{0}
\expandafter\def\csname ev@B-R/4B/g2/s2/target/diff-pp\endcsname{+1.06}
\expandafter\def\csname ev@B-R/4B/g2/s2/target/losses\endcsname{12}
\expandafter\def\csname ev@B-R/4B/g2/s2/target/outcome\endcsname{inconclusive}
\expandafter\def\csname ev@B-R/4B/g2/s2/target/p-recorded\endcsname{0.423}
\expandafter\def\csname ev@B-R/4B/g2/s2/target/protocol\endcsname{B-R}
\expandafter\def\csname ev@B-R/4B/g2/s2/target/task-losses-high\endcsname{12}
\expandafter\def\csname ev@B-R/4B/g2/s2/target/task-losses-low\endcsname{0}
\expandafter\def\csname ev@B-R/4B/g2/s2/target/task-p-high\endcsname{0.989}
\expandafter\def\csname ev@B-R/4B/g2/s2/target/task-p-low\endcsname{0.0064}
\expandafter\def\csname ev@B-R/4B/g2/s2/target/task-wins-high\endcsname{14}
\expandafter\def\csname ev@B-R/4B/g2/s2/target/task-wins-low\endcsname{1}
\expandafter\def\csname ev@B-R/4B/g2/s2/target/tasks\endcsname{47}
\expandafter\def\csname ev@B-R/4B/g2/s2/target/uncounted\endcsname{0}
\expandafter\def\csname ev@B-R/4B/g2/s2/target/uncounted-pct\endcsname{0.00}
\expandafter\def\csname ev@B-R/4B/g2/s2/target/unit\endcsname{attempt slot}
\expandafter\def\csname ev@B-R/4B/g2/s2/target/wins\endcsname{14}
\expandafter\def\csname ev@B-R/4B/g2/s3/exam/attempt-ci-hi-pp\endcsname{+4.31}
\expandafter\def\csname ev@B-R/4B/g2/s3/exam/attempt-ci-lo-pp\endcsname{-1.22}
\expandafter\def\csname ev@B-R/4B/g2/s3/exam/base-counted\endcsname{400}
\expandafter\def\csname ev@B-R/4B/g2/s3/exam/base-passes\endcsname{12}
\expandafter\def\csname ev@B-R/4B/g2/s3/exam/base-pct\endcsname{3.00}
\expandafter\def\csname ev@B-R/4B/g2/s3/exam/base-scheduled\endcsname{400}
\expandafter\def\csname ev@B-R/4B/g2/s3/exam/base-uncounted\endcsname{0}
\expandafter\def\csname ev@B-R/4B/g2/s3/exam/cand-counted\endcsname{400}
\expandafter\def\csname ev@B-R/4B/g2/s3/exam/cand-passes\endcsname{18}
\expandafter\def\csname ev@B-R/4B/g2/s3/exam/cand-pct\endcsname{4.50}
\expandafter\def\csname ev@B-R/4B/g2/s3/exam/cand-scheduled\endcsname{400}
\expandafter\def\csname ev@B-R/4B/g2/s3/exam/cand-uncounted\endcsname{0}
\expandafter\def\csname ev@B-R/4B/g2/s3/exam/diff-pp\endcsname{+1.50}
\expandafter\def\csname ev@B-R/4B/g2/s3/exam/losses\endcsname{7}
\expandafter\def\csname ev@B-R/4B/g2/s3/exam/outcome\endcsname{inconclusive}
\expandafter\def\csname ev@B-R/4B/g2/s3/exam/p-recorded\endcsname{0.132}
\expandafter\def\csname ev@B-R/4B/g2/s3/exam/protocol\endcsname{B-R}
\expandafter\def\csname ev@B-R/4B/g2/s3/exam/task-losses-high\endcsname{7}
\expandafter\def\csname ev@B-R/4B/g2/s3/exam/task-losses-low\endcsname{0}
\expandafter\def\csname ev@B-R/4B/g2/s3/exam/task-p-high\endcsname{0.91}
\expandafter\def\csname ev@B-R/4B/g2/s3/exam/task-p-low\endcsname{0.0037}
\expandafter\def\csname ev@B-R/4B/g2/s3/exam/task-wins-high\endcsname{13}
\expandafter\def\csname ev@B-R/4B/g2/s3/exam/task-wins-low\endcsname{2}
\expandafter\def\csname ev@B-R/4B/g2/s3/exam/tasks\endcsname{100}
\expandafter\def\csname ev@B-R/4B/g2/s3/exam/uncounted\endcsname{0}
\expandafter\def\csname ev@B-R/4B/g2/s3/exam/uncounted-pct\endcsname{0.00}
\expandafter\def\csname ev@B-R/4B/g2/s3/exam/unit\endcsname{attempt slot}
\expandafter\def\csname ev@B-R/4B/g2/s3/exam/wins\endcsname{13}
\expandafter\def\csname ev@B-R/4B/g2/s3/retention-g1/attempt-ci-hi-pp\endcsname{+17.56}
\expandafter\def\csname ev@B-R/4B/g2/s3/retention-g1/attempt-ci-lo-pp\endcsname{+0.11}
\expandafter\def\csname ev@B-R/4B/g2/s3/retention-g1/base-counted\endcsname{224}
\expandafter\def\csname ev@B-R/4B/g2/s3/retention-g1/base-passes\endcsname{68}
\expandafter\def\csname ev@B-R/4B/g2/s3/retention-g1/base-pct\endcsname{30.36}
\expandafter\def\csname ev@B-R/4B/g2/s3/retention-g1/base-scheduled\endcsname{224}
\expandafter\def\csname ev@B-R/4B/g2/s3/retention-g1/base-uncounted\endcsname{0}
\expandafter\def\csname ev@B-R/4B/g2/s3/retention-g1/cand-counted\endcsname{224}
\expandafter\def\csname ev@B-R/4B/g2/s3/retention-g1/cand-passes\endcsname{88}
\expandafter\def\csname ev@B-R/4B/g2/s3/retention-g1/cand-pct\endcsname{39.29}
\expandafter\def\csname ev@B-R/4B/g2/s3/retention-g1/cand-scheduled\endcsname{224}
\expandafter\def\csname ev@B-R/4B/g2/s3/retention-g1/cand-uncounted\endcsname{0}
\expandafter\def\csname ev@B-R/4B/g2/s3/retention-g1/diff-pp\endcsname{+8.93}
\expandafter\def\csname ev@B-R/4B/g2/s3/retention-g1/losses\endcsname{19}
\expandafter\def\csname ev@B-R/4B/g2/s3/retention-g1/outcome\endcsname{inconclusive}
\expandafter\def\csname ev@B-R/4B/g2/s3/retention-g1/p-recorded\endcsname{0.006}
\expandafter\def\csname ev@B-R/4B/g2/s3/retention-g1/protocol\endcsname{B-R}
\expandafter\def\csname ev@B-R/4B/g2/s3/retention-g1/task-losses-high\endcsname{19}
\expandafter\def\csname ev@B-R/4B/g2/s3/retention-g1/task-losses-low\endcsname{0}
\expandafter\def\csname ev@B-R/4B/g2/s3/retention-g1/task-p-high\endcsname{0.969}
\expandafter\def\csname ev@B-R/4B/g2/s3/retention-g1/task-p-low\endcsname{\ensuremath{7.0\times 10^{-8}}}
\expandafter\def\csname ev@B-R/4B/g2/s3/retention-g1/task-wins-high\endcsname{39}
\expandafter\def\csname ev@B-R/4B/g2/s3/retention-g1/task-wins-low\endcsname{5}
\expandafter\def\csname ev@B-R/4B/g2/s3/retention-g1/tasks\endcsname{56}
\expandafter\def\csname ev@B-R/4B/g2/s3/retention-g1/uncounted\endcsname{0}
\expandafter\def\csname ev@B-R/4B/g2/s3/retention-g1/uncounted-pct\endcsname{0.00}
\expandafter\def\csname ev@B-R/4B/g2/s3/retention-g1/unit\endcsname{attempt slot}
\expandafter\def\csname ev@B-R/4B/g2/s3/retention-g1/wins\endcsname{39}
\expandafter\def\csname ev@B-R/4B/g2/s3/target/attempt-ci-hi-pp\endcsname{+8.12}
\expandafter\def\csname ev@B-R/4B/g2/s3/target/attempt-ci-lo-pp\endcsname{-7.06}
\expandafter\def\csname ev@B-R/4B/g2/s3/target/base-counted\endcsname{188}
\expandafter\def\csname ev@B-R/4B/g2/s3/target/base-passes\endcsname{31}
\expandafter\def\csname ev@B-R/4B/g2/s3/target/base-pct\endcsname{16.49}
\expandafter\def\csname ev@B-R/4B/g2/s3/target/base-scheduled\endcsname{188}
\expandafter\def\csname ev@B-R/4B/g2/s3/target/base-uncounted\endcsname{0}
\expandafter\def\csname ev@B-R/4B/g2/s3/target/cand-counted\endcsname{188}
\expandafter\def\csname ev@B-R/4B/g2/s3/target/cand-passes\endcsname{32}
\expandafter\def\csname ev@B-R/4B/g2/s3/target/cand-pct\endcsname{17.02}
\expandafter\def\csname ev@B-R/4B/g2/s3/target/cand-scheduled\endcsname{188}
\expandafter\def\csname ev@B-R/4B/g2/s3/target/cand-uncounted\endcsname{0}
\expandafter\def\csname ev@B-R/4B/g2/s3/target/diff-pp\endcsname{+0.53}
\expandafter\def\csname ev@B-R/4B/g2/s3/target/losses\endcsname{16}
\expandafter\def\csname ev@B-R/4B/g2/s3/target/outcome\endcsname{inconclusive}
\expandafter\def\csname ev@B-R/4B/g2/s3/target/p-recorded\endcsname{0.5}
\expandafter\def\csname ev@B-R/4B/g2/s3/target/protocol\endcsname{B-R}
\expandafter\def\csname ev@B-R/4B/g2/s3/target/task-losses-high\endcsname{16}
\expandafter\def\csname ev@B-R/4B/g2/s3/target/task-losses-low\endcsname{0}
\expandafter\def\csname ev@B-R/4B/g2/s3/target/task-p-high\endcsname{0.998}
\expandafter\def\csname ev@B-R/4B/g2/s3/target/task-p-low\endcsname{0.0036}
\expandafter\def\csname ev@B-R/4B/g2/s3/target/task-wins-high\endcsname{17}
\expandafter\def\csname ev@B-R/4B/g2/s3/target/task-wins-low\endcsname{1}
\expandafter\def\csname ev@B-R/4B/g2/s3/target/tasks\endcsname{47}
\expandafter\def\csname ev@B-R/4B/g2/s3/target/uncounted\endcsname{0}
\expandafter\def\csname ev@B-R/4B/g2/s3/target/uncounted-pct\endcsname{0.00}
\expandafter\def\csname ev@B-R/4B/g2/s3/target/unit\endcsname{attempt slot}
\expandafter\def\csname ev@B-R/4B/g2/s3/target/wins\endcsname{17}
\expandafter\def\csname ev@B-R/4B/g2/s4/exam/attempt-ci-hi-pp\endcsname{+4.01}
\expandafter\def\csname ev@B-R/4B/g2/s4/exam/attempt-ci-lo-pp\endcsname{-1.43}
\expandafter\def\csname ev@B-R/4B/g2/s4/exam/base-counted\endcsname{400}
\expandafter\def\csname ev@B-R/4B/g2/s4/exam/base-passes\endcsname{12}
\expandafter\def\csname ev@B-R/4B/g2/s4/exam/base-pct\endcsname{3.00}
\expandafter\def\csname ev@B-R/4B/g2/s4/exam/base-scheduled\endcsname{400}
\expandafter\def\csname ev@B-R/4B/g2/s4/exam/base-uncounted\endcsname{0}
\expandafter\def\csname ev@B-R/4B/g2/s4/exam/cand-counted\endcsname{400}
\expandafter\def\csname ev@B-R/4B/g2/s4/exam/cand-passes\endcsname{17}
\expandafter\def\csname ev@B-R/4B/g2/s4/exam/cand-pct\endcsname{4.25}
\expandafter\def\csname ev@B-R/4B/g2/s4/exam/cand-scheduled\endcsname{400}
\expandafter\def\csname ev@B-R/4B/g2/s4/exam/cand-uncounted\endcsname{0}
\expandafter\def\csname ev@B-R/4B/g2/s4/exam/diff-pp\endcsname{+1.25}
\expandafter\def\csname ev@B-R/4B/g2/s4/exam/losses\endcsname{7}
\expandafter\def\csname ev@B-R/4B/g2/s4/exam/outcome\endcsname{inconclusive}
\expandafter\def\csname ev@B-R/4B/g2/s4/exam/p-recorded\endcsname{0.18}
\expandafter\def\csname ev@B-R/4B/g2/s4/exam/protocol\endcsname{B-R}
\expandafter\def\csname ev@B-R/4B/g2/s4/exam/task-losses-high\endcsname{7}
\expandafter\def\csname ev@B-R/4B/g2/s4/exam/task-losses-low\endcsname{0}
\expandafter\def\csname ev@B-R/4B/g2/s4/exam/task-p-high\endcsname{0.945}
\expandafter\def\csname ev@B-R/4B/g2/s4/exam/task-p-low\endcsname{0.0065}
\expandafter\def\csname ev@B-R/4B/g2/s4/exam/task-wins-high\endcsname{12}
\expandafter\def\csname ev@B-R/4B/g2/s4/exam/task-wins-low\endcsname{2}
\expandafter\def\csname ev@B-R/4B/g2/s4/exam/tasks\endcsname{100}
\expandafter\def\csname ev@B-R/4B/g2/s4/exam/uncounted\endcsname{0}
\expandafter\def\csname ev@B-R/4B/g2/s4/exam/uncounted-pct\endcsname{0.00}
\expandafter\def\csname ev@B-R/4B/g2/s4/exam/unit\endcsname{attempt slot}
\expandafter\def\csname ev@B-R/4B/g2/s4/exam/wins\endcsname{12}
\expandafter\def\csname ev@B-R/4B/g2/s4/retention-g1/attempt-ci-hi-pp\endcsname{+16.21}
\expandafter\def\csname ev@B-R/4B/g2/s4/retention-g1/attempt-ci-lo-pp\endcsname{-1.19}
\expandafter\def\csname ev@B-R/4B/g2/s4/retention-g1/base-counted\endcsname{224}
\expandafter\def\csname ev@B-R/4B/g2/s4/retention-g1/base-passes\endcsname{68}
\expandafter\def\csname ev@B-R/4B/g2/s4/retention-g1/base-pct\endcsname{30.36}
\expandafter\def\csname ev@B-R/4B/g2/s4/retention-g1/base-scheduled\endcsname{224}
\expandafter\def\csname ev@B-R/4B/g2/s4/retention-g1/base-uncounted\endcsname{0}
\expandafter\def\csname ev@B-R/4B/g2/s4/retention-g1/cand-counted\endcsname{224}
\expandafter\def\csname ev@B-R/4B/g2/s4/retention-g1/cand-passes\endcsname{85}
\expandafter\def\csname ev@B-R/4B/g2/s4/retention-g1/cand-pct\endcsname{37.95}
\expandafter\def\csname ev@B-R/4B/g2/s4/retention-g1/cand-scheduled\endcsname{224}
\expandafter\def\csname ev@B-R/4B/g2/s4/retention-g1/cand-uncounted\endcsname{0}
\expandafter\def\csname ev@B-R/4B/g2/s4/retention-g1/diff-pp\endcsname{+7.59}
\expandafter\def\csname ev@B-R/4B/g2/s4/retention-g1/losses\endcsname{17}
\expandafter\def\csname ev@B-R/4B/g2/s4/retention-g1/outcome\endcsname{inconclusive}
\expandafter\def\csname ev@B-R/4B/g2/s4/retention-g1/p-recorded\endcsname{0.012}
\expandafter\def\csname ev@B-R/4B/g2/s4/retention-g1/protocol\endcsname{B-R}
\expandafter\def\csname ev@B-R/4B/g2/s4/retention-g1/task-losses-high\endcsname{17}
\expandafter\def\csname ev@B-R/4B/g2/s4/retention-g1/task-losses-low\endcsname{0}
\expandafter\def\csname ev@B-R/4B/g2/s4/retention-g1/task-p-high\endcsname{0.962}
\expandafter\def\csname ev@B-R/4B/g2/s4/retention-g1/task-p-low\endcsname{\ensuremath{5.4\times 10^{-7}}}
\expandafter\def\csname ev@B-R/4B/g2/s4/retention-g1/task-wins-high\endcsname{34}
\expandafter\def\csname ev@B-R/4B/g2/s4/retention-g1/task-wins-low\endcsname{5}
\expandafter\def\csname ev@B-R/4B/g2/s4/retention-g1/tasks\endcsname{56}
\expandafter\def\csname ev@B-R/4B/g2/s4/retention-g1/uncounted\endcsname{0}
\expandafter\def\csname ev@B-R/4B/g2/s4/retention-g1/uncounted-pct\endcsname{0.00}
\expandafter\def\csname ev@B-R/4B/g2/s4/retention-g1/unit\endcsname{attempt slot}
\expandafter\def\csname ev@B-R/4B/g2/s4/retention-g1/wins\endcsname{34}
\expandafter\def\csname ev@B-R/4B/g2/s4/target/attempt-ci-hi-pp\endcsname{+13.24}
\expandafter\def\csname ev@B-R/4B/g2/s4/target/attempt-ci-lo-pp\endcsname{-2.66}
\expandafter\def\csname ev@B-R/4B/g2/s4/target/base-counted\endcsname{188}
\expandafter\def\csname ev@B-R/4B/g2/s4/target/base-passes\endcsname{31}
\expandafter\def\csname ev@B-R/4B/g2/s4/target/base-pct\endcsname{16.49}
\expandafter\def\csname ev@B-R/4B/g2/s4/target/base-scheduled\endcsname{188}
\expandafter\def\csname ev@B-R/4B/g2/s4/target/base-uncounted\endcsname{0}
\expandafter\def\csname ev@B-R/4B/g2/s4/target/cand-counted\endcsname{188}
\expandafter\def\csname ev@B-R/4B/g2/s4/target/cand-passes\endcsname{41}
\expandafter\def\csname ev@B-R/4B/g2/s4/target/cand-pct\endcsname{21.81}
\expandafter\def\csname ev@B-R/4B/g2/s4/target/cand-scheduled\endcsname{188}
\expandafter\def\csname ev@B-R/4B/g2/s4/target/cand-uncounted\endcsname{0}
\expandafter\def\csname ev@B-R/4B/g2/s4/target/diff-pp\endcsname{+5.32}
\expandafter\def\csname ev@B-R/4B/g2/s4/target/losses\endcsname{10}
\expandafter\def\csname ev@B-R/4B/g2/s4/target/outcome\endcsname{inconclusive}
\expandafter\def\csname ev@B-R/4B/g2/s4/target/p-recorded\endcsname{0.049}
\expandafter\def\csname ev@B-R/4B/g2/s4/target/protocol\endcsname{B-R}
\expandafter\def\csname ev@B-R/4B/g2/s4/target/task-losses-high\endcsname{10}
\expandafter\def\csname ev@B-R/4B/g2/s4/target/task-losses-low\endcsname{0}
\expandafter\def\csname ev@B-R/4B/g2/s4/target/task-p-high\endcsname{0.941}
\expandafter\def\csname ev@B-R/4B/g2/s4/target/task-p-low\endcsname{0.0002}
\expandafter\def\csname ev@B-R/4B/g2/s4/target/task-wins-high\endcsname{20}
\expandafter\def\csname ev@B-R/4B/g2/s4/target/task-wins-low\endcsname{3}
\expandafter\def\csname ev@B-R/4B/g2/s4/target/tasks\endcsname{47}
\expandafter\def\csname ev@B-R/4B/g2/s4/target/uncounted\endcsname{0}
\expandafter\def\csname ev@B-R/4B/g2/s4/target/uncounted-pct\endcsname{0.00}
\expandafter\def\csname ev@B-R/4B/g2/s4/target/unit\endcsname{attempt slot}
\expandafter\def\csname ev@B-R/4B/g2/s4/target/wins\endcsname{20}
\expandafter\def\csname ev@B-R/4B/g2/s5/exam/attempt-ci-hi-pp\endcsname{+2.82}
\expandafter\def\csname ev@B-R/4B/g2/s5/exam/attempt-ci-lo-pp\endcsname{-2.30}
\expandafter\def\csname ev@B-R/4B/g2/s5/exam/base-counted\endcsname{400}
\expandafter\def\csname ev@B-R/4B/g2/s5/exam/base-passes\endcsname{12}
\expandafter\def\csname ev@B-R/4B/g2/s5/exam/base-pct\endcsname{3.00}
\expandafter\def\csname ev@B-R/4B/g2/s5/exam/base-scheduled\endcsname{400}
\expandafter\def\csname ev@B-R/4B/g2/s5/exam/base-uncounted\endcsname{0}
\expandafter\def\csname ev@B-R/4B/g2/s5/exam/cand-counted\endcsname{400}
\expandafter\def\csname ev@B-R/4B/g2/s5/exam/cand-passes\endcsname{13}
\expandafter\def\csname ev@B-R/4B/g2/s5/exam/cand-pct\endcsname{3.25}
\expandafter\def\csname ev@B-R/4B/g2/s5/exam/cand-scheduled\endcsname{400}
\expandafter\def\csname ev@B-R/4B/g2/s5/exam/cand-uncounted\endcsname{0}
\expandafter\def\csname ev@B-R/4B/g2/s5/exam/diff-pp\endcsname{+0.25}
\expandafter\def\csname ev@B-R/4B/g2/s5/exam/losses\endcsname{10}
\expandafter\def\csname ev@B-R/4B/g2/s5/exam/outcome\endcsname{inconclusive}
\expandafter\def\csname ev@B-R/4B/g2/s5/exam/p-recorded\endcsname{0.5}
\expandafter\def\csname ev@B-R/4B/g2/s5/exam/protocol\endcsname{B-R}
\expandafter\def\csname ev@B-R/4B/g2/s5/exam/task-losses-high\endcsname{10}
\expandafter\def\csname ev@B-R/4B/g2/s5/exam/task-losses-low\endcsname{0}
\expandafter\def\csname ev@B-R/4B/g2/s5/exam/task-p-high\endcsname{0.989}
\expandafter\def\csname ev@B-R/4B/g2/s5/exam/task-p-low\endcsname{0.029}
\expandafter\def\csname ev@B-R/4B/g2/s5/exam/task-wins-high\endcsname{11}
\expandafter\def\csname ev@B-R/4B/g2/s5/exam/task-wins-low\endcsname{1}
\expandafter\def\csname ev@B-R/4B/g2/s5/exam/tasks\endcsname{100}
\expandafter\def\csname ev@B-R/4B/g2/s5/exam/uncounted\endcsname{0}
\expandafter\def\csname ev@B-R/4B/g2/s5/exam/uncounted-pct\endcsname{0.00}
\expandafter\def\csname ev@B-R/4B/g2/s5/exam/unit\endcsname{attempt slot}
\expandafter\def\csname ev@B-R/4B/g2/s5/exam/wins\endcsname{11}
\expandafter\def\csname ev@B-R/4B/g2/s5/retention-g1/attempt-ci-hi-pp\endcsname{+13.04}
\expandafter\def\csname ev@B-R/4B/g2/s5/retention-g1/attempt-ci-lo-pp\endcsname{-4.20}
\expandafter\def\csname ev@B-R/4B/g2/s5/retention-g1/base-counted\endcsname{224}
\expandafter\def\csname ev@B-R/4B/g2/s5/retention-g1/base-passes\endcsname{68}
\expandafter\def\csname ev@B-R/4B/g2/s5/retention-g1/base-pct\endcsname{30.36}
\expandafter\def\csname ev@B-R/4B/g2/s5/retention-g1/base-scheduled\endcsname{224}
\expandafter\def\csname ev@B-R/4B/g2/s5/retention-g1/base-uncounted\endcsname{0}
\expandafter\def\csname ev@B-R/4B/g2/s5/retention-g1/cand-counted\endcsname{224}
\expandafter\def\csname ev@B-R/4B/g2/s5/retention-g1/cand-passes\endcsname{78}
\expandafter\def\csname ev@B-R/4B/g2/s5/retention-g1/cand-pct\endcsname{34.82}
\expandafter\def\csname ev@B-R/4B/g2/s5/retention-g1/cand-scheduled\endcsname{224}
\expandafter\def\csname ev@B-R/4B/g2/s5/retention-g1/cand-uncounted\endcsname{0}
\expandafter\def\csname ev@B-R/4B/g2/s5/retention-g1/diff-pp\endcsname{+4.46}
\expandafter\def\csname ev@B-R/4B/g2/s5/retention-g1/losses\endcsname{17}
\expandafter\def\csname ev@B-R/4B/g2/s5/retention-g1/outcome\endcsname{inconclusive}
\expandafter\def\csname ev@B-R/4B/g2/s5/retention-g1/p-recorded\endcsname{0.087}
\expandafter\def\csname ev@B-R/4B/g2/s5/retention-g1/protocol\endcsname{B-R}
\expandafter\def\csname ev@B-R/4B/g2/s5/retention-g1/task-losses-high\endcsname{17}
\expandafter\def\csname ev@B-R/4B/g2/s5/retention-g1/task-losses-low\endcsname{0}
\expandafter\def\csname ev@B-R/4B/g2/s5/retention-g1/task-p-high\endcsname{0.989}
\expandafter\def\csname ev@B-R/4B/g2/s5/retention-g1/task-p-low\endcsname{\ensuremath{3.0\times 10^{-5}}}
\expandafter\def\csname ev@B-R/4B/g2/s5/retention-g1/task-wins-high\endcsname{27}
\expandafter\def\csname ev@B-R/4B/g2/s5/retention-g1/task-wins-low\endcsname{3}
\expandafter\def\csname ev@B-R/4B/g2/s5/retention-g1/tasks\endcsname{56}
\expandafter\def\csname ev@B-R/4B/g2/s5/retention-g1/uncounted\endcsname{0}
\expandafter\def\csname ev@B-R/4B/g2/s5/retention-g1/uncounted-pct\endcsname{0.00}
\expandafter\def\csname ev@B-R/4B/g2/s5/retention-g1/unit\endcsname{attempt slot}
\expandafter\def\csname ev@B-R/4B/g2/s5/retention-g1/wins\endcsname{27}
\expandafter\def\csname ev@B-R/4B/g2/s5/target/attempt-ci-hi-pp\endcsname{+4.07}
\expandafter\def\csname ev@B-R/4B/g2/s5/target/attempt-ci-lo-pp\endcsname{-10.44}
\expandafter\def\csname ev@B-R/4B/g2/s5/target/base-counted\endcsname{188}
\expandafter\def\csname ev@B-R/4B/g2/s5/target/base-passes\endcsname{31}
\expandafter\def\csname ev@B-R/4B/g2/s5/target/base-pct\endcsname{16.49}
\expandafter\def\csname ev@B-R/4B/g2/s5/target/base-scheduled\endcsname{188}
\expandafter\def\csname ev@B-R/4B/g2/s5/target/base-uncounted\endcsname{0}
\expandafter\def\csname ev@B-R/4B/g2/s5/target/cand-counted\endcsname{188}
\expandafter\def\csname ev@B-R/4B/g2/s5/target/cand-passes\endcsname{25}
\expandafter\def\csname ev@B-R/4B/g2/s5/target/cand-pct\endcsname{13.30}
\expandafter\def\csname ev@B-R/4B/g2/s5/target/cand-scheduled\endcsname{188}
\expandafter\def\csname ev@B-R/4B/g2/s5/target/cand-uncounted\endcsname{0}
\expandafter\def\csname ev@B-R/4B/g2/s5/target/diff-pp\endcsname{-3.19}
\expandafter\def\csname ev@B-R/4B/g2/s5/target/losses\endcsname{14}
\expandafter\def\csname ev@B-R/4B/g2/s5/target/outcome\endcsname{inconclusive}
\expandafter\def\csname ev@B-R/4B/g2/s5/target/p-recorded\endcsname{0.933}
\expandafter\def\csname ev@B-R/4B/g2/s5/target/protocol\endcsname{B-R}
\expandafter\def\csname ev@B-R/4B/g2/s5/target/task-losses-high\endcsname{14}
\expandafter\def\csname ev@B-R/4B/g2/s5/target/task-losses-low\endcsname{2}
\expandafter\def\csname ev@B-R/4B/g2/s5/target/task-p-high\endcsname{1}
\expandafter\def\csname ev@B-R/4B/g2/s5/target/task-p-low\endcsname{0.194}
\expandafter\def\csname ev@B-R/4B/g2/s5/target/task-wins-high\endcsname{8}
\expandafter\def\csname ev@B-R/4B/g2/s5/target/task-wins-low\endcsname{0}
\expandafter\def\csname ev@B-R/4B/g2/s5/target/tasks\endcsname{47}
\expandafter\def\csname ev@B-R/4B/g2/s5/target/uncounted\endcsname{0}
\expandafter\def\csname ev@B-R/4B/g2/s5/target/uncounted-pct\endcsname{0.00}
\expandafter\def\csname ev@B-R/4B/g2/s5/target/unit\endcsname{attempt slot}
\expandafter\def\csname ev@B-R/4B/g2/s5/target/wins\endcsname{8}
\expandafter\def\csname ev@B-R/4B/g2/unfinished-seeds\endcsname{0}
\expandafter\def\csname ev@B-R/4B/g3/pooled/exam/base-counted\endcsname{400}
\expandafter\def\csname ev@B-R/4B/g3/pooled/exam/base-passes\endcsname{12}
\expandafter\def\csname ev@B-R/4B/g3/pooled/exam/base-pct\endcsname{3.00}
\expandafter\def\csname ev@B-R/4B/g3/pooled/exam/base-scheduled\endcsname{400}
\expandafter\def\csname ev@B-R/4B/g3/pooled/exam/base-uncounted\endcsname{0}
\expandafter\def\csname ev@B-R/4B/g3/pooled/exam/cand-counted\endcsname{2000}
\expandafter\def\csname ev@B-R/4B/g3/pooled/exam/cand-passes\endcsname{72}
\expandafter\def\csname ev@B-R/4B/g3/pooled/exam/cand-pct\endcsname{3.60}
\expandafter\def\csname ev@B-R/4B/g3/pooled/exam/cand-scheduled\endcsname{2000}
\expandafter\def\csname ev@B-R/4B/g3/pooled/exam/cand-uncounted\endcsname{0}
\expandafter\def\csname ev@B-R/4B/g3/pooled/exam/diff-pp\endcsname{+0.60}
\expandafter\def\csname ev@B-R/4B/g3/pooled/exam/losses\endcsname{44}
\expandafter\def\csname ev@B-R/4B/g3/pooled/exam/outcome\endcsname{inconclusive}
\expandafter\def\csname ev@B-R/4B/g3/pooled/exam/p-recorded\endcsname{0.136}
\expandafter\def\csname ev@B-R/4B/g3/pooled/exam/protocol\endcsname{B-R}
\expandafter\def\csname ev@B-R/4B/g3/pooled/exam/seeds\endcsname{5}
\expandafter\def\csname ev@B-R/4B/g3/pooled/exam/seeds-negative\endcsname{0}
\expandafter\def\csname ev@B-R/4B/g3/pooled/exam/seeds-positive\endcsname{4}
\expandafter\def\csname ev@B-R/4B/g3/pooled/exam/tasks\endcsname{100}
\expandafter\def\csname ev@B-R/4B/g3/pooled/exam/uncounted\endcsname{0}
\expandafter\def\csname ev@B-R/4B/g3/pooled/exam/uncounted-pct\endcsname{0.00}
\expandafter\def\csname ev@B-R/4B/g3/pooled/exam/unit\endcsname{attempt slot, summed over training seeds}
\expandafter\def\csname ev@B-R/4B/g3/pooled/exam/wins\endcsname{56}
\expandafter\def\csname ev@B-R/4B/g3/pooled/retention-g1/base-counted\endcsname{224}
\expandafter\def\csname ev@B-R/4B/g3/pooled/retention-g1/base-passes\endcsname{68}
\expandafter\def\csname ev@B-R/4B/g3/pooled/retention-g1/base-pct\endcsname{30.36}
\expandafter\def\csname ev@B-R/4B/g3/pooled/retention-g1/base-scheduled\endcsname{224}
\expandafter\def\csname ev@B-R/4B/g3/pooled/retention-g1/base-uncounted\endcsname{0}
\expandafter\def\csname ev@B-R/4B/g3/pooled/retention-g1/cand-counted\endcsname{1120}
\expandafter\def\csname ev@B-R/4B/g3/pooled/retention-g1/cand-passes\endcsname{438}
\expandafter\def\csname ev@B-R/4B/g3/pooled/retention-g1/cand-pct\endcsname{39.11}
\expandafter\def\csname ev@B-R/4B/g3/pooled/retention-g1/cand-scheduled\endcsname{1120}
\expandafter\def\csname ev@B-R/4B/g3/pooled/retention-g1/cand-uncounted\endcsname{0}
\expandafter\def\csname ev@B-R/4B/g3/pooled/retention-g1/diff-pp\endcsname{+8.75}
\expandafter\def\csname ev@B-R/4B/g3/pooled/retention-g1/losses\endcsname{74}
\expandafter\def\csname ev@B-R/4B/g3/pooled/retention-g1/outcome\endcsname{inconclusive}
\expandafter\def\csname ev@B-R/4B/g3/pooled/retention-g1/p-recorded\endcsname{\ensuremath{1.8\times 10^{-10}}}
\expandafter\def\csname ev@B-R/4B/g3/pooled/retention-g1/protocol\endcsname{B-R}
\expandafter\def\csname ev@B-R/4B/g3/pooled/retention-g1/seeds\endcsname{5}
\expandafter\def\csname ev@B-R/4B/g3/pooled/retention-g1/seeds-negative\endcsname{0}
\expandafter\def\csname ev@B-R/4B/g3/pooled/retention-g1/seeds-positive\endcsname{5}
\expandafter\def\csname ev@B-R/4B/g3/pooled/retention-g1/tasks\endcsname{56}
\expandafter\def\csname ev@B-R/4B/g3/pooled/retention-g1/uncounted\endcsname{0}
\expandafter\def\csname ev@B-R/4B/g3/pooled/retention-g1/uncounted-pct\endcsname{0.00}
\expandafter\def\csname ev@B-R/4B/g3/pooled/retention-g1/unit\endcsname{attempt slot, summed over training seeds}
\expandafter\def\csname ev@B-R/4B/g3/pooled/retention-g1/wins\endcsname{172}
\expandafter\def\csname ev@B-R/4B/g3/pooled/retention-g2/base-counted\endcsname{188}
\expandafter\def\csname ev@B-R/4B/g3/pooled/retention-g2/base-passes\endcsname{31}
\expandafter\def\csname ev@B-R/4B/g3/pooled/retention-g2/base-pct\endcsname{16.49}
\expandafter\def\csname ev@B-R/4B/g3/pooled/retention-g2/base-scheduled\endcsname{188}
\expandafter\def\csname ev@B-R/4B/g3/pooled/retention-g2/base-uncounted\endcsname{0}
\expandafter\def\csname ev@B-R/4B/g3/pooled/retention-g2/cand-counted\endcsname{940}
\expandafter\def\csname ev@B-R/4B/g3/pooled/retention-g2/cand-passes\endcsname{187}
\expandafter\def\csname ev@B-R/4B/g3/pooled/retention-g2/cand-pct\endcsname{19.89}
\expandafter\def\csname ev@B-R/4B/g3/pooled/retention-g2/cand-scheduled\endcsname{940}
\expandafter\def\csname ev@B-R/4B/g3/pooled/retention-g2/cand-uncounted\endcsname{0}
\expandafter\def\csname ev@B-R/4B/g3/pooled/retention-g2/diff-pp\endcsname{+3.40}
\expandafter\def\csname ev@B-R/4B/g3/pooled/retention-g2/losses\endcsname{56}
\expandafter\def\csname ev@B-R/4B/g3/pooled/retention-g2/outcome\endcsname{inconclusive}
\expandafter\def\csname ev@B-R/4B/g3/pooled/retention-g2/p-recorded\endcsname{0.0048}
\expandafter\def\csname ev@B-R/4B/g3/pooled/retention-g2/protocol\endcsname{B-R}
\expandafter\def\csname ev@B-R/4B/g3/pooled/retention-g2/seeds\endcsname{5}
\expandafter\def\csname ev@B-R/4B/g3/pooled/retention-g2/seeds-negative\endcsname{1}
\expandafter\def\csname ev@B-R/4B/g3/pooled/retention-g2/seeds-positive\endcsname{4}
\expandafter\def\csname ev@B-R/4B/g3/pooled/retention-g2/tasks\endcsname{47}
\expandafter\def\csname ev@B-R/4B/g3/pooled/retention-g2/uncounted\endcsname{0}
\expandafter\def\csname ev@B-R/4B/g3/pooled/retention-g2/uncounted-pct\endcsname{0.00}
\expandafter\def\csname ev@B-R/4B/g3/pooled/retention-g2/unit\endcsname{attempt slot, summed over training seeds}
\expandafter\def\csname ev@B-R/4B/g3/pooled/retention-g2/wins\endcsname{88}
\expandafter\def\csname ev@B-R/4B/g3/pooled/target/base-counted\endcsname{196}
\expandafter\def\csname ev@B-R/4B/g3/pooled/target/base-passes\endcsname{27}
\expandafter\def\csname ev@B-R/4B/g3/pooled/target/base-pct\endcsname{13.78}
\expandafter\def\csname ev@B-R/4B/g3/pooled/target/base-scheduled\endcsname{196}
\expandafter\def\csname ev@B-R/4B/g3/pooled/target/base-uncounted\endcsname{0}
\expandafter\def\csname ev@B-R/4B/g3/pooled/target/cand-counted\endcsname{980}
\expandafter\def\csname ev@B-R/4B/g3/pooled/target/cand-passes\endcsname{186}
\expandafter\def\csname ev@B-R/4B/g3/pooled/target/cand-pct\endcsname{18.98}
\expandafter\def\csname ev@B-R/4B/g3/pooled/target/cand-scheduled\endcsname{980}
\expandafter\def\csname ev@B-R/4B/g3/pooled/target/cand-uncounted\endcsname{0}
\expandafter\def\csname ev@B-R/4B/g3/pooled/target/diff-pp\endcsname{+5.20}
\expandafter\def\csname ev@B-R/4B/g3/pooled/target/losses\endcsname{51}
\expandafter\def\csname ev@B-R/4B/g3/pooled/target/outcome\endcsname{inconclusive}
\expandafter\def\csname ev@B-R/4B/g3/pooled/target/p-recorded\endcsname{\ensuremath{2.3\times 10^{-5}}}
\expandafter\def\csname ev@B-R/4B/g3/pooled/target/protocol\endcsname{B-R}
\expandafter\def\csname ev@B-R/4B/g3/pooled/target/seeds\endcsname{5}
\expandafter\def\csname ev@B-R/4B/g3/pooled/target/seeds-negative\endcsname{0}
\expandafter\def\csname ev@B-R/4B/g3/pooled/target/seeds-positive\endcsname{5}
\expandafter\def\csname ev@B-R/4B/g3/pooled/target/tasks\endcsname{49}
\expandafter\def\csname ev@B-R/4B/g3/pooled/target/uncounted\endcsname{0}
\expandafter\def\csname ev@B-R/4B/g3/pooled/target/uncounted-pct\endcsname{0.00}
\expandafter\def\csname ev@B-R/4B/g3/pooled/target/unit\endcsname{attempt slot, summed over training seeds}
\expandafter\def\csname ev@B-R/4B/g3/pooled/target/wins\endcsname{102}
\expandafter\def\csname ev@B-R/4B/g3/s1/exam/attempt-ci-hi-pp\endcsname{+2.82}
\expandafter\def\csname ev@B-R/4B/g3/s1/exam/attempt-ci-lo-pp\endcsname{-2.30}
\expandafter\def\csname ev@B-R/4B/g3/s1/exam/base-counted\endcsname{400}
\expandafter\def\csname ev@B-R/4B/g3/s1/exam/base-passes\endcsname{12}
\expandafter\def\csname ev@B-R/4B/g3/s1/exam/base-pct\endcsname{3.00}
\expandafter\def\csname ev@B-R/4B/g3/s1/exam/base-scheduled\endcsname{400}
\expandafter\def\csname ev@B-R/4B/g3/s1/exam/base-uncounted\endcsname{0}
\expandafter\def\csname ev@B-R/4B/g3/s1/exam/cand-counted\endcsname{400}
\expandafter\def\csname ev@B-R/4B/g3/s1/exam/cand-passes\endcsname{13}
\expandafter\def\csname ev@B-R/4B/g3/s1/exam/cand-pct\endcsname{3.25}
\expandafter\def\csname ev@B-R/4B/g3/s1/exam/cand-scheduled\endcsname{400}
\expandafter\def\csname ev@B-R/4B/g3/s1/exam/cand-uncounted\endcsname{0}
\expandafter\def\csname ev@B-R/4B/g3/s1/exam/diff-pp\endcsname{+0.25}
\expandafter\def\csname ev@B-R/4B/g3/s1/exam/losses\endcsname{10}
\expandafter\def\csname ev@B-R/4B/g3/s1/exam/outcome\endcsname{inconclusive}
\expandafter\def\csname ev@B-R/4B/g3/s1/exam/p-recorded\endcsname{0.5}
\expandafter\def\csname ev@B-R/4B/g3/s1/exam/protocol\endcsname{B-R}
\expandafter\def\csname ev@B-R/4B/g3/s1/exam/task-losses-high\endcsname{10}
\expandafter\def\csname ev@B-R/4B/g3/s1/exam/task-losses-low\endcsname{0}
\expandafter\def\csname ev@B-R/4B/g3/s1/exam/task-p-high\endcsname{0.989}
\expandafter\def\csname ev@B-R/4B/g3/s1/exam/task-p-low\endcsname{0.029}
\expandafter\def\csname ev@B-R/4B/g3/s1/exam/task-wins-high\endcsname{11}
\expandafter\def\csname ev@B-R/4B/g3/s1/exam/task-wins-low\endcsname{1}
\expandafter\def\csname ev@B-R/4B/g3/s1/exam/tasks\endcsname{100}
\expandafter\def\csname ev@B-R/4B/g3/s1/exam/uncounted\endcsname{0}
\expandafter\def\csname ev@B-R/4B/g3/s1/exam/uncounted-pct\endcsname{0.00}
\expandafter\def\csname ev@B-R/4B/g3/s1/exam/unit\endcsname{attempt slot}
\expandafter\def\csname ev@B-R/4B/g3/s1/exam/wins\endcsname{11}
\expandafter\def\csname ev@B-R/4B/g3/s1/retention-g1/attempt-ci-hi-pp\endcsname{+15.30}
\expandafter\def\csname ev@B-R/4B/g3/s1/retention-g1/attempt-ci-lo-pp\endcsname{-2.05}
\expandafter\def\csname ev@B-R/4B/g3/s1/retention-g1/base-counted\endcsname{224}
\expandafter\def\csname ev@B-R/4B/g3/s1/retention-g1/base-passes\endcsname{68}
\expandafter\def\csname ev@B-R/4B/g3/s1/retention-g1/base-pct\endcsname{30.36}
\expandafter\def\csname ev@B-R/4B/g3/s1/retention-g1/base-scheduled\endcsname{224}
\expandafter\def\csname ev@B-R/4B/g3/s1/retention-g1/base-uncounted\endcsname{0}
\expandafter\def\csname ev@B-R/4B/g3/s1/retention-g1/cand-counted\endcsname{224}
\expandafter\def\csname ev@B-R/4B/g3/s1/retention-g1/cand-passes\endcsname{83}
\expandafter\def\csname ev@B-R/4B/g3/s1/retention-g1/cand-pct\endcsname{37.05}
\expandafter\def\csname ev@B-R/4B/g3/s1/retention-g1/cand-scheduled\endcsname{224}
\expandafter\def\csname ev@B-R/4B/g3/s1/retention-g1/cand-uncounted\endcsname{0}
\expandafter\def\csname ev@B-R/4B/g3/s1/retention-g1/diff-pp\endcsname{+6.70}
\expandafter\def\csname ev@B-R/4B/g3/s1/retention-g1/losses\endcsname{13}
\expandafter\def\csname ev@B-R/4B/g3/s1/retention-g1/outcome\endcsname{inconclusive}
\expandafter\def\csname ev@B-R/4B/g3/s1/retention-g1/p-recorded\endcsname{0.014}
\expandafter\def\csname ev@B-R/4B/g3/s1/retention-g1/protocol\endcsname{B-R}
\expandafter\def\csname ev@B-R/4B/g3/s1/retention-g1/task-losses-high\endcsname{13}
\expandafter\def\csname ev@B-R/4B/g3/s1/retention-g1/task-losses-low\endcsname{0}
\expandafter\def\csname ev@B-R/4B/g3/s1/retention-g1/task-p-high\endcsname{0.942}
\expandafter\def\csname ev@B-R/4B/g3/s1/retention-g1/task-p-low\endcsname{\ensuremath{4.2\times 10^{-6}}}
\expandafter\def\csname ev@B-R/4B/g3/s1/retention-g1/task-wins-high\endcsname{28}
\expandafter\def\csname ev@B-R/4B/g3/s1/retention-g1/task-wins-low\endcsname{4}
\expandafter\def\csname ev@B-R/4B/g3/s1/retention-g1/tasks\endcsname{56}
\expandafter\def\csname ev@B-R/4B/g3/s1/retention-g1/uncounted\endcsname{0}
\expandafter\def\csname ev@B-R/4B/g3/s1/retention-g1/uncounted-pct\endcsname{0.00}
\expandafter\def\csname ev@B-R/4B/g3/s1/retention-g1/unit\endcsname{attempt slot}
\expandafter\def\csname ev@B-R/4B/g3/s1/retention-g1/wins\endcsname{28}
\expandafter\def\csname ev@B-R/4B/g3/s1/retention-g2/attempt-ci-hi-pp\endcsname{+6.97}
\expandafter\def\csname ev@B-R/4B/g3/s1/retention-g2/attempt-ci-lo-pp\endcsname{-8.03}
\expandafter\def\csname ev@B-R/4B/g3/s1/retention-g2/base-counted\endcsname{188}
\expandafter\def\csname ev@B-R/4B/g3/s1/retention-g2/base-passes\endcsname{31}
\expandafter\def\csname ev@B-R/4B/g3/s1/retention-g2/base-pct\endcsname{16.49}
\expandafter\def\csname ev@B-R/4B/g3/s1/retention-g2/base-scheduled\endcsname{188}
\expandafter\def\csname ev@B-R/4B/g3/s1/retention-g2/base-uncounted\endcsname{0}
\expandafter\def\csname ev@B-R/4B/g3/s1/retention-g2/cand-counted\endcsname{188}
\expandafter\def\csname ev@B-R/4B/g3/s1/retention-g2/cand-passes\endcsname{30}
\expandafter\def\csname ev@B-R/4B/g3/s1/retention-g2/cand-pct\endcsname{15.96}
\expandafter\def\csname ev@B-R/4B/g3/s1/retention-g2/cand-scheduled\endcsname{188}
\expandafter\def\csname ev@B-R/4B/g3/s1/retention-g2/cand-uncounted\endcsname{0}
\expandafter\def\csname ev@B-R/4B/g3/s1/retention-g2/diff-pp\endcsname{-0.53}
\expandafter\def\csname ev@B-R/4B/g3/s1/retention-g2/losses\endcsname{12}
\expandafter\def\csname ev@B-R/4B/g3/s1/retention-g2/outcome\endcsname{inconclusive}
\expandafter\def\csname ev@B-R/4B/g3/s1/retention-g2/p-recorded\endcsname{0.661}
\expandafter\def\csname ev@B-R/4B/g3/s1/retention-g2/protocol\endcsname{B-R}
\expandafter\def\csname ev@B-R/4B/g3/s1/retention-g2/task-losses-high\endcsname{12}
\expandafter\def\csname ev@B-R/4B/g3/s1/retention-g2/task-losses-low\endcsname{1}
\expandafter\def\csname ev@B-R/4B/g3/s1/retention-g2/task-p-high\endcsname{1}
\expandafter\def\csname ev@B-R/4B/g3/s1/retention-g2/task-p-low\endcsname{0.029}
\expandafter\def\csname ev@B-R/4B/g3/s1/retention-g2/task-wins-high\endcsname{11}
\expandafter\def\csname ev@B-R/4B/g3/s1/retention-g2/task-wins-low\endcsname{0}
\expandafter\def\csname ev@B-R/4B/g3/s1/retention-g2/tasks\endcsname{47}
\expandafter\def\csname ev@B-R/4B/g3/s1/retention-g2/uncounted\endcsname{0}
\expandafter\def\csname ev@B-R/4B/g3/s1/retention-g2/uncounted-pct\endcsname{0.00}
\expandafter\def\csname ev@B-R/4B/g3/s1/retention-g2/unit\endcsname{attempt slot}
\expandafter\def\csname ev@B-R/4B/g3/s1/retention-g2/wins\endcsname{11}
\expandafter\def\csname ev@B-R/4B/g3/s1/target/attempt-ci-hi-pp\endcsname{+14.62}
\expandafter\def\csname ev@B-R/4B/g3/s1/target/attempt-ci-lo-pp\endcsname{-0.38}
\expandafter\def\csname ev@B-R/4B/g3/s1/target/base-counted\endcsname{196}
\expandafter\def\csname ev@B-R/4B/g3/s1/target/base-passes\endcsname{27}
\expandafter\def\csname ev@B-R/4B/g3/s1/target/base-pct\endcsname{13.78}
\expandafter\def\csname ev@B-R/4B/g3/s1/target/base-scheduled\endcsname{196}
\expandafter\def\csname ev@B-R/4B/g3/s1/target/base-uncounted\endcsname{0}
\expandafter\def\csname ev@B-R/4B/g3/s1/target/cand-counted\endcsname{196}
\expandafter\def\csname ev@B-R/4B/g3/s1/target/cand-passes\endcsname{41}
\expandafter\def\csname ev@B-R/4B/g3/s1/target/cand-pct\endcsname{20.92}
\expandafter\def\csname ev@B-R/4B/g3/s1/target/cand-scheduled\endcsname{196}
\expandafter\def\csname ev@B-R/4B/g3/s1/target/cand-uncounted\endcsname{0}
\expandafter\def\csname ev@B-R/4B/g3/s1/target/diff-pp\endcsname{+7.14}
\expandafter\def\csname ev@B-R/4B/g3/s1/target/losses\endcsname{10}
\expandafter\def\csname ev@B-R/4B/g3/s1/target/outcome\endcsname{inconclusive}
\expandafter\def\csname ev@B-R/4B/g3/s1/target/p-recorded\endcsname{0.012}
\expandafter\def\csname ev@B-R/4B/g3/s1/target/protocol\endcsname{B-R}
\expandafter\def\csname ev@B-R/4B/g3/s1/target/task-losses-high\endcsname{10}
\expandafter\def\csname ev@B-R/4B/g3/s1/target/task-losses-low\endcsname{0}
\expandafter\def\csname ev@B-R/4B/g3/s1/target/task-p-high\endcsname{0.895}
\expandafter\def\csname ev@B-R/4B/g3/s1/target/task-p-low\endcsname{\ensuremath{1.8\times 10^{-5}}}
\expandafter\def\csname ev@B-R/4B/g3/s1/target/task-wins-high\endcsname{24}
\expandafter\def\csname ev@B-R/4B/g3/s1/target/task-wins-low\endcsname{4}
\expandafter\def\csname ev@B-R/4B/g3/s1/target/tasks\endcsname{49}
\expandafter\def\csname ev@B-R/4B/g3/s1/target/uncounted\endcsname{0}
\expandafter\def\csname ev@B-R/4B/g3/s1/target/uncounted-pct\endcsname{0.00}
\expandafter\def\csname ev@B-R/4B/g3/s1/target/unit\endcsname{attempt slot}
\expandafter\def\csname ev@B-R/4B/g3/s1/target/wins\endcsname{24}
\expandafter\def\csname ev@B-R/4B/g3/s2/exam/attempt-ci-hi-pp\endcsname{+3.72}
\expandafter\def\csname ev@B-R/4B/g3/s2/exam/attempt-ci-lo-pp\endcsname{-1.65}
\expandafter\def\csname ev@B-R/4B/g3/s2/exam/base-counted\endcsname{400}
\expandafter\def\csname ev@B-R/4B/g3/s2/exam/base-passes\endcsname{12}
\expandafter\def\csname ev@B-R/4B/g3/s2/exam/base-pct\endcsname{3.00}
\expandafter\def\csname ev@B-R/4B/g3/s2/exam/base-scheduled\endcsname{400}
\expandafter\def\csname ev@B-R/4B/g3/s2/exam/base-uncounted\endcsname{0}
\expandafter\def\csname ev@B-R/4B/g3/s2/exam/cand-counted\endcsname{400}
\expandafter\def\csname ev@B-R/4B/g3/s2/exam/cand-passes\endcsname{16}
\expandafter\def\csname ev@B-R/4B/g3/s2/exam/cand-pct\endcsname{4.00}
\expandafter\def\csname ev@B-R/4B/g3/s2/exam/cand-scheduled\endcsname{400}
\expandafter\def\csname ev@B-R/4B/g3/s2/exam/cand-uncounted\endcsname{0}
\expandafter\def\csname ev@B-R/4B/g3/s2/exam/diff-pp\endcsname{+1.00}
\expandafter\def\csname ev@B-R/4B/g3/s2/exam/losses\endcsname{10}
\expandafter\def\csname ev@B-R/4B/g3/s2/exam/outcome\endcsname{inconclusive}
\expandafter\def\csname ev@B-R/4B/g3/s2/exam/p-recorded\endcsname{0.271}
\expandafter\def\csname ev@B-R/4B/g3/s2/exam/protocol\endcsname{B-R}
\expandafter\def\csname ev@B-R/4B/g3/s2/exam/task-losses-high\endcsname{10}
\expandafter\def\csname ev@B-R/4B/g3/s2/exam/task-losses-low\endcsname{0}
\expandafter\def\csname ev@B-R/4B/g3/s2/exam/task-p-high\endcsname{0.971}
\expandafter\def\csname ev@B-R/4B/g3/s2/exam/task-p-low\endcsname{0.0064}
\expandafter\def\csname ev@B-R/4B/g3/s2/exam/task-wins-high\endcsname{14}
\expandafter\def\csname ev@B-R/4B/g3/s2/exam/task-wins-low\endcsname{1}
\expandafter\def\csname ev@B-R/4B/g3/s2/exam/tasks\endcsname{100}
\expandafter\def\csname ev@B-R/4B/g3/s2/exam/uncounted\endcsname{0}
\expandafter\def\csname ev@B-R/4B/g3/s2/exam/uncounted-pct\endcsname{0.00}
\expandafter\def\csname ev@B-R/4B/g3/s2/exam/unit\endcsname{attempt slot}
\expandafter\def\csname ev@B-R/4B/g3/s2/exam/wins\endcsname{14}
\expandafter\def\csname ev@B-R/4B/g3/s2/retention-g1/attempt-ci-hi-pp\endcsname{+11.67}
\expandafter\def\csname ev@B-R/4B/g3/s2/retention-g1/attempt-ci-lo-pp\endcsname{-5.48}
\expandafter\def\csname ev@B-R/4B/g3/s2/retention-g1/base-counted\endcsname{224}
\expandafter\def\csname ev@B-R/4B/g3/s2/retention-g1/base-passes\endcsname{68}
\expandafter\def\csname ev@B-R/4B/g3/s2/retention-g1/base-pct\endcsname{30.36}
\expandafter\def\csname ev@B-R/4B/g3/s2/retention-g1/base-scheduled\endcsname{224}
\expandafter\def\csname ev@B-R/4B/g3/s2/retention-g1/base-uncounted\endcsname{0}
\expandafter\def\csname ev@B-R/4B/g3/s2/retention-g1/cand-counted\endcsname{224}
\expandafter\def\csname ev@B-R/4B/g3/s2/retention-g1/cand-passes\endcsname{75}
\expandafter\def\csname ev@B-R/4B/g3/s2/retention-g1/cand-pct\endcsname{33.48}
\expandafter\def\csname ev@B-R/4B/g3/s2/retention-g1/cand-scheduled\endcsname{224}
\expandafter\def\csname ev@B-R/4B/g3/s2/retention-g1/cand-uncounted\endcsname{0}
\expandafter\def\csname ev@B-R/4B/g3/s2/retention-g1/diff-pp\endcsname{+3.12}
\expandafter\def\csname ev@B-R/4B/g3/s2/retention-g1/losses\endcsname{22}
\expandafter\def\csname ev@B-R/4B/g3/s2/retention-g1/outcome\endcsname{inconclusive}
\expandafter\def\csname ev@B-R/4B/g3/s2/retention-g1/p-recorded\endcsname{0.201}
\expandafter\def\csname ev@B-R/4B/g3/s2/retention-g1/protocol\endcsname{B-R}
\expandafter\def\csname ev@B-R/4B/g3/s2/retention-g1/task-losses-high\endcsname{22}
\expandafter\def\csname ev@B-R/4B/g3/s2/retention-g1/task-losses-low\endcsname{0}
\expandafter\def\csname ev@B-R/4B/g3/s2/retention-g1/task-p-high\endcsname{0.998}
\expandafter\def\csname ev@B-R/4B/g3/s2/retention-g1/task-p-low\endcsname{\ensuremath{5.7\times 10^{-5}}}
\expandafter\def\csname ev@B-R/4B/g3/s2/retention-g1/task-wins-high\endcsname{29}
\expandafter\def\csname ev@B-R/4B/g3/s2/retention-g1/task-wins-low\endcsname{2}
\expandafter\def\csname ev@B-R/4B/g3/s2/retention-g1/tasks\endcsname{56}
\expandafter\def\csname ev@B-R/4B/g3/s2/retention-g1/uncounted\endcsname{0}
\expandafter\def\csname ev@B-R/4B/g3/s2/retention-g1/uncounted-pct\endcsname{0.00}
\expandafter\def\csname ev@B-R/4B/g3/s2/retention-g1/unit\endcsname{attempt slot}
\expandafter\def\csname ev@B-R/4B/g3/s2/retention-g1/wins\endcsname{29}
\expandafter\def\csname ev@B-R/4B/g3/s2/retention-g2/attempt-ci-hi-pp\endcsname{+13.24}
\expandafter\def\csname ev@B-R/4B/g3/s2/retention-g2/attempt-ci-lo-pp\endcsname{-2.66}
\expandafter\def\csname ev@B-R/4B/g3/s2/retention-g2/base-counted\endcsname{188}
\expandafter\def\csname ev@B-R/4B/g3/s2/retention-g2/base-passes\endcsname{31}
\expandafter\def\csname ev@B-R/4B/g3/s2/retention-g2/base-pct\endcsname{16.49}
\expandafter\def\csname ev@B-R/4B/g3/s2/retention-g2/base-scheduled\endcsname{188}
\expandafter\def\csname ev@B-R/4B/g3/s2/retention-g2/base-uncounted\endcsname{0}
\expandafter\def\csname ev@B-R/4B/g3/s2/retention-g2/cand-counted\endcsname{188}
\expandafter\def\csname ev@B-R/4B/g3/s2/retention-g2/cand-passes\endcsname{41}
\expandafter\def\csname ev@B-R/4B/g3/s2/retention-g2/cand-pct\endcsname{21.81}
\expandafter\def\csname ev@B-R/4B/g3/s2/retention-g2/cand-scheduled\endcsname{188}
\expandafter\def\csname ev@B-R/4B/g3/s2/retention-g2/cand-uncounted\endcsname{0}
\expandafter\def\csname ev@B-R/4B/g3/s2/retention-g2/diff-pp\endcsname{+5.32}
\expandafter\def\csname ev@B-R/4B/g3/s2/retention-g2/losses\endcsname{12}
\expandafter\def\csname ev@B-R/4B/g3/s2/retention-g2/outcome\endcsname{inconclusive}
\expandafter\def\csname ev@B-R/4B/g3/s2/retention-g2/p-recorded\endcsname{0.061}
\expandafter\def\csname ev@B-R/4B/g3/s2/retention-g2/protocol\endcsname{B-R}
\expandafter\def\csname ev@B-R/4B/g3/s2/retention-g2/task-losses-high\endcsname{12}
\expandafter\def\csname ev@B-R/4B/g3/s2/retention-g2/task-losses-low\endcsname{0}
\expandafter\def\csname ev@B-R/4B/g3/s2/retention-g2/task-p-high\endcsname{0.952}
\expandafter\def\csname ev@B-R/4B/g3/s2/retention-g2/task-p-low\endcsname{\ensuremath{7.8\times 10^{-5}}}
\expandafter\def\csname ev@B-R/4B/g3/s2/retention-g2/task-wins-high\endcsname{22}
\expandafter\def\csname ev@B-R/4B/g3/s2/retention-g2/task-wins-low\endcsname{3}
\expandafter\def\csname ev@B-R/4B/g3/s2/retention-g2/tasks\endcsname{47}
\expandafter\def\csname ev@B-R/4B/g3/s2/retention-g2/uncounted\endcsname{0}
\expandafter\def\csname ev@B-R/4B/g3/s2/retention-g2/uncounted-pct\endcsname{0.00}
\expandafter\def\csname ev@B-R/4B/g3/s2/retention-g2/unit\endcsname{attempt slot}
\expandafter\def\csname ev@B-R/4B/g3/s2/retention-g2/wins\endcsname{22}
\expandafter\def\csname ev@B-R/4B/g3/s2/target/attempt-ci-hi-pp\endcsname{+10.79}
\expandafter\def\csname ev@B-R/4B/g3/s2/target/attempt-ci-lo-pp\endcsname{-3.66}
\expandafter\def\csname ev@B-R/4B/g3/s2/target/base-counted\endcsname{196}
\expandafter\def\csname ev@B-R/4B/g3/s2/target/base-passes\endcsname{27}
\expandafter\def\csname ev@B-R/4B/g3/s2/target/base-pct\endcsname{13.78}
\expandafter\def\csname ev@B-R/4B/g3/s2/target/base-scheduled\endcsname{196}
\expandafter\def\csname ev@B-R/4B/g3/s2/target/base-uncounted\endcsname{0}
\expandafter\def\csname ev@B-R/4B/g3/s2/target/cand-counted\endcsname{196}
\expandafter\def\csname ev@B-R/4B/g3/s2/target/cand-passes\endcsname{34}
\expandafter\def\csname ev@B-R/4B/g3/s2/target/cand-pct\endcsname{17.35}
\expandafter\def\csname ev@B-R/4B/g3/s2/target/cand-scheduled\endcsname{196}
\expandafter\def\csname ev@B-R/4B/g3/s2/target/cand-uncounted\endcsname{0}
\expandafter\def\csname ev@B-R/4B/g3/s2/target/diff-pp\endcsname{+3.57}
\expandafter\def\csname ev@B-R/4B/g3/s2/target/losses\endcsname{9}
\expandafter\def\csname ev@B-R/4B/g3/s2/target/outcome\endcsname{inconclusive}
\expandafter\def\csname ev@B-R/4B/g3/s2/target/p-recorded\endcsname{0.115}
\expandafter\def\csname ev@B-R/4B/g3/s2/target/protocol\endcsname{B-R}
\expandafter\def\csname ev@B-R/4B/g3/s2/target/task-losses-high\endcsname{9}
\expandafter\def\csname ev@B-R/4B/g3/s2/target/task-losses-low\endcsname{0}
\expandafter\def\csname ev@B-R/4B/g3/s2/target/task-p-high\endcsname{0.954}
\expandafter\def\csname ev@B-R/4B/g3/s2/target/task-p-low\endcsname{0.0012}
\expandafter\def\csname ev@B-R/4B/g3/s2/target/task-wins-high\endcsname{16}
\expandafter\def\csname ev@B-R/4B/g3/s2/target/task-wins-low\endcsname{2}
\expandafter\def\csname ev@B-R/4B/g3/s2/target/tasks\endcsname{49}
\expandafter\def\csname ev@B-R/4B/g3/s2/target/uncounted\endcsname{0}
\expandafter\def\csname ev@B-R/4B/g3/s2/target/uncounted-pct\endcsname{0.00}
\expandafter\def\csname ev@B-R/4B/g3/s2/target/unit\endcsname{attempt slot}
\expandafter\def\csname ev@B-R/4B/g3/s2/target/wins\endcsname{16}
\expandafter\def\csname ev@B-R/4B/g3/s3/exam/attempt-ci-hi-pp\endcsname{+3.12}
\expandafter\def\csname ev@B-R/4B/g3/s3/exam/attempt-ci-lo-pp\endcsname{-2.08}
\expandafter\def\csname ev@B-R/4B/g3/s3/exam/base-counted\endcsname{400}
\expandafter\def\csname ev@B-R/4B/g3/s3/exam/base-passes\endcsname{12}
\expandafter\def\csname ev@B-R/4B/g3/s3/exam/base-pct\endcsname{3.00}
\expandafter\def\csname ev@B-R/4B/g3/s3/exam/base-scheduled\endcsname{400}
\expandafter\def\csname ev@B-R/4B/g3/s3/exam/base-uncounted\endcsname{0}
\expandafter\def\csname ev@B-R/4B/g3/s3/exam/cand-counted\endcsname{400}
\expandafter\def\csname ev@B-R/4B/g3/s3/exam/cand-passes\endcsname{14}
\expandafter\def\csname ev@B-R/4B/g3/s3/exam/cand-pct\endcsname{3.50}
\expandafter\def\csname ev@B-R/4B/g3/s3/exam/cand-scheduled\endcsname{400}
\expandafter\def\csname ev@B-R/4B/g3/s3/exam/cand-uncounted\endcsname{0}
\expandafter\def\csname ev@B-R/4B/g3/s3/exam/diff-pp\endcsname{+0.50}
\expandafter\def\csname ev@B-R/4B/g3/s3/exam/losses\endcsname{8}
\expandafter\def\csname ev@B-R/4B/g3/s3/exam/outcome\endcsname{inconclusive}
\expandafter\def\csname ev@B-R/4B/g3/s3/exam/p-recorded\endcsname{0.407}
\expandafter\def\csname ev@B-R/4B/g3/s3/exam/protocol\endcsname{B-R}
\expandafter\def\csname ev@B-R/4B/g3/s3/exam/task-losses-high\endcsname{8}
\expandafter\def\csname ev@B-R/4B/g3/s3/exam/task-losses-low\endcsname{0}
\expandafter\def\csname ev@B-R/4B/g3/s3/exam/task-p-high\endcsname{0.967}
\expandafter\def\csname ev@B-R/4B/g3/s3/exam/task-p-low\endcsname{0.019}
\expandafter\def\csname ev@B-R/4B/g3/s3/exam/task-wins-high\endcsname{10}
\expandafter\def\csname ev@B-R/4B/g3/s3/exam/task-wins-low\endcsname{1}
\expandafter\def\csname ev@B-R/4B/g3/s3/exam/tasks\endcsname{100}
\expandafter\def\csname ev@B-R/4B/g3/s3/exam/uncounted\endcsname{0}
\expandafter\def\csname ev@B-R/4B/g3/s3/exam/uncounted-pct\endcsname{0.00}
\expandafter\def\csname ev@B-R/4B/g3/s3/exam/unit\endcsname{attempt slot}
\expandafter\def\csname ev@B-R/4B/g3/s3/exam/wins\endcsname{10}
\expandafter\def\csname ev@B-R/4B/g3/s3/retention-g1/attempt-ci-hi-pp\endcsname{+21.15}
\expandafter\def\csname ev@B-R/4B/g3/s3/retention-g1/attempt-ci-lo-pp\endcsname{+3.58}
\expandafter\def\csname ev@B-R/4B/g3/s3/retention-g1/base-counted\endcsname{224}
\expandafter\def\csname ev@B-R/4B/g3/s3/retention-g1/base-passes\endcsname{68}
\expandafter\def\csname ev@B-R/4B/g3/s3/retention-g1/base-pct\endcsname{30.36}
\expandafter\def\csname ev@B-R/4B/g3/s3/retention-g1/base-scheduled\endcsname{224}
\expandafter\def\csname ev@B-R/4B/g3/s3/retention-g1/base-uncounted\endcsname{0}
\expandafter\def\csname ev@B-R/4B/g3/s3/retention-g1/cand-counted\endcsname{224}
\expandafter\def\csname ev@B-R/4B/g3/s3/retention-g1/cand-passes\endcsname{96}
\expandafter\def\csname ev@B-R/4B/g3/s3/retention-g1/cand-pct\endcsname{42.86}
\expandafter\def\csname ev@B-R/4B/g3/s3/retention-g1/cand-scheduled\endcsname{224}
\expandafter\def\csname ev@B-R/4B/g3/s3/retention-g1/cand-uncounted\endcsname{0}
\expandafter\def\csname ev@B-R/4B/g3/s3/retention-g1/diff-pp\endcsname{+12.50}
\expandafter\def\csname ev@B-R/4B/g3/s3/retention-g1/losses\endcsname{12}
\expandafter\def\csname ev@B-R/4B/g3/s3/retention-g1/outcome\endcsname{inconclusive}
\expandafter\def\csname ev@B-R/4B/g3/s3/retention-g1/p-recorded\endcsname{\ensuremath{6.4\times 10^{-5}}}
\expandafter\def\csname ev@B-R/4B/g3/s3/retention-g1/protocol\endcsname{B-R}
\expandafter\def\csname ev@B-R/4B/g3/s3/retention-g1/task-losses-high\endcsname{12}
\expandafter\def\csname ev@B-R/4B/g3/s3/retention-g1/task-losses-low\endcsname{0}
\expandafter\def\csname ev@B-R/4B/g3/s3/retention-g1/task-p-high\endcsname{0.738}
\expandafter\def\csname ev@B-R/4B/g3/s3/retention-g1/task-p-low\endcsname{\ensuremath{1.5\times 10^{-9}}}
\expandafter\def\csname ev@B-R/4B/g3/s3/retention-g1/task-wins-high\endcsname{40}
\expandafter\def\csname ev@B-R/4B/g3/s3/retention-g1/task-wins-low\endcsname{7}
\expandafter\def\csname ev@B-R/4B/g3/s3/retention-g1/tasks\endcsname{56}
\expandafter\def\csname ev@B-R/4B/g3/s3/retention-g1/uncounted\endcsname{0}
\expandafter\def\csname ev@B-R/4B/g3/s3/retention-g1/uncounted-pct\endcsname{0.00}
\expandafter\def\csname ev@B-R/4B/g3/s3/retention-g1/unit\endcsname{attempt slot}
\expandafter\def\csname ev@B-R/4B/g3/s3/retention-g1/wins\endcsname{40}
\expandafter\def\csname ev@B-R/4B/g3/s3/retention-g2/attempt-ci-hi-pp\endcsname{+11.55}
\expandafter\def\csname ev@B-R/4B/g3/s3/retention-g2/attempt-ci-lo-pp\endcsname{-4.14}
\expandafter\def\csname ev@B-R/4B/g3/s3/retention-g2/base-counted\endcsname{188}
\expandafter\def\csname ev@B-R/4B/g3/s3/retention-g2/base-passes\endcsname{31}
\expandafter\def\csname ev@B-R/4B/g3/s3/retention-g2/base-pct\endcsname{16.49}
\expandafter\def\csname ev@B-R/4B/g3/s3/retention-g2/base-scheduled\endcsname{188}
\expandafter\def\csname ev@B-R/4B/g3/s3/retention-g2/base-uncounted\endcsname{0}
\expandafter\def\csname ev@B-R/4B/g3/s3/retention-g2/cand-counted\endcsname{188}
\expandafter\def\csname ev@B-R/4B/g3/s3/retention-g2/cand-passes\endcsname{38}
\expandafter\def\csname ev@B-R/4B/g3/s3/retention-g2/cand-pct\endcsname{20.21}
\expandafter\def\csname ev@B-R/4B/g3/s3/retention-g2/cand-scheduled\endcsname{188}
\expandafter\def\csname ev@B-R/4B/g3/s3/retention-g2/cand-uncounted\endcsname{0}
\expandafter\def\csname ev@B-R/4B/g3/s3/retention-g2/diff-pp\endcsname{+3.72}
\expandafter\def\csname ev@B-R/4B/g3/s3/retention-g2/losses\endcsname{9}
\expandafter\def\csname ev@B-R/4B/g3/s3/retention-g2/outcome\endcsname{inconclusive}
\expandafter\def\csname ev@B-R/4B/g3/s3/retention-g2/p-recorded\endcsname{0.115}
\expandafter\def\csname ev@B-R/4B/g3/s3/retention-g2/protocol\endcsname{B-R}
\expandafter\def\csname ev@B-R/4B/g3/s3/retention-g2/task-losses-high\endcsname{9}
\expandafter\def\csname ev@B-R/4B/g3/s3/retention-g2/task-losses-low\endcsname{0}
\expandafter\def\csname ev@B-R/4B/g3/s3/retention-g2/task-p-high\endcsname{0.954}
\expandafter\def\csname ev@B-R/4B/g3/s3/retention-g2/task-p-low\endcsname{0.0012}
\expandafter\def\csname ev@B-R/4B/g3/s3/retention-g2/task-wins-high\endcsname{16}
\expandafter\def\csname ev@B-R/4B/g3/s3/retention-g2/task-wins-low\endcsname{2}
\expandafter\def\csname ev@B-R/4B/g3/s3/retention-g2/tasks\endcsname{47}
\expandafter\def\csname ev@B-R/4B/g3/s3/retention-g2/uncounted\endcsname{0}
\expandafter\def\csname ev@B-R/4B/g3/s3/retention-g2/uncounted-pct\endcsname{0.00}
\expandafter\def\csname ev@B-R/4B/g3/s3/retention-g2/unit\endcsname{attempt slot}
\expandafter\def\csname ev@B-R/4B/g3/s3/retention-g2/wins\endcsname{16}
\expandafter\def\csname ev@B-R/4B/g3/s3/target/attempt-ci-hi-pp\endcsname{+10.24}
\expandafter\def\csname ev@B-R/4B/g3/s3/target/attempt-ci-lo-pp\endcsname{-4.12}
\expandafter\def\csname ev@B-R/4B/g3/s3/target/base-counted\endcsname{196}
\expandafter\def\csname ev@B-R/4B/g3/s3/target/base-passes\endcsname{27}
\expandafter\def\csname ev@B-R/4B/g3/s3/target/base-pct\endcsname{13.78}
\expandafter\def\csname ev@B-R/4B/g3/s3/target/base-scheduled\endcsname{196}
\expandafter\def\csname ev@B-R/4B/g3/s3/target/base-uncounted\endcsname{0}
\expandafter\def\csname ev@B-R/4B/g3/s3/target/cand-counted\endcsname{196}
\expandafter\def\csname ev@B-R/4B/g3/s3/target/cand-passes\endcsname{33}
\expandafter\def\csname ev@B-R/4B/g3/s3/target/cand-pct\endcsname{16.84}
\expandafter\def\csname ev@B-R/4B/g3/s3/target/cand-scheduled\endcsname{196}
\expandafter\def\csname ev@B-R/4B/g3/s3/target/cand-uncounted\endcsname{0}
\expandafter\def\csname ev@B-R/4B/g3/s3/target/diff-pp\endcsname{+3.06}
\expandafter\def\csname ev@B-R/4B/g3/s3/target/losses\endcsname{10}
\expandafter\def\csname ev@B-R/4B/g3/s3/target/outcome\endcsname{inconclusive}
\expandafter\def\csname ev@B-R/4B/g3/s3/target/p-recorded\endcsname{0.163}
\expandafter\def\csname ev@B-R/4B/g3/s3/target/protocol\endcsname{B-R}
\expandafter\def\csname ev@B-R/4B/g3/s3/target/task-losses-high\endcsname{10}
\expandafter\def\csname ev@B-R/4B/g3/s3/target/task-losses-low\endcsname{0}
\expandafter\def\csname ev@B-R/4B/g3/s3/target/task-p-high\endcsname{0.971}
\expandafter\def\csname ev@B-R/4B/g3/s3/target/task-p-low\endcsname{0.0021}
\expandafter\def\csname ev@B-R/4B/g3/s3/target/task-wins-high\endcsname{16}
\expandafter\def\csname ev@B-R/4B/g3/s3/target/task-wins-low\endcsname{2}
\expandafter\def\csname ev@B-R/4B/g3/s3/target/tasks\endcsname{49}
\expandafter\def\csname ev@B-R/4B/g3/s3/target/uncounted\endcsname{0}
\expandafter\def\csname ev@B-R/4B/g3/s3/target/uncounted-pct\endcsname{0.00}
\expandafter\def\csname ev@B-R/4B/g3/s3/target/unit\endcsname{attempt slot}
\expandafter\def\csname ev@B-R/4B/g3/s3/target/wins\endcsname{16}
\expandafter\def\csname ev@B-R/4B/g3/s4/exam/attempt-ci-hi-pp\endcsname{+4.01}
\expandafter\def\csname ev@B-R/4B/g3/s4/exam/attempt-ci-lo-pp\endcsname{-1.43}
\expandafter\def\csname ev@B-R/4B/g3/s4/exam/base-counted\endcsname{400}
\expandafter\def\csname ev@B-R/4B/g3/s4/exam/base-passes\endcsname{12}
\expandafter\def\csname ev@B-R/4B/g3/s4/exam/base-pct\endcsname{3.00}
\expandafter\def\csname ev@B-R/4B/g3/s4/exam/base-scheduled\endcsname{400}
\expandafter\def\csname ev@B-R/4B/g3/s4/exam/base-uncounted\endcsname{0}
\expandafter\def\csname ev@B-R/4B/g3/s4/exam/cand-counted\endcsname{400}
\expandafter\def\csname ev@B-R/4B/g3/s4/exam/cand-passes\endcsname{17}
\expandafter\def\csname ev@B-R/4B/g3/s4/exam/cand-pct\endcsname{4.25}
\expandafter\def\csname ev@B-R/4B/g3/s4/exam/cand-scheduled\endcsname{400}
\expandafter\def\csname ev@B-R/4B/g3/s4/exam/cand-uncounted\endcsname{0}
\expandafter\def\csname ev@B-R/4B/g3/s4/exam/diff-pp\endcsname{+1.25}
\expandafter\def\csname ev@B-R/4B/g3/s4/exam/losses\endcsname{7}
\expandafter\def\csname ev@B-R/4B/g3/s4/exam/outcome\endcsname{inconclusive}
\expandafter\def\csname ev@B-R/4B/g3/s4/exam/p-recorded\endcsname{0.18}
\expandafter\def\csname ev@B-R/4B/g3/s4/exam/protocol\endcsname{B-R}
\expandafter\def\csname ev@B-R/4B/g3/s4/exam/task-losses-high\endcsname{7}
\expandafter\def\csname ev@B-R/4B/g3/s4/exam/task-losses-low\endcsname{0}
\expandafter\def\csname ev@B-R/4B/g3/s4/exam/task-p-high\endcsname{0.945}
\expandafter\def\csname ev@B-R/4B/g3/s4/exam/task-p-low\endcsname{0.0065}
\expandafter\def\csname ev@B-R/4B/g3/s4/exam/task-wins-high\endcsname{12}
\expandafter\def\csname ev@B-R/4B/g3/s4/exam/task-wins-low\endcsname{2}
\expandafter\def\csname ev@B-R/4B/g3/s4/exam/tasks\endcsname{100}
\expandafter\def\csname ev@B-R/4B/g3/s4/exam/uncounted\endcsname{0}
\expandafter\def\csname ev@B-R/4B/g3/s4/exam/uncounted-pct\endcsname{0.00}
\expandafter\def\csname ev@B-R/4B/g3/s4/exam/unit\endcsname{attempt slot}
\expandafter\def\csname ev@B-R/4B/g3/s4/exam/wins\endcsname{12}
\expandafter\def\csname ev@B-R/4B/g3/s4/retention-g1/attempt-ci-hi-pp\endcsname{+18.01}
\expandafter\def\csname ev@B-R/4B/g3/s4/retention-g1/attempt-ci-lo-pp\endcsname{+0.54}
\expandafter\def\csname ev@B-R/4B/g3/s4/retention-g1/base-counted\endcsname{224}
\expandafter\def\csname ev@B-R/4B/g3/s4/retention-g1/base-passes\endcsname{68}
\expandafter\def\csname ev@B-R/4B/g3/s4/retention-g1/base-pct\endcsname{30.36}
\expandafter\def\csname ev@B-R/4B/g3/s4/retention-g1/base-scheduled\endcsname{224}
\expandafter\def\csname ev@B-R/4B/g3/s4/retention-g1/base-uncounted\endcsname{0}
\expandafter\def\csname ev@B-R/4B/g3/s4/retention-g1/cand-counted\endcsname{224}
\expandafter\def\csname ev@B-R/4B/g3/s4/retention-g1/cand-passes\endcsname{89}
\expandafter\def\csname ev@B-R/4B/g3/s4/retention-g1/cand-pct\endcsname{39.73}
\expandafter\def\csname ev@B-R/4B/g3/s4/retention-g1/cand-scheduled\endcsname{224}
\expandafter\def\csname ev@B-R/4B/g3/s4/retention-g1/cand-uncounted\endcsname{0}
\expandafter\def\csname ev@B-R/4B/g3/s4/retention-g1/diff-pp\endcsname{+9.38}
\expandafter\def\csname ev@B-R/4B/g3/s4/retention-g1/losses\endcsname{16}
\expandafter\def\csname ev@B-R/4B/g3/s4/retention-g1/outcome\endcsname{inconclusive}
\expandafter\def\csname ev@B-R/4B/g3/s4/retention-g1/p-recorded\endcsname{0.0027}
\expandafter\def\csname ev@B-R/4B/g3/s4/retention-g1/protocol\endcsname{B-R}
\expandafter\def\csname ev@B-R/4B/g3/s4/retention-g1/task-losses-high\endcsname{16}
\expandafter\def\csname ev@B-R/4B/g3/s4/retention-g1/task-losses-low\endcsname{0}
\expandafter\def\csname ev@B-R/4B/g3/s4/retention-g1/task-p-high\endcsname{0.924}
\expandafter\def\csname ev@B-R/4B/g3/s4/retention-g1/task-p-low\endcsname{\ensuremath{5.1\times 10^{-8}}}
\expandafter\def\csname ev@B-R/4B/g3/s4/retention-g1/task-wins-high\endcsname{37}
\expandafter\def\csname ev@B-R/4B/g3/s4/retention-g1/task-wins-low\endcsname{6}
\expandafter\def\csname ev@B-R/4B/g3/s4/retention-g1/tasks\endcsname{56}
\expandafter\def\csname ev@B-R/4B/g3/s4/retention-g1/uncounted\endcsname{0}
\expandafter\def\csname ev@B-R/4B/g3/s4/retention-g1/uncounted-pct\endcsname{0.00}
\expandafter\def\csname ev@B-R/4B/g3/s4/retention-g1/unit\endcsname{attempt slot}
\expandafter\def\csname ev@B-R/4B/g3/s4/retention-g1/wins\endcsname{37}
\expandafter\def\csname ev@B-R/4B/g3/s4/retention-g2/attempt-ci-hi-pp\endcsname{+11.55}
\expandafter\def\csname ev@B-R/4B/g3/s4/retention-g2/attempt-ci-lo-pp\endcsname{-4.14}
\expandafter\def\csname ev@B-R/4B/g3/s4/retention-g2/base-counted\endcsname{188}
\expandafter\def\csname ev@B-R/4B/g3/s4/retention-g2/base-passes\endcsname{31}
\expandafter\def\csname ev@B-R/4B/g3/s4/retention-g2/base-pct\endcsname{16.49}
\expandafter\def\csname ev@B-R/4B/g3/s4/retention-g2/base-scheduled\endcsname{188}
\expandafter\def\csname ev@B-R/4B/g3/s4/retention-g2/base-uncounted\endcsname{0}
\expandafter\def\csname ev@B-R/4B/g3/s4/retention-g2/cand-counted\endcsname{188}
\expandafter\def\csname ev@B-R/4B/g3/s4/retention-g2/cand-passes\endcsname{38}
\expandafter\def\csname ev@B-R/4B/g3/s4/retention-g2/cand-pct\endcsname{20.21}
\expandafter\def\csname ev@B-R/4B/g3/s4/retention-g2/cand-scheduled\endcsname{188}
\expandafter\def\csname ev@B-R/4B/g3/s4/retention-g2/cand-uncounted\endcsname{0}
\expandafter\def\csname ev@B-R/4B/g3/s4/retention-g2/diff-pp\endcsname{+3.72}
\expandafter\def\csname ev@B-R/4B/g3/s4/retention-g2/losses\endcsname{11}
\expandafter\def\csname ev@B-R/4B/g3/s4/retention-g2/outcome\endcsname{inconclusive}
\expandafter\def\csname ev@B-R/4B/g3/s4/retention-g2/p-recorded\endcsname{0.132}
\expandafter\def\csname ev@B-R/4B/g3/s4/retention-g2/protocol\endcsname{B-R}
\expandafter\def\csname ev@B-R/4B/g3/s4/retention-g2/task-losses-high\endcsname{11}
\expandafter\def\csname ev@B-R/4B/g3/s4/retention-g2/task-losses-low\endcsname{0}
\expandafter\def\csname ev@B-R/4B/g3/s4/retention-g2/task-p-high\endcsname{0.962}
\expandafter\def\csname ev@B-R/4B/g3/s4/retention-g2/task-p-low\endcsname{0.00074}
\expandafter\def\csname ev@B-R/4B/g3/s4/retention-g2/task-wins-high\endcsname{18}
\expandafter\def\csname ev@B-R/4B/g3/s4/retention-g2/task-wins-low\endcsname{2}
\expandafter\def\csname ev@B-R/4B/g3/s4/retention-g2/tasks\endcsname{47}
\expandafter\def\csname ev@B-R/4B/g3/s4/retention-g2/uncounted\endcsname{0}
\expandafter\def\csname ev@B-R/4B/g3/s4/retention-g2/uncounted-pct\endcsname{0.00}
\expandafter\def\csname ev@B-R/4B/g3/s4/retention-g2/unit\endcsname{attempt slot}
\expandafter\def\csname ev@B-R/4B/g3/s4/retention-g2/wins\endcsname{18}
\expandafter\def\csname ev@B-R/4B/g3/s4/target/attempt-ci-hi-pp\endcsname{+12.43}
\expandafter\def\csname ev@B-R/4B/g3/s4/target/attempt-ci-lo-pp\endcsname{-2.26}
\expandafter\def\csname ev@B-R/4B/g3/s4/target/base-counted\endcsname{196}
\expandafter\def\csname ev@B-R/4B/g3/s4/target/base-passes\endcsname{27}
\expandafter\def\csname ev@B-R/4B/g3/s4/target/base-pct\endcsname{13.78}
\expandafter\def\csname ev@B-R/4B/g3/s4/target/base-scheduled\endcsname{196}
\expandafter\def\csname ev@B-R/4B/g3/s4/target/base-uncounted\endcsname{0}
\expandafter\def\csname ev@B-R/4B/g3/s4/target/cand-counted\endcsname{196}
\expandafter\def\csname ev@B-R/4B/g3/s4/target/cand-passes\endcsname{37}
\expandafter\def\csname ev@B-R/4B/g3/s4/target/cand-pct\endcsname{18.88}
\expandafter\def\csname ev@B-R/4B/g3/s4/target/cand-scheduled\endcsname{196}
\expandafter\def\csname ev@B-R/4B/g3/s4/target/cand-uncounted\endcsname{0}
\expandafter\def\csname ev@B-R/4B/g3/s4/target/diff-pp\endcsname{+5.10}
\expandafter\def\csname ev@B-R/4B/g3/s4/target/losses\endcsname{10}
\expandafter\def\csname ev@B-R/4B/g3/s4/target/outcome\endcsname{inconclusive}
\expandafter\def\csname ev@B-R/4B/g3/s4/target/p-recorded\endcsname{0.049}
\expandafter\def\csname ev@B-R/4B/g3/s4/target/protocol\endcsname{B-R}
\expandafter\def\csname ev@B-R/4B/g3/s4/target/task-losses-high\endcsname{10}
\expandafter\def\csname ev@B-R/4B/g3/s4/target/task-losses-low\endcsname{0}
\expandafter\def\csname ev@B-R/4B/g3/s4/target/task-p-high\endcsname{0.941}
\expandafter\def\csname ev@B-R/4B/g3/s4/target/task-p-low\endcsname{0.0002}
\expandafter\def\csname ev@B-R/4B/g3/s4/target/task-wins-high\endcsname{20}
\expandafter\def\csname ev@B-R/4B/g3/s4/target/task-wins-low\endcsname{3}
\expandafter\def\csname ev@B-R/4B/g3/s4/target/tasks\endcsname{49}
\expandafter\def\csname ev@B-R/4B/g3/s4/target/uncounted\endcsname{0}
\expandafter\def\csname ev@B-R/4B/g3/s4/target/uncounted-pct\endcsname{0.00}
\expandafter\def\csname ev@B-R/4B/g3/s4/target/unit\endcsname{attempt slot}
\expandafter\def\csname ev@B-R/4B/g3/s4/target/wins\endcsname{20}
\expandafter\def\csname ev@B-R/4B/g3/s5/exam/attempt-ci-hi-pp\endcsname{+2.52}
\expandafter\def\csname ev@B-R/4B/g3/s5/exam/attempt-ci-lo-pp\endcsname{-2.52}
\expandafter\def\csname ev@B-R/4B/g3/s5/exam/base-counted\endcsname{400}
\expandafter\def\csname ev@B-R/4B/g3/s5/exam/base-passes\endcsname{12}
\expandafter\def\csname ev@B-R/4B/g3/s5/exam/base-pct\endcsname{3.00}
\expandafter\def\csname ev@B-R/4B/g3/s5/exam/base-scheduled\endcsname{400}
\expandafter\def\csname ev@B-R/4B/g3/s5/exam/base-uncounted\endcsname{0}
\expandafter\def\csname ev@B-R/4B/g3/s5/exam/cand-counted\endcsname{400}
\expandafter\def\csname ev@B-R/4B/g3/s5/exam/cand-passes\endcsname{12}
\expandafter\def\csname ev@B-R/4B/g3/s5/exam/cand-pct\endcsname{3.00}
\expandafter\def\csname ev@B-R/4B/g3/s5/exam/cand-scheduled\endcsname{400}
\expandafter\def\csname ev@B-R/4B/g3/s5/exam/cand-uncounted\endcsname{0}
\expandafter\def\csname ev@B-R/4B/g3/s5/exam/diff-pp\endcsname{+0.00}
\expandafter\def\csname ev@B-R/4B/g3/s5/exam/losses\endcsname{9}
\expandafter\def\csname ev@B-R/4B/g3/s5/exam/outcome\endcsname{inconclusive}
\expandafter\def\csname ev@B-R/4B/g3/s5/exam/p-recorded\endcsname{0.593}
\expandafter\def\csname ev@B-R/4B/g3/s5/exam/protocol\endcsname{B-R}
\expandafter\def\csname ev@B-R/4B/g3/s5/exam/task-losses-high\endcsname{9}
\expandafter\def\csname ev@B-R/4B/g3/s5/exam/task-losses-low\endcsname{0}
\expandafter\def\csname ev@B-R/4B/g3/s5/exam/task-p-high\endcsname{1}
\expandafter\def\csname ev@B-R/4B/g3/s5/exam/task-p-low\endcsname{0.055}
\expandafter\def\csname ev@B-R/4B/g3/s5/exam/task-wins-high\endcsname{9}
\expandafter\def\csname ev@B-R/4B/g3/s5/exam/task-wins-low\endcsname{0}
\expandafter\def\csname ev@B-R/4B/g3/s5/exam/tasks\endcsname{100}
\expandafter\def\csname ev@B-R/4B/g3/s5/exam/uncounted\endcsname{0}
\expandafter\def\csname ev@B-R/4B/g3/s5/exam/uncounted-pct\endcsname{0.00}
\expandafter\def\csname ev@B-R/4B/g3/s5/exam/unit\endcsname{attempt slot}
\expandafter\def\csname ev@B-R/4B/g3/s5/exam/wins\endcsname{9}
\expandafter\def\csname ev@B-R/4B/g3/s5/retention-g1/attempt-ci-hi-pp\endcsname{+20.70}
\expandafter\def\csname ev@B-R/4B/g3/s5/retention-g1/attempt-ci-lo-pp\endcsname{+3.14}
\expandafter\def\csname ev@B-R/4B/g3/s5/retention-g1/base-counted\endcsname{224}
\expandafter\def\csname ev@B-R/4B/g3/s5/retention-g1/base-passes\endcsname{68}
\expandafter\def\csname ev@B-R/4B/g3/s5/retention-g1/base-pct\endcsname{30.36}
\expandafter\def\csname ev@B-R/4B/g3/s5/retention-g1/base-scheduled\endcsname{224}
\expandafter\def\csname ev@B-R/4B/g3/s5/retention-g1/base-uncounted\endcsname{0}
\expandafter\def\csname ev@B-R/4B/g3/s5/retention-g1/cand-counted\endcsname{224}
\expandafter\def\csname ev@B-R/4B/g3/s5/retention-g1/cand-passes\endcsname{95}
\expandafter\def\csname ev@B-R/4B/g3/s5/retention-g1/cand-pct\endcsname{42.41}
\expandafter\def\csname ev@B-R/4B/g3/s5/retention-g1/cand-scheduled\endcsname{224}
\expandafter\def\csname ev@B-R/4B/g3/s5/retention-g1/cand-uncounted\endcsname{0}
\expandafter\def\csname ev@B-R/4B/g3/s5/retention-g1/diff-pp\endcsname{+12.05}
\expandafter\def\csname ev@B-R/4B/g3/s5/retention-g1/losses\endcsname{11}
\expandafter\def\csname ev@B-R/4B/g3/s5/retention-g1/outcome\endcsname{inconclusive}
\expandafter\def\csname ev@B-R/4B/g3/s5/retention-g1/p-recorded\endcsname{\ensuremath{7.1\times 10^{-5}}}
\expandafter\def\csname ev@B-R/4B/g3/s5/retention-g1/protocol\endcsname{B-R}
\expandafter\def\csname ev@B-R/4B/g3/s5/retention-g1/task-losses-high\endcsname{11}
\expandafter\def\csname ev@B-R/4B/g3/s5/retention-g1/task-losses-low\endcsname{0}
\expandafter\def\csname ev@B-R/4B/g3/s5/retention-g1/task-p-high\endcsname{0.668}
\expandafter\def\csname ev@B-R/4B/g3/s5/retention-g1/task-p-low\endcsname{\ensuremath{5.1\times 10^{-9}}}
\expandafter\def\csname ev@B-R/4B/g3/s5/retention-g1/task-wins-high\endcsname{38}
\expandafter\def\csname ev@B-R/4B/g3/s5/retention-g1/task-wins-low\endcsname{7}
\expandafter\def\csname ev@B-R/4B/g3/s5/retention-g1/tasks\endcsname{56}
\expandafter\def\csname ev@B-R/4B/g3/s5/retention-g1/uncounted\endcsname{0}
\expandafter\def\csname ev@B-R/4B/g3/s5/retention-g1/uncounted-pct\endcsname{0.00}
\expandafter\def\csname ev@B-R/4B/g3/s5/retention-g1/unit\endcsname{attempt slot}
\expandafter\def\csname ev@B-R/4B/g3/s5/retention-g1/wins\endcsname{38}
\expandafter\def\csname ev@B-R/4B/g3/s5/retention-g2/attempt-ci-hi-pp\endcsname{+12.68}
\expandafter\def\csname ev@B-R/4B/g3/s5/retention-g2/attempt-ci-lo-pp\endcsname{-3.16}
\expandafter\def\csname ev@B-R/4B/g3/s5/retention-g2/base-counted\endcsname{188}
\expandafter\def\csname ev@B-R/4B/g3/s5/retention-g2/base-passes\endcsname{31}
\expandafter\def\csname ev@B-R/4B/g3/s5/retention-g2/base-pct\endcsname{16.49}
\expandafter\def\csname ev@B-R/4B/g3/s5/retention-g2/base-scheduled\endcsname{188}
\expandafter\def\csname ev@B-R/4B/g3/s5/retention-g2/base-uncounted\endcsname{0}
\expandafter\def\csname ev@B-R/4B/g3/s5/retention-g2/cand-counted\endcsname{188}
\expandafter\def\csname ev@B-R/4B/g3/s5/retention-g2/cand-passes\endcsname{40}
\expandafter\def\csname ev@B-R/4B/g3/s5/retention-g2/cand-pct\endcsname{21.28}
\expandafter\def\csname ev@B-R/4B/g3/s5/retention-g2/cand-scheduled\endcsname{188}
\expandafter\def\csname ev@B-R/4B/g3/s5/retention-g2/cand-uncounted\endcsname{0}
\expandafter\def\csname ev@B-R/4B/g3/s5/retention-g2/diff-pp\endcsname{+4.79}
\expandafter\def\csname ev@B-R/4B/g3/s5/retention-g2/losses\endcsname{12}
\expandafter\def\csname ev@B-R/4B/g3/s5/retention-g2/outcome\endcsname{inconclusive}
\expandafter\def\csname ev@B-R/4B/g3/s5/retention-g2/p-recorded\endcsname{0.081}
\expandafter\def\csname ev@B-R/4B/g3/s5/retention-g2/protocol\endcsname{B-R}
\expandafter\def\csname ev@B-R/4B/g3/s5/retention-g2/task-losses-high\endcsname{12}
\expandafter\def\csname ev@B-R/4B/g3/s5/retention-g2/task-losses-low\endcsname{0}
\expandafter\def\csname ev@B-R/4B/g3/s5/retention-g2/task-p-high\endcsname{0.962}
\expandafter\def\csname ev@B-R/4B/g3/s5/retention-g2/task-p-low\endcsname{0.00014}
\expandafter\def\csname ev@B-R/4B/g3/s5/retention-g2/task-wins-high\endcsname{21}
\expandafter\def\csname ev@B-R/4B/g3/s5/retention-g2/task-wins-low\endcsname{3}
\expandafter\def\csname ev@B-R/4B/g3/s5/retention-g2/tasks\endcsname{47}
\expandafter\def\csname ev@B-R/4B/g3/s5/retention-g2/uncounted\endcsname{0}
\expandafter\def\csname ev@B-R/4B/g3/s5/retention-g2/uncounted-pct\endcsname{0.00}
\expandafter\def\csname ev@B-R/4B/g3/s5/retention-g2/unit\endcsname{attempt slot}
\expandafter\def\csname ev@B-R/4B/g3/s5/retention-g2/wins\endcsname{21}
\expandafter\def\csname ev@B-R/4B/g3/s5/target/attempt-ci-hi-pp\endcsname{+14.62}
\expandafter\def\csname ev@B-R/4B/g3/s5/target/attempt-ci-lo-pp\endcsname{-0.38}
\expandafter\def\csname ev@B-R/4B/g3/s5/target/base-counted\endcsname{196}
\expandafter\def\csname ev@B-R/4B/g3/s5/target/base-passes\endcsname{27}
\expandafter\def\csname ev@B-R/4B/g3/s5/target/base-pct\endcsname{13.78}
\expandafter\def\csname ev@B-R/4B/g3/s5/target/base-scheduled\endcsname{196}
\expandafter\def\csname ev@B-R/4B/g3/s5/target/base-uncounted\endcsname{0}
\expandafter\def\csname ev@B-R/4B/g3/s5/target/cand-counted\endcsname{196}
\expandafter\def\csname ev@B-R/4B/g3/s5/target/cand-passes\endcsname{41}
\expandafter\def\csname ev@B-R/4B/g3/s5/target/cand-pct\endcsname{20.92}
\expandafter\def\csname ev@B-R/4B/g3/s5/target/cand-scheduled\endcsname{196}
\expandafter\def\csname ev@B-R/4B/g3/s5/target/cand-uncounted\endcsname{0}
\expandafter\def\csname ev@B-R/4B/g3/s5/target/diff-pp\endcsname{+7.14}
\expandafter\def\csname ev@B-R/4B/g3/s5/target/losses\endcsname{12}
\expandafter\def\csname ev@B-R/4B/g3/s5/target/outcome\endcsname{inconclusive}
\expandafter\def\csname ev@B-R/4B/g3/s5/target/p-recorded\endcsname{0.017}
\expandafter\def\csname ev@B-R/4B/g3/s5/target/protocol\endcsname{B-R}
\expandafter\def\csname ev@B-R/4B/g3/s5/target/task-losses-high\endcsname{12}
\expandafter\def\csname ev@B-R/4B/g3/s5/target/task-losses-low\endcsname{0}
\expandafter\def\csname ev@B-R/4B/g3/s5/target/task-p-high\endcsname{0.916}
\expandafter\def\csname ev@B-R/4B/g3/s5/target/task-p-low\endcsname{\ensuremath{7.6\times 10^{-6}}}
\expandafter\def\csname ev@B-R/4B/g3/s5/target/task-wins-high\endcsname{26}
\expandafter\def\csname ev@B-R/4B/g3/s5/target/task-wins-low\endcsname{4}
\expandafter\def\csname ev@B-R/4B/g3/s5/target/tasks\endcsname{49}
\expandafter\def\csname ev@B-R/4B/g3/s5/target/uncounted\endcsname{0}
\expandafter\def\csname ev@B-R/4B/g3/s5/target/uncounted-pct\endcsname{0.00}
\expandafter\def\csname ev@B-R/4B/g3/s5/target/unit\endcsname{attempt slot}
\expandafter\def\csname ev@B-R/4B/g3/s5/target/wins\endcsname{26}
\expandafter\def\csname ev@B-R/4B/g3/unfinished-seeds\endcsname{0}
\expandafter\def\csname ev@B-R/8B/g2/pooled/exam/base-counted\endcsname{400}
\expandafter\def\csname ev@B-R/8B/g2/pooled/exam/base-passes\endcsname{30}
\expandafter\def\csname ev@B-R/8B/g2/pooled/exam/base-pct\endcsname{7.50}
\expandafter\def\csname ev@B-R/8B/g2/pooled/exam/base-scheduled\endcsname{400}
\expandafter\def\csname ev@B-R/8B/g2/pooled/exam/base-uncounted\endcsname{0}
\expandafter\def\csname ev@B-R/8B/g2/pooled/exam/cand-counted\endcsname{1600}
\expandafter\def\csname ev@B-R/8B/g2/pooled/exam/cand-passes\endcsname{119}
\expandafter\def\csname ev@B-R/8B/g2/pooled/exam/cand-pct\endcsname{7.44}
\expandafter\def\csname ev@B-R/8B/g2/pooled/exam/cand-scheduled\endcsname{1600}
\expandafter\def\csname ev@B-R/8B/g2/pooled/exam/cand-uncounted\endcsname{0}
\expandafter\def\csname ev@B-R/8B/g2/pooled/exam/diff-pp\endcsname{-0.06}
\expandafter\def\csname ev@B-R/8B/g2/pooled/exam/losses\endcsname{64}
\expandafter\def\csname ev@B-R/8B/g2/pooled/exam/outcome\endcsname{inconclusive}
\expandafter\def\csname ev@B-R/8B/g2/pooled/exam/p-recorded\endcsname{0.57}
\expandafter\def\csname ev@B-R/8B/g2/pooled/exam/protocol\endcsname{B-R}
\expandafter\def\csname ev@B-R/8B/g2/pooled/exam/seeds\endcsname{4}
\expandafter\def\csname ev@B-R/8B/g2/pooled/exam/seeds-negative\endcsname{2}
\expandafter\def\csname ev@B-R/8B/g2/pooled/exam/seeds-positive\endcsname{1}
\expandafter\def\csname ev@B-R/8B/g2/pooled/exam/tasks\endcsname{100}
\expandafter\def\csname ev@B-R/8B/g2/pooled/exam/uncounted\endcsname{0}
\expandafter\def\csname ev@B-R/8B/g2/pooled/exam/uncounted-pct\endcsname{0.00}
\expandafter\def\csname ev@B-R/8B/g2/pooled/exam/unit\endcsname{attempt slot, summed over training seeds}
\expandafter\def\csname ev@B-R/8B/g2/pooled/exam/wins\endcsname{63}
\expandafter\def\csname ev@B-R/8B/g2/pooled/retention-g1/base-counted\endcsname{224}
\expandafter\def\csname ev@B-R/8B/g2/pooled/retention-g1/base-passes\endcsname{86}
\expandafter\def\csname ev@B-R/8B/g2/pooled/retention-g1/base-pct\endcsname{38.39}
\expandafter\def\csname ev@B-R/8B/g2/pooled/retention-g1/base-scheduled\endcsname{224}
\expandafter\def\csname ev@B-R/8B/g2/pooled/retention-g1/base-uncounted\endcsname{0}
\expandafter\def\csname ev@B-R/8B/g2/pooled/retention-g1/cand-counted\endcsname{896}
\expandafter\def\csname ev@B-R/8B/g2/pooled/retention-g1/cand-passes\endcsname{375}
\expandafter\def\csname ev@B-R/8B/g2/pooled/retention-g1/cand-pct\endcsname{41.85}
\expandafter\def\csname ev@B-R/8B/g2/pooled/retention-g1/cand-scheduled\endcsname{896}
\expandafter\def\csname ev@B-R/8B/g2/pooled/retention-g1/cand-uncounted\endcsname{0}
\expandafter\def\csname ev@B-R/8B/g2/pooled/retention-g1/diff-pp\endcsname{+3.46}
\expandafter\def\csname ev@B-R/8B/g2/pooled/retention-g1/losses\endcsname{84}
\expandafter\def\csname ev@B-R/8B/g2/pooled/retention-g1/outcome\endcsname{inconclusive}
\expandafter\def\csname ev@B-R/8B/g2/pooled/retention-g1/p-recorded\endcsname{0.017}
\expandafter\def\csname ev@B-R/8B/g2/pooled/retention-g1/protocol\endcsname{B-R}
\expandafter\def\csname ev@B-R/8B/g2/pooled/retention-g1/seeds\endcsname{4}
\expandafter\def\csname ev@B-R/8B/g2/pooled/retention-g1/seeds-negative\endcsname{0}
\expandafter\def\csname ev@B-R/8B/g2/pooled/retention-g1/seeds-positive\endcsname{4}
\expandafter\def\csname ev@B-R/8B/g2/pooled/retention-g1/tasks\endcsname{56}
\expandafter\def\csname ev@B-R/8B/g2/pooled/retention-g1/uncounted\endcsname{0}
\expandafter\def\csname ev@B-R/8B/g2/pooled/retention-g1/uncounted-pct\endcsname{0.00}
\expandafter\def\csname ev@B-R/8B/g2/pooled/retention-g1/unit\endcsname{attempt slot, summed over training seeds}
\expandafter\def\csname ev@B-R/8B/g2/pooled/retention-g1/wins\endcsname{115}
\expandafter\def\csname ev@B-R/8B/g2/pooled/target/base-counted\endcsname{188}
\expandafter\def\csname ev@B-R/8B/g2/pooled/target/base-passes\endcsname{40}
\expandafter\def\csname ev@B-R/8B/g2/pooled/target/base-pct\endcsname{21.28}
\expandafter\def\csname ev@B-R/8B/g2/pooled/target/base-scheduled\endcsname{188}
\expandafter\def\csname ev@B-R/8B/g2/pooled/target/base-uncounted\endcsname{0}
\expandafter\def\csname ev@B-R/8B/g2/pooled/target/cand-counted\endcsname{752}
\expandafter\def\csname ev@B-R/8B/g2/pooled/target/cand-passes\endcsname{206}
\expandafter\def\csname ev@B-R/8B/g2/pooled/target/cand-pct\endcsname{27.39}
\expandafter\def\csname ev@B-R/8B/g2/pooled/target/cand-scheduled\endcsname{752}
\expandafter\def\csname ev@B-R/8B/g2/pooled/target/cand-uncounted\endcsname{0}
\expandafter\def\csname ev@B-R/8B/g2/pooled/target/diff-pp\endcsname{+6.12}
\expandafter\def\csname ev@B-R/8B/g2/pooled/target/losses\endcsname{53}
\expandafter\def\csname ev@B-R/8B/g2/pooled/target/outcome\endcsname{inconclusive}
\expandafter\def\csname ev@B-R/8B/g2/pooled/target/p-recorded\endcsname{0.00012}
\expandafter\def\csname ev@B-R/8B/g2/pooled/target/protocol\endcsname{B-R}
\expandafter\def\csname ev@B-R/8B/g2/pooled/target/seeds\endcsname{4}
\expandafter\def\csname ev@B-R/8B/g2/pooled/target/seeds-negative\endcsname{0}
\expandafter\def\csname ev@B-R/8B/g2/pooled/target/seeds-positive\endcsname{4}
\expandafter\def\csname ev@B-R/8B/g2/pooled/target/tasks\endcsname{47}
\expandafter\def\csname ev@B-R/8B/g2/pooled/target/uncounted\endcsname{0}
\expandafter\def\csname ev@B-R/8B/g2/pooled/target/uncounted-pct\endcsname{0.00}
\expandafter\def\csname ev@B-R/8B/g2/pooled/target/unit\endcsname{attempt slot, summed over training seeds}
\expandafter\def\csname ev@B-R/8B/g2/pooled/target/wins\endcsname{99}
\expandafter\def\csname ev@B-R/8B/g2/s1/exam/attempt-ci-hi-pp\endcsname{+3.72}
\expandafter\def\csname ev@B-R/8B/g2/s1/exam/attempt-ci-lo-pp\endcsname{-3.72}
\expandafter\def\csname ev@B-R/8B/g2/s1/exam/base-counted\endcsname{400}
\expandafter\def\csname ev@B-R/8B/g2/s1/exam/base-passes\endcsname{30}
\expandafter\def\csname ev@B-R/8B/g2/s1/exam/base-pct\endcsname{7.50}
\expandafter\def\csname ev@B-R/8B/g2/s1/exam/base-scheduled\endcsname{400}
\expandafter\def\csname ev@B-R/8B/g2/s1/exam/base-uncounted\endcsname{0}
\expandafter\def\csname ev@B-R/8B/g2/s1/exam/cand-counted\endcsname{400}
\expandafter\def\csname ev@B-R/8B/g2/s1/exam/cand-passes\endcsname{30}
\expandafter\def\csname ev@B-R/8B/g2/s1/exam/cand-pct\endcsname{7.50}
\expandafter\def\csname ev@B-R/8B/g2/s1/exam/cand-scheduled\endcsname{400}
\expandafter\def\csname ev@B-R/8B/g2/s1/exam/cand-uncounted\endcsname{0}
\expandafter\def\csname ev@B-R/8B/g2/s1/exam/diff-pp\endcsname{+0.00}
\expandafter\def\csname ev@B-R/8B/g2/s1/exam/losses\endcsname{17}
\expandafter\def\csname ev@B-R/8B/g2/s1/exam/outcome\endcsname{inconclusive}
\expandafter\def\csname ev@B-R/8B/g2/s1/exam/p-recorded\endcsname{0.568}
\expandafter\def\csname ev@B-R/8B/g2/s1/exam/protocol\endcsname{B-R}
\expandafter\def\csname ev@B-R/8B/g2/s1/exam/task-losses-high\endcsname{17}
\expandafter\def\csname ev@B-R/8B/g2/s1/exam/task-losses-low\endcsname{0}
\expandafter\def\csname ev@B-R/8B/g2/s1/exam/task-p-high\endcsname{1}
\expandafter\def\csname ev@B-R/8B/g2/s1/exam/task-p-low\endcsname{0.0059}
\expandafter\def\csname ev@B-R/8B/g2/s1/exam/task-wins-high\endcsname{17}
\expandafter\def\csname ev@B-R/8B/g2/s1/exam/task-wins-low\endcsname{0}
\expandafter\def\csname ev@B-R/8B/g2/s1/exam/tasks\endcsname{100}
\expandafter\def\csname ev@B-R/8B/g2/s1/exam/uncounted\endcsname{0}
\expandafter\def\csname ev@B-R/8B/g2/s1/exam/uncounted-pct\endcsname{0.00}
\expandafter\def\csname ev@B-R/8B/g2/s1/exam/unit\endcsname{attempt slot}
\expandafter\def\csname ev@B-R/8B/g2/s1/exam/wins\endcsname{17}
\expandafter\def\csname ev@B-R/8B/g2/s1/retention-g1/attempt-ci-hi-pp\endcsname{+11.64}
\expandafter\def\csname ev@B-R/8B/g2/s1/retention-g1/attempt-ci-lo-pp\endcsname{-6.34}
\expandafter\def\csname ev@B-R/8B/g2/s1/retention-g1/base-counted\endcsname{224}
\expandafter\def\csname ev@B-R/8B/g2/s1/retention-g1/base-passes\endcsname{86}
\expandafter\def\csname ev@B-R/8B/g2/s1/retention-g1/base-pct\endcsname{38.39}
\expandafter\def\csname ev@B-R/8B/g2/s1/retention-g1/base-scheduled\endcsname{224}
\expandafter\def\csname ev@B-R/8B/g2/s1/retention-g1/base-uncounted\endcsname{0}
\expandafter\def\csname ev@B-R/8B/g2/s1/retention-g1/cand-counted\endcsname{224}
\expandafter\def\csname ev@B-R/8B/g2/s1/retention-g1/cand-passes\endcsname{92}
\expandafter\def\csname ev@B-R/8B/g2/s1/retention-g1/cand-pct\endcsname{41.07}
\expandafter\def\csname ev@B-R/8B/g2/s1/retention-g1/cand-scheduled\endcsname{224}
\expandafter\def\csname ev@B-R/8B/g2/s1/retention-g1/cand-uncounted\endcsname{0}
\expandafter\def\csname ev@B-R/8B/g2/s1/retention-g1/diff-pp\endcsname{+2.68}
\expandafter\def\csname ev@B-R/8B/g2/s1/retention-g1/losses\endcsname{23}
\expandafter\def\csname ev@B-R/8B/g2/s1/retention-g1/outcome\endcsname{inconclusive}
\expandafter\def\csname ev@B-R/8B/g2/s1/retention-g1/p-recorded\endcsname{0.244}
\expandafter\def\csname ev@B-R/8B/g2/s1/retention-g1/protocol\endcsname{B-R}
\expandafter\def\csname ev@B-R/8B/g2/s1/retention-g1/task-losses-high\endcsname{23}
\expandafter\def\csname ev@B-R/8B/g2/s1/retention-g1/task-losses-low\endcsname{0}
\expandafter\def\csname ev@B-R/8B/g2/s1/retention-g1/task-p-high\endcsname{0.999}
\expandafter\def\csname ev@B-R/8B/g2/s1/retention-g1/task-p-low\endcsname{\ensuremath{5.8\times 10^{-5}}}
\expandafter\def\csname ev@B-R/8B/g2/s1/retention-g1/task-wins-high\endcsname{29}
\expandafter\def\csname ev@B-R/8B/g2/s1/retention-g1/task-wins-low\endcsname{2}
\expandafter\def\csname ev@B-R/8B/g2/s1/retention-g1/tasks\endcsname{56}
\expandafter\def\csname ev@B-R/8B/g2/s1/retention-g1/uncounted\endcsname{0}
\expandafter\def\csname ev@B-R/8B/g2/s1/retention-g1/uncounted-pct\endcsname{0.00}
\expandafter\def\csname ev@B-R/8B/g2/s1/retention-g1/unit\endcsname{attempt slot}
\expandafter\def\csname ev@B-R/8B/g2/s1/retention-g1/wins\endcsname{29}
\expandafter\def\csname ev@B-R/8B/g2/s1/target/attempt-ci-hi-pp\endcsname{+16.62}
\expandafter\def\csname ev@B-R/8B/g2/s1/target/attempt-ci-lo-pp\endcsname{-0.81}
\expandafter\def\csname ev@B-R/8B/g2/s1/target/base-counted\endcsname{188}
\expandafter\def\csname ev@B-R/8B/g2/s1/target/base-passes\endcsname{40}
\expandafter\def\csname ev@B-R/8B/g2/s1/target/base-pct\endcsname{21.28}
\expandafter\def\csname ev@B-R/8B/g2/s1/target/base-scheduled\endcsname{188}
\expandafter\def\csname ev@B-R/8B/g2/s1/target/base-uncounted\endcsname{0}
\expandafter\def\csname ev@B-R/8B/g2/s1/target/cand-counted\endcsname{188}
\expandafter\def\csname ev@B-R/8B/g2/s1/target/cand-passes\endcsname{55}
\expandafter\def\csname ev@B-R/8B/g2/s1/target/cand-pct\endcsname{29.26}
\expandafter\def\csname ev@B-R/8B/g2/s1/target/cand-scheduled\endcsname{188}
\expandafter\def\csname ev@B-R/8B/g2/s1/target/cand-uncounted\endcsname{0}
\expandafter\def\csname ev@B-R/8B/g2/s1/target/diff-pp\endcsname{+7.98}
\expandafter\def\csname ev@B-R/8B/g2/s1/target/losses\endcsname{14}
\expandafter\def\csname ev@B-R/8B/g2/s1/target/outcome\endcsname{inconclusive}
\expandafter\def\csname ev@B-R/8B/g2/s1/target/p-recorded\endcsname{0.016}
\expandafter\def\csname ev@B-R/8B/g2/s1/target/protocol\endcsname{B-R}
\expandafter\def\csname ev@B-R/8B/g2/s1/target/task-losses-high\endcsname{14}
\expandafter\def\csname ev@B-R/8B/g2/s1/target/task-losses-low\endcsname{0}
\expandafter\def\csname ev@B-R/8B/g2/s1/target/task-p-high\endcsname{0.942}
\expandafter\def\csname ev@B-R/8B/g2/s1/target/task-p-low\endcsname{\ensuremath{4.2\times 10^{-6}}}
\expandafter\def\csname ev@B-R/8B/g2/s1/target/task-wins-high\endcsname{29}
\expandafter\def\csname ev@B-R/8B/g2/s1/target/task-wins-low\endcsname{4}
\expandafter\def\csname ev@B-R/8B/g2/s1/target/tasks\endcsname{47}
\expandafter\def\csname ev@B-R/8B/g2/s1/target/uncounted\endcsname{0}
\expandafter\def\csname ev@B-R/8B/g2/s1/target/uncounted-pct\endcsname{0.00}
\expandafter\def\csname ev@B-R/8B/g2/s1/target/unit\endcsname{attempt slot}
\expandafter\def\csname ev@B-R/8B/g2/s1/target/wins\endcsname{29}
\expandafter\def\csname ev@B-R/8B/g2/s2/exam/attempt-ci-hi-pp\endcsname{+2.88}
\expandafter\def\csname ev@B-R/8B/g2/s2/exam/attempt-ci-lo-pp\endcsname{-4.40}
\expandafter\def\csname ev@B-R/8B/g2/s2/exam/base-counted\endcsname{400}
\expandafter\def\csname ev@B-R/8B/g2/s2/exam/base-passes\endcsname{30}
\expandafter\def\csname ev@B-R/8B/g2/s2/exam/base-pct\endcsname{7.50}
\expandafter\def\csname ev@B-R/8B/g2/s2/exam/base-scheduled\endcsname{400}
\expandafter\def\csname ev@B-R/8B/g2/s2/exam/base-uncounted\endcsname{0}
\expandafter\def\csname ev@B-R/8B/g2/s2/exam/cand-counted\endcsname{400}
\expandafter\def\csname ev@B-R/8B/g2/s2/exam/cand-passes\endcsname{27}
\expandafter\def\csname ev@B-R/8B/g2/s2/exam/cand-pct\endcsname{6.75}
\expandafter\def\csname ev@B-R/8B/g2/s2/exam/cand-scheduled\endcsname{400}
\expandafter\def\csname ev@B-R/8B/g2/s2/exam/cand-uncounted\endcsname{0}
\expandafter\def\csname ev@B-R/8B/g2/s2/exam/diff-pp\endcsname{-0.75}
\expandafter\def\csname ev@B-R/8B/g2/s2/exam/losses\endcsname{14}
\expandafter\def\csname ev@B-R/8B/g2/s2/exam/outcome\endcsname{inconclusive}
\expandafter\def\csname ev@B-R/8B/g2/s2/exam/p-recorded\endcsname{0.788}
\expandafter\def\csname ev@B-R/8B/g2/s2/exam/protocol\endcsname{B-R}
\expandafter\def\csname ev@B-R/8B/g2/s2/exam/task-losses-high\endcsname{14}
\expandafter\def\csname ev@B-R/8B/g2/s2/exam/task-losses-low\endcsname{1}
\expandafter\def\csname ev@B-R/8B/g2/s2/exam/task-p-high\endcsname{1}
\expandafter\def\csname ev@B-R/8B/g2/s2/exam/task-p-low\endcsname{0.059}
\expandafter\def\csname ev@B-R/8B/g2/s2/exam/task-wins-high\endcsname{11}
\expandafter\def\csname ev@B-R/8B/g2/s2/exam/task-wins-low\endcsname{0}
\expandafter\def\csname ev@B-R/8B/g2/s2/exam/tasks\endcsname{100}
\expandafter\def\csname ev@B-R/8B/g2/s2/exam/uncounted\endcsname{0}
\expandafter\def\csname ev@B-R/8B/g2/s2/exam/uncounted-pct\endcsname{0.00}
\expandafter\def\csname ev@B-R/8B/g2/s2/exam/unit\endcsname{attempt slot}
\expandafter\def\csname ev@B-R/8B/g2/s2/exam/wins\endcsname{11}
\expandafter\def\csname ev@B-R/8B/g2/s2/retention-g1/attempt-ci-hi-pp\endcsname{+11.19}
\expandafter\def\csname ev@B-R/8B/g2/s2/retention-g1/attempt-ci-lo-pp\endcsname{-6.78}
\expandafter\def\csname ev@B-R/8B/g2/s2/retention-g1/base-counted\endcsname{224}
\expandafter\def\csname ev@B-R/8B/g2/s2/retention-g1/base-passes\endcsname{86}
\expandafter\def\csname ev@B-R/8B/g2/s2/retention-g1/base-pct\endcsname{38.39}
\expandafter\def\csname ev@B-R/8B/g2/s2/retention-g1/base-scheduled\endcsname{224}
\expandafter\def\csname ev@B-R/8B/g2/s2/retention-g1/base-uncounted\endcsname{0}
\expandafter\def\csname ev@B-R/8B/g2/s2/retention-g1/cand-counted\endcsname{224}
\expandafter\def\csname ev@B-R/8B/g2/s2/retention-g1/cand-passes\endcsname{91}
\expandafter\def\csname ev@B-R/8B/g2/s2/retention-g1/cand-pct\endcsname{40.62}
\expandafter\def\csname ev@B-R/8B/g2/s2/retention-g1/cand-scheduled\endcsname{224}
\expandafter\def\csname ev@B-R/8B/g2/s2/retention-g1/cand-uncounted\endcsname{0}
\expandafter\def\csname ev@B-R/8B/g2/s2/retention-g1/diff-pp\endcsname{+2.23}
\expandafter\def\csname ev@B-R/8B/g2/s2/retention-g1/losses\endcsname{23}
\expandafter\def\csname ev@B-R/8B/g2/s2/retention-g1/outcome\endcsname{inconclusive}
\expandafter\def\csname ev@B-R/8B/g2/s2/retention-g1/p-recorded\endcsname{0.288}
\expandafter\def\csname ev@B-R/8B/g2/s2/retention-g1/protocol\endcsname{B-R}
\expandafter\def\csname ev@B-R/8B/g2/s2/retention-g1/task-losses-high\endcsname{23}
\expandafter\def\csname ev@B-R/8B/g2/s2/retention-g1/task-losses-low\endcsname{0}
\expandafter\def\csname ev@B-R/8B/g2/s2/retention-g1/task-p-high\endcsname{0.999}
\expandafter\def\csname ev@B-R/8B/g2/s2/retention-g1/task-p-low\endcsname{\ensuremath{9.8\times 10^{-5}}}
\expandafter\def\csname ev@B-R/8B/g2/s2/retention-g1/task-wins-high\endcsname{28}
\expandafter\def\csname ev@B-R/8B/g2/s2/retention-g1/task-wins-low\endcsname{2}
\expandafter\def\csname ev@B-R/8B/g2/s2/retention-g1/tasks\endcsname{56}
\expandafter\def\csname ev@B-R/8B/g2/s2/retention-g1/uncounted\endcsname{0}
\expandafter\def\csname ev@B-R/8B/g2/s2/retention-g1/uncounted-pct\endcsname{0.00}
\expandafter\def\csname ev@B-R/8B/g2/s2/retention-g1/unit\endcsname{attempt slot}
\expandafter\def\csname ev@B-R/8B/g2/s2/retention-g1/wins\endcsname{28}
\expandafter\def\csname ev@B-R/8B/g2/s2/target/attempt-ci-hi-pp\endcsname{+14.41}
\expandafter\def\csname ev@B-R/8B/g2/s2/target/attempt-ci-lo-pp\endcsname{-2.82}
\expandafter\def\csname ev@B-R/8B/g2/s2/target/base-counted\endcsname{188}
\expandafter\def\csname ev@B-R/8B/g2/s2/target/base-passes\endcsname{40}
\expandafter\def\csname ev@B-R/8B/g2/s2/target/base-pct\endcsname{21.28}
\expandafter\def\csname ev@B-R/8B/g2/s2/target/base-scheduled\endcsname{188}
\expandafter\def\csname ev@B-R/8B/g2/s2/target/base-uncounted\endcsname{0}
\expandafter\def\csname ev@B-R/8B/g2/s2/target/cand-counted\endcsname{188}
\expandafter\def\csname ev@B-R/8B/g2/s2/target/cand-passes\endcsname{51}
\expandafter\def\csname ev@B-R/8B/g2/s2/target/cand-pct\endcsname{27.13}
\expandafter\def\csname ev@B-R/8B/g2/s2/target/cand-scheduled\endcsname{188}
\expandafter\def\csname ev@B-R/8B/g2/s2/target/cand-uncounted\endcsname{0}
\expandafter\def\csname ev@B-R/8B/g2/s2/target/diff-pp\endcsname{+5.85}
\expandafter\def\csname ev@B-R/8B/g2/s2/target/losses\endcsname{17}
\expandafter\def\csname ev@B-R/8B/g2/s2/target/outcome\endcsname{inconclusive}
\expandafter\def\csname ev@B-R/8B/g2/s2/target/p-recorded\endcsname{0.068}
\expandafter\def\csname ev@B-R/8B/g2/s2/target/protocol\endcsname{B-R}
\expandafter\def\csname ev@B-R/8B/g2/s2/target/task-losses-high\endcsname{17}
\expandafter\def\csname ev@B-R/8B/g2/s2/target/task-losses-low\endcsname{0}
\expandafter\def\csname ev@B-R/8B/g2/s2/target/task-p-high\endcsname{0.989}
\expandafter\def\csname ev@B-R/8B/g2/s2/target/task-p-low\endcsname{\ensuremath{1.7\times 10^{-5}}}
\expandafter\def\csname ev@B-R/8B/g2/s2/target/task-wins-high\endcsname{28}
\expandafter\def\csname ev@B-R/8B/g2/s2/target/task-wins-low\endcsname{3}
\expandafter\def\csname ev@B-R/8B/g2/s2/target/tasks\endcsname{47}
\expandafter\def\csname ev@B-R/8B/g2/s2/target/uncounted\endcsname{0}
\expandafter\def\csname ev@B-R/8B/g2/s2/target/uncounted-pct\endcsname{0.00}
\expandafter\def\csname ev@B-R/8B/g2/s2/target/unit\endcsname{attempt slot}
\expandafter\def\csname ev@B-R/8B/g2/s2/target/wins\endcsname{28}
\expandafter\def\csname ev@B-R/8B/g2/s3/exam/attempt-ci-hi-pp\endcsname{+3.44}
\expandafter\def\csname ev@B-R/8B/g2/s3/exam/attempt-ci-lo-pp\endcsname{-3.95}
\expandafter\def\csname ev@B-R/8B/g2/s3/exam/base-counted\endcsname{400}
\expandafter\def\csname ev@B-R/8B/g2/s3/exam/base-passes\endcsname{30}
\expandafter\def\csname ev@B-R/8B/g2/s3/exam/base-pct\endcsname{7.50}
\expandafter\def\csname ev@B-R/8B/g2/s3/exam/base-scheduled\endcsname{400}
\expandafter\def\csname ev@B-R/8B/g2/s3/exam/base-uncounted\endcsname{0}
\expandafter\def\csname ev@B-R/8B/g2/s3/exam/cand-counted\endcsname{400}
\expandafter\def\csname ev@B-R/8B/g2/s3/exam/cand-passes\endcsname{29}
\expandafter\def\csname ev@B-R/8B/g2/s3/exam/cand-pct\endcsname{7.25}
\expandafter\def\csname ev@B-R/8B/g2/s3/exam/cand-scheduled\endcsname{400}
\expandafter\def\csname ev@B-R/8B/g2/s3/exam/cand-uncounted\endcsname{0}
\expandafter\def\csname ev@B-R/8B/g2/s3/exam/diff-pp\endcsname{-0.25}
\expandafter\def\csname ev@B-R/8B/g2/s3/exam/losses\endcsname{17}
\expandafter\def\csname ev@B-R/8B/g2/s3/exam/outcome\endcsname{inconclusive}
\expandafter\def\csname ev@B-R/8B/g2/s3/exam/p-recorded\endcsname{0.636}
\expandafter\def\csname ev@B-R/8B/g2/s3/exam/protocol\endcsname{B-R}
\expandafter\def\csname ev@B-R/8B/g2/s3/exam/task-losses-high\endcsname{17}
\expandafter\def\csname ev@B-R/8B/g2/s3/exam/task-losses-low\endcsname{1}
\expandafter\def\csname ev@B-R/8B/g2/s3/exam/task-p-high\endcsname{1}
\expandafter\def\csname ev@B-R/8B/g2/s3/exam/task-p-low\endcsname{0.0096}
\expandafter\def\csname ev@B-R/8B/g2/s3/exam/task-wins-high\endcsname{16}
\expandafter\def\csname ev@B-R/8B/g2/s3/exam/task-wins-low\endcsname{0}
\expandafter\def\csname ev@B-R/8B/g2/s3/exam/tasks\endcsname{100}
\expandafter\def\csname ev@B-R/8B/g2/s3/exam/uncounted\endcsname{0}
\expandafter\def\csname ev@B-R/8B/g2/s3/exam/uncounted-pct\endcsname{0.00}
\expandafter\def\csname ev@B-R/8B/g2/s3/exam/unit\endcsname{attempt slot}
\expandafter\def\csname ev@B-R/8B/g2/s3/exam/wins\endcsname{16}
\expandafter\def\csname ev@B-R/8B/g2/s3/retention-g1/attempt-ci-hi-pp\endcsname{+13.43}
\expandafter\def\csname ev@B-R/8B/g2/s3/retention-g1/attempt-ci-lo-pp\endcsname{-4.60}
\expandafter\def\csname ev@B-R/8B/g2/s3/retention-g1/base-counted\endcsname{224}
\expandafter\def\csname ev@B-R/8B/g2/s3/retention-g1/base-passes\endcsname{86}
\expandafter\def\csname ev@B-R/8B/g2/s3/retention-g1/base-pct\endcsname{38.39}
\expandafter\def\csname ev@B-R/8B/g2/s3/retention-g1/base-scheduled\endcsname{224}
\expandafter\def\csname ev@B-R/8B/g2/s3/retention-g1/base-uncounted\endcsname{0}
\expandafter\def\csname ev@B-R/8B/g2/s3/retention-g1/cand-counted\endcsname{224}
\expandafter\def\csname ev@B-R/8B/g2/s3/retention-g1/cand-passes\endcsname{96}
\expandafter\def\csname ev@B-R/8B/g2/s3/retention-g1/cand-pct\endcsname{42.86}
\expandafter\def\csname ev@B-R/8B/g2/s3/retention-g1/cand-scheduled\endcsname{224}
\expandafter\def\csname ev@B-R/8B/g2/s3/retention-g1/cand-uncounted\endcsname{0}
\expandafter\def\csname ev@B-R/8B/g2/s3/retention-g1/diff-pp\endcsname{+4.46}
\expandafter\def\csname ev@B-R/8B/g2/s3/retention-g1/losses\endcsname{20}
\expandafter\def\csname ev@B-R/8B/g2/s3/retention-g1/outcome\endcsname{inconclusive}
\expandafter\def\csname ev@B-R/8B/g2/s3/retention-g1/p-recorded\endcsname{0.101}
\expandafter\def\csname ev@B-R/8B/g2/s3/retention-g1/protocol\endcsname{B-R}
\expandafter\def\csname ev@B-R/8B/g2/s3/retention-g1/task-losses-high\endcsname{20}
\expandafter\def\csname ev@B-R/8B/g2/s3/retention-g1/task-losses-low\endcsname{0}
\expandafter\def\csname ev@B-R/8B/g2/s3/retention-g1/task-p-high\endcsname{0.994}
\expandafter\def\csname ev@B-R/8B/g2/s3/retention-g1/task-p-low\endcsname{\ensuremath{1.1\times 10^{-5}}}
\expandafter\def\csname ev@B-R/8B/g2/s3/retention-g1/task-wins-high\endcsname{30}
\expandafter\def\csname ev@B-R/8B/g2/s3/retention-g1/task-wins-low\endcsname{3}
\expandafter\def\csname ev@B-R/8B/g2/s3/retention-g1/tasks\endcsname{56}
\expandafter\def\csname ev@B-R/8B/g2/s3/retention-g1/uncounted\endcsname{0}
\expandafter\def\csname ev@B-R/8B/g2/s3/retention-g1/uncounted-pct\endcsname{0.00}
\expandafter\def\csname ev@B-R/8B/g2/s3/retention-g1/unit\endcsname{attempt slot}
\expandafter\def\csname ev@B-R/8B/g2/s3/retention-g1/wins\endcsname{30}
\expandafter\def\csname ev@B-R/8B/g2/s3/target/attempt-ci-hi-pp\endcsname{+14.41}
\expandafter\def\csname ev@B-R/8B/g2/s3/target/attempt-ci-lo-pp\endcsname{-2.82}
\expandafter\def\csname ev@B-R/8B/g2/s3/target/base-counted\endcsname{188}
\expandafter\def\csname ev@B-R/8B/g2/s3/target/base-passes\endcsname{40}
\expandafter\def\csname ev@B-R/8B/g2/s3/target/base-pct\endcsname{21.28}
\expandafter\def\csname ev@B-R/8B/g2/s3/target/base-scheduled\endcsname{188}
\expandafter\def\csname ev@B-R/8B/g2/s3/target/base-uncounted\endcsname{0}
\expandafter\def\csname ev@B-R/8B/g2/s3/target/cand-counted\endcsname{188}
\expandafter\def\csname ev@B-R/8B/g2/s3/target/cand-passes\endcsname{51}
\expandafter\def\csname ev@B-R/8B/g2/s3/target/cand-pct\endcsname{27.13}
\expandafter\def\csname ev@B-R/8B/g2/s3/target/cand-scheduled\endcsname{188}
\expandafter\def\csname ev@B-R/8B/g2/s3/target/cand-uncounted\endcsname{0}
\expandafter\def\csname ev@B-R/8B/g2/s3/target/diff-pp\endcsname{+5.85}
\expandafter\def\csname ev@B-R/8B/g2/s3/target/losses\endcsname{11}
\expandafter\def\csname ev@B-R/8B/g2/s3/target/outcome\endcsname{inconclusive}
\expandafter\def\csname ev@B-R/8B/g2/s3/target/p-recorded\endcsname{0.04}
\expandafter\def\csname ev@B-R/8B/g2/s3/target/protocol\endcsname{B-R}
\expandafter\def\csname ev@B-R/8B/g2/s3/target/task-losses-high\endcsname{11}
\expandafter\def\csname ev@B-R/8B/g2/s3/target/task-losses-low\endcsname{0}
\expandafter\def\csname ev@B-R/8B/g2/s3/target/task-p-high\endcsname{0.928}
\expandafter\def\csname ev@B-R/8B/g2/s3/target/task-p-low\endcsname{\ensuremath{7.8\times 10^{-5}}}
\expandafter\def\csname ev@B-R/8B/g2/s3/target/task-wins-high\endcsname{22}
\expandafter\def\csname ev@B-R/8B/g2/s3/target/task-wins-low\endcsname{3}
\expandafter\def\csname ev@B-R/8B/g2/s3/target/tasks\endcsname{47}
\expandafter\def\csname ev@B-R/8B/g2/s3/target/uncounted\endcsname{0}
\expandafter\def\csname ev@B-R/8B/g2/s3/target/uncounted-pct\endcsname{0.00}
\expandafter\def\csname ev@B-R/8B/g2/s3/target/unit\endcsname{attempt slot}
\expandafter\def\csname ev@B-R/8B/g2/s3/target/wins\endcsname{22}
\expandafter\def\csname ev@B-R/8B/g2/s4/exam/attempt-ci-hi-pp\endcsname{+4.56}
\expandafter\def\csname ev@B-R/8B/g2/s4/exam/attempt-ci-lo-pp\endcsname{-3.04}
\expandafter\def\csname ev@B-R/8B/g2/s4/exam/base-counted\endcsname{400}
\expandafter\def\csname ev@B-R/8B/g2/s4/exam/base-passes\endcsname{30}
\expandafter\def\csname ev@B-R/8B/g2/s4/exam/base-pct\endcsname{7.50}
\expandafter\def\csname ev@B-R/8B/g2/s4/exam/base-scheduled\endcsname{400}
\expandafter\def\csname ev@B-R/8B/g2/s4/exam/base-uncounted\endcsname{0}
\expandafter\def\csname ev@B-R/8B/g2/s4/exam/cand-counted\endcsname{400}
\expandafter\def\csname ev@B-R/8B/g2/s4/exam/cand-passes\endcsname{33}
\expandafter\def\csname ev@B-R/8B/g2/s4/exam/cand-pct\endcsname{8.25}
\expandafter\def\csname ev@B-R/8B/g2/s4/exam/cand-scheduled\endcsname{400}
\expandafter\def\csname ev@B-R/8B/g2/s4/exam/cand-uncounted\endcsname{0}
\expandafter\def\csname ev@B-R/8B/g2/s4/exam/diff-pp\endcsname{+0.75}
\expandafter\def\csname ev@B-R/8B/g2/s4/exam/losses\endcsname{16}
\expandafter\def\csname ev@B-R/8B/g2/s4/exam/outcome\endcsname{inconclusive}
\expandafter\def\csname ev@B-R/8B/g2/s4/exam/p-recorded\endcsname{0.368}
\expandafter\def\csname ev@B-R/8B/g2/s4/exam/protocol\endcsname{B-R}
\expandafter\def\csname ev@B-R/8B/g2/s4/exam/task-losses-high\endcsname{16}
\expandafter\def\csname ev@B-R/8B/g2/s4/exam/task-losses-low\endcsname{0}
\expandafter\def\csname ev@B-R/8B/g2/s4/exam/task-p-high\endcsname{0.996}
\expandafter\def\csname ev@B-R/8B/g2/s4/exam/task-p-low\endcsname{0.0013}
\expandafter\def\csname ev@B-R/8B/g2/s4/exam/task-wins-high\endcsname{19}
\expandafter\def\csname ev@B-R/8B/g2/s4/exam/task-wins-low\endcsname{1}
\expandafter\def\csname ev@B-R/8B/g2/s4/exam/tasks\endcsname{100}
\expandafter\def\csname ev@B-R/8B/g2/s4/exam/uncounted\endcsname{0}
\expandafter\def\csname ev@B-R/8B/g2/s4/exam/uncounted-pct\endcsname{0.00}
\expandafter\def\csname ev@B-R/8B/g2/s4/exam/unit\endcsname{attempt slot}
\expandafter\def\csname ev@B-R/8B/g2/s4/exam/wins\endcsname{19}
\expandafter\def\csname ev@B-R/8B/g2/s4/retention-g1/attempt-ci-hi-pp\endcsname{+13.43}
\expandafter\def\csname ev@B-R/8B/g2/s4/retention-g1/attempt-ci-lo-pp\endcsname{-4.60}
\expandafter\def\csname ev@B-R/8B/g2/s4/retention-g1/base-counted\endcsname{224}
\expandafter\def\csname ev@B-R/8B/g2/s4/retention-g1/base-passes\endcsname{86}
\expandafter\def\csname ev@B-R/8B/g2/s4/retention-g1/base-pct\endcsname{38.39}
\expandafter\def\csname ev@B-R/8B/g2/s4/retention-g1/base-scheduled\endcsname{224}
\expandafter\def\csname ev@B-R/8B/g2/s4/retention-g1/base-uncounted\endcsname{0}
\expandafter\def\csname ev@B-R/8B/g2/s4/retention-g1/cand-counted\endcsname{224}
\expandafter\def\csname ev@B-R/8B/g2/s4/retention-g1/cand-passes\endcsname{96}
\expandafter\def\csname ev@B-R/8B/g2/s4/retention-g1/cand-pct\endcsname{42.86}
\expandafter\def\csname ev@B-R/8B/g2/s4/retention-g1/cand-scheduled\endcsname{224}
\expandafter\def\csname ev@B-R/8B/g2/s4/retention-g1/cand-uncounted\endcsname{0}
\expandafter\def\csname ev@B-R/8B/g2/s4/retention-g1/diff-pp\endcsname{+4.46}
\expandafter\def\csname ev@B-R/8B/g2/s4/retention-g1/losses\endcsname{18}
\expandafter\def\csname ev@B-R/8B/g2/s4/retention-g1/outcome\endcsname{inconclusive}
\expandafter\def\csname ev@B-R/8B/g2/s4/retention-g1/p-recorded\endcsname{0.092}
\expandafter\def\csname ev@B-R/8B/g2/s4/retention-g1/protocol\endcsname{B-R}
\expandafter\def\csname ev@B-R/8B/g2/s4/retention-g1/task-losses-high\endcsname{18}
\expandafter\def\csname ev@B-R/8B/g2/s4/retention-g1/task-losses-low\endcsname{0}
\expandafter\def\csname ev@B-R/8B/g2/s4/retention-g1/task-p-high\endcsname{0.993}
\expandafter\def\csname ev@B-R/8B/g2/s4/retention-g1/task-p-low\endcsname{\ensuremath{3.0\times 10^{-5}}}
\expandafter\def\csname ev@B-R/8B/g2/s4/retention-g1/task-wins-high\endcsname{28}
\expandafter\def\csname ev@B-R/8B/g2/s4/retention-g1/task-wins-low\endcsname{3}
\expandafter\def\csname ev@B-R/8B/g2/s4/retention-g1/tasks\endcsname{56}
\expandafter\def\csname ev@B-R/8B/g2/s4/retention-g1/uncounted\endcsname{0}
\expandafter\def\csname ev@B-R/8B/g2/s4/retention-g1/uncounted-pct\endcsname{0.00}
\expandafter\def\csname ev@B-R/8B/g2/s4/retention-g1/unit\endcsname{attempt slot}
\expandafter\def\csname ev@B-R/8B/g2/s4/retention-g1/wins\endcsname{28}
\expandafter\def\csname ev@B-R/8B/g2/s4/target/attempt-ci-hi-pp\endcsname{+13.30}
\expandafter\def\csname ev@B-R/8B/g2/s4/target/attempt-ci-lo-pp\endcsname{-3.81}
\expandafter\def\csname ev@B-R/8B/g2/s4/target/base-counted\endcsname{188}
\expandafter\def\csname ev@B-R/8B/g2/s4/target/base-passes\endcsname{40}
\expandafter\def\csname ev@B-R/8B/g2/s4/target/base-pct\endcsname{21.28}
\expandafter\def\csname ev@B-R/8B/g2/s4/target/base-scheduled\endcsname{188}
\expandafter\def\csname ev@B-R/8B/g2/s4/target/base-uncounted\endcsname{0}
\expandafter\def\csname ev@B-R/8B/g2/s4/target/cand-counted\endcsname{188}
\expandafter\def\csname ev@B-R/8B/g2/s4/target/cand-passes\endcsname{49}
\expandafter\def\csname ev@B-R/8B/g2/s4/target/cand-pct\endcsname{26.06}
\expandafter\def\csname ev@B-R/8B/g2/s4/target/cand-scheduled\endcsname{188}
\expandafter\def\csname ev@B-R/8B/g2/s4/target/cand-uncounted\endcsname{0}
\expandafter\def\csname ev@B-R/8B/g2/s4/target/diff-pp\endcsname{+4.79}
\expandafter\def\csname ev@B-R/8B/g2/s4/target/losses\endcsname{11}
\expandafter\def\csname ev@B-R/8B/g2/s4/target/outcome\endcsname{inconclusive}
\expandafter\def\csname ev@B-R/8B/g2/s4/target/p-recorded\endcsname{0.075}
\expandafter\def\csname ev@B-R/8B/g2/s4/target/protocol\endcsname{B-R}
\expandafter\def\csname ev@B-R/8B/g2/s4/target/task-losses-high\endcsname{11}
\expandafter\def\csname ev@B-R/8B/g2/s4/target/task-losses-low\endcsname{0}
\expandafter\def\csname ev@B-R/8B/g2/s4/target/task-p-high\endcsname{0.962}
\expandafter\def\csname ev@B-R/8B/g2/s4/target/task-p-low\endcsname{0.00024}
\expandafter\def\csname ev@B-R/8B/g2/s4/target/task-wins-high\endcsname{20}
\expandafter\def\csname ev@B-R/8B/g2/s4/target/task-wins-low\endcsname{3}
\expandafter\def\csname ev@B-R/8B/g2/s4/target/tasks\endcsname{47}
\expandafter\def\csname ev@B-R/8B/g2/s4/target/uncounted\endcsname{0}
\expandafter\def\csname ev@B-R/8B/g2/s4/target/uncounted-pct\endcsname{0.00}
\expandafter\def\csname ev@B-R/8B/g2/s4/target/unit\endcsname{attempt slot}
\expandafter\def\csname ev@B-R/8B/g2/s4/target/wins\endcsname{20}
\expandafter\def\csname ev@B-R/8B/g2/unfinished-seeds\endcsname{0}
\expandafter\def\csname ev@B-R/8B/g3/pooled/exam/base-counted\endcsname{400}
\expandafter\def\csname ev@B-R/8B/g3/pooled/exam/base-passes\endcsname{32}
\expandafter\def\csname ev@B-R/8B/g3/pooled/exam/base-pct\endcsname{8.00}
\expandafter\def\csname ev@B-R/8B/g3/pooled/exam/base-scheduled\endcsname{400}
\expandafter\def\csname ev@B-R/8B/g3/pooled/exam/base-uncounted\endcsname{0}
\expandafter\def\csname ev@B-R/8B/g3/pooled/exam/cand-counted\endcsname{1200}
\expandafter\def\csname ev@B-R/8B/g3/pooled/exam/cand-passes\endcsname{105}
\expandafter\def\csname ev@B-R/8B/g3/pooled/exam/cand-pct\endcsname{8.75}
\expandafter\def\csname ev@B-R/8B/g3/pooled/exam/cand-scheduled\endcsname{1200}
\expandafter\def\csname ev@B-R/8B/g3/pooled/exam/cand-uncounted\endcsname{0}
\expandafter\def\csname ev@B-R/8B/g3/pooled/exam/diff-pp\endcsname{+0.75}
\expandafter\def\csname ev@B-R/8B/g3/pooled/exam/losses\endcsname{57}
\expandafter\def\csname ev@B-R/8B/g3/pooled/exam/outcome\endcsname{inconclusive}
\expandafter\def\csname ev@B-R/8B/g3/pooled/exam/p-recorded\endcsname{0.235}
\expandafter\def\csname ev@B-R/8B/g3/pooled/exam/protocol\endcsname{B-R}
\expandafter\def\csname ev@B-R/8B/g3/pooled/exam/seeds\endcsname{3}
\expandafter\def\csname ev@B-R/8B/g3/pooled/exam/seeds-negative\endcsname{0}
\expandafter\def\csname ev@B-R/8B/g3/pooled/exam/seeds-positive\endcsname{2}
\expandafter\def\csname ev@B-R/8B/g3/pooled/exam/tasks\endcsname{100}
\expandafter\def\csname ev@B-R/8B/g3/pooled/exam/uncounted\endcsname{0}
\expandafter\def\csname ev@B-R/8B/g3/pooled/exam/uncounted-pct\endcsname{0.00}
\expandafter\def\csname ev@B-R/8B/g3/pooled/exam/unit\endcsname{attempt slot, summed over training seeds}
\expandafter\def\csname ev@B-R/8B/g3/pooled/exam/wins\endcsname{66}
\expandafter\def\csname ev@B-R/8B/g3/pooled/retention-g1/base-counted\endcsname{224}
\expandafter\def\csname ev@B-R/8B/g3/pooled/retention-g1/base-passes\endcsname{105}
\expandafter\def\csname ev@B-R/8B/g3/pooled/retention-g1/base-pct\endcsname{46.88}
\expandafter\def\csname ev@B-R/8B/g3/pooled/retention-g1/base-scheduled\endcsname{224}
\expandafter\def\csname ev@B-R/8B/g3/pooled/retention-g1/base-uncounted\endcsname{0}
\expandafter\def\csname ev@B-R/8B/g3/pooled/retention-g1/cand-counted\endcsname{672}
\expandafter\def\csname ev@B-R/8B/g3/pooled/retention-g1/cand-passes\endcsname{319}
\expandafter\def\csname ev@B-R/8B/g3/pooled/retention-g1/cand-pct\endcsname{47.47}
\expandafter\def\csname ev@B-R/8B/g3/pooled/retention-g1/cand-scheduled\endcsname{672}
\expandafter\def\csname ev@B-R/8B/g3/pooled/retention-g1/cand-uncounted\endcsname{0}
\expandafter\def\csname ev@B-R/8B/g3/pooled/retention-g1/diff-pp\endcsname{+0.60}
\expandafter\def\csname ev@B-R/8B/g3/pooled/retention-g1/losses\endcsname{48}
\expandafter\def\csname ev@B-R/8B/g3/pooled/retention-g1/outcome\endcsname{inconclusive}
\expandafter\def\csname ev@B-R/8B/g3/pooled/retention-g1/p-recorded\endcsname{0.382}
\expandafter\def\csname ev@B-R/8B/g3/pooled/retention-g1/protocol\endcsname{B-R}
\expandafter\def\csname ev@B-R/8B/g3/pooled/retention-g1/seeds\endcsname{3}
\expandafter\def\csname ev@B-R/8B/g3/pooled/retention-g1/seeds-negative\endcsname{1}
\expandafter\def\csname ev@B-R/8B/g3/pooled/retention-g1/seeds-positive\endcsname{2}
\expandafter\def\csname ev@B-R/8B/g3/pooled/retention-g1/tasks\endcsname{56}
\expandafter\def\csname ev@B-R/8B/g3/pooled/retention-g1/uncounted\endcsname{0}
\expandafter\def\csname ev@B-R/8B/g3/pooled/retention-g1/uncounted-pct\endcsname{0.00}
\expandafter\def\csname ev@B-R/8B/g3/pooled/retention-g1/unit\endcsname{attempt slot, summed over training seeds}
\expandafter\def\csname ev@B-R/8B/g3/pooled/retention-g1/wins\endcsname{52}
\expandafter\def\csname ev@B-R/8B/g3/pooled/retention-g2/base-counted\endcsname{188}
\expandafter\def\csname ev@B-R/8B/g3/pooled/retention-g2/base-passes\endcsname{54}
\expandafter\def\csname ev@B-R/8B/g3/pooled/retention-g2/base-pct\endcsname{28.72}
\expandafter\def\csname ev@B-R/8B/g3/pooled/retention-g2/base-scheduled\endcsname{188}
\expandafter\def\csname ev@B-R/8B/g3/pooled/retention-g2/base-uncounted\endcsname{0}
\expandafter\def\csname ev@B-R/8B/g3/pooled/retention-g2/cand-counted\endcsname{564}
\expandafter\def\csname ev@B-R/8B/g3/pooled/retention-g2/cand-passes\endcsname{164}
\expandafter\def\csname ev@B-R/8B/g3/pooled/retention-g2/cand-pct\endcsname{29.08}
\expandafter\def\csname ev@B-R/8B/g3/pooled/retention-g2/cand-scheduled\endcsname{564}
\expandafter\def\csname ev@B-R/8B/g3/pooled/retention-g2/cand-uncounted\endcsname{0}
\expandafter\def\csname ev@B-R/8B/g3/pooled/retention-g2/diff-pp\endcsname{+0.35}
\expandafter\def\csname ev@B-R/8B/g3/pooled/retention-g2/losses\endcsname{59}
\expandafter\def\csname ev@B-R/8B/g3/pooled/retention-g2/outcome\endcsname{inconclusive}
\expandafter\def\csname ev@B-R/8B/g3/pooled/retention-g2/p-recorded\endcsname{0.464}
\expandafter\def\csname ev@B-R/8B/g3/pooled/retention-g2/protocol\endcsname{B-R}
\expandafter\def\csname ev@B-R/8B/g3/pooled/retention-g2/seeds\endcsname{3}
\expandafter\def\csname ev@B-R/8B/g3/pooled/retention-g2/seeds-negative\endcsname{1}
\expandafter\def\csname ev@B-R/8B/g3/pooled/retention-g2/seeds-positive\endcsname{2}
\expandafter\def\csname ev@B-R/8B/g3/pooled/retention-g2/tasks\endcsname{47}
\expandafter\def\csname ev@B-R/8B/g3/pooled/retention-g2/uncounted\endcsname{0}
\expandafter\def\csname ev@B-R/8B/g3/pooled/retention-g2/uncounted-pct\endcsname{0.00}
\expandafter\def\csname ev@B-R/8B/g3/pooled/retention-g2/unit\endcsname{attempt slot, summed over training seeds}
\expandafter\def\csname ev@B-R/8B/g3/pooled/retention-g2/wins\endcsname{61}
\expandafter\def\csname ev@B-R/8B/g3/pooled/target/base-counted\endcsname{196}
\expandafter\def\csname ev@B-R/8B/g3/pooled/target/base-passes\endcsname{45}
\expandafter\def\csname ev@B-R/8B/g3/pooled/target/base-pct\endcsname{22.96}
\expandafter\def\csname ev@B-R/8B/g3/pooled/target/base-scheduled\endcsname{196}
\expandafter\def\csname ev@B-R/8B/g3/pooled/target/base-uncounted\endcsname{0}
\expandafter\def\csname ev@B-R/8B/g3/pooled/target/cand-counted\endcsname{588}
\expandafter\def\csname ev@B-R/8B/g3/pooled/target/cand-passes\endcsname{164}
\expandafter\def\csname ev@B-R/8B/g3/pooled/target/cand-pct\endcsname{27.89}
\expandafter\def\csname ev@B-R/8B/g3/pooled/target/cand-scheduled\endcsname{588}
\expandafter\def\csname ev@B-R/8B/g3/pooled/target/cand-uncounted\endcsname{0}
\expandafter\def\csname ev@B-R/8B/g3/pooled/target/diff-pp\endcsname{+4.93}
\expandafter\def\csname ev@B-R/8B/g3/pooled/target/losses\endcsname{35}
\expandafter\def\csname ev@B-R/8B/g3/pooled/target/outcome\endcsname{inconclusive}
\expandafter\def\csname ev@B-R/8B/g3/pooled/target/p-recorded\endcsname{0.0023}
\expandafter\def\csname ev@B-R/8B/g3/pooled/target/protocol\endcsname{B-R}
\expandafter\def\csname ev@B-R/8B/g3/pooled/target/seeds\endcsname{3}
\expandafter\def\csname ev@B-R/8B/g3/pooled/target/seeds-negative\endcsname{0}
\expandafter\def\csname ev@B-R/8B/g3/pooled/target/seeds-positive\endcsname{3}
\expandafter\def\csname ev@B-R/8B/g3/pooled/target/tasks\endcsname{49}
\expandafter\def\csname ev@B-R/8B/g3/pooled/target/uncounted\endcsname{0}
\expandafter\def\csname ev@B-R/8B/g3/pooled/target/uncounted-pct\endcsname{0.00}
\expandafter\def\csname ev@B-R/8B/g3/pooled/target/unit\endcsname{attempt slot, summed over training seeds}
\expandafter\def\csname ev@B-R/8B/g3/pooled/target/wins\endcsname{64}
\expandafter\def\csname ev@B-R/8B/g3/s1/exam/attempt-ci-hi-pp\endcsname{+3.83}
\expandafter\def\csname ev@B-R/8B/g3/s1/exam/attempt-ci-lo-pp\endcsname{-3.83}
\expandafter\def\csname ev@B-R/8B/g3/s1/exam/base-counted\endcsname{400}
\expandafter\def\csname ev@B-R/8B/g3/s1/exam/base-passes\endcsname{32}
\expandafter\def\csname ev@B-R/8B/g3/s1/exam/base-pct\endcsname{8.00}
\expandafter\def\csname ev@B-R/8B/g3/s1/exam/base-scheduled\endcsname{400}
\expandafter\def\csname ev@B-R/8B/g3/s1/exam/base-uncounted\endcsname{0}
\expandafter\def\csname ev@B-R/8B/g3/s1/exam/cand-counted\endcsname{400}
\expandafter\def\csname ev@B-R/8B/g3/s1/exam/cand-passes\endcsname{32}
\expandafter\def\csname ev@B-R/8B/g3/s1/exam/cand-pct\endcsname{8.00}
\expandafter\def\csname ev@B-R/8B/g3/s1/exam/cand-scheduled\endcsname{400}
\expandafter\def\csname ev@B-R/8B/g3/s1/exam/cand-uncounted\endcsname{0}
\expandafter\def\csname ev@B-R/8B/g3/s1/exam/diff-pp\endcsname{+0.00}
\expandafter\def\csname ev@B-R/8B/g3/s1/exam/losses\endcsname{18}
\expandafter\def\csname ev@B-R/8B/g3/s1/exam/outcome\endcsname{inconclusive}
\expandafter\def\csname ev@B-R/8B/g3/s1/exam/p-recorded\endcsname{0.566}
\expandafter\def\csname ev@B-R/8B/g3/s1/exam/protocol\endcsname{B-R}
\expandafter\def\csname ev@B-R/8B/g3/s1/exam/task-losses-high\endcsname{18}
\expandafter\def\csname ev@B-R/8B/g3/s1/exam/task-losses-low\endcsname{0}
\expandafter\def\csname ev@B-R/8B/g3/s1/exam/task-p-high\endcsname{1}
\expandafter\def\csname ev@B-R/8B/g3/s1/exam/task-p-low\endcsname{0.0053}
\expandafter\def\csname ev@B-R/8B/g3/s1/exam/task-wins-high\endcsname{18}
\expandafter\def\csname ev@B-R/8B/g3/s1/exam/task-wins-low\endcsname{0}
\expandafter\def\csname ev@B-R/8B/g3/s1/exam/tasks\endcsname{100}
\expandafter\def\csname ev@B-R/8B/g3/s1/exam/uncounted\endcsname{0}
\expandafter\def\csname ev@B-R/8B/g3/s1/exam/uncounted-pct\endcsname{0.00}
\expandafter\def\csname ev@B-R/8B/g3/s1/exam/unit\endcsname{attempt slot}
\expandafter\def\csname ev@B-R/8B/g3/s1/exam/wins\endcsname{18}
\expandafter\def\csname ev@B-R/8B/g3/s1/retention-g1/attempt-ci-hi-pp\endcsname{+10.94}
\expandafter\def\csname ev@B-R/8B/g3/s1/retention-g1/attempt-ci-lo-pp\endcsname{-7.41}
\expandafter\def\csname ev@B-R/8B/g3/s1/retention-g1/base-counted\endcsname{224}
\expandafter\def\csname ev@B-R/8B/g3/s1/retention-g1/base-passes\endcsname{105}
\expandafter\def\csname ev@B-R/8B/g3/s1/retention-g1/base-pct\endcsname{46.88}
\expandafter\def\csname ev@B-R/8B/g3/s1/retention-g1/base-scheduled\endcsname{224}
\expandafter\def\csname ev@B-R/8B/g3/s1/retention-g1/base-uncounted\endcsname{0}
\expandafter\def\csname ev@B-R/8B/g3/s1/retention-g1/cand-counted\endcsname{224}
\expandafter\def\csname ev@B-R/8B/g3/s1/retention-g1/cand-passes\endcsname{109}
\expandafter\def\csname ev@B-R/8B/g3/s1/retention-g1/cand-pct\endcsname{48.66}
\expandafter\def\csname ev@B-R/8B/g3/s1/retention-g1/cand-scheduled\endcsname{224}
\expandafter\def\csname ev@B-R/8B/g3/s1/retention-g1/cand-uncounted\endcsname{0}
\expandafter\def\csname ev@B-R/8B/g3/s1/retention-g1/diff-pp\endcsname{+1.79}
\expandafter\def\csname ev@B-R/8B/g3/s1/retention-g1/losses\endcsname{14}
\expandafter\def\csname ev@B-R/8B/g3/s1/retention-g1/outcome\endcsname{inconclusive}
\expandafter\def\csname ev@B-R/8B/g3/s1/retention-g1/p-recorded\endcsname{0.298}
\expandafter\def\csname ev@B-R/8B/g3/s1/retention-g1/protocol\endcsname{B-R}
\expandafter\def\csname ev@B-R/8B/g3/s1/retention-g1/task-losses-high\endcsname{14}
\expandafter\def\csname ev@B-R/8B/g3/s1/retention-g1/task-losses-low\endcsname{0}
\expandafter\def\csname ev@B-R/8B/g3/s1/retention-g1/task-p-high\endcsname{0.99}
\expandafter\def\csname ev@B-R/8B/g3/s1/retention-g1/task-p-low\endcsname{0.0022}
\expandafter\def\csname ev@B-R/8B/g3/s1/retention-g1/task-wins-high\endcsname{18}
\expandafter\def\csname ev@B-R/8B/g3/s1/retention-g1/task-wins-low\endcsname{1}
\expandafter\def\csname ev@B-R/8B/g3/s1/retention-g1/tasks\endcsname{56}
\expandafter\def\csname ev@B-R/8B/g3/s1/retention-g1/uncounted\endcsname{0}
\expandafter\def\csname ev@B-R/8B/g3/s1/retention-g1/uncounted-pct\endcsname{0.00}
\expandafter\def\csname ev@B-R/8B/g3/s1/retention-g1/unit\endcsname{attempt slot}
\expandafter\def\csname ev@B-R/8B/g3/s1/retention-g1/wins\endcsname{18}
\expandafter\def\csname ev@B-R/8B/g3/s1/retention-g2/attempt-ci-hi-pp\endcsname{+6.34}
\expandafter\def\csname ev@B-R/8B/g3/s1/retention-g2/attempt-ci-lo-pp\endcsname{-11.60}
\expandafter\def\csname ev@B-R/8B/g3/s1/retention-g2/base-counted\endcsname{188}
\expandafter\def\csname ev@B-R/8B/g3/s1/retention-g2/base-passes\endcsname{54}
\expandafter\def\csname ev@B-R/8B/g3/s1/retention-g2/base-pct\endcsname{28.72}
\expandafter\def\csname ev@B-R/8B/g3/s1/retention-g2/base-scheduled\endcsname{188}
\expandafter\def\csname ev@B-R/8B/g3/s1/retention-g2/base-uncounted\endcsname{0}
\expandafter\def\csname ev@B-R/8B/g3/s1/retention-g2/cand-counted\endcsname{188}
\expandafter\def\csname ev@B-R/8B/g3/s1/retention-g2/cand-passes\endcsname{49}
\expandafter\def\csname ev@B-R/8B/g3/s1/retention-g2/cand-pct\endcsname{26.06}
\expandafter\def\csname ev@B-R/8B/g3/s1/retention-g2/cand-scheduled\endcsname{188}
\expandafter\def\csname ev@B-R/8B/g3/s1/retention-g2/cand-uncounted\endcsname{0}
\expandafter\def\csname ev@B-R/8B/g3/s1/retention-g2/diff-pp\endcsname{-2.66}
\expandafter\def\csname ev@B-R/8B/g3/s1/retention-g2/losses\endcsname{19}
\expandafter\def\csname ev@B-R/8B/g3/s1/retention-g2/outcome\endcsname{inconclusive}
\expandafter\def\csname ev@B-R/8B/g3/s1/retention-g2/p-recorded\endcsname{0.852}
\expandafter\def\csname ev@B-R/8B/g3/s1/retention-g2/protocol\endcsname{B-R}
\expandafter\def\csname ev@B-R/8B/g3/s1/retention-g2/task-losses-high\endcsname{19}
\expandafter\def\csname ev@B-R/8B/g3/s1/retention-g2/task-losses-low\endcsname{2}
\expandafter\def\csname ev@B-R/8B/g3/s1/retention-g2/task-p-high\endcsname{1}
\expandafter\def\csname ev@B-R/8B/g3/s1/retention-g2/task-p-low\endcsname{0.032}
\expandafter\def\csname ev@B-R/8B/g3/s1/retention-g2/task-wins-high\endcsname{14}
\expandafter\def\csname ev@B-R/8B/g3/s1/retention-g2/task-wins-low\endcsname{0}
\expandafter\def\csname ev@B-R/8B/g3/s1/retention-g2/tasks\endcsname{47}
\expandafter\def\csname ev@B-R/8B/g3/s1/retention-g2/uncounted\endcsname{0}
\expandafter\def\csname ev@B-R/8B/g3/s1/retention-g2/uncounted-pct\endcsname{0.00}
\expandafter\def\csname ev@B-R/8B/g3/s1/retention-g2/unit\endcsname{attempt slot}
\expandafter\def\csname ev@B-R/8B/g3/s1/retention-g2/wins\endcsname{14}
\expandafter\def\csname ev@B-R/8B/g3/s1/target/attempt-ci-hi-pp\endcsname{+15.75}
\expandafter\def\csname ev@B-R/8B/g3/s1/target/attempt-ci-lo-pp\endcsname{-1.60}
\expandafter\def\csname ev@B-R/8B/g3/s1/target/base-counted\endcsname{196}
\expandafter\def\csname ev@B-R/8B/g3/s1/target/base-passes\endcsname{45}
\expandafter\def\csname ev@B-R/8B/g3/s1/target/base-pct\endcsname{22.96}
\expandafter\def\csname ev@B-R/8B/g3/s1/target/base-scheduled\endcsname{196}
\expandafter\def\csname ev@B-R/8B/g3/s1/target/base-uncounted\endcsname{0}
\expandafter\def\csname ev@B-R/8B/g3/s1/target/cand-counted\endcsname{196}
\expandafter\def\csname ev@B-R/8B/g3/s1/target/cand-passes\endcsname{59}
\expandafter\def\csname ev@B-R/8B/g3/s1/target/cand-pct\endcsname{30.10}
\expandafter\def\csname ev@B-R/8B/g3/s1/target/cand-scheduled\endcsname{196}
\expandafter\def\csname ev@B-R/8B/g3/s1/target/cand-uncounted\endcsname{0}
\expandafter\def\csname ev@B-R/8B/g3/s1/target/diff-pp\endcsname{+7.14}
\expandafter\def\csname ev@B-R/8B/g3/s1/target/losses\endcsname{11}
\expandafter\def\csname ev@B-R/8B/g3/s1/target/outcome\endcsname{inconclusive}
\expandafter\def\csname ev@B-R/8B/g3/s1/target/p-recorded\endcsname{0.014}
\expandafter\def\csname ev@B-R/8B/g3/s1/target/protocol\endcsname{B-R}
\expandafter\def\csname ev@B-R/8B/g3/s1/target/task-losses-high\endcsname{11}
\expandafter\def\csname ev@B-R/8B/g3/s1/target/task-losses-low\endcsname{0}
\expandafter\def\csname ev@B-R/8B/g3/s1/target/task-p-high\endcsname{0.895}
\expandafter\def\csname ev@B-R/8B/g3/s1/target/task-p-low\endcsname{\ensuremath{1.4\times 10^{-5}}}
\expandafter\def\csname ev@B-R/8B/g3/s1/target/task-wins-high\endcsname{25}
\expandafter\def\csname ev@B-R/8B/g3/s1/target/task-wins-low\endcsname{4}
\expandafter\def\csname ev@B-R/8B/g3/s1/target/tasks\endcsname{49}
\expandafter\def\csname ev@B-R/8B/g3/s1/target/uncounted\endcsname{0}
\expandafter\def\csname ev@B-R/8B/g3/s1/target/uncounted-pct\endcsname{0.00}
\expandafter\def\csname ev@B-R/8B/g3/s1/target/unit\endcsname{attempt slot}
\expandafter\def\csname ev@B-R/8B/g3/s1/target/wins\endcsname{25}
\expandafter\def\csname ev@B-R/8B/g3/s2/exam/attempt-ci-hi-pp\endcsname{+5.21}
\expandafter\def\csname ev@B-R/8B/g3/s2/exam/attempt-ci-lo-pp\endcsname{-2.69}
\expandafter\def\csname ev@B-R/8B/g3/s2/exam/base-counted\endcsname{400}
\expandafter\def\csname ev@B-R/8B/g3/s2/exam/base-passes\endcsname{32}
\expandafter\def\csname ev@B-R/8B/g3/s2/exam/base-pct\endcsname{8.00}
\expandafter\def\csname ev@B-R/8B/g3/s2/exam/base-scheduled\endcsname{400}
\expandafter\def\csname ev@B-R/8B/g3/s2/exam/base-uncounted\endcsname{0}
\expandafter\def\csname ev@B-R/8B/g3/s2/exam/cand-counted\endcsname{400}
\expandafter\def\csname ev@B-R/8B/g3/s2/exam/cand-passes\endcsname{37}
\expandafter\def\csname ev@B-R/8B/g3/s2/exam/cand-pct\endcsname{9.25}
\expandafter\def\csname ev@B-R/8B/g3/s2/exam/cand-scheduled\endcsname{400}
\expandafter\def\csname ev@B-R/8B/g3/s2/exam/cand-uncounted\endcsname{0}
\expandafter\def\csname ev@B-R/8B/g3/s2/exam/diff-pp\endcsname{+1.25}
\expandafter\def\csname ev@B-R/8B/g3/s2/exam/losses\endcsname{18}
\expandafter\def\csname ev@B-R/8B/g3/s2/exam/outcome\endcsname{inconclusive}
\expandafter\def\csname ev@B-R/8B/g3/s2/exam/p-recorded\endcsname{0.266}
\expandafter\def\csname ev@B-R/8B/g3/s2/exam/protocol\endcsname{B-R}
\expandafter\def\csname ev@B-R/8B/g3/s2/exam/task-losses-high\endcsname{18}
\expandafter\def\csname ev@B-R/8B/g3/s2/exam/task-losses-low\endcsname{0}
\expandafter\def\csname ev@B-R/8B/g3/s2/exam/task-p-high\endcsname{0.997}
\expandafter\def\csname ev@B-R/8B/g3/s2/exam/task-p-low\endcsname{0.00046}
\expandafter\def\csname ev@B-R/8B/g3/s2/exam/task-wins-high\endcsname{23}
\expandafter\def\csname ev@B-R/8B/g3/s2/exam/task-wins-low\endcsname{2}
\expandafter\def\csname ev@B-R/8B/g3/s2/exam/tasks\endcsname{100}
\expandafter\def\csname ev@B-R/8B/g3/s2/exam/uncounted\endcsname{0}
\expandafter\def\csname ev@B-R/8B/g3/s2/exam/uncounted-pct\endcsname{0.00}
\expandafter\def\csname ev@B-R/8B/g3/s2/exam/unit\endcsname{attempt slot}
\expandafter\def\csname ev@B-R/8B/g3/s2/exam/wins\endcsname{23}
\expandafter\def\csname ev@B-R/8B/g3/s2/retention-g1/attempt-ci-hi-pp\endcsname{+9.61}
\expandafter\def\csname ev@B-R/8B/g3/s2/retention-g1/attempt-ci-lo-pp\endcsname{-8.73}
\expandafter\def\csname ev@B-R/8B/g3/s2/retention-g1/base-counted\endcsname{224}
\expandafter\def\csname ev@B-R/8B/g3/s2/retention-g1/base-passes\endcsname{105}
\expandafter\def\csname ev@B-R/8B/g3/s2/retention-g1/base-pct\endcsname{46.88}
\expandafter\def\csname ev@B-R/8B/g3/s2/retention-g1/base-scheduled\endcsname{224}
\expandafter\def\csname ev@B-R/8B/g3/s2/retention-g1/base-uncounted\endcsname{0}
\expandafter\def\csname ev@B-R/8B/g3/s2/retention-g1/cand-counted\endcsname{224}
\expandafter\def\csname ev@B-R/8B/g3/s2/retention-g1/cand-passes\endcsname{106}
\expandafter\def\csname ev@B-R/8B/g3/s2/retention-g1/cand-pct\endcsname{47.32}
\expandafter\def\csname ev@B-R/8B/g3/s2/retention-g1/cand-scheduled\endcsname{224}
\expandafter\def\csname ev@B-R/8B/g3/s2/retention-g1/cand-uncounted\endcsname{0}
\expandafter\def\csname ev@B-R/8B/g3/s2/retention-g1/diff-pp\endcsname{+0.45}
\expandafter\def\csname ev@B-R/8B/g3/s2/retention-g1/losses\endcsname{16}
\expandafter\def\csname ev@B-R/8B/g3/s2/retention-g1/outcome\endcsname{inconclusive}
\expandafter\def\csname ev@B-R/8B/g3/s2/retention-g1/p-recorded\endcsname{0.5}
\expandafter\def\csname ev@B-R/8B/g3/s2/retention-g1/protocol\endcsname{B-R}
\expandafter\def\csname ev@B-R/8B/g3/s2/retention-g1/task-losses-high\endcsname{16}
\expandafter\def\csname ev@B-R/8B/g3/s2/retention-g1/task-losses-low\endcsname{0}
\expandafter\def\csname ev@B-R/8B/g3/s2/retention-g1/task-p-high\endcsname{0.998}
\expandafter\def\csname ev@B-R/8B/g3/s2/retention-g1/task-p-low\endcsname{0.0036}
\expandafter\def\csname ev@B-R/8B/g3/s2/retention-g1/task-wins-high\endcsname{17}
\expandafter\def\csname ev@B-R/8B/g3/s2/retention-g1/task-wins-low\endcsname{1}
\expandafter\def\csname ev@B-R/8B/g3/s2/retention-g1/tasks\endcsname{56}
\expandafter\def\csname ev@B-R/8B/g3/s2/retention-g1/uncounted\endcsname{0}
\expandafter\def\csname ev@B-R/8B/g3/s2/retention-g1/uncounted-pct\endcsname{0.00}
\expandafter\def\csname ev@B-R/8B/g3/s2/retention-g1/unit\endcsname{attempt slot}
\expandafter\def\csname ev@B-R/8B/g3/s2/retention-g1/wins\endcsname{17}
\expandafter\def\csname ev@B-R/8B/g3/s2/retention-g2/attempt-ci-hi-pp\endcsname{+10.74}
\expandafter\def\csname ev@B-R/8B/g3/s2/retention-g2/attempt-ci-lo-pp\endcsname{-7.59}
\expandafter\def\csname ev@B-R/8B/g3/s2/retention-g2/base-counted\endcsname{188}
\expandafter\def\csname ev@B-R/8B/g3/s2/retention-g2/base-passes\endcsname{54}
\expandafter\def\csname ev@B-R/8B/g3/s2/retention-g2/base-pct\endcsname{28.72}
\expandafter\def\csname ev@B-R/8B/g3/s2/retention-g2/base-scheduled\endcsname{188}
\expandafter\def\csname ev@B-R/8B/g3/s2/retention-g2/base-uncounted\endcsname{0}
\expandafter\def\csname ev@B-R/8B/g3/s2/retention-g2/cand-counted\endcsname{188}
\expandafter\def\csname ev@B-R/8B/g3/s2/retention-g2/cand-passes\endcsname{57}
\expandafter\def\csname ev@B-R/8B/g3/s2/retention-g2/cand-pct\endcsname{30.32}
\expandafter\def\csname ev@B-R/8B/g3/s2/retention-g2/cand-scheduled\endcsname{188}
\expandafter\def\csname ev@B-R/8B/g3/s2/retention-g2/cand-uncounted\endcsname{0}
\expandafter\def\csname ev@B-R/8B/g3/s2/retention-g2/diff-pp\endcsname{+1.60}
\expandafter\def\csname ev@B-R/8B/g3/s2/retention-g2/losses\endcsname{20}
\expandafter\def\csname ev@B-R/8B/g3/s2/retention-g2/outcome\endcsname{inconclusive}
\expandafter\def\csname ev@B-R/8B/g3/s2/retention-g2/p-recorded\endcsname{0.38}
\expandafter\def\csname ev@B-R/8B/g3/s2/retention-g2/protocol\endcsname{B-R}
\expandafter\def\csname ev@B-R/8B/g3/s2/retention-g2/task-losses-high\endcsname{20}
\expandafter\def\csname ev@B-R/8B/g3/s2/retention-g2/task-losses-low\endcsname{0}
\expandafter\def\csname ev@B-R/8B/g3/s2/retention-g2/task-p-high\endcsname{0.999}
\expandafter\def\csname ev@B-R/8B/g3/s2/retention-g2/task-p-low\endcsname{0.00046}
\expandafter\def\csname ev@B-R/8B/g3/s2/retention-g2/task-wins-high\endcsname{23}
\expandafter\def\csname ev@B-R/8B/g3/s2/retention-g2/task-wins-low\endcsname{1}
\expandafter\def\csname ev@B-R/8B/g3/s2/retention-g2/tasks\endcsname{47}
\expandafter\def\csname ev@B-R/8B/g3/s2/retention-g2/uncounted\endcsname{0}
\expandafter\def\csname ev@B-R/8B/g3/s2/retention-g2/uncounted-pct\endcsname{0.00}
\expandafter\def\csname ev@B-R/8B/g3/s2/retention-g2/unit\endcsname{attempt slot}
\expandafter\def\csname ev@B-R/8B/g3/s2/retention-g2/wins\endcsname{23}
\expandafter\def\csname ev@B-R/8B/g3/s2/target/attempt-ci-hi-pp\endcsname{+11.52}
\expandafter\def\csname ev@B-R/8B/g3/s2/target/attempt-ci-lo-pp\endcsname{-5.45}
\expandafter\def\csname ev@B-R/8B/g3/s2/target/base-counted\endcsname{196}
\expandafter\def\csname ev@B-R/8B/g3/s2/target/base-passes\endcsname{45}
\expandafter\def\csname ev@B-R/8B/g3/s2/target/base-pct\endcsname{22.96}
\expandafter\def\csname ev@B-R/8B/g3/s2/target/base-scheduled\endcsname{196}
\expandafter\def\csname ev@B-R/8B/g3/s2/target/base-uncounted\endcsname{0}
\expandafter\def\csname ev@B-R/8B/g3/s2/target/cand-counted\endcsname{196}
\expandafter\def\csname ev@B-R/8B/g3/s2/target/cand-passes\endcsname{51}
\expandafter\def\csname ev@B-R/8B/g3/s2/target/cand-pct\endcsname{26.02}
\expandafter\def\csname ev@B-R/8B/g3/s2/target/cand-scheduled\endcsname{196}
\expandafter\def\csname ev@B-R/8B/g3/s2/target/cand-uncounted\endcsname{0}
\expandafter\def\csname ev@B-R/8B/g3/s2/target/diff-pp\endcsname{+3.06}
\expandafter\def\csname ev@B-R/8B/g3/s2/target/losses\endcsname{11}
\expandafter\def\csname ev@B-R/8B/g3/s2/target/outcome\endcsname{inconclusive}
\expandafter\def\csname ev@B-R/8B/g3/s2/target/p-recorded\endcsname{0.172}
\expandafter\def\csname ev@B-R/8B/g3/s2/target/protocol\endcsname{B-R}
\expandafter\def\csname ev@B-R/8B/g3/s2/target/task-losses-high\endcsname{11}
\expandafter\def\csname ev@B-R/8B/g3/s2/target/task-losses-low\endcsname{0}
\expandafter\def\csname ev@B-R/8B/g3/s2/target/task-p-high\endcsname{0.971}
\expandafter\def\csname ev@B-R/8B/g3/s2/target/task-p-low\endcsname{0.0013}
\expandafter\def\csname ev@B-R/8B/g3/s2/target/task-wins-high\endcsname{17}
\expandafter\def\csname ev@B-R/8B/g3/s2/target/task-wins-low\endcsname{2}
\expandafter\def\csname ev@B-R/8B/g3/s2/target/tasks\endcsname{49}
\expandafter\def\csname ev@B-R/8B/g3/s2/target/uncounted\endcsname{0}
\expandafter\def\csname ev@B-R/8B/g3/s2/target/uncounted-pct\endcsname{0.00}
\expandafter\def\csname ev@B-R/8B/g3/s2/target/unit\endcsname{attempt slot}
\expandafter\def\csname ev@B-R/8B/g3/s2/target/wins\endcsname{17}
\expandafter\def\csname ev@B-R/8B/g3/s3/exam/attempt-ci-hi-pp\endcsname{+4.93}
\expandafter\def\csname ev@B-R/8B/g3/s3/exam/attempt-ci-lo-pp\endcsname{-2.92}
\expandafter\def\csname ev@B-R/8B/g3/s3/exam/base-counted\endcsname{400}
\expandafter\def\csname ev@B-R/8B/g3/s3/exam/base-passes\endcsname{32}
\expandafter\def\csname ev@B-R/8B/g3/s3/exam/base-pct\endcsname{8.00}
\expandafter\def\csname ev@B-R/8B/g3/s3/exam/base-scheduled\endcsname{400}
\expandafter\def\csname ev@B-R/8B/g3/s3/exam/base-uncounted\endcsname{0}
\expandafter\def\csname ev@B-R/8B/g3/s3/exam/cand-counted\endcsname{400}
\expandafter\def\csname ev@B-R/8B/g3/s3/exam/cand-passes\endcsname{36}
\expandafter\def\csname ev@B-R/8B/g3/s3/exam/cand-pct\endcsname{9.00}
\expandafter\def\csname ev@B-R/8B/g3/s3/exam/cand-scheduled\endcsname{400}
\expandafter\def\csname ev@B-R/8B/g3/s3/exam/cand-uncounted\endcsname{0}
\expandafter\def\csname ev@B-R/8B/g3/s3/exam/diff-pp\endcsname{+1.00}
\expandafter\def\csname ev@B-R/8B/g3/s3/exam/losses\endcsname{21}
\expandafter\def\csname ev@B-R/8B/g3/s3/exam/outcome\endcsname{inconclusive}
\expandafter\def\csname ev@B-R/8B/g3/s3/exam/p-recorded\endcsname{0.329}
\expandafter\def\csname ev@B-R/8B/g3/s3/exam/protocol\endcsname{B-R}
\expandafter\def\csname ev@B-R/8B/g3/s3/exam/task-losses-high\endcsname{21}
\expandafter\def\csname ev@B-R/8B/g3/s3/exam/task-losses-low\endcsname{0}
\expandafter\def\csname ev@B-R/8B/g3/s3/exam/task-p-high\endcsname{0.999}
\expandafter\def\csname ev@B-R/8B/g3/s3/exam/task-p-low\endcsname{0.00027}
\expandafter\def\csname ev@B-R/8B/g3/s3/exam/task-wins-high\endcsname{25}
\expandafter\def\csname ev@B-R/8B/g3/s3/exam/task-wins-low\endcsname{1}
\expandafter\def\csname ev@B-R/8B/g3/s3/exam/tasks\endcsname{100}
\expandafter\def\csname ev@B-R/8B/g3/s3/exam/uncounted\endcsname{0}
\expandafter\def\csname ev@B-R/8B/g3/s3/exam/uncounted-pct\endcsname{0.00}
\expandafter\def\csname ev@B-R/8B/g3/s3/exam/unit\endcsname{attempt slot}
\expandafter\def\csname ev@B-R/8B/g3/s3/exam/wins\endcsname{25}
\expandafter\def\csname ev@B-R/8B/g3/s3/retention-g1/attempt-ci-hi-pp\endcsname{+8.72}
\expandafter\def\csname ev@B-R/8B/g3/s3/retention-g1/attempt-ci-lo-pp\endcsname{-9.60}
\expandafter\def\csname ev@B-R/8B/g3/s3/retention-g1/base-counted\endcsname{224}
\expandafter\def\csname ev@B-R/8B/g3/s3/retention-g1/base-passes\endcsname{105}
\expandafter\def\csname ev@B-R/8B/g3/s3/retention-g1/base-pct\endcsname{46.88}
\expandafter\def\csname ev@B-R/8B/g3/s3/retention-g1/base-scheduled\endcsname{224}
\expandafter\def\csname ev@B-R/8B/g3/s3/retention-g1/base-uncounted\endcsname{0}
\expandafter\def\csname ev@B-R/8B/g3/s3/retention-g1/cand-counted\endcsname{224}
\expandafter\def\csname ev@B-R/8B/g3/s3/retention-g1/cand-passes\endcsname{104}
\expandafter\def\csname ev@B-R/8B/g3/s3/retention-g1/cand-pct\endcsname{46.43}
\expandafter\def\csname ev@B-R/8B/g3/s3/retention-g1/cand-scheduled\endcsname{224}
\expandafter\def\csname ev@B-R/8B/g3/s3/retention-g1/cand-uncounted\endcsname{0}
\expandafter\def\csname ev@B-R/8B/g3/s3/retention-g1/diff-pp\endcsname{-0.45}
\expandafter\def\csname ev@B-R/8B/g3/s3/retention-g1/losses\endcsname{18}
\expandafter\def\csname ev@B-R/8B/g3/s3/retention-g1/outcome\endcsname{inconclusive}
\expandafter\def\csname ev@B-R/8B/g3/s3/retention-g1/p-recorded\endcsname{0.632}
\expandafter\def\csname ev@B-R/8B/g3/s3/retention-g1/protocol\endcsname{B-R}
\expandafter\def\csname ev@B-R/8B/g3/s3/retention-g1/task-losses-high\endcsname{18}
\expandafter\def\csname ev@B-R/8B/g3/s3/retention-g1/task-losses-low\endcsname{1}
\expandafter\def\csname ev@B-R/8B/g3/s3/retention-g1/task-p-high\endcsname{1}
\expandafter\def\csname ev@B-R/8B/g3/s3/retention-g1/task-p-low\endcsname{0.0085}
\expandafter\def\csname ev@B-R/8B/g3/s3/retention-g1/task-wins-high\endcsname{17}
\expandafter\def\csname ev@B-R/8B/g3/s3/retention-g1/task-wins-low\endcsname{0}
\expandafter\def\csname ev@B-R/8B/g3/s3/retention-g1/tasks\endcsname{56}
\expandafter\def\csname ev@B-R/8B/g3/s3/retention-g1/uncounted\endcsname{0}
\expandafter\def\csname ev@B-R/8B/g3/s3/retention-g1/uncounted-pct\endcsname{0.00}
\expandafter\def\csname ev@B-R/8B/g3/s3/retention-g1/unit\endcsname{attempt slot}
\expandafter\def\csname ev@B-R/8B/g3/s3/retention-g1/wins\endcsname{17}
\expandafter\def\csname ev@B-R/8B/g3/s3/retention-g2/attempt-ci-hi-pp\endcsname{+11.29}
\expandafter\def\csname ev@B-R/8B/g3/s3/retention-g2/attempt-ci-lo-pp\endcsname{-7.08}
\expandafter\def\csname ev@B-R/8B/g3/s3/retention-g2/base-counted\endcsname{188}
\expandafter\def\csname ev@B-R/8B/g3/s3/retention-g2/base-passes\endcsname{54}
\expandafter\def\csname ev@B-R/8B/g3/s3/retention-g2/base-pct\endcsname{28.72}
\expandafter\def\csname ev@B-R/8B/g3/s3/retention-g2/base-scheduled\endcsname{188}
\expandafter\def\csname ev@B-R/8B/g3/s3/retention-g2/base-uncounted\endcsname{0}
\expandafter\def\csname ev@B-R/8B/g3/s3/retention-g2/cand-counted\endcsname{188}
\expandafter\def\csname ev@B-R/8B/g3/s3/retention-g2/cand-passes\endcsname{58}
\expandafter\def\csname ev@B-R/8B/g3/s3/retention-g2/cand-pct\endcsname{30.85}
\expandafter\def\csname ev@B-R/8B/g3/s3/retention-g2/cand-scheduled\endcsname{188}
\expandafter\def\csname ev@B-R/8B/g3/s3/retention-g2/cand-uncounted\endcsname{0}
\expandafter\def\csname ev@B-R/8B/g3/s3/retention-g2/diff-pp\endcsname{+2.13}
\expandafter\def\csname ev@B-R/8B/g3/s3/retention-g2/losses\endcsname{20}
\expandafter\def\csname ev@B-R/8B/g3/s3/retention-g2/outcome\endcsname{inconclusive}
\expandafter\def\csname ev@B-R/8B/g3/s3/retention-g2/p-recorded\endcsname{0.326}
\expandafter\def\csname ev@B-R/8B/g3/s3/retention-g2/protocol\endcsname{B-R}
\expandafter\def\csname ev@B-R/8B/g3/s3/retention-g2/task-losses-high\endcsname{20}
\expandafter\def\csname ev@B-R/8B/g3/s3/retention-g2/task-losses-low\endcsname{0}
\expandafter\def\csname ev@B-R/8B/g3/s3/retention-g2/task-p-high\endcsname{0.999}
\expandafter\def\csname ev@B-R/8B/g3/s3/retention-g2/task-p-low\endcsname{0.00027}
\expandafter\def\csname ev@B-R/8B/g3/s3/retention-g2/task-wins-high\endcsname{24}
\expandafter\def\csname ev@B-R/8B/g3/s3/retention-g2/task-wins-low\endcsname{1}
\expandafter\def\csname ev@B-R/8B/g3/s3/retention-g2/tasks\endcsname{47}
\expandafter\def\csname ev@B-R/8B/g3/s3/retention-g2/uncounted\endcsname{0}
\expandafter\def\csname ev@B-R/8B/g3/s3/retention-g2/uncounted-pct\endcsname{0.00}
\expandafter\def\csname ev@B-R/8B/g3/s3/retention-g2/unit\endcsname{attempt slot}
\expandafter\def\csname ev@B-R/8B/g3/s3/retention-g2/wins\endcsname{24}
\expandafter\def\csname ev@B-R/8B/g3/s3/target/attempt-ci-hi-pp\endcsname{+13.11}
\expandafter\def\csname ev@B-R/8B/g3/s3/target/attempt-ci-lo-pp\endcsname{-4.01}
\expandafter\def\csname ev@B-R/8B/g3/s3/target/base-counted\endcsname{196}
\expandafter\def\csname ev@B-R/8B/g3/s3/target/base-passes\endcsname{45}
\expandafter\def\csname ev@B-R/8B/g3/s3/target/base-pct\endcsname{22.96}
\expandafter\def\csname ev@B-R/8B/g3/s3/target/base-scheduled\endcsname{196}
\expandafter\def\csname ev@B-R/8B/g3/s3/target/base-uncounted\endcsname{0}
\expandafter\def\csname ev@B-R/8B/g3/s3/target/cand-counted\endcsname{196}
\expandafter\def\csname ev@B-R/8B/g3/s3/target/cand-passes\endcsname{54}
\expandafter\def\csname ev@B-R/8B/g3/s3/target/cand-pct\endcsname{27.55}
\expandafter\def\csname ev@B-R/8B/g3/s3/target/cand-scheduled\endcsname{196}
\expandafter\def\csname ev@B-R/8B/g3/s3/target/cand-uncounted\endcsname{0}
\expandafter\def\csname ev@B-R/8B/g3/s3/target/diff-pp\endcsname{+4.59}
\expandafter\def\csname ev@B-R/8B/g3/s3/target/losses\endcsname{13}
\expandafter\def\csname ev@B-R/8B/g3/s3/target/outcome\endcsname{inconclusive}
\expandafter\def\csname ev@B-R/8B/g3/s3/target/p-recorded\endcsname{0.088}
\expandafter\def\csname ev@B-R/8B/g3/s3/target/protocol\endcsname{B-R}
\expandafter\def\csname ev@B-R/8B/g3/s3/target/task-losses-high\endcsname{13}
\expandafter\def\csname ev@B-R/8B/g3/s3/target/task-losses-low\endcsname{0}
\expandafter\def\csname ev@B-R/8B/g3/s3/target/task-p-high\endcsname{0.968}
\expandafter\def\csname ev@B-R/8B/g3/s3/target/task-p-low\endcsname{0.00014}
\expandafter\def\csname ev@B-R/8B/g3/s3/target/task-wins-high\endcsname{22}
\expandafter\def\csname ev@B-R/8B/g3/s3/target/task-wins-low\endcsname{3}
\expandafter\def\csname ev@B-R/8B/g3/s3/target/tasks\endcsname{49}
\expandafter\def\csname ev@B-R/8B/g3/s3/target/uncounted\endcsname{0}
\expandafter\def\csname ev@B-R/8B/g3/s3/target/uncounted-pct\endcsname{0.00}
\expandafter\def\csname ev@B-R/8B/g3/s3/target/unit\endcsname{attempt slot}
\expandafter\def\csname ev@B-R/8B/g3/s3/target/wins\endcsname{22}
\expandafter\def\csname ev@B-R/8B/g3/unfinished-seeds\endcsname{1}
\expandafter\def\csname ev@B/4B/g1/completion-claims/eval-exam/baseline/false\endcsname{293}
\expandafter\def\csname ev@B/4B/g1/completion-claims/eval-exam/baseline/true\endcsname{0}
\expandafter\def\csname ev@B/4B/g1/completion-claims/eval-exam/candidate/false\endcsname{373}
\expandafter\def\csname ev@B/4B/g1/completion-claims/eval-exam/candidate/true\endcsname{11}
\expandafter\def\csname ev@B/4B/g1/completion-claims/eval-unseen/baseline/false\endcsname{160}
\expandafter\def\csname ev@B/4B/g1/completion-claims/eval-unseen/baseline/true\endcsname{6}
\expandafter\def\csname ev@B/4B/g1/completion-claims/eval-unseen/candidate/false\endcsname{157}
\expandafter\def\csname ev@B/4B/g1/completion-claims/eval-unseen/candidate/true\endcsname{64}
\expandafter\def\csname ev@B/4B/g1/exam/attempt-ci-hi-pp\endcsname{+4.86}
\expandafter\def\csname ev@B/4B/g1/exam/attempt-ci-lo-pp\endcsname{+1.21}
\expandafter\def\csname ev@B/4B/g1/exam/base-counted\endcsname{397}
\expandafter\def\csname ev@B/4B/g1/exam/base-passes\endcsname{0}
\expandafter\def\csname ev@B/4B/g1/exam/base-pct\endcsname{0.00}
\expandafter\def\csname ev@B/4B/g1/exam/base-scheduled\endcsname{400}
\expandafter\def\csname ev@B/4B/g1/exam/base-uncounted\endcsname{3}
\expandafter\def\csname ev@B/4B/g1/exam/cand-counted\endcsname{400}
\expandafter\def\csname ev@B/4B/g1/exam/cand-passes\endcsname{11}
\expandafter\def\csname ev@B/4B/g1/exam/cand-pct\endcsname{2.75}
\expandafter\def\csname ev@B/4B/g1/exam/cand-scheduled\endcsname{400}
\expandafter\def\csname ev@B/4B/g1/exam/cand-uncounted\endcsname{0}
\expandafter\def\csname ev@B/4B/g1/exam/diff-pp\endcsname{+2.75}
\expandafter\def\csname ev@B/4B/g1/exam/losses\endcsname{0}
\expandafter\def\csname ev@B/4B/g1/exam/outcome\endcsname{inconclusive}
\expandafter\def\csname ev@B/4B/g1/exam/p-recorded\endcsname{0.00049}
\expandafter\def\csname ev@B/4B/g1/exam/protocol\endcsname{B}
\expandafter\def\csname ev@B/4B/g1/exam/task-losses-high\endcsname{0}
\expandafter\def\csname ev@B/4B/g1/exam/task-losses-low\endcsname{0}
\expandafter\def\csname ev@B/4B/g1/exam/task-p-high\endcsname{0.125}
\expandafter\def\csname ev@B/4B/g1/exam/task-p-low\endcsname{0.00049}
\expandafter\def\csname ev@B/4B/g1/exam/task-wins-high\endcsname{11}
\expandafter\def\csname ev@B/4B/g1/exam/task-wins-low\endcsname{3}
\expandafter\def\csname ev@B/4B/g1/exam/tasks\endcsname{100}
\expandafter\def\csname ev@B/4B/g1/exam/uncounted\endcsname{3}
\expandafter\def\csname ev@B/4B/g1/exam/uncounted-pct\endcsname{0.75}
\expandafter\def\csname ev@B/4B/g1/exam/unit\endcsname{attempt slot}
\expandafter\def\csname ev@B/4B/g1/exam/wins\endcsname{11}
\expandafter\def\csname ev@B/4B/g1/target/attempt-ci-hi-pp\endcsname{+32.28}
\expandafter\def\csname ev@B/4B/g1/target/attempt-ci-lo-pp\endcsname{+19.56}
\expandafter\def\csname ev@B/4B/g1/target/base-counted\endcsname{222}
\expandafter\def\csname ev@B/4B/g1/target/base-passes\endcsname{6}
\expandafter\def\csname ev@B/4B/g1/target/base-pct\endcsname{2.70}
\expandafter\def\csname ev@B/4B/g1/target/base-scheduled\endcsname{224}
\expandafter\def\csname ev@B/4B/g1/target/base-uncounted\endcsname{2}
\expandafter\def\csname ev@B/4B/g1/target/cand-counted\endcsname{224}
\expandafter\def\csname ev@B/4B/g1/target/cand-passes\endcsname{64}
\expandafter\def\csname ev@B/4B/g1/target/cand-pct\endcsname{28.57}
\expandafter\def\csname ev@B/4B/g1/target/cand-scheduled\endcsname{224}
\expandafter\def\csname ev@B/4B/g1/target/cand-uncounted\endcsname{0}
\expandafter\def\csname ev@B/4B/g1/target/diff-pp\endcsname{+25.87}
\expandafter\def\csname ev@B/4B/g1/target/losses\endcsname{4}
\expandafter\def\csname ev@B/4B/g1/target/outcome\endcsname{improvement supported}
\expandafter\def\csname ev@B/4B/g1/target/p-recorded\endcsname{\ensuremath{1.0\times 10^{-14}}}
\expandafter\def\csname ev@B/4B/g1/target/promoted\endcsname{yes}
\expandafter\def\csname ev@B/4B/g1/target/protocol\endcsname{B}
\expandafter\def\csname ev@B/4B/g1/target/task-losses-high\endcsname{4}
\expandafter\def\csname ev@B/4B/g1/target/task-losses-low\endcsname{0}
\expandafter\def\csname ev@B/4B/g1/target/task-p-high\endcsname{0.0059}
\expandafter\def\csname ev@B/4B/g1/target/task-p-low\endcsname{\ensuremath{1.4\times 10^{-17}}}
\expandafter\def\csname ev@B/4B/g1/target/task-wins-high\endcsname{56}
\expandafter\def\csname ev@B/4B/g1/target/task-wins-low\endcsname{15}
\expandafter\def\csname ev@B/4B/g1/target/tasks\endcsname{56}
\expandafter\def\csname ev@B/4B/g1/target/uncounted\endcsname{2}
\expandafter\def\csname ev@B/4B/g1/target/uncounted-pct\endcsname{0.89}
\expandafter\def\csname ev@B/4B/g1/target/unit\endcsname{attempt slot}
\expandafter\def\csname ev@B/4B/g1/target/wins\endcsname{62}
\expandafter\def\csname ev@B/4B/g2/exam/attempt-ci-hi-pp\endcsname{+2.21}
\expandafter\def\csname ev@B/4B/g2/exam/attempt-ci-lo-pp\endcsname{-2.73}
\expandafter\def\csname ev@B/4B/g2/exam/base-counted\endcsname{400}
\expandafter\def\csname ev@B/4B/g2/exam/base-passes\endcsname{12}
\expandafter\def\csname ev@B/4B/g2/exam/base-pct\endcsname{3.00}
\expandafter\def\csname ev@B/4B/g2/exam/base-scheduled\endcsname{400}
\expandafter\def\csname ev@B/4B/g2/exam/base-uncounted\endcsname{0}
\expandafter\def\csname ev@B/4B/g2/exam/cand-counted\endcsname{400}
\expandafter\def\csname ev@B/4B/g2/exam/cand-passes\endcsname{11}
\expandafter\def\csname ev@B/4B/g2/exam/cand-pct\endcsname{2.75}
\expandafter\def\csname ev@B/4B/g2/exam/cand-scheduled\endcsname{400}
\expandafter\def\csname ev@B/4B/g2/exam/cand-uncounted\endcsname{0}
\expandafter\def\csname ev@B/4B/g2/exam/diff-pp\endcsname{-0.25}
\expandafter\def\csname ev@B/4B/g2/exam/losses\endcsname{11}
\expandafter\def\csname ev@B/4B/g2/exam/outcome\endcsname{inconclusive}
\expandafter\def\csname ev@B/4B/g2/exam/p-recorded\endcsname{0.668}
\expandafter\def\csname ev@B/4B/g2/exam/protocol\endcsname{B}
\expandafter\def\csname ev@B/4B/g2/exam/task-losses-high\endcsname{11}
\expandafter\def\csname ev@B/4B/g2/exam/task-losses-low\endcsname{1}
\expandafter\def\csname ev@B/4B/g2/exam/task-p-high\endcsname{1}
\expandafter\def\csname ev@B/4B/g2/exam/task-p-low\endcsname{0.046}
\expandafter\def\csname ev@B/4B/g2/exam/task-wins-high\endcsname{10}
\expandafter\def\csname ev@B/4B/g2/exam/task-wins-low\endcsname{0}
\expandafter\def\csname ev@B/4B/g2/exam/tasks\endcsname{100}
\expandafter\def\csname ev@B/4B/g2/exam/uncounted\endcsname{0}
\expandafter\def\csname ev@B/4B/g2/exam/uncounted-pct\endcsname{0.00}
\expandafter\def\csname ev@B/4B/g2/exam/unit\endcsname{attempt slot}
\expandafter\def\csname ev@B/4B/g2/exam/wins\endcsname{10}
\expandafter\def\csname ev@B/4B/g2/retention-g1/attempt-ci-hi-pp\endcsname{+12.13}
\expandafter\def\csname ev@B/4B/g2/retention-g1/attempt-ci-lo-pp\endcsname{-5.06}
\expandafter\def\csname ev@B/4B/g2/retention-g1/base-counted\endcsname{224}
\expandafter\def\csname ev@B/4B/g2/retention-g1/base-passes\endcsname{68}
\expandafter\def\csname ev@B/4B/g2/retention-g1/base-pct\endcsname{30.36}
\expandafter\def\csname ev@B/4B/g2/retention-g1/base-scheduled\endcsname{224}
\expandafter\def\csname ev@B/4B/g2/retention-g1/base-uncounted\endcsname{0}
\expandafter\def\csname ev@B/4B/g2/retention-g1/cand-counted\endcsname{224}
\expandafter\def\csname ev@B/4B/g2/retention-g1/cand-passes\endcsname{76}
\expandafter\def\csname ev@B/4B/g2/retention-g1/cand-pct\endcsname{33.93}
\expandafter\def\csname ev@B/4B/g2/retention-g1/cand-scheduled\endcsname{224}
\expandafter\def\csname ev@B/4B/g2/retention-g1/cand-uncounted\endcsname{0}
\expandafter\def\csname ev@B/4B/g2/retention-g1/diff-pp\endcsname{+3.57}
\expandafter\def\csname ev@B/4B/g2/retention-g1/gate-passed\endcsname{yes}
\expandafter\def\csname ev@B/4B/g2/retention-g1/losses\endcsname{19}
\expandafter\def\csname ev@B/4B/g2/retention-g1/outcome\endcsname{inconclusive}
\expandafter\def\csname ev@B/4B/g2/retention-g1/p-recorded\endcsname{0.151}
\expandafter\def\csname ev@B/4B/g2/retention-g1/protocol\endcsname{B}
\expandafter\def\csname ev@B/4B/g2/retention-g1/task-losses-high\endcsname{19}
\expandafter\def\csname ev@B/4B/g2/retention-g1/task-losses-low\endcsname{0}
\expandafter\def\csname ev@B/4B/g2/retention-g1/task-p-high\endcsname{0.995}
\expandafter\def\csname ev@B/4B/g2/retention-g1/task-p-low\endcsname{\ensuremath{5.7\times 10^{-5}}}
\expandafter\def\csname ev@B/4B/g2/retention-g1/task-wins-high\endcsname{27}
\expandafter\def\csname ev@B/4B/g2/retention-g1/task-wins-low\endcsname{2}
\expandafter\def\csname ev@B/4B/g2/retention-g1/tasks\endcsname{56}
\expandafter\def\csname ev@B/4B/g2/retention-g1/uncounted\endcsname{0}
\expandafter\def\csname ev@B/4B/g2/retention-g1/uncounted-pct\endcsname{0.00}
\expandafter\def\csname ev@B/4B/g2/retention-g1/unit\endcsname{attempt slot}
\expandafter\def\csname ev@B/4B/g2/retention-g1/wins\endcsname{27}
\expandafter\def\csname ev@B/4B/g2/target/attempt-ci-hi-pp\endcsname{+9.84}
\expandafter\def\csname ev@B/4B/g2/target/attempt-ci-lo-pp\endcsname{-5.60}
\expandafter\def\csname ev@B/4B/g2/target/base-counted\endcsname{188}
\expandafter\def\csname ev@B/4B/g2/target/base-passes\endcsname{31}
\expandafter\def\csname ev@B/4B/g2/target/base-pct\endcsname{16.49}
\expandafter\def\csname ev@B/4B/g2/target/base-scheduled\endcsname{188}
\expandafter\def\csname ev@B/4B/g2/target/base-uncounted\endcsname{0}
\expandafter\def\csname ev@B/4B/g2/target/cand-counted\endcsname{188}
\expandafter\def\csname ev@B/4B/g2/target/cand-passes\endcsname{35}
\expandafter\def\csname ev@B/4B/g2/target/cand-pct\endcsname{18.62}
\expandafter\def\csname ev@B/4B/g2/target/cand-scheduled\endcsname{188}
\expandafter\def\csname ev@B/4B/g2/target/cand-uncounted\endcsname{0}
\expandafter\def\csname ev@B/4B/g2/target/diff-pp\endcsname{+2.13}
\expandafter\def\csname ev@B/4B/g2/target/losses\endcsname{13}
\expandafter\def\csname ev@B/4B/g2/target/outcome\endcsname{inconclusive}
\expandafter\def\csname ev@B/4B/g2/target/p-recorded\endcsname{0.292}
\expandafter\def\csname ev@B/4B/g2/target/promoted\endcsname{no}
\expandafter\def\csname ev@B/4B/g2/target/protocol\endcsname{B}
\expandafter\def\csname ev@B/4B/g2/target/task-losses-high\endcsname{13}
\expandafter\def\csname ev@B/4B/g2/target/task-losses-low\endcsname{0}
\expandafter\def\csname ev@B/4B/g2/target/task-p-high\endcsname{0.989}
\expandafter\def\csname ev@B/4B/g2/target/task-p-low\endcsname{0.0022}
\expandafter\def\csname ev@B/4B/g2/target/task-wins-high\endcsname{17}
\expandafter\def\csname ev@B/4B/g2/target/task-wins-low\endcsname{1}
\expandafter\def\csname ev@B/4B/g2/target/tasks\endcsname{47}
\expandafter\def\csname ev@B/4B/g2/target/uncounted\endcsname{0}
\expandafter\def\csname ev@B/4B/g2/target/uncounted-pct\endcsname{0.00}
\expandafter\def\csname ev@B/4B/g2/target/unit\endcsname{attempt slot}
\expandafter\def\csname ev@B/4B/g2/target/wins\endcsname{17}
\expandafter\def\csname ev@B/4B/g3/exam/attempt-ci-hi-pp\endcsname{+2.82}
\expandafter\def\csname ev@B/4B/g3/exam/attempt-ci-lo-pp\endcsname{-2.30}
\expandafter\def\csname ev@B/4B/g3/exam/base-counted\endcsname{400}
\expandafter\def\csname ev@B/4B/g3/exam/base-passes\endcsname{12}
\expandafter\def\csname ev@B/4B/g3/exam/base-pct\endcsname{3.00}
\expandafter\def\csname ev@B/4B/g3/exam/base-scheduled\endcsname{400}
\expandafter\def\csname ev@B/4B/g3/exam/base-uncounted\endcsname{0}
\expandafter\def\csname ev@B/4B/g3/exam/cand-counted\endcsname{400}
\expandafter\def\csname ev@B/4B/g3/exam/cand-passes\endcsname{13}
\expandafter\def\csname ev@B/4B/g3/exam/cand-pct\endcsname{3.25}
\expandafter\def\csname ev@B/4B/g3/exam/cand-scheduled\endcsname{400}
\expandafter\def\csname ev@B/4B/g3/exam/cand-uncounted\endcsname{0}
\expandafter\def\csname ev@B/4B/g3/exam/diff-pp\endcsname{+0.25}
\expandafter\def\csname ev@B/4B/g3/exam/losses\endcsname{10}
\expandafter\def\csname ev@B/4B/g3/exam/outcome\endcsname{inconclusive}
\expandafter\def\csname ev@B/4B/g3/exam/p-recorded\endcsname{0.5}
\expandafter\def\csname ev@B/4B/g3/exam/protocol\endcsname{B}
\expandafter\def\csname ev@B/4B/g3/exam/task-losses-high\endcsname{10}
\expandafter\def\csname ev@B/4B/g3/exam/task-losses-low\endcsname{0}
\expandafter\def\csname ev@B/4B/g3/exam/task-p-high\endcsname{0.989}
\expandafter\def\csname ev@B/4B/g3/exam/task-p-low\endcsname{0.029}
\expandafter\def\csname ev@B/4B/g3/exam/task-wins-high\endcsname{11}
\expandafter\def\csname ev@B/4B/g3/exam/task-wins-low\endcsname{1}
\expandafter\def\csname ev@B/4B/g3/exam/tasks\endcsname{100}
\expandafter\def\csname ev@B/4B/g3/exam/uncounted\endcsname{0}
\expandafter\def\csname ev@B/4B/g3/exam/uncounted-pct\endcsname{0.00}
\expandafter\def\csname ev@B/4B/g3/exam/unit\endcsname{attempt slot}
\expandafter\def\csname ev@B/4B/g3/exam/wins\endcsname{11}
\expandafter\def\csname ev@B/4B/g3/retention-g1/attempt-ci-hi-pp\endcsname{+15.30}
\expandafter\def\csname ev@B/4B/g3/retention-g1/attempt-ci-lo-pp\endcsname{-2.05}
\expandafter\def\csname ev@B/4B/g3/retention-g1/base-counted\endcsname{224}
\expandafter\def\csname ev@B/4B/g3/retention-g1/base-passes\endcsname{68}
\expandafter\def\csname ev@B/4B/g3/retention-g1/base-pct\endcsname{30.36}
\expandafter\def\csname ev@B/4B/g3/retention-g1/base-scheduled\endcsname{224}
\expandafter\def\csname ev@B/4B/g3/retention-g1/base-uncounted\endcsname{0}
\expandafter\def\csname ev@B/4B/g3/retention-g1/cand-counted\endcsname{224}
\expandafter\def\csname ev@B/4B/g3/retention-g1/cand-passes\endcsname{83}
\expandafter\def\csname ev@B/4B/g3/retention-g1/cand-pct\endcsname{37.05}
\expandafter\def\csname ev@B/4B/g3/retention-g1/cand-scheduled\endcsname{224}
\expandafter\def\csname ev@B/4B/g3/retention-g1/cand-uncounted\endcsname{0}
\expandafter\def\csname ev@B/4B/g3/retention-g1/diff-pp\endcsname{+6.70}
\expandafter\def\csname ev@B/4B/g3/retention-g1/gate-passed\endcsname{yes}
\expandafter\def\csname ev@B/4B/g3/retention-g1/losses\endcsname{13}
\expandafter\def\csname ev@B/4B/g3/retention-g1/outcome\endcsname{inconclusive}
\expandafter\def\csname ev@B/4B/g3/retention-g1/p-recorded\endcsname{0.014}
\expandafter\def\csname ev@B/4B/g3/retention-g1/protocol\endcsname{B}
\expandafter\def\csname ev@B/4B/g3/retention-g1/task-losses-high\endcsname{13}
\expandafter\def\csname ev@B/4B/g3/retention-g1/task-losses-low\endcsname{0}
\expandafter\def\csname ev@B/4B/g3/retention-g1/task-p-high\endcsname{0.942}
\expandafter\def\csname ev@B/4B/g3/retention-g1/task-p-low\endcsname{\ensuremath{4.2\times 10^{-6}}}
\expandafter\def\csname ev@B/4B/g3/retention-g1/task-wins-high\endcsname{28}
\expandafter\def\csname ev@B/4B/g3/retention-g1/task-wins-low\endcsname{4}
\expandafter\def\csname ev@B/4B/g3/retention-g1/tasks\endcsname{56}
\expandafter\def\csname ev@B/4B/g3/retention-g1/uncounted\endcsname{0}
\expandafter\def\csname ev@B/4B/g3/retention-g1/uncounted-pct\endcsname{0.00}
\expandafter\def\csname ev@B/4B/g3/retention-g1/unit\endcsname{attempt slot}
\expandafter\def\csname ev@B/4B/g3/retention-g1/wins\endcsname{28}
\expandafter\def\csname ev@B/4B/g3/retention-g2/attempt-ci-hi-pp\endcsname{+6.97}
\expandafter\def\csname ev@B/4B/g3/retention-g2/attempt-ci-lo-pp\endcsname{-8.03}
\expandafter\def\csname ev@B/4B/g3/retention-g2/base-counted\endcsname{188}
\expandafter\def\csname ev@B/4B/g3/retention-g2/base-passes\endcsname{31}
\expandafter\def\csname ev@B/4B/g3/retention-g2/base-pct\endcsname{16.49}
\expandafter\def\csname ev@B/4B/g3/retention-g2/base-scheduled\endcsname{188}
\expandafter\def\csname ev@B/4B/g3/retention-g2/base-uncounted\endcsname{0}
\expandafter\def\csname ev@B/4B/g3/retention-g2/cand-counted\endcsname{188}
\expandafter\def\csname ev@B/4B/g3/retention-g2/cand-passes\endcsname{30}
\expandafter\def\csname ev@B/4B/g3/retention-g2/cand-pct\endcsname{15.96}
\expandafter\def\csname ev@B/4B/g3/retention-g2/cand-scheduled\endcsname{188}
\expandafter\def\csname ev@B/4B/g3/retention-g2/cand-uncounted\endcsname{0}
\expandafter\def\csname ev@B/4B/g3/retention-g2/diff-pp\endcsname{-0.53}
\expandafter\def\csname ev@B/4B/g3/retention-g2/gate-passed\endcsname{yes}
\expandafter\def\csname ev@B/4B/g3/retention-g2/losses\endcsname{12}
\expandafter\def\csname ev@B/4B/g3/retention-g2/outcome\endcsname{inconclusive}
\expandafter\def\csname ev@B/4B/g3/retention-g2/p-recorded\endcsname{0.661}
\expandafter\def\csname ev@B/4B/g3/retention-g2/protocol\endcsname{B}
\expandafter\def\csname ev@B/4B/g3/retention-g2/task-losses-high\endcsname{12}
\expandafter\def\csname ev@B/4B/g3/retention-g2/task-losses-low\endcsname{1}
\expandafter\def\csname ev@B/4B/g3/retention-g2/task-p-high\endcsname{1}
\expandafter\def\csname ev@B/4B/g3/retention-g2/task-p-low\endcsname{0.029}
\expandafter\def\csname ev@B/4B/g3/retention-g2/task-wins-high\endcsname{11}
\expandafter\def\csname ev@B/4B/g3/retention-g2/task-wins-low\endcsname{0}
\expandafter\def\csname ev@B/4B/g3/retention-g2/tasks\endcsname{47}
\expandafter\def\csname ev@B/4B/g3/retention-g2/uncounted\endcsname{0}
\expandafter\def\csname ev@B/4B/g3/retention-g2/uncounted-pct\endcsname{0.00}
\expandafter\def\csname ev@B/4B/g3/retention-g2/unit\endcsname{attempt slot}
\expandafter\def\csname ev@B/4B/g3/retention-g2/wins\endcsname{11}
\expandafter\def\csname ev@B/4B/g3/target/attempt-ci-hi-pp\endcsname{+14.62}
\expandafter\def\csname ev@B/4B/g3/target/attempt-ci-lo-pp\endcsname{-0.38}
\expandafter\def\csname ev@B/4B/g3/target/base-counted\endcsname{196}
\expandafter\def\csname ev@B/4B/g3/target/base-passes\endcsname{27}
\expandafter\def\csname ev@B/4B/g3/target/base-pct\endcsname{13.78}
\expandafter\def\csname ev@B/4B/g3/target/base-scheduled\endcsname{196}
\expandafter\def\csname ev@B/4B/g3/target/base-uncounted\endcsname{0}
\expandafter\def\csname ev@B/4B/g3/target/cand-counted\endcsname{196}
\expandafter\def\csname ev@B/4B/g3/target/cand-passes\endcsname{41}
\expandafter\def\csname ev@B/4B/g3/target/cand-pct\endcsname{20.92}
\expandafter\def\csname ev@B/4B/g3/target/cand-scheduled\endcsname{196}
\expandafter\def\csname ev@B/4B/g3/target/cand-uncounted\endcsname{0}
\expandafter\def\csname ev@B/4B/g3/target/diff-pp\endcsname{+7.14}
\expandafter\def\csname ev@B/4B/g3/target/losses\endcsname{10}
\expandafter\def\csname ev@B/4B/g3/target/outcome\endcsname{inconclusive}
\expandafter\def\csname ev@B/4B/g3/target/p-recorded\endcsname{0.012}
\expandafter\def\csname ev@B/4B/g3/target/promoted\endcsname{yes}
\expandafter\def\csname ev@B/4B/g3/target/protocol\endcsname{B}
\expandafter\def\csname ev@B/4B/g3/target/task-losses-high\endcsname{10}
\expandafter\def\csname ev@B/4B/g3/target/task-losses-low\endcsname{0}
\expandafter\def\csname ev@B/4B/g3/target/task-p-high\endcsname{0.895}
\expandafter\def\csname ev@B/4B/g3/target/task-p-low\endcsname{\ensuremath{1.8\times 10^{-5}}}
\expandafter\def\csname ev@B/4B/g3/target/task-wins-high\endcsname{24}
\expandafter\def\csname ev@B/4B/g3/target/task-wins-low\endcsname{4}
\expandafter\def\csname ev@B/4B/g3/target/tasks\endcsname{49}
\expandafter\def\csname ev@B/4B/g3/target/uncounted\endcsname{0}
\expandafter\def\csname ev@B/4B/g3/target/uncounted-pct\endcsname{0.00}
\expandafter\def\csname ev@B/4B/g3/target/unit\endcsname{attempt slot}
\expandafter\def\csname ev@B/4B/g3/target/wins\endcsname{24}
\expandafter\def\csname ev@B/4B/generations\endcsname{3}
\expandafter\def\csname ev@B/4B/promoted-generations\endcsname{2}
\expandafter\def\csname ev@B/4B/rejected-generations\endcsname{1}
\expandafter\def\csname ev@B/8B/g1/completion-claims/eval-exam/baseline/false\endcsname{249}
\expandafter\def\csname ev@B/8B/g1/completion-claims/eval-exam/baseline/true\endcsname{0}
\expandafter\def\csname ev@B/8B/g1/completion-claims/eval-exam/candidate/false\endcsname{321}
\expandafter\def\csname ev@B/8B/g1/completion-claims/eval-exam/candidate/true\endcsname{25}
\expandafter\def\csname ev@B/8B/g1/completion-claims/eval-unseen/baseline/false\endcsname{164}
\expandafter\def\csname ev@B/8B/g1/completion-claims/eval-unseen/baseline/true\endcsname{9}
\expandafter\def\csname ev@B/8B/g1/completion-claims/eval-unseen/candidate/false\endcsname{134}
\expandafter\def\csname ev@B/8B/g1/completion-claims/eval-unseen/candidate/true\endcsname{85}
\expandafter\def\csname ev@B/8B/g1/exam/attempt-ci-hi-pp\endcsname{+9.06}
\expandafter\def\csname ev@B/8B/g1/exam/attempt-ci-lo-pp\endcsname{+4.04}
\expandafter\def\csname ev@B/8B/g1/exam/base-counted\endcsname{391}
\expandafter\def\csname ev@B/8B/g1/exam/base-passes\endcsname{0}
\expandafter\def\csname ev@B/8B/g1/exam/base-pct\endcsname{0.00}
\expandafter\def\csname ev@B/8B/g1/exam/base-scheduled\endcsname{400}
\expandafter\def\csname ev@B/8B/g1/exam/base-uncounted\endcsname{9}
\expandafter\def\csname ev@B/8B/g1/exam/cand-counted\endcsname{400}
\expandafter\def\csname ev@B/8B/g1/exam/cand-passes\endcsname{25}
\expandafter\def\csname ev@B/8B/g1/exam/cand-pct\endcsname{6.25}
\expandafter\def\csname ev@B/8B/g1/exam/cand-scheduled\endcsname{400}
\expandafter\def\csname ev@B/8B/g1/exam/cand-uncounted\endcsname{0}
\expandafter\def\csname ev@B/8B/g1/exam/diff-pp\endcsname{+6.25}
\expandafter\def\csname ev@B/8B/g1/exam/losses\endcsname{0}
\expandafter\def\csname ev@B/8B/g1/exam/outcome\endcsname{excluded}
\expandafter\def\csname ev@B/8B/g1/exam/p-recorded\endcsname{\ensuremath{3.0\times 10^{-8}}}
\expandafter\def\csname ev@B/8B/g1/exam/protocol\endcsname{B}
\expandafter\def\csname ev@B/8B/g1/exam/task-losses-high\endcsname{0}
\expandafter\def\csname ev@B/8B/g1/exam/task-losses-low\endcsname{0}
\expandafter\def\csname ev@B/8B/g1/exam/task-p-high\endcsname{0.0078}
\expandafter\def\csname ev@B/8B/g1/exam/task-p-low\endcsname{\ensuremath{3.0\times 10^{-8}}}
\expandafter\def\csname ev@B/8B/g1/exam/task-wins-high\endcsname{25}
\expandafter\def\csname ev@B/8B/g1/exam/task-wins-low\endcsname{7}
\expandafter\def\csname ev@B/8B/g1/exam/tasks\endcsname{100}
\expandafter\def\csname ev@B/8B/g1/exam/uncounted\endcsname{9}
\expandafter\def\csname ev@B/8B/g1/exam/uncounted-pct\endcsname{2.25}
\expandafter\def\csname ev@B/8B/g1/exam/unit\endcsname{attempt slot}
\expandafter\def\csname ev@B/8B/g1/exam/wins\endcsname{25}
\expandafter\def\csname ev@B/8B/g1/target/attempt-ci-hi-pp\endcsname{+40.70}
\expandafter\def\csname ev@B/8B/g1/target/attempt-ci-lo-pp\endcsname{+26.92}
\expandafter\def\csname ev@B/8B/g1/target/base-counted\endcsname{224}
\expandafter\def\csname ev@B/8B/g1/target/base-passes\endcsname{9}
\expandafter\def\csname ev@B/8B/g1/target/base-pct\endcsname{4.02}
\expandafter\def\csname ev@B/8B/g1/target/base-scheduled\endcsname{224}
\expandafter\def\csname ev@B/8B/g1/target/base-uncounted\endcsname{0}
\expandafter\def\csname ev@B/8B/g1/target/cand-counted\endcsname{224}
\expandafter\def\csname ev@B/8B/g1/target/cand-passes\endcsname{85}
\expandafter\def\csname ev@B/8B/g1/target/cand-pct\endcsname{37.95}
\expandafter\def\csname ev@B/8B/g1/target/cand-scheduled\endcsname{224}
\expandafter\def\csname ev@B/8B/g1/target/cand-uncounted\endcsname{0}
\expandafter\def\csname ev@B/8B/g1/target/diff-pp\endcsname{+33.93}
\expandafter\def\csname ev@B/8B/g1/target/losses\endcsname{3}
\expandafter\def\csname ev@B/8B/g1/target/outcome\endcsname{improvement supported}
\expandafter\def\csname ev@B/8B/g1/target/p-recorded\endcsname{\ensuremath{1.9\times 10^{-20}}}
\expandafter\def\csname ev@B/8B/g1/target/promoted\endcsname{yes}
\expandafter\def\csname ev@B/8B/g1/target/protocol\endcsname{B}
\expandafter\def\csname ev@B/8B/g1/target/task-losses-high\endcsname{3}
\expandafter\def\csname ev@B/8B/g1/target/task-losses-low\endcsname{0}
\expandafter\def\csname ev@B/8B/g1/target/task-p-high\endcsname{0.00024}
\expandafter\def\csname ev@B/8B/g1/target/task-p-low\endcsname{\ensuremath{1.4\times 10^{-17}}}
\expandafter\def\csname ev@B/8B/g1/target/task-wins-high\endcsname{56}
\expandafter\def\csname ev@B/8B/g1/target/task-wins-low\endcsname{19}
\expandafter\def\csname ev@B/8B/g1/target/tasks\endcsname{56}
\expandafter\def\csname ev@B/8B/g1/target/uncounted\endcsname{0}
\expandafter\def\csname ev@B/8B/g1/target/uncounted-pct\endcsname{0.00}
\expandafter\def\csname ev@B/8B/g1/target/unit\endcsname{attempt slot}
\expandafter\def\csname ev@B/8B/g1/target/wins\endcsname{79}
\expandafter\def\csname ev@B/8B/g2/exam/attempt-ci-hi-pp\endcsname{+3.72}
\expandafter\def\csname ev@B/8B/g2/exam/attempt-ci-lo-pp\endcsname{-3.72}
\expandafter\def\csname ev@B/8B/g2/exam/base-counted\endcsname{400}
\expandafter\def\csname ev@B/8B/g2/exam/base-passes\endcsname{30}
\expandafter\def\csname ev@B/8B/g2/exam/base-pct\endcsname{7.50}
\expandafter\def\csname ev@B/8B/g2/exam/base-scheduled\endcsname{400}
\expandafter\def\csname ev@B/8B/g2/exam/base-uncounted\endcsname{0}
\expandafter\def\csname ev@B/8B/g2/exam/cand-counted\endcsname{400}
\expandafter\def\csname ev@B/8B/g2/exam/cand-passes\endcsname{30}
\expandafter\def\csname ev@B/8B/g2/exam/cand-pct\endcsname{7.50}
\expandafter\def\csname ev@B/8B/g2/exam/cand-scheduled\endcsname{400}
\expandafter\def\csname ev@B/8B/g2/exam/cand-uncounted\endcsname{0}
\expandafter\def\csname ev@B/8B/g2/exam/diff-pp\endcsname{+0.00}
\expandafter\def\csname ev@B/8B/g2/exam/losses\endcsname{17}
\expandafter\def\csname ev@B/8B/g2/exam/outcome\endcsname{inconclusive}
\expandafter\def\csname ev@B/8B/g2/exam/p-recorded\endcsname{0.568}
\expandafter\def\csname ev@B/8B/g2/exam/protocol\endcsname{B}
\expandafter\def\csname ev@B/8B/g2/exam/task-losses-high\endcsname{17}
\expandafter\def\csname ev@B/8B/g2/exam/task-losses-low\endcsname{0}
\expandafter\def\csname ev@B/8B/g2/exam/task-p-high\endcsname{1}
\expandafter\def\csname ev@B/8B/g2/exam/task-p-low\endcsname{0.0059}
\expandafter\def\csname ev@B/8B/g2/exam/task-wins-high\endcsname{17}
\expandafter\def\csname ev@B/8B/g2/exam/task-wins-low\endcsname{0}
\expandafter\def\csname ev@B/8B/g2/exam/tasks\endcsname{100}
\expandafter\def\csname ev@B/8B/g2/exam/uncounted\endcsname{0}
\expandafter\def\csname ev@B/8B/g2/exam/uncounted-pct\endcsname{0.00}
\expandafter\def\csname ev@B/8B/g2/exam/unit\endcsname{attempt slot}
\expandafter\def\csname ev@B/8B/g2/exam/wins\endcsname{17}
\expandafter\def\csname ev@B/8B/g2/retention-g1/attempt-ci-hi-pp\endcsname{+11.64}
\expandafter\def\csname ev@B/8B/g2/retention-g1/attempt-ci-lo-pp\endcsname{-6.34}
\expandafter\def\csname ev@B/8B/g2/retention-g1/base-counted\endcsname{224}
\expandafter\def\csname ev@B/8B/g2/retention-g1/base-passes\endcsname{86}
\expandafter\def\csname ev@B/8B/g2/retention-g1/base-pct\endcsname{38.39}
\expandafter\def\csname ev@B/8B/g2/retention-g1/base-scheduled\endcsname{224}
\expandafter\def\csname ev@B/8B/g2/retention-g1/base-uncounted\endcsname{0}
\expandafter\def\csname ev@B/8B/g2/retention-g1/cand-counted\endcsname{224}
\expandafter\def\csname ev@B/8B/g2/retention-g1/cand-passes\endcsname{92}
\expandafter\def\csname ev@B/8B/g2/retention-g1/cand-pct\endcsname{41.07}
\expandafter\def\csname ev@B/8B/g2/retention-g1/cand-scheduled\endcsname{224}
\expandafter\def\csname ev@B/8B/g2/retention-g1/cand-uncounted\endcsname{0}
\expandafter\def\csname ev@B/8B/g2/retention-g1/diff-pp\endcsname{+2.68}
\expandafter\def\csname ev@B/8B/g2/retention-g1/gate-passed\endcsname{yes}
\expandafter\def\csname ev@B/8B/g2/retention-g1/losses\endcsname{23}
\expandafter\def\csname ev@B/8B/g2/retention-g1/outcome\endcsname{inconclusive}
\expandafter\def\csname ev@B/8B/g2/retention-g1/p-recorded\endcsname{0.244}
\expandafter\def\csname ev@B/8B/g2/retention-g1/protocol\endcsname{B}
\expandafter\def\csname ev@B/8B/g2/retention-g1/task-losses-high\endcsname{23}
\expandafter\def\csname ev@B/8B/g2/retention-g1/task-losses-low\endcsname{0}
\expandafter\def\csname ev@B/8B/g2/retention-g1/task-p-high\endcsname{0.999}
\expandafter\def\csname ev@B/8B/g2/retention-g1/task-p-low\endcsname{\ensuremath{5.8\times 10^{-5}}}
\expandafter\def\csname ev@B/8B/g2/retention-g1/task-wins-high\endcsname{29}
\expandafter\def\csname ev@B/8B/g2/retention-g1/task-wins-low\endcsname{2}
\expandafter\def\csname ev@B/8B/g2/retention-g1/tasks\endcsname{56}
\expandafter\def\csname ev@B/8B/g2/retention-g1/uncounted\endcsname{0}
\expandafter\def\csname ev@B/8B/g2/retention-g1/uncounted-pct\endcsname{0.00}
\expandafter\def\csname ev@B/8B/g2/retention-g1/unit\endcsname{attempt slot}
\expandafter\def\csname ev@B/8B/g2/retention-g1/wins\endcsname{29}
\expandafter\def\csname ev@B/8B/g2/target/attempt-ci-hi-pp\endcsname{+16.62}
\expandafter\def\csname ev@B/8B/g2/target/attempt-ci-lo-pp\endcsname{-0.81}
\expandafter\def\csname ev@B/8B/g2/target/base-counted\endcsname{188}
\expandafter\def\csname ev@B/8B/g2/target/base-passes\endcsname{40}
\expandafter\def\csname ev@B/8B/g2/target/base-pct\endcsname{21.28}
\expandafter\def\csname ev@B/8B/g2/target/base-scheduled\endcsname{188}
\expandafter\def\csname ev@B/8B/g2/target/base-uncounted\endcsname{0}
\expandafter\def\csname ev@B/8B/g2/target/cand-counted\endcsname{188}
\expandafter\def\csname ev@B/8B/g2/target/cand-passes\endcsname{55}
\expandafter\def\csname ev@B/8B/g2/target/cand-pct\endcsname{29.26}
\expandafter\def\csname ev@B/8B/g2/target/cand-scheduled\endcsname{188}
\expandafter\def\csname ev@B/8B/g2/target/cand-uncounted\endcsname{0}
\expandafter\def\csname ev@B/8B/g2/target/diff-pp\endcsname{+7.98}
\expandafter\def\csname ev@B/8B/g2/target/losses\endcsname{14}
\expandafter\def\csname ev@B/8B/g2/target/outcome\endcsname{inconclusive}
\expandafter\def\csname ev@B/8B/g2/target/p-recorded\endcsname{0.016}
\expandafter\def\csname ev@B/8B/g2/target/promoted\endcsname{yes}
\expandafter\def\csname ev@B/8B/g2/target/protocol\endcsname{B}
\expandafter\def\csname ev@B/8B/g2/target/task-losses-high\endcsname{14}
\expandafter\def\csname ev@B/8B/g2/target/task-losses-low\endcsname{0}
\expandafter\def\csname ev@B/8B/g2/target/task-p-high\endcsname{0.942}
\expandafter\def\csname ev@B/8B/g2/target/task-p-low\endcsname{\ensuremath{4.2\times 10^{-6}}}
\expandafter\def\csname ev@B/8B/g2/target/task-wins-high\endcsname{29}
\expandafter\def\csname ev@B/8B/g2/target/task-wins-low\endcsname{4}
\expandafter\def\csname ev@B/8B/g2/target/tasks\endcsname{47}
\expandafter\def\csname ev@B/8B/g2/target/uncounted\endcsname{0}
\expandafter\def\csname ev@B/8B/g2/target/uncounted-pct\endcsname{0.00}
\expandafter\def\csname ev@B/8B/g2/target/unit\endcsname{attempt slot}
\expandafter\def\csname ev@B/8B/g2/target/wins\endcsname{29}
\expandafter\def\csname ev@B/8B/g3/exam/attempt-ci-hi-pp\endcsname{+3.83}
\expandafter\def\csname ev@B/8B/g3/exam/attempt-ci-lo-pp\endcsname{-3.83}
\expandafter\def\csname ev@B/8B/g3/exam/base-counted\endcsname{400}
\expandafter\def\csname ev@B/8B/g3/exam/base-passes\endcsname{32}
\expandafter\def\csname ev@B/8B/g3/exam/base-pct\endcsname{8.00}
\expandafter\def\csname ev@B/8B/g3/exam/base-scheduled\endcsname{400}
\expandafter\def\csname ev@B/8B/g3/exam/base-uncounted\endcsname{0}
\expandafter\def\csname ev@B/8B/g3/exam/cand-counted\endcsname{400}
\expandafter\def\csname ev@B/8B/g3/exam/cand-passes\endcsname{32}
\expandafter\def\csname ev@B/8B/g3/exam/cand-pct\endcsname{8.00}
\expandafter\def\csname ev@B/8B/g3/exam/cand-scheduled\endcsname{400}
\expandafter\def\csname ev@B/8B/g3/exam/cand-uncounted\endcsname{0}
\expandafter\def\csname ev@B/8B/g3/exam/diff-pp\endcsname{+0.00}
\expandafter\def\csname ev@B/8B/g3/exam/losses\endcsname{18}
\expandafter\def\csname ev@B/8B/g3/exam/outcome\endcsname{inconclusive}
\expandafter\def\csname ev@B/8B/g3/exam/p-recorded\endcsname{0.566}
\expandafter\def\csname ev@B/8B/g3/exam/protocol\endcsname{B}
\expandafter\def\csname ev@B/8B/g3/exam/task-losses-high\endcsname{18}
\expandafter\def\csname ev@B/8B/g3/exam/task-losses-low\endcsname{0}
\expandafter\def\csname ev@B/8B/g3/exam/task-p-high\endcsname{1}
\expandafter\def\csname ev@B/8B/g3/exam/task-p-low\endcsname{0.0053}
\expandafter\def\csname ev@B/8B/g3/exam/task-wins-high\endcsname{18}
\expandafter\def\csname ev@B/8B/g3/exam/task-wins-low\endcsname{0}
\expandafter\def\csname ev@B/8B/g3/exam/tasks\endcsname{100}
\expandafter\def\csname ev@B/8B/g3/exam/uncounted\endcsname{0}
\expandafter\def\csname ev@B/8B/g3/exam/uncounted-pct\endcsname{0.00}
\expandafter\def\csname ev@B/8B/g3/exam/unit\endcsname{attempt slot}
\expandafter\def\csname ev@B/8B/g3/exam/wins\endcsname{18}
\expandafter\def\csname ev@B/8B/g3/retention-g1/attempt-ci-hi-pp\endcsname{+10.94}
\expandafter\def\csname ev@B/8B/g3/retention-g1/attempt-ci-lo-pp\endcsname{-7.41}
\expandafter\def\csname ev@B/8B/g3/retention-g1/base-counted\endcsname{224}
\expandafter\def\csname ev@B/8B/g3/retention-g1/base-passes\endcsname{105}
\expandafter\def\csname ev@B/8B/g3/retention-g1/base-pct\endcsname{46.88}
\expandafter\def\csname ev@B/8B/g3/retention-g1/base-scheduled\endcsname{224}
\expandafter\def\csname ev@B/8B/g3/retention-g1/base-uncounted\endcsname{0}
\expandafter\def\csname ev@B/8B/g3/retention-g1/cand-counted\endcsname{224}
\expandafter\def\csname ev@B/8B/g3/retention-g1/cand-passes\endcsname{109}
\expandafter\def\csname ev@B/8B/g3/retention-g1/cand-pct\endcsname{48.66}
\expandafter\def\csname ev@B/8B/g3/retention-g1/cand-scheduled\endcsname{224}
\expandafter\def\csname ev@B/8B/g3/retention-g1/cand-uncounted\endcsname{0}
\expandafter\def\csname ev@B/8B/g3/retention-g1/diff-pp\endcsname{+1.79}
\expandafter\def\csname ev@B/8B/g3/retention-g1/gate-passed\endcsname{yes}
\expandafter\def\csname ev@B/8B/g3/retention-g1/losses\endcsname{14}
\expandafter\def\csname ev@B/8B/g3/retention-g1/outcome\endcsname{inconclusive}
\expandafter\def\csname ev@B/8B/g3/retention-g1/p-recorded\endcsname{0.298}
\expandafter\def\csname ev@B/8B/g3/retention-g1/protocol\endcsname{B}
\expandafter\def\csname ev@B/8B/g3/retention-g1/task-losses-high\endcsname{14}
\expandafter\def\csname ev@B/8B/g3/retention-g1/task-losses-low\endcsname{0}
\expandafter\def\csname ev@B/8B/g3/retention-g1/task-p-high\endcsname{0.99}
\expandafter\def\csname ev@B/8B/g3/retention-g1/task-p-low\endcsname{0.0022}
\expandafter\def\csname ev@B/8B/g3/retention-g1/task-wins-high\endcsname{18}
\expandafter\def\csname ev@B/8B/g3/retention-g1/task-wins-low\endcsname{1}
\expandafter\def\csname ev@B/8B/g3/retention-g1/tasks\endcsname{56}
\expandafter\def\csname ev@B/8B/g3/retention-g1/uncounted\endcsname{0}
\expandafter\def\csname ev@B/8B/g3/retention-g1/uncounted-pct\endcsname{0.00}
\expandafter\def\csname ev@B/8B/g3/retention-g1/unit\endcsname{attempt slot}
\expandafter\def\csname ev@B/8B/g3/retention-g1/wins\endcsname{18}
\expandafter\def\csname ev@B/8B/g3/retention-g2/attempt-ci-hi-pp\endcsname{+6.34}
\expandafter\def\csname ev@B/8B/g3/retention-g2/attempt-ci-lo-pp\endcsname{-11.60}
\expandafter\def\csname ev@B/8B/g3/retention-g2/base-counted\endcsname{188}
\expandafter\def\csname ev@B/8B/g3/retention-g2/base-passes\endcsname{54}
\expandafter\def\csname ev@B/8B/g3/retention-g2/base-pct\endcsname{28.72}
\expandafter\def\csname ev@B/8B/g3/retention-g2/base-scheduled\endcsname{188}
\expandafter\def\csname ev@B/8B/g3/retention-g2/base-uncounted\endcsname{0}
\expandafter\def\csname ev@B/8B/g3/retention-g2/cand-counted\endcsname{188}
\expandafter\def\csname ev@B/8B/g3/retention-g2/cand-passes\endcsname{49}
\expandafter\def\csname ev@B/8B/g3/retention-g2/cand-pct\endcsname{26.06}
\expandafter\def\csname ev@B/8B/g3/retention-g2/cand-scheduled\endcsname{188}
\expandafter\def\csname ev@B/8B/g3/retention-g2/cand-uncounted\endcsname{0}
\expandafter\def\csname ev@B/8B/g3/retention-g2/diff-pp\endcsname{-2.66}
\expandafter\def\csname ev@B/8B/g3/retention-g2/gate-passed\endcsname{yes}
\expandafter\def\csname ev@B/8B/g3/retention-g2/losses\endcsname{19}
\expandafter\def\csname ev@B/8B/g3/retention-g2/outcome\endcsname{inconclusive}
\expandafter\def\csname ev@B/8B/g3/retention-g2/p-recorded\endcsname{0.852}
\expandafter\def\csname ev@B/8B/g3/retention-g2/protocol\endcsname{B}
\expandafter\def\csname ev@B/8B/g3/retention-g2/task-losses-high\endcsname{19}
\expandafter\def\csname ev@B/8B/g3/retention-g2/task-losses-low\endcsname{2}
\expandafter\def\csname ev@B/8B/g3/retention-g2/task-p-high\endcsname{1}
\expandafter\def\csname ev@B/8B/g3/retention-g2/task-p-low\endcsname{0.032}
\expandafter\def\csname ev@B/8B/g3/retention-g2/task-wins-high\endcsname{14}
\expandafter\def\csname ev@B/8B/g3/retention-g2/task-wins-low\endcsname{0}
\expandafter\def\csname ev@B/8B/g3/retention-g2/tasks\endcsname{47}
\expandafter\def\csname ev@B/8B/g3/retention-g2/uncounted\endcsname{0}
\expandafter\def\csname ev@B/8B/g3/retention-g2/uncounted-pct\endcsname{0.00}
\expandafter\def\csname ev@B/8B/g3/retention-g2/unit\endcsname{attempt slot}
\expandafter\def\csname ev@B/8B/g3/retention-g2/wins\endcsname{14}
\expandafter\def\csname ev@B/8B/g3/target/attempt-ci-hi-pp\endcsname{+15.75}
\expandafter\def\csname ev@B/8B/g3/target/attempt-ci-lo-pp\endcsname{-1.60}
\expandafter\def\csname ev@B/8B/g3/target/base-counted\endcsname{196}
\expandafter\def\csname ev@B/8B/g3/target/base-passes\endcsname{45}
\expandafter\def\csname ev@B/8B/g3/target/base-pct\endcsname{22.96}
\expandafter\def\csname ev@B/8B/g3/target/base-scheduled\endcsname{196}
\expandafter\def\csname ev@B/8B/g3/target/base-uncounted\endcsname{0}
\expandafter\def\csname ev@B/8B/g3/target/cand-counted\endcsname{196}
\expandafter\def\csname ev@B/8B/g3/target/cand-passes\endcsname{59}
\expandafter\def\csname ev@B/8B/g3/target/cand-pct\endcsname{30.10}
\expandafter\def\csname ev@B/8B/g3/target/cand-scheduled\endcsname{196}
\expandafter\def\csname ev@B/8B/g3/target/cand-uncounted\endcsname{0}
\expandafter\def\csname ev@B/8B/g3/target/diff-pp\endcsname{+7.14}
\expandafter\def\csname ev@B/8B/g3/target/losses\endcsname{11}
\expandafter\def\csname ev@B/8B/g3/target/outcome\endcsname{inconclusive}
\expandafter\def\csname ev@B/8B/g3/target/p-recorded\endcsname{0.014}
\expandafter\def\csname ev@B/8B/g3/target/promoted\endcsname{yes}
\expandafter\def\csname ev@B/8B/g3/target/protocol\endcsname{B}
\expandafter\def\csname ev@B/8B/g3/target/task-losses-high\endcsname{11}
\expandafter\def\csname ev@B/8B/g3/target/task-losses-low\endcsname{0}
\expandafter\def\csname ev@B/8B/g3/target/task-p-high\endcsname{0.895}
\expandafter\def\csname ev@B/8B/g3/target/task-p-low\endcsname{\ensuremath{1.4\times 10^{-5}}}
\expandafter\def\csname ev@B/8B/g3/target/task-wins-high\endcsname{25}
\expandafter\def\csname ev@B/8B/g3/target/task-wins-low\endcsname{4}
\expandafter\def\csname ev@B/8B/g3/target/tasks\endcsname{49}
\expandafter\def\csname ev@B/8B/g3/target/uncounted\endcsname{0}
\expandafter\def\csname ev@B/8B/g3/target/uncounted-pct\endcsname{0.00}
\expandafter\def\csname ev@B/8B/g3/target/unit\endcsname{attempt slot}
\expandafter\def\csname ev@B/8B/g3/target/wins\endcsname{25}
\expandafter\def\csname ev@B/8B/generations\endcsname{3}
\expandafter\def\csname ev@B/8B/promoted-generations\endcsname{3}
\expandafter\def\csname ev@B/8B/rejected-generations\endcsname{0}
\expandafter\def\csname ev@arms/b/exam-vs-control/base-counted\endcsname{1200}
\expandafter\def\csname ev@arms/b/exam-vs-control/base-passes\endcsname{0}
\expandafter\def\csname ev@arms/b/exam-vs-control/base-pct\endcsname{0.00}
\expandafter\def\csname ev@arms/b/exam-vs-control/cand-counted\endcsname{1200}
\expandafter\def\csname ev@arms/b/exam-vs-control/cand-passes\endcsname{22}
\expandafter\def\csname ev@arms/b/exam-vs-control/cand-pct\endcsname{1.83}
\expandafter\def\csname ev@arms/b/exam-vs-control/ci-hi-pp\endcsname{+3.17}
\expandafter\def\csname ev@arms/b/exam-vs-control/ci-lo-pp\endcsname{+0.75}
\expandafter\def\csname ev@arms/b/exam-vs-control/diff-pp\endcsname{+1.83}
\expandafter\def\csname ev@arms/b/exam-vs-control/losses\endcsname{0}
\expandafter\def\csname ev@arms/b/exam-vs-control/outcome\endcsname{improvement supported}
\expandafter\def\csname ev@arms/b/exam-vs-control/p-recorded\endcsname{0.00098}
\expandafter\def\csname ev@arms/b/exam-vs-control/protocol\endcsname{arms}
\expandafter\def\csname ev@arms/b/exam-vs-control/task-p\endcsname{0.00098}
\expandafter\def\csname ev@arms/b/exam-vs-control/tasks\endcsname{100}
\expandafter\def\csname ev@arms/b/exam-vs-control/ties\endcsname{90}
\expandafter\def\csname ev@arms/b/exam-vs-control/uncounted\endcsname{11}
\expandafter\def\csname ev@arms/b/exam-vs-control/uncounted-pct\endcsname{1.38}
\expandafter\def\csname ev@arms/b/exam-vs-control/unit\endcsname{task}
\expandafter\def\csname ev@arms/b/exam-vs-control/wins\endcsname{10}
\expandafter\def\csname ev@arms/b/exam/base-counted\endcsname{1174}
\expandafter\def\csname ev@arms/b/exam/base-passes\endcsname{0}
\expandafter\def\csname ev@arms/b/exam/base-pct\endcsname{0.00}
\expandafter\def\csname ev@arms/b/exam/base-scheduled\endcsname{1200}
\expandafter\def\csname ev@arms/b/exam/base-uncounted\endcsname{26}
\expandafter\def\csname ev@arms/b/exam/base-uncounted-pct\endcsname{2.17}
\expandafter\def\csname ev@arms/b/exam/cand-counted\endcsname{1200}
\expandafter\def\csname ev@arms/b/exam/cand-passes\endcsname{22}
\expandafter\def\csname ev@arms/b/exam/cand-pct\endcsname{1.83}
\expandafter\def\csname ev@arms/b/exam/ci-hi-pp\endcsname{+3.19}
\expandafter\def\csname ev@arms/b/exam/ci-lo-pp\endcsname{+0.76}
\expandafter\def\csname ev@arms/b/exam/diff-pp\endcsname{+1.87}
\expandafter\def\csname ev@arms/b/exam/losses\endcsname{0}
\expandafter\def\csname ev@arms/b/exam/outcome\endcsname{improvement supported}
\expandafter\def\csname ev@arms/b/exam/p-recorded\endcsname{0.00098}
\expandafter\def\csname ev@arms/b/exam/protocol\endcsname{arms}
\expandafter\def\csname ev@arms/b/exam/task-p\endcsname{0.00098}
\expandafter\def\csname ev@arms/b/exam/tasks\endcsname{100}
\expandafter\def\csname ev@arms/b/exam/tasks-solved\endcsname{10}
\expandafter\def\csname ev@arms/b/exam/ties\endcsname{90}
\expandafter\def\csname ev@arms/b/exam/uncounted\endcsname{11}
\expandafter\def\csname ev@arms/b/exam/uncounted-pct\endcsname{1.38}
\expandafter\def\csname ev@arms/b/exam/unit\endcsname{task}
\expandafter\def\csname ev@arms/b/exam/wins\endcsname{10}
\expandafter\def\csname ev@arms/b/target-vs-control/base-counted\endcsname{1128}
\expandafter\def\csname ev@arms/b/target-vs-control/base-passes\endcsname{9}
\expandafter\def\csname ev@arms/b/target-vs-control/base-pct\endcsname{0.80}
\expandafter\def\csname ev@arms/b/target-vs-control/cand-counted\endcsname{1128}
\expandafter\def\csname ev@arms/b/target-vs-control/cand-passes\endcsname{705}
\expandafter\def\csname ev@arms/b/target-vs-control/cand-pct\endcsname{62.50}
\expandafter\def\csname ev@arms/b/target-vs-control/ci-hi-pp\endcsname{+68.26}
\expandafter\def\csname ev@arms/b/target-vs-control/ci-lo-pp\endcsname{+54.88}
\expandafter\def\csname ev@arms/b/target-vs-control/diff-pp\endcsname{+61.70}
\expandafter\def\csname ev@arms/b/target-vs-control/losses\endcsname{0}
\expandafter\def\csname ev@arms/b/target-vs-control/outcome\endcsname{improvement supported}
\expandafter\def\csname ev@arms/b/target-vs-control/p-recorded\endcsname{\ensuremath{1.3\times 10^{-26}}}
\expandafter\def\csname ev@arms/b/target-vs-control/protocol\endcsname{arms}
\expandafter\def\csname ev@arms/b/target-vs-control/task-p\endcsname{\ensuremath{1.3\times 10^{-26}}}
\expandafter\def\csname ev@arms/b/target-vs-control/tasks\endcsname{94}
\expandafter\def\csname ev@arms/b/target-vs-control/ties\endcsname{8}
\expandafter\def\csname ev@arms/b/target-vs-control/uncounted\endcsname{0}
\expandafter\def\csname ev@arms/b/target-vs-control/uncounted-pct\endcsname{0.00}
\expandafter\def\csname ev@arms/b/target-vs-control/unit\endcsname{task}
\expandafter\def\csname ev@arms/b/target-vs-control/wins\endcsname{86}
\expandafter\def\csname ev@arms/b/target/base-counted\endcsname{1128}
\expandafter\def\csname ev@arms/b/target/base-passes\endcsname{126}
\expandafter\def\csname ev@arms/b/target/base-pct\endcsname{11.17}
\expandafter\def\csname ev@arms/b/target/base-scheduled\endcsname{1128}
\expandafter\def\csname ev@arms/b/target/base-uncounted\endcsname{0}
\expandafter\def\csname ev@arms/b/target/base-uncounted-pct\endcsname{0.00}
\expandafter\def\csname ev@arms/b/target/cand-counted\endcsname{1128}
\expandafter\def\csname ev@arms/b/target/cand-passes\endcsname{705}
\expandafter\def\csname ev@arms/b/target/cand-pct\endcsname{62.50}
\expandafter\def\csname ev@arms/b/target/ci-hi-pp\endcsname{+58.16}
\expandafter\def\csname ev@arms/b/target/ci-lo-pp\endcsname{+44.68}
\expandafter\def\csname ev@arms/b/target/diff-pp\endcsname{+51.33}
\expandafter\def\csname ev@arms/b/target/losses\endcsname{1}
\expandafter\def\csname ev@arms/b/target/outcome\endcsname{improvement supported}
\expandafter\def\csname ev@arms/b/target/p-recorded\endcsname{\ensuremath{8.7\times 10^{-24}}}
\expandafter\def\csname ev@arms/b/target/protocol\endcsname{arms}
\expandafter\def\csname ev@arms/b/target/task-p\endcsname{\ensuremath{8.7\times 10^{-24}}}
\expandafter\def\csname ev@arms/b/target/tasks\endcsname{94}
\expandafter\def\csname ev@arms/b/target/tasks-solved\endcsname{86}
\expandafter\def\csname ev@arms/b/target/ties\endcsname{11}
\expandafter\def\csname ev@arms/b/target/uncounted\endcsname{0}
\expandafter\def\csname ev@arms/b/target/uncounted-pct\endcsname{0.00}
\expandafter\def\csname ev@arms/b/target/unit\endcsname{task}
\expandafter\def\csname ev@arms/b/target/wins\endcsname{82}
\expandafter\def\csname ev@arms/c-vs-b/exam/base-counted\endcsname{1200}
\expandafter\def\csname ev@arms/c-vs-b/exam/base-passes\endcsname{22}
\expandafter\def\csname ev@arms/c-vs-b/exam/base-pct\endcsname{1.83}
\expandafter\def\csname ev@arms/c-vs-b/exam/cand-counted\endcsname{1200}
\expandafter\def\csname ev@arms/c-vs-b/exam/cand-passes\endcsname{10}
\expandafter\def\csname ev@arms/c-vs-b/exam/cand-pct\endcsname{0.83}
\expandafter\def\csname ev@arms/c-vs-b/exam/ci-hi-pp\endcsname{-0.17}
\expandafter\def\csname ev@arms/c-vs-b/exam/ci-lo-pp\endcsname{-1.92}
\expandafter\def\csname ev@arms/c-vs-b/exam/diff-pp\endcsname{-1.00}
\expandafter\def\csname ev@arms/c-vs-b/exam/losses\endcsname{8}
\expandafter\def\csname ev@arms/c-vs-b/exam/outcome\endcsname{regression supported}
\expandafter\def\csname ev@arms/c-vs-b/exam/p-recorded\endcsname{0.989}
\expandafter\def\csname ev@arms/c-vs-b/exam/p-two-sided\endcsname{0.109}
\expandafter\def\csname ev@arms/c-vs-b/exam/protocol\endcsname{arms}
\expandafter\def\csname ev@arms/c-vs-b/exam/task-p\endcsname{0.989}
\expandafter\def\csname ev@arms/c-vs-b/exam/tasks\endcsname{100}
\expandafter\def\csname ev@arms/c-vs-b/exam/ties\endcsname{90}
\expandafter\def\csname ev@arms/c-vs-b/exam/uncounted\endcsname{6}
\expandafter\def\csname ev@arms/c-vs-b/exam/uncounted-pct\endcsname{0.75}
\expandafter\def\csname ev@arms/c-vs-b/exam/unit\endcsname{task}
\expandafter\def\csname ev@arms/c-vs-b/exam/wins\endcsname{2}
\expandafter\def\csname ev@arms/c-vs-b/target/base-counted\endcsname{1128}
\expandafter\def\csname ev@arms/c-vs-b/target/base-passes\endcsname{705}
\expandafter\def\csname ev@arms/c-vs-b/target/base-pct\endcsname{62.50}
\expandafter\def\csname ev@arms/c-vs-b/target/cand-counted\endcsname{1128}
\expandafter\def\csname ev@arms/c-vs-b/target/cand-passes\endcsname{642}
\expandafter\def\csname ev@arms/c-vs-b/target/cand-pct\endcsname{56.91}
\expandafter\def\csname ev@arms/c-vs-b/target/ci-hi-pp\endcsname{-1.60}
\expandafter\def\csname ev@arms/c-vs-b/target/ci-lo-pp\endcsname{-9.84}
\expandafter\def\csname ev@arms/c-vs-b/target/diff-pp\endcsname{-5.59}
\expandafter\def\csname ev@arms/c-vs-b/target/losses\endcsname{42}
\expandafter\def\csname ev@arms/c-vs-b/target/outcome\endcsname{regression supported}
\expandafter\def\csname ev@arms/c-vs-b/target/p-recorded\endcsname{0.991}
\expandafter\def\csname ev@arms/c-vs-b/target/p-two-sided\endcsname{0.036}
\expandafter\def\csname ev@arms/c-vs-b/target/protocol\endcsname{arms}
\expandafter\def\csname ev@arms/c-vs-b/target/task-p\endcsname{0.991}
\expandafter\def\csname ev@arms/c-vs-b/target/tasks\endcsname{94}
\expandafter\def\csname ev@arms/c-vs-b/target/ties\endcsname{28}
\expandafter\def\csname ev@arms/c-vs-b/target/uncounted\endcsname{2}
\expandafter\def\csname ev@arms/c-vs-b/target/uncounted-pct\endcsname{0.27}
\expandafter\def\csname ev@arms/c-vs-b/target/unit\endcsname{task}
\expandafter\def\csname ev@arms/c-vs-b/target/wins\endcsname{24}
\expandafter\def\csname ev@arms/c/exam-vs-control/base-counted\endcsname{1200}
\expandafter\def\csname ev@arms/c/exam-vs-control/base-passes\endcsname{0}
\expandafter\def\csname ev@arms/c/exam-vs-control/base-pct\endcsname{0.00}
\expandafter\def\csname ev@arms/c/exam-vs-control/cand-counted\endcsname{1200}
\expandafter\def\csname ev@arms/c/exam-vs-control/cand-passes\endcsname{10}
\expandafter\def\csname ev@arms/c/exam-vs-control/cand-pct\endcsname{0.83}
\expandafter\def\csname ev@arms/c/exam-vs-control/ci-hi-pp\endcsname{+1.67}
\expandafter\def\csname ev@arms/c/exam-vs-control/ci-lo-pp\endcsname{+0.25}
\expandafter\def\csname ev@arms/c/exam-vs-control/diff-pp\endcsname{+0.83}
\expandafter\def\csname ev@arms/c/exam-vs-control/losses\endcsname{0}
\expandafter\def\csname ev@arms/c/exam-vs-control/outcome\endcsname{improvement supported}
\expandafter\def\csname ev@arms/c/exam-vs-control/p-recorded\endcsname{0.016}
\expandafter\def\csname ev@arms/c/exam-vs-control/protocol\endcsname{arms}
\expandafter\def\csname ev@arms/c/exam-vs-control/task-p\endcsname{0.016}
\expandafter\def\csname ev@arms/c/exam-vs-control/tasks\endcsname{100}
\expandafter\def\csname ev@arms/c/exam-vs-control/ties\endcsname{94}
\expandafter\def\csname ev@arms/c/exam-vs-control/uncounted\endcsname{6}
\expandafter\def\csname ev@arms/c/exam-vs-control/uncounted-pct\endcsname{0.75}
\expandafter\def\csname ev@arms/c/exam-vs-control/unit\endcsname{task}
\expandafter\def\csname ev@arms/c/exam-vs-control/wins\endcsname{6}
\expandafter\def\csname ev@arms/c/exam/base-counted\endcsname{1188}
\expandafter\def\csname ev@arms/c/exam/base-passes\endcsname{0}
\expandafter\def\csname ev@arms/c/exam/base-pct\endcsname{0.00}
\expandafter\def\csname ev@arms/c/exam/base-scheduled\endcsname{1200}
\expandafter\def\csname ev@arms/c/exam/base-uncounted\endcsname{12}
\expandafter\def\csname ev@arms/c/exam/base-uncounted-pct\endcsname{1.00}
\expandafter\def\csname ev@arms/c/exam/cand-counted\endcsname{1200}
\expandafter\def\csname ev@arms/c/exam/cand-passes\endcsname{10}
\expandafter\def\csname ev@arms/c/exam/cand-pct\endcsname{0.83}
\expandafter\def\csname ev@arms/c/exam/ci-hi-pp\endcsname{+1.67}
\expandafter\def\csname ev@arms/c/exam/ci-lo-pp\endcsname{+0.25}
\expandafter\def\csname ev@arms/c/exam/diff-pp\endcsname{+0.84}
\expandafter\def\csname ev@arms/c/exam/losses\endcsname{0}
\expandafter\def\csname ev@arms/c/exam/outcome\endcsname{improvement supported}
\expandafter\def\csname ev@arms/c/exam/p-recorded\endcsname{0.016}
\expandafter\def\csname ev@arms/c/exam/protocol\endcsname{arms}
\expandafter\def\csname ev@arms/c/exam/task-p\endcsname{0.016}
\expandafter\def\csname ev@arms/c/exam/tasks\endcsname{100}
\expandafter\def\csname ev@arms/c/exam/tasks-solved\endcsname{6}
\expandafter\def\csname ev@arms/c/exam/ties\endcsname{94}
\expandafter\def\csname ev@arms/c/exam/uncounted\endcsname{6}
\expandafter\def\csname ev@arms/c/exam/uncounted-pct\endcsname{0.75}
\expandafter\def\csname ev@arms/c/exam/unit\endcsname{task}
\expandafter\def\csname ev@arms/c/exam/wins\endcsname{6}
\expandafter\def\csname ev@arms/c/target-vs-control/base-counted\endcsname{1128}
\expandafter\def\csname ev@arms/c/target-vs-control/base-passes\endcsname{9}
\expandafter\def\csname ev@arms/c/target-vs-control/base-pct\endcsname{0.80}
\expandafter\def\csname ev@arms/c/target-vs-control/cand-counted\endcsname{1128}
\expandafter\def\csname ev@arms/c/target-vs-control/cand-passes\endcsname{642}
\expandafter\def\csname ev@arms/c/target-vs-control/cand-pct\endcsname{56.91}
\expandafter\def\csname ev@arms/c/target-vs-control/ci-hi-pp\endcsname{+63.21}
\expandafter\def\csname ev@arms/c/target-vs-control/ci-lo-pp\endcsname{+49.02}
\expandafter\def\csname ev@arms/c/target-vs-control/diff-pp\endcsname{+56.12}
\expandafter\def\csname ev@arms/c/target-vs-control/losses\endcsname{0}
\expandafter\def\csname ev@arms/c/target-vs-control/outcome\endcsname{improvement supported}
\expandafter\def\csname ev@arms/c/target-vs-control/p-recorded\endcsname{\ensuremath{5.2\times 10^{-26}}}
\expandafter\def\csname ev@arms/c/target-vs-control/protocol\endcsname{arms}
\expandafter\def\csname ev@arms/c/target-vs-control/task-p\endcsname{\ensuremath{5.2\times 10^{-26}}}
\expandafter\def\csname ev@arms/c/target-vs-control/tasks\endcsname{94}
\expandafter\def\csname ev@arms/c/target-vs-control/ties\endcsname{10}
\expandafter\def\csname ev@arms/c/target-vs-control/uncounted\endcsname{2}
\expandafter\def\csname ev@arms/c/target-vs-control/uncounted-pct\endcsname{0.27}
\expandafter\def\csname ev@arms/c/target-vs-control/unit\endcsname{task}
\expandafter\def\csname ev@arms/c/target-vs-control/wins\endcsname{84}
\expandafter\def\csname ev@arms/c/target/base-counted\endcsname{1126}
\expandafter\def\csname ev@arms/c/target/base-passes\endcsname{113}
\expandafter\def\csname ev@arms/c/target/base-pct\endcsname{10.04}
\expandafter\def\csname ev@arms/c/target/base-scheduled\endcsname{1128}
\expandafter\def\csname ev@arms/c/target/base-uncounted\endcsname{2}
\expandafter\def\csname ev@arms/c/target/base-uncounted-pct\endcsname{0.18}
\expandafter\def\csname ev@arms/c/target/cand-counted\endcsname{1128}
\expandafter\def\csname ev@arms/c/target/cand-passes\endcsname{642}
\expandafter\def\csname ev@arms/c/target/cand-pct\endcsname{56.91}
\expandafter\def\csname ev@arms/c/target/ci-hi-pp\endcsname{+54.17}
\expandafter\def\csname ev@arms/c/target/ci-lo-pp\endcsname{+39.72}
\expandafter\def\csname ev@arms/c/target/diff-pp\endcsname{+46.98}
\expandafter\def\csname ev@arms/c/target/losses\endcsname{3}
\expandafter\def\csname ev@arms/c/target/outcome\endcsname{improvement supported}
\expandafter\def\csname ev@arms/c/target/p-recorded\endcsname{\ensuremath{1.9\times 10^{-20}}}
\expandafter\def\csname ev@arms/c/target/protocol\endcsname{arms}
\expandafter\def\csname ev@arms/c/target/task-p\endcsname{\ensuremath{1.9\times 10^{-20}}}
\expandafter\def\csname ev@arms/c/target/tasks\endcsname{94}
\expandafter\def\csname ev@arms/c/target/tasks-solved\endcsname{84}
\expandafter\def\csname ev@arms/c/target/ties\endcsname{12}
\expandafter\def\csname ev@arms/c/target/uncounted\endcsname{2}
\expandafter\def\csname ev@arms/c/target/uncounted-pct\endcsname{0.27}
\expandafter\def\csname ev@arms/c/target/unit\endcsname{task}
\expandafter\def\csname ev@arms/c/target/wins\endcsname{79}
\expandafter\def\csname ev@arms/d-vs-b/exam/base-counted\endcsname{1200}
\expandafter\def\csname ev@arms/d-vs-b/exam/base-passes\endcsname{22}
\expandafter\def\csname ev@arms/d-vs-b/exam/base-pct\endcsname{1.83}
\expandafter\def\csname ev@arms/d-vs-b/exam/cand-counted\endcsname{1200}
\expandafter\def\csname ev@arms/d-vs-b/exam/cand-passes\endcsname{22}
\expandafter\def\csname ev@arms/d-vs-b/exam/cand-pct\endcsname{1.83}
\expandafter\def\csname ev@arms/d-vs-b/exam/ci-hi-pp\endcsname{+0.83}
\expandafter\def\csname ev@arms/d-vs-b/exam/ci-lo-pp\endcsname{-0.83}
\expandafter\def\csname ev@arms/d-vs-b/exam/diff-pp\endcsname{+0.00}
\expandafter\def\csname ev@arms/d-vs-b/exam/losses\endcsname{5}
\expandafter\def\csname ev@arms/d-vs-b/exam/outcome\endcsname{inconclusive}
\expandafter\def\csname ev@arms/d-vs-b/exam/p-recorded\endcsname{0.387}
\expandafter\def\csname ev@arms/d-vs-b/exam/p-two-sided\endcsname{0.774}
\expandafter\def\csname ev@arms/d-vs-b/exam/protocol\endcsname{arms}
\expandafter\def\csname ev@arms/d-vs-b/exam/task-p\endcsname{0.387}
\expandafter\def\csname ev@arms/d-vs-b/exam/tasks\endcsname{100}
\expandafter\def\csname ev@arms/d-vs-b/exam/ties\endcsname{88}
\expandafter\def\csname ev@arms/d-vs-b/exam/uncounted\endcsname{5}
\expandafter\def\csname ev@arms/d-vs-b/exam/uncounted-pct\endcsname{0.62}
\expandafter\def\csname ev@arms/d-vs-b/exam/unit\endcsname{task}
\expandafter\def\csname ev@arms/d-vs-b/exam/wins\endcsname{7}
\expandafter\def\csname ev@arms/d-vs-b/target/base-counted\endcsname{1128}
\expandafter\def\csname ev@arms/d-vs-b/target/base-passes\endcsname{705}
\expandafter\def\csname ev@arms/d-vs-b/target/base-pct\endcsname{62.50}
\expandafter\def\csname ev@arms/d-vs-b/target/cand-counted\endcsname{1128}
\expandafter\def\csname ev@arms/d-vs-b/target/cand-passes\endcsname{723}
\expandafter\def\csname ev@arms/d-vs-b/target/cand-pct\endcsname{64.10}
\expandafter\def\csname ev@arms/d-vs-b/target/ci-hi-pp\endcsname{+4.96}
\expandafter\def\csname ev@arms/d-vs-b/target/ci-lo-pp\endcsname{-1.77}
\expandafter\def\csname ev@arms/d-vs-b/target/diff-pp\endcsname{+1.60}
\expandafter\def\csname ev@arms/d-vs-b/target/losses\endcsname{24}
\expandafter\def\csname ev@arms/d-vs-b/target/outcome\endcsname{inconclusive}
\expandafter\def\csname ev@arms/d-vs-b/target/p-recorded\endcsname{0.062}
\expandafter\def\csname ev@arms/d-vs-b/target/p-two-sided\endcsname{0.124}
\expandafter\def\csname ev@arms/d-vs-b/target/protocol\endcsname{arms}
\expandafter\def\csname ev@arms/d-vs-b/target/task-p\endcsname{0.062}
\expandafter\def\csname ev@arms/d-vs-b/target/tasks\endcsname{94}
\expandafter\def\csname ev@arms/d-vs-b/target/ties\endcsname{33}
\expandafter\def\csname ev@arms/d-vs-b/target/uncounted\endcsname{0}
\expandafter\def\csname ev@arms/d-vs-b/target/uncounted-pct\endcsname{0.00}
\expandafter\def\csname ev@arms/d-vs-b/target/unit\endcsname{task}
\expandafter\def\csname ev@arms/d-vs-b/target/wins\endcsname{37}
\expandafter\def\csname ev@arms/d/exam-vs-control/base-counted\endcsname{1200}
\expandafter\def\csname ev@arms/d/exam-vs-control/base-passes\endcsname{0}
\expandafter\def\csname ev@arms/d/exam-vs-control/base-pct\endcsname{0.00}
\expandafter\def\csname ev@arms/d/exam-vs-control/cand-counted\endcsname{1200}
\expandafter\def\csname ev@arms/d/exam-vs-control/cand-passes\endcsname{22}
\expandafter\def\csname ev@arms/d/exam-vs-control/cand-pct\endcsname{1.83}
\expandafter\def\csname ev@arms/d/exam-vs-control/ci-hi-pp\endcsname{+3.17}
\expandafter\def\csname ev@arms/d/exam-vs-control/ci-lo-pp\endcsname{+0.75}
\expandafter\def\csname ev@arms/d/exam-vs-control/diff-pp\endcsname{+1.83}
\expandafter\def\csname ev@arms/d/exam-vs-control/losses\endcsname{0}
\expandafter\def\csname ev@arms/d/exam-vs-control/outcome\endcsname{improvement supported}
\expandafter\def\csname ev@arms/d/exam-vs-control/p-recorded\endcsname{0.00049}
\expandafter\def\csname ev@arms/d/exam-vs-control/protocol\endcsname{arms}
\expandafter\def\csname ev@arms/d/exam-vs-control/task-p\endcsname{0.00049}
\expandafter\def\csname ev@arms/d/exam-vs-control/tasks\endcsname{100}
\expandafter\def\csname ev@arms/d/exam-vs-control/ties\endcsname{89}
\expandafter\def\csname ev@arms/d/exam-vs-control/uncounted\endcsname{5}
\expandafter\def\csname ev@arms/d/exam-vs-control/uncounted-pct\endcsname{0.62}
\expandafter\def\csname ev@arms/d/exam-vs-control/unit\endcsname{task}
\expandafter\def\csname ev@arms/d/exam-vs-control/wins\endcsname{11}
\expandafter\def\csname ev@arms/d/exam/base-counted\endcsname{1187}
\expandafter\def\csname ev@arms/d/exam/base-passes\endcsname{0}
\expandafter\def\csname ev@arms/d/exam/base-pct\endcsname{0.00}
\expandafter\def\csname ev@arms/d/exam/base-scheduled\endcsname{1200}
\expandafter\def\csname ev@arms/d/exam/base-uncounted\endcsname{13}
\expandafter\def\csname ev@arms/d/exam/base-uncounted-pct\endcsname{1.08}
\expandafter\def\csname ev@arms/d/exam/cand-counted\endcsname{1200}
\expandafter\def\csname ev@arms/d/exam/cand-passes\endcsname{22}
\expandafter\def\csname ev@arms/d/exam/cand-pct\endcsname{1.83}
\expandafter\def\csname ev@arms/d/exam/ci-hi-pp\endcsname{+3.20}
\expandafter\def\csname ev@arms/d/exam/ci-lo-pp\endcsname{+0.75}
\expandafter\def\csname ev@arms/d/exam/diff-pp\endcsname{+1.85}
\expandafter\def\csname ev@arms/d/exam/losses\endcsname{0}
\expandafter\def\csname ev@arms/d/exam/outcome\endcsname{improvement supported}
\expandafter\def\csname ev@arms/d/exam/p-recorded\endcsname{0.00049}
\expandafter\def\csname ev@arms/d/exam/protocol\endcsname{arms}
\expandafter\def\csname ev@arms/d/exam/task-p\endcsname{0.00049}
\expandafter\def\csname ev@arms/d/exam/tasks\endcsname{100}
\expandafter\def\csname ev@arms/d/exam/tasks-solved\endcsname{11}
\expandafter\def\csname ev@arms/d/exam/ties\endcsname{89}
\expandafter\def\csname ev@arms/d/exam/uncounted\endcsname{5}
\expandafter\def\csname ev@arms/d/exam/uncounted-pct\endcsname{0.62}
\expandafter\def\csname ev@arms/d/exam/unit\endcsname{task}
\expandafter\def\csname ev@arms/d/exam/wins\endcsname{11}
\expandafter\def\csname ev@arms/d/target-vs-control/base-counted\endcsname{1128}
\expandafter\def\csname ev@arms/d/target-vs-control/base-passes\endcsname{9}
\expandafter\def\csname ev@arms/d/target-vs-control/base-pct\endcsname{0.80}
\expandafter\def\csname ev@arms/d/target-vs-control/cand-counted\endcsname{1128}
\expandafter\def\csname ev@arms/d/target-vs-control/cand-passes\endcsname{723}
\expandafter\def\csname ev@arms/d/target-vs-control/cand-pct\endcsname{64.10}
\expandafter\def\csname ev@arms/d/target-vs-control/ci-hi-pp\endcsname{+69.95}
\expandafter\def\csname ev@arms/d/target-vs-control/ci-lo-pp\endcsname{+56.38}
\expandafter\def\csname ev@arms/d/target-vs-control/diff-pp\endcsname{+63.30}
\expandafter\def\csname ev@arms/d/target-vs-control/losses\endcsname{0}
\expandafter\def\csname ev@arms/d/target-vs-control/outcome\endcsname{improvement supported}
\expandafter\def\csname ev@arms/d/target-vs-control/p-recorded\endcsname{\ensuremath{6.5\times 10^{-27}}}
\expandafter\def\csname ev@arms/d/target-vs-control/protocol\endcsname{arms}
\expandafter\def\csname ev@arms/d/target-vs-control/task-p\endcsname{\ensuremath{6.5\times 10^{-27}}}
\expandafter\def\csname ev@arms/d/target-vs-control/tasks\endcsname{94}
\expandafter\def\csname ev@arms/d/target-vs-control/ties\endcsname{7}
\expandafter\def\csname ev@arms/d/target-vs-control/uncounted\endcsname{0}
\expandafter\def\csname ev@arms/d/target-vs-control/uncounted-pct\endcsname{0.00}
\expandafter\def\csname ev@arms/d/target-vs-control/unit\endcsname{task}
\expandafter\def\csname ev@arms/d/target-vs-control/wins\endcsname{87}
\expandafter\def\csname ev@arms/d/target/base-counted\endcsname{1128}
\expandafter\def\csname ev@arms/d/target/base-passes\endcsname{119}
\expandafter\def\csname ev@arms/d/target/base-pct\endcsname{10.55}
\expandafter\def\csname ev@arms/d/target/base-scheduled\endcsname{1128}
\expandafter\def\csname ev@arms/d/target/base-uncounted\endcsname{0}
\expandafter\def\csname ev@arms/d/target/base-uncounted-pct\endcsname{0.00}
\expandafter\def\csname ev@arms/d/target/cand-counted\endcsname{1128}
\expandafter\def\csname ev@arms/d/target/cand-passes\endcsname{723}
\expandafter\def\csname ev@arms/d/target/cand-pct\endcsname{64.10}
\expandafter\def\csname ev@arms/d/target/ci-hi-pp\endcsname{+60.46}
\expandafter\def\csname ev@arms/d/target/ci-lo-pp\endcsname{+46.28}
\expandafter\def\csname ev@arms/d/target/diff-pp\endcsname{+53.55}
\expandafter\def\csname ev@arms/d/target/losses\endcsname{1}
\expandafter\def\csname ev@arms/d/target/outcome\endcsname{improvement supported}
\expandafter\def\csname ev@arms/d/target/p-recorded\endcsname{\ensuremath{2.2\times 10^{-24}}}
\expandafter\def\csname ev@arms/d/target/protocol\endcsname{arms}
\expandafter\def\csname ev@arms/d/target/task-p\endcsname{\ensuremath{2.2\times 10^{-24}}}
\expandafter\def\csname ev@arms/d/target/tasks\endcsname{94}
\expandafter\def\csname ev@arms/d/target/tasks-solved\endcsname{87}
\expandafter\def\csname ev@arms/d/target/ties\endcsname{9}
\expandafter\def\csname ev@arms/d/target/uncounted\endcsname{0}
\expandafter\def\csname ev@arms/d/target/uncounted-pct\endcsname{0.00}
\expandafter\def\csname ev@arms/d/target/unit\endcsname{task}
\expandafter\def\csname ev@arms/d/target/wins\endcsname{84}
\expandafter\def\csname ev@arms/f-vs-base/exam/base-counted\endcsname{1177}
\expandafter\def\csname ev@arms/f-vs-base/exam/base-passes\endcsname{0}
\expandafter\def\csname ev@arms/f-vs-base/exam/base-pct\endcsname{0.00}
\expandafter\def\csname ev@arms/f-vs-base/exam/cand-counted\endcsname{1200}
\expandafter\def\csname ev@arms/f-vs-base/exam/cand-passes\endcsname{0}
\expandafter\def\csname ev@arms/f-vs-base/exam/cand-pct\endcsname{0.00}
\expandafter\def\csname ev@arms/f-vs-base/exam/ci-hi-pp\endcsname{+0.00}
\expandafter\def\csname ev@arms/f-vs-base/exam/ci-lo-pp\endcsname{+0.00}
\expandafter\def\csname ev@arms/f-vs-base/exam/diff-pp\endcsname{+0.00}
\expandafter\def\csname ev@arms/f-vs-base/exam/losses\endcsname{0}
\expandafter\def\csname ev@arms/f-vs-base/exam/outcome\endcsname{inconclusive}
\expandafter\def\csname ev@arms/f-vs-base/exam/p-recorded\endcsname{1}
\expandafter\def\csname ev@arms/f-vs-base/exam/p-two-sided\endcsname{1}
\expandafter\def\csname ev@arms/f-vs-base/exam/protocol\endcsname{arms}
\expandafter\def\csname ev@arms/f-vs-base/exam/task-p\endcsname{1}
\expandafter\def\csname ev@arms/f-vs-base/exam/tasks\endcsname{100}
\expandafter\def\csname ev@arms/f-vs-base/exam/ties\endcsname{100}
\expandafter\def\csname ev@arms/f-vs-base/exam/uncounted\endcsname{11}
\expandafter\def\csname ev@arms/f-vs-base/exam/uncounted-pct\endcsname{1.38}
\expandafter\def\csname ev@arms/f-vs-base/exam/unit\endcsname{task}
\expandafter\def\csname ev@arms/f-vs-base/exam/wins\endcsname{0}
\expandafter\def\csname ev@arms/f-vs-base/target/base-counted\endcsname{1128}
\expandafter\def\csname ev@arms/f-vs-base/target/base-passes\endcsname{133}
\expandafter\def\csname ev@arms/f-vs-base/target/base-pct\endcsname{11.79}
\expandafter\def\csname ev@arms/f-vs-base/target/cand-counted\endcsname{1128}
\expandafter\def\csname ev@arms/f-vs-base/target/cand-passes\endcsname{9}
\expandafter\def\csname ev@arms/f-vs-base/target/cand-pct\endcsname{0.80}
\expandafter\def\csname ev@arms/f-vs-base/target/ci-hi-pp\endcsname{-6.38}
\expandafter\def\csname ev@arms/f-vs-base/target/ci-lo-pp\endcsname{-16.31}
\expandafter\def\csname ev@arms/f-vs-base/target/diff-pp\endcsname{-10.99}
\expandafter\def\csname ev@arms/f-vs-base/target/losses\endcsname{26}
\expandafter\def\csname ev@arms/f-vs-base/target/outcome\endcsname{regression supported}
\expandafter\def\csname ev@arms/f-vs-base/target/p-recorded\endcsname{1}
\expandafter\def\csname ev@arms/f-vs-base/target/p-two-sided\endcsname{\ensuremath{3.0\times 10^{-6}}}
\expandafter\def\csname ev@arms/f-vs-base/target/protocol\endcsname{arms}
\expandafter\def\csname ev@arms/f-vs-base/target/task-p\endcsname{1}
\expandafter\def\csname ev@arms/f-vs-base/target/tasks\endcsname{94}
\expandafter\def\csname ev@arms/f-vs-base/target/ties\endcsname{66}
\expandafter\def\csname ev@arms/f-vs-base/target/uncounted\endcsname{0}
\expandafter\def\csname ev@arms/f-vs-base/target/uncounted-pct\endcsname{0.00}
\expandafter\def\csname ev@arms/f-vs-base/target/unit\endcsname{task}
\expandafter\def\csname ev@arms/f-vs-base/target/wins\endcsname{2}
\expandafter\def\csname ev@arms/g-vs-b/exam/base-counted\endcsname{1200}
\expandafter\def\csname ev@arms/g-vs-b/exam/base-passes\endcsname{22}
\expandafter\def\csname ev@arms/g-vs-b/exam/base-pct\endcsname{1.83}
\expandafter\def\csname ev@arms/g-vs-b/exam/cand-counted\endcsname{1200}
\expandafter\def\csname ev@arms/g-vs-b/exam/cand-passes\endcsname{42}
\expandafter\def\csname ev@arms/g-vs-b/exam/cand-pct\endcsname{3.50}
\expandafter\def\csname ev@arms/g-vs-b/exam/ci-hi-pp\endcsname{+3.08}
\expandafter\def\csname ev@arms/g-vs-b/exam/ci-lo-pp\endcsname{+0.25}
\expandafter\def\csname ev@arms/g-vs-b/exam/diff-pp\endcsname{+1.67}
\expandafter\def\csname ev@arms/g-vs-b/exam/losses\endcsname{5}
\expandafter\def\csname ev@arms/g-vs-b/exam/outcome\endcsname{improvement supported}
\expandafter\def\csname ev@arms/g-vs-b/exam/p-recorded\endcsname{0.0012}
\expandafter\def\csname ev@arms/g-vs-b/exam/p-two-sided\endcsname{0.0025}
\expandafter\def\csname ev@arms/g-vs-b/exam/protocol\endcsname{arms}
\expandafter\def\csname ev@arms/g-vs-b/exam/task-p\endcsname{0.0012}
\expandafter\def\csname ev@arms/g-vs-b/exam/tasks\endcsname{100}
\expandafter\def\csname ev@arms/g-vs-b/exam/ties\endcsname{74}
\expandafter\def\csname ev@arms/g-vs-b/exam/uncounted\endcsname{10}
\expandafter\def\csname ev@arms/g-vs-b/exam/uncounted-pct\endcsname{1.25}
\expandafter\def\csname ev@arms/g-vs-b/exam/unit\endcsname{task}
\expandafter\def\csname ev@arms/g-vs-b/exam/wins\endcsname{21}
\expandafter\def\csname ev@arms/g/exam-vs-control/base-counted\endcsname{1200}
\expandafter\def\csname ev@arms/g/exam-vs-control/base-passes\endcsname{0}
\expandafter\def\csname ev@arms/g/exam-vs-control/base-pct\endcsname{0.00}
\expandafter\def\csname ev@arms/g/exam-vs-control/cand-counted\endcsname{1200}
\expandafter\def\csname ev@arms/g/exam-vs-control/cand-passes\endcsname{42}
\expandafter\def\csname ev@arms/g/exam-vs-control/cand-pct\endcsname{3.50}
\expandafter\def\csname ev@arms/g/exam-vs-control/ci-hi-pp\endcsname{+5.08}
\expandafter\def\csname ev@arms/g/exam-vs-control/ci-lo-pp\endcsname{+2.17}
\expandafter\def\csname ev@arms/g/exam-vs-control/diff-pp\endcsname{+3.50}
\expandafter\def\csname ev@arms/g/exam-vs-control/losses\endcsname{0}
\expandafter\def\csname ev@arms/g/exam-vs-control/outcome\endcsname{improvement supported}
\expandafter\def\csname ev@arms/g/exam-vs-control/p-recorded\endcsname{\ensuremath{3.0\times 10^{-8}}}
\expandafter\def\csname ev@arms/g/exam-vs-control/protocol\endcsname{arms}
\expandafter\def\csname ev@arms/g/exam-vs-control/task-p\endcsname{\ensuremath{3.0\times 10^{-8}}}
\expandafter\def\csname ev@arms/g/exam-vs-control/tasks\endcsname{100}
\expandafter\def\csname ev@arms/g/exam-vs-control/ties\endcsname{75}
\expandafter\def\csname ev@arms/g/exam-vs-control/uncounted\endcsname{10}
\expandafter\def\csname ev@arms/g/exam-vs-control/uncounted-pct\endcsname{1.25}
\expandafter\def\csname ev@arms/g/exam-vs-control/unit\endcsname{task}
\expandafter\def\csname ev@arms/g/exam-vs-control/wins\endcsname{25}
\expandafter\def\csname ev@arms/g/exam/base-counted\endcsname{1175}
\expandafter\def\csname ev@arms/g/exam/base-passes\endcsname{0}
\expandafter\def\csname ev@arms/g/exam/base-pct\endcsname{0.00}
\expandafter\def\csname ev@arms/g/exam/base-scheduled\endcsname{1200}
\expandafter\def\csname ev@arms/g/exam/base-uncounted\endcsname{25}
\expandafter\def\csname ev@arms/g/exam/base-uncounted-pct\endcsname{2.08}
\expandafter\def\csname ev@arms/g/exam/cand-counted\endcsname{1200}
\expandafter\def\csname ev@arms/g/exam/cand-passes\endcsname{42}
\expandafter\def\csname ev@arms/g/exam/cand-pct\endcsname{3.50}
\expandafter\def\csname ev@arms/g/exam/ci-hi-pp\endcsname{+5.08}
\expandafter\def\csname ev@arms/g/exam/ci-lo-pp\endcsname{+2.08}
\expandafter\def\csname ev@arms/g/exam/diff-pp\endcsname{+3.49}
\expandafter\def\csname ev@arms/g/exam/losses\endcsname{0}
\expandafter\def\csname ev@arms/g/exam/outcome\endcsname{improvement supported}
\expandafter\def\csname ev@arms/g/exam/p-recorded\endcsname{\ensuremath{6.0\times 10^{-8}}}
\expandafter\def\csname ev@arms/g/exam/protocol\endcsname{arms}
\expandafter\def\csname ev@arms/g/exam/task-p\endcsname{\ensuremath{6.0\times 10^{-8}}}
\expandafter\def\csname ev@arms/g/exam/tasks\endcsname{100}
\expandafter\def\csname ev@arms/g/exam/tasks-solved\endcsname{25}
\expandafter\def\csname ev@arms/g/exam/ties\endcsname{76}
\expandafter\def\csname ev@arms/g/exam/uncounted\endcsname{10}
\expandafter\def\csname ev@arms/g/exam/uncounted-pct\endcsname{1.25}
\expandafter\def\csname ev@arms/g/exam/unit\endcsname{task}
\expandafter\def\csname ev@arms/g/exam/wins\endcsname{24}
\expandafter\def\csname ev@arms/g/target/base-counted\endcsname{612}
\expandafter\def\csname ev@arms/g/target/base-passes\endcsname{22}
\expandafter\def\csname ev@arms/g/target/base-pct\endcsname{3.59}
\expandafter\def\csname ev@arms/g/target/base-scheduled\endcsname{612}
\expandafter\def\csname ev@arms/g/target/base-uncounted\endcsname{0}
\expandafter\def\csname ev@arms/g/target/base-uncounted-pct\endcsname{0.00}
\expandafter\def\csname ev@arms/g/target/cand-counted\endcsname{612}
\expandafter\def\csname ev@arms/g/target/cand-passes\endcsname{127}
\expandafter\def\csname ev@arms/g/target/cand-pct\endcsname{20.75}
\expandafter\def\csname ev@arms/g/target/ci-hi-pp\endcsname{+23.69}
\expandafter\def\csname ev@arms/g/target/ci-lo-pp\endcsname{+11.11}
\expandafter\def\csname ev@arms/g/target/diff-pp\endcsname{+17.16}
\expandafter\def\csname ev@arms/g/target/losses\endcsname{0}
\expandafter\def\csname ev@arms/g/target/outcome\endcsname{improvement supported}
\expandafter\def\csname ev@arms/g/target/p-recorded\endcsname{\ensuremath{3.0\times 10^{-8}}}
\expandafter\def\csname ev@arms/g/target/protocol\endcsname{arms}
\expandafter\def\csname ev@arms/g/target/task-p\endcsname{\ensuremath{3.0\times 10^{-8}}}
\expandafter\def\csname ev@arms/g/target/tasks\endcsname{51}
\expandafter\def\csname ev@arms/g/target/tasks-solved\endcsname{26}
\expandafter\def\csname ev@arms/g/target/ties\endcsname{26}
\expandafter\def\csname ev@arms/g/target/uncounted\endcsname{0}
\expandafter\def\csname ev@arms/g/target/uncounted-pct\endcsname{0.00}
\expandafter\def\csname ev@arms/g/target/unit\endcsname{task}
\expandafter\def\csname ev@arms/g/target/wins\endcsname{25}
\expandafter\def\csname ev@arms/s/exam/base-counted\endcsname{1184}
\expandafter\def\csname ev@arms/s/exam/base-passes\endcsname{0}
\expandafter\def\csname ev@arms/s/exam/base-pct\endcsname{0.00}
\expandafter\def\csname ev@arms/s/exam/base-scheduled\endcsname{1200}
\expandafter\def\csname ev@arms/s/exam/base-uncounted\endcsname{16}
\expandafter\def\csname ev@arms/s/exam/base-uncounted-pct\endcsname{1.33}
\expandafter\def\csname ev@arms/s/exam/cand-counted\endcsname{1166}
\expandafter\def\csname ev@arms/s/exam/cand-passes\endcsname{0}
\expandafter\def\csname ev@arms/s/exam/cand-pct\endcsname{0.00}
\expandafter\def\csname ev@arms/s/exam/ci-hi-pp\endcsname{+0.00}
\expandafter\def\csname ev@arms/s/exam/ci-lo-pp\endcsname{+0.00}
\expandafter\def\csname ev@arms/s/exam/diff-pp\endcsname{+0.00}
\expandafter\def\csname ev@arms/s/exam/losses\endcsname{0}
\expandafter\def\csname ev@arms/s/exam/outcome\endcsname{excluded}
\expandafter\def\csname ev@arms/s/exam/p-recorded\endcsname{1}
\expandafter\def\csname ev@arms/s/exam/protocol\endcsname{arms}
\expandafter\def\csname ev@arms/s/exam/task-p\endcsname{1}
\expandafter\def\csname ev@arms/s/exam/tasks\endcsname{100}
\expandafter\def\csname ev@arms/s/exam/tasks-solved\endcsname{0}
\expandafter\def\csname ev@arms/s/exam/ties\endcsname{100}
\expandafter\def\csname ev@arms/s/exam/uncounted\endcsname{21}
\expandafter\def\csname ev@arms/s/exam/uncounted-pct\endcsname{2.62}
\expandafter\def\csname ev@arms/s/exam/unit\endcsname{task}
\expandafter\def\csname ev@arms/s/exam/wins\endcsname{0}
\expandafter\def\csname ev@calibration/4B/attempts\endcsname{1156}
\expandafter\def\csname ev@calibration/4B/passes\endcsname{0}
\expandafter\def\csname ev@calibration/8B/attempts\endcsname{1118}
\expandafter\def\csname ev@calibration/8B/passes\endcsname{0}
\expandafter\def\csname ev@claim/eng-accept-reject/abstract\endcsname{no}
\expandafter\def\csname ev@claim/eng-accept-reject/observed\endcsname{yes}
\expandafter\def\csname ev@claim/eng-accept-reject/status\endcsname{exercised}
\expandafter\def\csname ev@claim/eng-generate-tasks/abstract\endcsname{no}
\expandafter\def\csname ev@claim/eng-generate-tasks/observed\endcsname{yes}
\expandafter\def\csname ev@claim/eng-generate-tasks/status\endcsname{exercised}
\expandafter\def\csname ev@claim/eng-independent-evaluation/abstract\endcsname{no}
\expandafter\def\csname ev@claim/eng-independent-evaluation/observed\endcsname{yes}
\expandafter\def\csname ev@claim/eng-independent-evaluation/status\endcsname{exercised}
\expandafter\def\csname ev@claim/eng-measure-earlier/abstract\endcsname{no}
\expandafter\def\csname ev@claim/eng-measure-earlier/observed\endcsname{yes}
\expandafter\def\csname ev@claim/eng-measure-earlier/status\endcsname{exercised}
\expandafter\def\csname ev@claim/eng-reject-invalid/abstract\endcsname{no}
\expandafter\def\csname ev@claim/eng-reject-invalid/observed\endcsname{yes}
\expandafter\def\csname ev@claim/eng-reject-invalid/status\endcsname{exercised}
\expandafter\def\csname ev@claim/eng-successive-generations/abstract\endcsname{as confirmed}
\expandafter\def\csname ev@claim/eng-successive-generations/observed\endcsname{yes}
\expandafter\def\csname ev@claim/eng-successive-generations/status\endcsname{exercised}
\expandafter\def\csname ev@claim/eng-train-adapters/abstract\endcsname{no}
\expandafter\def\csname ev@claim/eng-train-adapters/observed\endcsname{yes}
\expandafter\def\csname ev@claim/eng-train-adapters/status\endcsname{exercised}
\expandafter\def\csname ev@claim/eng-training-material/abstract\endcsname{no}
\expandafter\def\csname ev@claim/eng-training-material/observed\endcsname{yes}
\expandafter\def\csname ev@claim/eng-training-material/status\endcsname{exercised}
\expandafter\def\csname ev@claim/eng-verified-solutions/abstract\endcsname{no}
\expandafter\def\csname ev@claim/eng-verified-solutions/observed\endcsname{yes}
\expandafter\def\csname ev@claim/eng-verified-solutions/status\endcsname{exercised}
\expandafter\def\csname ev@claim/learning-efficiency/abstract\endcsname{as not established}
\expandafter\def\csname ev@claim/learning-efficiency/observed\endcsname{no}
\expandafter\def\csname ev@claim/learning-efficiency/status\endcsname{inconclusive}
\expandafter\def\csname ev@claim/lua-early-steps/abstract\endcsname{as confirmed}
\expandafter\def\csname ev@claim/lua-early-steps/observed\endcsname{yes}
\expandafter\def\csname ev@claim/lua-early-steps/status\endcsname{improvement supported}
\expandafter\def\csname ev@claim/lua-late-steps/abstract\endcsname{as not established}
\expandafter\def\csname ev@claim/lua-late-steps/observed\endcsname{yes}
\expandafter\def\csname ev@claim/lua-late-steps/status\endcsname{inconclusive}
\expandafter\def\csname ev@claim/lua-new-families/abstract\endcsname{as confirmed}
\expandafter\def\csname ev@claim/lua-new-families/observed\endcsname{yes}
\expandafter\def\csname ev@claim/lua-new-families/status\endcsname{improvement supported}
\expandafter\def\csname ev@claim/lua-total-gain/abstract\endcsname{as confirmed}
\expandafter\def\csname ev@claim/lua-total-gain/observed\endcsname{yes}
\expandafter\def\csname ev@claim/lua-total-gain/status\endcsname{improvement supported}
\expandafter\def\csname ev@claim/vim-a-exam-4b-g1/abstract\endcsname{no}
\expandafter\def\csname ev@claim/vim-a-exam-4b-g1/observed\endcsname{yes}
\expandafter\def\csname ev@claim/vim-a-exam-4b-g1/status\endcsname{improvement supported}
\expandafter\def\csname ev@claim/vim-a-exam-8b-g1/abstract\endcsname{no}
\expandafter\def\csname ev@claim/vim-a-exam-8b-g1/observed\endcsname{yes}
\expandafter\def\csname ev@claim/vim-a-exam-8b-g1/status\endcsname{excluded}
\expandafter\def\csname ev@claim/vim-a-exam-g2/abstract\endcsname{no}
\expandafter\def\csname ev@claim/vim-a-exam-g2/observed\endcsname{yes}
\expandafter\def\csname ev@claim/vim-a-exam-g2/status\endcsname{inconclusive}
\expandafter\def\csname ev@claim/vim-a-retention/abstract\endcsname{no}
\expandafter\def\csname ev@claim/vim-a-retention/observed\endcsname{yes}
\expandafter\def\csname ev@claim/vim-a-retention/status\endcsname{retention supported}
\expandafter\def\csname ev@claim/vim-b-exam-4b-g1/abstract\endcsname{as not established}
\expandafter\def\csname ev@claim/vim-b-exam-4b-g1/observed\endcsname{yes}
\expandafter\def\csname ev@claim/vim-b-exam-4b-g1/status\endcsname{inconclusive}
\expandafter\def\csname ev@claim/vim-b-exam-8b-g1/abstract\endcsname{descriptive only}
\expandafter\def\csname ev@claim/vim-b-exam-8b-g1/observed\endcsname{yes}
\expandafter\def\csname ev@claim/vim-b-exam-8b-g1/status\endcsname{excluded}
\expandafter\def\csname ev@claim/vim-b-exam-steps/abstract\endcsname{as not established}
\expandafter\def\csname ev@claim/vim-b-exam-steps/observed\endcsname{no}
\expandafter\def\csname ev@claim/vim-b-exam-steps/status\endcsname{inconclusive}
\expandafter\def\csname ev@claim/vim-b-heldout-g1/abstract\endcsname{as confirmed}
\expandafter\def\csname ev@claim/vim-b-heldout-g1/observed\endcsname{yes}
\expandafter\def\csname ev@claim/vim-b-heldout-g1/status\endcsname{improvement supported}
\expandafter\def\csname ev@claim/vim-b-heldout-steps-4b/abstract\endcsname{as not established}
\expandafter\def\csname ev@claim/vim-b-heldout-steps-4b/observed\endcsname{yes}
\expandafter\def\csname ev@claim/vim-b-heldout-steps-4b/status\endcsname{inconclusive}
\expandafter\def\csname ev@claim/vim-b-heldout-steps-8b/abstract\endcsname{as not established}
\expandafter\def\csname ev@claim/vim-b-heldout-steps-8b/observed\endcsname{yes}
\expandafter\def\csname ev@claim/vim-b-heldout-steps-8b/status\endcsname{inconclusive}
\expandafter\def\csname ev@claim/vim-b-retention/abstract\endcsname{as not established}
\expandafter\def\csname ev@claim/vim-b-retention/observed\endcsname{yes}
\expandafter\def\csname ev@claim/vim-b-retention/status\endcsname{inconclusive}
\expandafter\def\csname ev@claim/vim-br-exam/abstract\endcsname{as not established}
\expandafter\def\csname ev@claim/vim-br-exam/observed\endcsname{no}
\expandafter\def\csname ev@claim/vim-br-exam/status\endcsname{inconclusive}
\expandafter\def\csname ev@claim/vim-br-heldout/abstract\endcsname{as not established}
\expandafter\def\csname ev@claim/vim-br-heldout/observed\endcsname{yes}
\expandafter\def\csname ev@claim/vim-br-heldout/status\endcsname{inconclusive}
\expandafter\def\csname ev@claim/vim-heldout-arms/abstract\endcsname{no}
\expandafter\def\csname ev@claim/vim-heldout-arms/observed\endcsname{yes}
\expandafter\def\csname ev@claim/vim-heldout-arms/status\endcsname{improvement supported}
\expandafter\def\csname ev@claim/vim-learner-only/abstract\endcsname{descriptive only}
\expandafter\def\csname ev@claim/vim-learner-only/observed\endcsname{no}
\expandafter\def\csname ev@claim/vim-learner-only/status\endcsname{excluded}
\expandafter\def\csname ev@claim/vim-outcome-first/abstract\endcsname{as confirmed}
\expandafter\def\csname ev@claim/vim-outcome-first/observed\endcsname{yes}
\expandafter\def\csname ev@claim/vim-outcome-first/status\endcsname{improvement supported}
\expandafter\def\csname ev@claim/vim-practice-retention-4b/abstract\endcsname{no}
\expandafter\def\csname ev@claim/vim-practice-retention-4b/observed\endcsname{yes}
\expandafter\def\csname ev@claim/vim-practice-retention-4b/status\endcsname{retention supported}
\expandafter\def\csname ev@claim/vim-practice-retention-8b/abstract\endcsname{no}
\expandafter\def\csname ev@claim/vim-practice-retention-8b/observed\endcsname{yes}
\expandafter\def\csname ev@claim/vim-practice-retention-8b/status\endcsname{retention supported}
\expandafter\def\csname ev@claim/vim-practice-transfer-4b/abstract\endcsname{as confirmed}
\expandafter\def\csname ev@claim/vim-practice-transfer-4b/observed\endcsname{yes}
\expandafter\def\csname ev@claim/vim-practice-transfer-4b/status\endcsname{improvement supported}
\expandafter\def\csname ev@claim/vim-practice-transfer-8b/abstract\endcsname{as confirmed}
\expandafter\def\csname ev@claim/vim-practice-transfer-8b/observed\endcsname{yes}
\expandafter\def\csname ev@claim/vim-practice-transfer-8b/status\endcsname{improvement supported}
\expandafter\def\csname ev@claim/vim-shuffled-control/abstract\endcsname{as not established}
\expandafter\def\csname ev@claim/vim-shuffled-control/observed\endcsname{no}
\expandafter\def\csname ev@claim/vim-shuffled-control/status\endcsname{inconclusive}
\expandafter\def\csname ev@claim/vim-transfer-arms/abstract\endcsname{as confirmed}
\expandafter\def\csname ev@claim/vim-transfer-arms/observed\endcsname{yes}
\expandafter\def\csname ev@claim/vim-transfer-arms/status\endcsname{improvement supported}
\expandafter\def\csname ev@compute/arms/b/row-tokens-millions\endcsname{82.6}
\expandafter\def\csname ev@compute/arms/b/selected-step\endcsname{800/800/800}
\expandafter\def\csname ev@compute/arms/b/steps-completed\endcsname{800/800/800}
\expandafter\def\csname ev@compute/arms/b/train-gpu-hours\endcsname{2.56}
\expandafter\def\csname ev@compute/arms/b/train-minutes-per-seed\endcsname{51}
\expandafter\def\csname ev@compute/arms/c/row-tokens-millions\endcsname{47.1}
\expandafter\def\csname ev@compute/arms/c/selected-step\endcsname{360/320/360}
\expandafter\def\csname ev@compute/arms/c/steps-completed\endcsname{600/560/600}
\expandafter\def\csname ev@compute/arms/c/train-gpu-hours\endcsname{0.86}
\expandafter\def\csname ev@compute/arms/c/train-minutes-per-seed\endcsname{17}
\expandafter\def\csname ev@compute/arms/d/row-tokens-millions\endcsname{82.3}
\expandafter\def\csname ev@compute/arms/d/selected-step\endcsname{800/800/800}
\expandafter\def\csname ev@compute/arms/d/steps-completed\endcsname{800/800/800}
\expandafter\def\csname ev@compute/arms/d/train-gpu-hours\endcsname{2.53}
\expandafter\def\csname ev@compute/arms/d/train-minutes-per-seed\endcsname{51}
\expandafter\def\csname ev@compute/arms/device\endcsname{NVIDIA GH200 120GB}
\expandafter\def\csname ev@compute/arms/f/row-tokens-millions\endcsname{83.4}
\expandafter\def\csname ev@compute/arms/f/selected-step\endcsname{800/800/800}
\expandafter\def\csname ev@compute/arms/f/steps-completed\endcsname{800/800/800}
\expandafter\def\csname ev@compute/arms/f/train-gpu-hours\endcsname{2.55}
\expandafter\def\csname ev@compute/arms/f/train-minutes-per-seed\endcsname{51}
\expandafter\def\csname ev@compute/arms/g/row-tokens-millions\endcsname{80.2}
\expandafter\def\csname ev@compute/arms/g/selected-step\endcsname{520/520/520}
\expandafter\def\csname ev@compute/arms/g/steps-completed\endcsname{760/760/760}
\expandafter\def\csname ev@compute/arms/g/train-gpu-hours\endcsname{2.05}
\expandafter\def\csname ev@compute/arms/g/train-minutes-per-seed\endcsname{41}
\expandafter\def\csname ev@compute/arms/runs\endcsname{15}
\expandafter\def\csname ev@compute/arms/total-steps-completed\endcsname{11240}
\expandafter\def\csname ev@compute/arms/total/train-gpu-hours\endcsname{10.55}
\expandafter\def\csname ev@compute/eval/A/attempts-scheduled\endcsname{7896}
\expandafter\def\csname ev@compute/eval/B-R/attempts-scheduled\endcsname{11732}
\expandafter\def\csname ev@compute/eval/B/attempts-scheduled\endcsname{9776}
\expandafter\def\csname ev@compute/lua/g1/gpu-hours\endcsname{2.99}
\expandafter\def\csname ev@compute/lua/g1/proposal-calls\endcsname{7040}
\expandafter\def\csname ev@compute/lua/g1/proposed\endcsname{5100}
\expandafter\def\csname ev@compute/lua/g1/steps\endcsname{1000}
\expandafter\def\csname ev@compute/lua/g1/trained-records\endcsname{4000}
\expandafter\def\csname ev@compute/lua/g2/gpu-hours\endcsname{2.25}
\expandafter\def\csname ev@compute/lua/g2/proposal-calls\endcsname{12352}
\expandafter\def\csname ev@compute/lua/g2/proposed\endcsname{9300}
\expandafter\def\csname ev@compute/lua/g2/steps\endcsname{2000}
\expandafter\def\csname ev@compute/lua/g2/trained-records\endcsname{8000}
\expandafter\def\csname ev@compute/lua/g3/gpu-hours\endcsname{4.31}
\expandafter\def\csname ev@compute/lua/g3/proposal-calls\endcsname{29248}
\expandafter\def\csname ev@compute/lua/g3/proposed\endcsname{18600}
\expandafter\def\csname ev@compute/lua/g3/steps\endcsname{4000}
\expandafter\def\csname ev@compute/lua/g3/trained-records\endcsname{16000}
\expandafter\def\csname ev@compute/lua/g4/gpu-hours\endcsname{11.90}
\expandafter\def\csname ev@compute/lua/g4/proposal-calls\endcsname{107584}
\expandafter\def\csname ev@compute/lua/g4/proposed\endcsname{39506}
\expandafter\def\csname ev@compute/lua/g4/steps\endcsname{8000}
\expandafter\def\csname ev@compute/lua/g4/trained-records\endcsname{32000}
\expandafter\def\csname ev@compute/lua/total/gpu-hours\endcsname{21.4}
\expandafter\def\csname ev@compute/missing/count\endcsname{12}
\expandafter\def\csname ev@compute/rollouts/backward2/oracle/budget-exhausted\endcsname{0}
\expandafter\def\csname ev@compute/rollouts/backward2/oracle/counted\endcsname{467}
\expandafter\def\csname ev@compute/rollouts/backward2/oracle/physical-attempts\endcsname{467}
\expandafter\def\csname ev@compute/rollouts/backward2/oracle/timeouts\endcsname{0}
\expandafter\def\csname ev@compute/rollouts/backward2/oracle/trials\endcsname{467}
\expandafter\def\csname ev@compute/rollouts/backward2/oracle/verified\endcsname{467}
\expandafter\def\csname ev@compute/rollouts/backward2/reproduce-attempts/budget-exhausted\endcsname{325}
\expandafter\def\csname ev@compute/rollouts/backward2/reproduce-attempts/counted\endcsname{2805}
\expandafter\def\csname ev@compute/rollouts/backward2/reproduce-attempts/physical-attempts\endcsname{3799}
\expandafter\def\csname ev@compute/rollouts/backward2/reproduce-attempts/timeouts\endcsname{1498}
\expandafter\def\csname ev@compute/rollouts/backward2/reproduce-attempts/trials\endcsname{2820}
\expandafter\def\csname ev@compute/rollouts/backward2/reproduce-attempts/verified\endcsname{1387}
\expandafter\def\csname ev@compute/rollouts/backward2/student-attempts/budget-exhausted\endcsname{196}
\expandafter\def\csname ev@compute/rollouts/backward2/student-attempts/counted\endcsname{3688}
\expandafter\def\csname ev@compute/rollouts/backward2/student-attempts/physical-attempts\endcsname{4091}
\expandafter\def\csname ev@compute/rollouts/backward2/student-attempts/timeouts\endcsname{39}
\expandafter\def\csname ev@compute/rollouts/backward2/student-attempts/trials\endcsname{3736}
\expandafter\def\csname ev@compute/rollouts/backward2/student-attempts/verified\endcsname{116}
\expandafter\def\csname ev@compute/rollouts/backward2/teacher-attempts/budget-exhausted\endcsname{136}
\expandafter\def\csname ev@compute/rollouts/backward2/teacher-attempts/counted\endcsname{1868}
\expandafter\def\csname ev@compute/rollouts/backward2/teacher-attempts/physical-attempts\endcsname{1875}
\expandafter\def\csname ev@compute/rollouts/backward2/teacher-attempts/timeouts\endcsname{154}
\expandafter\def\csname ev@compute/rollouts/backward2/teacher-attempts/trials\endcsname{1868}
\expandafter\def\csname ev@compute/rollouts/backward2/teacher-attempts/verified\endcsname{1348}
\expandafter\def\csname ev@compute/rollouts/material2/oracle/budget-exhausted\endcsname{0}
\expandafter\def\csname ev@compute/rollouts/material2/oracle/counted\endcsname{1030}
\expandafter\def\csname ev@compute/rollouts/material2/oracle/physical-attempts\endcsname{1030}
\expandafter\def\csname ev@compute/rollouts/material2/oracle/timeouts\endcsname{0}
\expandafter\def\csname ev@compute/rollouts/material2/oracle/trials\endcsname{1030}
\expandafter\def\csname ev@compute/rollouts/material2/oracle/verified\endcsname{1026}
\expandafter\def\csname ev@compute/rollouts/material2/student-attempts/budget-exhausted\endcsname{303}
\expandafter\def\csname ev@compute/rollouts/material2/student-attempts/counted\endcsname{8204}
\expandafter\def\csname ev@compute/rollouts/material2/student-attempts/physical-attempts\endcsname{8533}
\expandafter\def\csname ev@compute/rollouts/material2/student-attempts/timeouts\endcsname{183}
\expandafter\def\csname ev@compute/rollouts/material2/student-attempts/trials\endcsname{8208}
\expandafter\def\csname ev@compute/rollouts/material2/student-attempts/verified\endcsname{824}
\expandafter\def\csname ev@compute/rollouts/material2/teacher-attempts/budget-exhausted\endcsname{75}
\expandafter\def\csname ev@compute/rollouts/material2/teacher-attempts/counted\endcsname{4095}
\expandafter\def\csname ev@compute/rollouts/material2/teacher-attempts/physical-attempts\endcsname{4113}
\expandafter\def\csname ev@compute/rollouts/material2/teacher-attempts/timeouts\endcsname{240}
\expandafter\def\csname ev@compute/rollouts/material2/teacher-attempts/trials\endcsname{4104}
\expandafter\def\csname ev@compute/rollouts/material2/teacher-attempts/verified\endcsname{3319}
\expandafter\def\csname ev@const/alpha\endcsname{0.05}
\expandafter\def\csname ev@const/attempts-per-task\endcsname{4}
\expandafter\def\csname ev@const/exams\endcsname{100}
\expandafter\def\csname ev@const/retention-margin-pp\endcsname{5}
\expandafter\def\csname ev@const/uncounted-bound-pct\endcsname{2}
\expandafter\def\csname ev@funnel/backward/admission/admitted\endcsname{467}
\expandafter\def\csname ev@funnel/backward/admission/refused\endcsname{56}
\expandafter\def\csname ev@funnel/backward/oracle/counted\endcsname{467}
\expandafter\def\csname ev@funnel/backward/oracle/pass-pct\endcsname{100.0}
\expandafter\def\csname ev@funnel/backward/oracle/passed\endcsname{467}
\expandafter\def\csname ev@funnel/backward/oracle/tasks\endcsname{467}
\expandafter\def\csname ev@funnel/backward/oracle/tasks-solved\endcsname{467}
\expandafter\def\csname ev@funnel/backward/proposals/admitted\endcsname{705}
\expandafter\def\csname ev@funnel/backward/proposals/proposed\endcsname{771}
\expandafter\def\csname ev@funnel/backward/proposals/refused\endcsname{66}
\expandafter\def\csname ev@funnel/backward/reproduce-attempts/counted\endcsname{2805}
\expandafter\def\csname ev@funnel/backward/reproduce-attempts/pass-pct\endcsname{49.4}
\expandafter\def\csname ev@funnel/backward/reproduce-attempts/passed\endcsname{1387}
\expandafter\def\csname ev@funnel/backward/reproduce-attempts/tasks\endcsname{705}
\expandafter\def\csname ev@funnel/backward/reproduce-attempts/tasks-solved\endcsname{562}
\expandafter\def\csname ev@funnel/backward/stage/describe-items/items\endcsname{523}
\expandafter\def\csname ev@funnel/backward/stage/describe-items/rejected\endcsname{182}
\expandafter\def\csname ev@funnel/backward/stage/described/items\endcsname{523}
\expandafter\def\csname ev@funnel/backward/stage/described/rejected\endcsname{0}
\expandafter\def\csname ev@funnel/backward/stage/proposed/items\endcsname{259}
\expandafter\def\csname ev@funnel/backward/stage/proposed/rejected\endcsname{4}
\expandafter\def\csname ev@funnel/backward/stage/sections/items\endcsname{263}
\expandafter\def\csname ev@funnel/backward/stage/sections/rejected\endcsname{0}
\expandafter\def\csname ev@funnel/backward/student-attempts/counted\endcsname{3688}
\expandafter\def\csname ev@funnel/backward/student-attempts/pass-pct\endcsname{3.1}
\expandafter\def\csname ev@funnel/backward/student-attempts/passed\endcsname{116}
\expandafter\def\csname ev@funnel/backward/student-attempts/tasks\endcsname{467}
\expandafter\def\csname ev@funnel/backward/student-attempts/tasks-solved\endcsname{36}
\expandafter\def\csname ev@funnel/backward/teacher-attempts/counted\endcsname{1868}
\expandafter\def\csname ev@funnel/backward/teacher-attempts/pass-pct\endcsname{72.2}
\expandafter\def\csname ev@funnel/backward/teacher-attempts/passed\endcsname{1348}
\expandafter\def\csname ev@funnel/backward/teacher-attempts/tasks\endcsname{467}
\expandafter\def\csname ev@funnel/backward/teacher-attempts/tasks-solved\endcsname{410}
\expandafter\def\csname ev@funnel/forward/admission/admitted\endcsname{1030}
\expandafter\def\csname ev@funnel/forward/admission/candidates\endcsname{6705}
\expandafter\def\csname ev@funnel/forward/admission/malformed\endcsname{349}
\expandafter\def\csname ev@funnel/forward/admission/refused\endcsname{5675}
\expandafter\def\csname ev@funnel/forward/oracle/counted\endcsname{1030}
\expandafter\def\csname ev@funnel/forward/oracle/pass-pct\endcsname{99.6}
\expandafter\def\csname ev@funnel/forward/oracle/passed\endcsname{1026}
\expandafter\def\csname ev@funnel/forward/oracle/tasks\endcsname{1030}
\expandafter\def\csname ev@funnel/forward/oracle/tasks-solved\endcsname{1026}
\expandafter\def\csname ev@funnel/forward/stage/fixed/items\endcsname{1139}
\expandafter\def\csname ev@funnel/forward/stage/fixed/rejected\endcsname{1}
\expandafter\def\csname ev@funnel/forward/stage/fixes/items\endcsname{1140}
\expandafter\def\csname ev@funnel/forward/stage/fixes/rejected\endcsname{0}
\expandafter\def\csname ev@funnel/forward/stage/proposed-b/items\endcsname{255}
\expandafter\def\csname ev@funnel/forward/stage/proposed-b/rejected\endcsname{8}
\expandafter\def\csname ev@funnel/forward/stage/proposed-c/items\endcsname{255}
\expandafter\def\csname ev@funnel/forward/stage/proposed-c/rejected\endcsname{8}
\expandafter\def\csname ev@funnel/forward/stage/proposed/items\endcsname{257}
\expandafter\def\csname ev@funnel/forward/stage/proposed/rejected\endcsname{6}
\expandafter\def\csname ev@funnel/forward/stage/refused/items\endcsname{2341}
\expandafter\def\csname ev@funnel/forward/stage/refused/rejected\endcsname{0}
\expandafter\def\csname ev@funnel/forward/stage/repaired/items\endcsname{2336}
\expandafter\def\csname ev@funnel/forward/stage/repaired/rejected\endcsname{5}
\expandafter\def\csname ev@funnel/forward/stage/reviewed-fixed/items\endcsname{2074}
\expandafter\def\csname ev@funnel/forward/stage/reviewed-fixed/rejected\endcsname{18}
\expandafter\def\csname ev@funnel/forward/stage/reviewed/items\endcsname{4385}
\expandafter\def\csname ev@funnel/forward/stage/reviewed/rejected\endcsname{49}
\expandafter\def\csname ev@funnel/forward/stage/sections/items\endcsname{263}
\expandafter\def\csname ev@funnel/forward/stage/sections/rejected\endcsname{0}
\expandafter\def\csname ev@funnel/forward/stage/to-review-fixed/items\endcsname{2092}
\expandafter\def\csname ev@funnel/forward/stage/to-review-fixed/rejected\endcsname{0}
\expandafter\def\csname ev@funnel/forward/stage/to-review/items\endcsname{4434}
\expandafter\def\csname ev@funnel/forward/stage/to-review/rejected\endcsname{0}
\expandafter\def\csname ev@funnel/forward/student-attempts/counted\endcsname{8204}
\expandafter\def\csname ev@funnel/forward/student-attempts/pass-pct\endcsname{10.0}
\expandafter\def\csname ev@funnel/forward/student-attempts/passed\endcsname{824}
\expandafter\def\csname ev@funnel/forward/student-attempts/tasks\endcsname{1026}
\expandafter\def\csname ev@funnel/forward/student-attempts/tasks-solved\endcsname{231}
\expandafter\def\csname ev@funnel/forward/teacher-attempts/counted\endcsname{4095}
\expandafter\def\csname ev@funnel/forward/teacher-attempts/pass-pct\endcsname{81.1}
\expandafter\def\csname ev@funnel/forward/teacher-attempts/passed\endcsname{3319}
\expandafter\def\csname ev@funnel/forward/teacher-attempts/tasks\endcsname{1026}
\expandafter\def\csname ev@funnel/forward/teacher-attempts/tasks-solved\endcsname{944}
\expandafter\def\csname ev@lua/criterion/G1\endcsname{does not hold}
\expandafter\def\csname ev@lua/criterion/G2\endcsname{holds}
\expandafter\def\csname ev@lua/criterion/G3\endcsname{holds}
\expandafter\def\csname ev@lua/gen4-over-base/after-pct\endcsname{13.24}
\expandafter\def\csname ev@lua/gen4-over-base/before-pct\endcsname{5.05}
\expandafter\def\csname ev@lua/gen4-over-base/ci-hi-pp\endcsname{+11.52}
\expandafter\def\csname ev@lua/gen4-over-base/ci-lo-pp\endcsname{+5.12}
\expandafter\def\csname ev@lua/gen4-over-base/diff-pp\endcsname{+8.19}
\expandafter\def\csname ev@lua/gen4-over-base/families\endcsname{136}
\expandafter\def\csname ev@lua/gen4-over-base/items\endcsname{2040}
\expandafter\def\csname ev@lua/gen4-over-base/new-families/after-pct\endcsname{4.63}
\expandafter\def\csname ev@lua/gen4-over-base/new-families/before-pct\endcsname{1.38}
\expandafter\def\csname ev@lua/gen4-over-base/new-families/ci-hi-pp\endcsname{+7.10}
\expandafter\def\csname ev@lua/gen4-over-base/new-families/ci-lo-pp\endcsname{+0.66}
\expandafter\def\csname ev@lua/gen4-over-base/new-families/diff-pp\endcsname{+3.25}
\expandafter\def\csname ev@lua/gen4-over-base/new-families/families\endcsname{58}
\expandafter\def\csname ev@lua/gen4-over-base/new-families/items\endcsname{870}
\expandafter\def\csname ev@lua/gen4-over-base/new-families/outcome\endcsname{improvement supported}
\expandafter\def\csname ev@lua/gen4-over-base/new-families/protocol\endcsname{lua}
\expandafter\def\csname ev@lua/gen4-over-base/new-families/unit\endcsname{program family and lineage}
\expandafter\def\csname ev@lua/gen4-over-base/old-families/after-pct\endcsname{19.64}
\expandafter\def\csname ev@lua/gen4-over-base/old-families/before-pct\endcsname{7.78}
\expandafter\def\csname ev@lua/gen4-over-base/old-families/ci-hi-pp\endcsname{+17.09}
\expandafter\def\csname ev@lua/gen4-over-base/old-families/ci-lo-pp\endcsname{+7.01}
\expandafter\def\csname ev@lua/gen4-over-base/old-families/diff-pp\endcsname{+11.86}
\expandafter\def\csname ev@lua/gen4-over-base/old-families/families\endcsname{78}
\expandafter\def\csname ev@lua/gen4-over-base/old-families/items\endcsname{1170}
\expandafter\def\csname ev@lua/gen4-over-base/old-families/outcome\endcsname{improvement supported}
\expandafter\def\csname ev@lua/gen4-over-base/old-families/protocol\endcsname{lua}
\expandafter\def\csname ev@lua/gen4-over-base/old-families/unit\endcsname{program family and lineage}
\expandafter\def\csname ev@lua/gen4-over-base/outcome\endcsname{improvement supported}
\expandafter\def\csname ev@lua/gen4-over-base/protocol\endcsname{lua}
\expandafter\def\csname ev@lua/gen4-over-base/unit\endcsname{program family and lineage}
\expandafter\def\csname ev@lua/group/base/lineages\endcsname{1}
\expandafter\def\csname ev@lua/group/base/max-pct\endcsname{5.0}
\expandafter\def\csname ev@lua/group/base/min-pct\endcsname{5.0}
\expandafter\def\csname ev@lua/group/base/pct\endcsname{5.05}
\expandafter\def\csname ev@lua/group/gen1/lineages\endcsname{4}
\expandafter\def\csname ev@lua/group/gen1/max-pct\endcsname{11.1}
\expandafter\def\csname ev@lua/group/gen1/min-pct\endcsname{9.0}
\expandafter\def\csname ev@lua/group/gen1/pct\endcsname{10.12}
\expandafter\def\csname ev@lua/group/gen2/lineages\endcsname{4}
\expandafter\def\csname ev@lua/group/gen2/max-pct\endcsname{12.7}
\expandafter\def\csname ev@lua/group/gen2/min-pct\endcsname{11.1}
\expandafter\def\csname ev@lua/group/gen2/pct\endcsname{11.72}
\expandafter\def\csname ev@lua/group/gen3/lineages\endcsname{4}
\expandafter\def\csname ev@lua/group/gen3/max-pct\endcsname{13.9}
\expandafter\def\csname ev@lua/group/gen3/min-pct\endcsname{12.1}
\expandafter\def\csname ev@lua/group/gen3/pct\endcsname{12.95}
\expandafter\def\csname ev@lua/group/gen4/lineages\endcsname{4}
\expandafter\def\csname ev@lua/group/gen4/max-pct\endcsname{14.0}
\expandafter\def\csname ev@lua/group/gen4/min-pct\endcsname{12.3}
\expandafter\def\csname ev@lua/group/gen4/pct\endcsname{13.24}
\expandafter\def\csname ev@lua/step1/after-pct\endcsname{10.12}
\expandafter\def\csname ev@lua/step1/before-pct\endcsname{5.05}
\expandafter\def\csname ev@lua/step1/ci-hi-pp\endcsname{+7.76}
\expandafter\def\csname ev@lua/step1/ci-lo-pp\endcsname{+2.60}
\expandafter\def\csname ev@lua/step1/diff-pp\endcsname{+5.07}
\expandafter\def\csname ev@lua/step1/families\endcsname{136}
\expandafter\def\csname ev@lua/step1/items\endcsname{2040}
\expandafter\def\csname ev@lua/step1/outcome\endcsname{improvement supported}
\expandafter\def\csname ev@lua/step1/protocol\endcsname{lua}
\expandafter\def\csname ev@lua/step1/unit\endcsname{program family and lineage}
\expandafter\def\csname ev@lua/step2/after-pct\endcsname{11.72}
\expandafter\def\csname ev@lua/step2/before-pct\endcsname{10.12}
\expandafter\def\csname ev@lua/step2/ci-hi-pp\endcsname{+3.33}
\expandafter\def\csname ev@lua/step2/ci-lo-pp\endcsname{+0.12}
\expandafter\def\csname ev@lua/step2/diff-pp\endcsname{+1.59}
\expandafter\def\csname ev@lua/step2/families\endcsname{136}
\expandafter\def\csname ev@lua/step2/items\endcsname{2040}
\expandafter\def\csname ev@lua/step2/outcome\endcsname{improvement supported}
\expandafter\def\csname ev@lua/step2/protocol\endcsname{lua}
\expandafter\def\csname ev@lua/step2/unit\endcsname{program family and lineage}
\expandafter\def\csname ev@lua/step3/after-pct\endcsname{12.95}
\expandafter\def\csname ev@lua/step3/before-pct\endcsname{11.72}
\expandafter\def\csname ev@lua/step3/ci-hi-pp\endcsname{+2.98}
\expandafter\def\csname ev@lua/step3/ci-lo-pp\endcsname{-0.36}
\expandafter\def\csname ev@lua/step3/diff-pp\endcsname{+1.24}
\expandafter\def\csname ev@lua/step3/families\endcsname{136}
\expandafter\def\csname ev@lua/step3/items\endcsname{2040}
\expandafter\def\csname ev@lua/step3/outcome\endcsname{inconclusive}
\expandafter\def\csname ev@lua/step3/protocol\endcsname{lua}
\expandafter\def\csname ev@lua/step3/unit\endcsname{program family and lineage}
\expandafter\def\csname ev@lua/step4/after-pct\endcsname{13.24}
\expandafter\def\csname ev@lua/step4/before-pct\endcsname{12.95}
\expandafter\def\csname ev@lua/step4/ci-hi-pp\endcsname{+2.24}
\expandafter\def\csname ev@lua/step4/ci-lo-pp\endcsname{-1.54}
\expandafter\def\csname ev@lua/step4/diff-pp\endcsname{+0.28}
\expandafter\def\csname ev@lua/step4/families\endcsname{136}
\expandafter\def\csname ev@lua/step4/items\endcsname{2040}
\expandafter\def\csname ev@lua/step4/outcome\endcsname{inconclusive}
\expandafter\def\csname ev@lua/step4/protocol\endcsname{lua}
\expandafter\def\csname ev@lua/step4/unit\endcsname{program family and lineage}
\expandafter\def\csname ev@material/b/records\endcsname{3739}
\expandafter\def\csname ev@material/b/selected-step\endcsname{800}
\expandafter\def\csname ev@material/b/student-train-records\endcsname{660}
\expandafter\def\csname ev@material/b/supervised-characters\endcsname{6454922}
\expandafter\def\csname ev@material/b/supervised-messages\endcsname{20211}
\expandafter\def\csname ev@material/b/tasks\endcsname{857}
\expandafter\def\csname ev@material/b/teacher-train-records\endcsname{2699}
\expandafter\def\csname ev@material/b/train-records\endcsname{3359}
\expandafter\def\csname ev@material/b/validation-records\endcsname{380}
\expandafter\def\csname ev@material/c/records\endcsname{857}
\expandafter\def\csname ev@material/c/student-train-records\endcsname{142}
\expandafter\def\csname ev@material/c/supervised-characters\endcsname{1154407}
\expandafter\def\csname ev@material/c/supervised-messages\endcsname{3518}
\expandafter\def\csname ev@material/c/tasks\endcsname{857}
\expandafter\def\csname ev@material/c/teacher-train-records\endcsname{629}
\expandafter\def\csname ev@material/c/train-records\endcsname{771}
\expandafter\def\csname ev@material/c/validation-records\endcsname{86}
\expandafter\def\csname ev@material/d/records\endcsname{3739}
\expandafter\def\csname ev@material/d/selected-step\endcsname{800}
\expandafter\def\csname ev@material/d/student-train-records\endcsname{660}
\expandafter\def\csname ev@material/d/supervised-characters\endcsname{4828473}
\expandafter\def\csname ev@material/d/supervised-messages\endcsname{14275}
\expandafter\def\csname ev@material/d/tasks\endcsname{857}
\expandafter\def\csname ev@material/d/teacher-train-records\endcsname{2699}
\expandafter\def\csname ev@material/d/train-records\endcsname{3359}
\expandafter\def\csname ev@material/d/validation-records\endcsname{380}
\expandafter\def\csname ev@material/f/records\endcsname{3739}
\expandafter\def\csname ev@material/f/selected-step\endcsname{800}
\expandafter\def\csname ev@material/f/student-train-records\endcsname{660}
\expandafter\def\csname ev@material/f/supervised-characters\endcsname{6454922}
\expandafter\def\csname ev@material/f/supervised-messages\endcsname{20211}
\expandafter\def\csname ev@material/f/tasks\endcsname{857}
\expandafter\def\csname ev@material/f/teacher-train-records\endcsname{2699}
\expandafter\def\csname ev@material/f/train-records\endcsname{3359}
\expandafter\def\csname ev@material/f/validation-records\endcsname{380}
\expandafter\def\csname ev@material/g/records\endcsname{1230}
\expandafter\def\csname ev@material/g/selected-step\endcsname{520}
\expandafter\def\csname ev@material/g/student-train-records\endcsname{86}
\expandafter\def\csname ev@material/g/supervised-characters\endcsname{3116155}
\expandafter\def\csname ev@material/g/supervised-messages\endcsname{7584}
\expandafter\def\csname ev@material/g/tasks\endcsname{348}
\expandafter\def\csname ev@material/g/teacher-train-records\endcsname{1014}
\expandafter\def\csname ev@material/g/train-records\endcsname{1100}
\expandafter\def\csname ev@material/g/validation-records\endcsname{130}
\expandafter\def\csname ev@material/s/records\endcsname{727}
\expandafter\def\csname ev@material/s/selected-step\endcsname{640}
\expandafter\def\csname ev@material/s/student-train-records\endcsname{660}
\expandafter\def\csname ev@material/s/supervised-characters\endcsname{629243}
\expandafter\def\csname ev@material/s/supervised-messages\endcsname{1955}
\expandafter\def\csname ev@material/s/tasks\endcsname{200}
\expandafter\def\csname ev@material/s/teacher-train-records\endcsname{0}
\expandafter\def\csname ev@material/s/train-records\endcsname{660}
\expandafter\def\csname ev@material/s/validation-records\endcsname{67}
\expandafter\def\csname ev@practice-transfer/4B/base-pct-high\endcsname{16}
\expandafter\def\csname ev@practice-transfer/4B/base-pct-low\endcsname{15}
\expandafter\def\csname ev@practice-transfer/4B/cand-pct-high\endcsname{41}
\expandafter\def\csname ev@practice-transfer/4B/cand-pct-low\endcsname{37}
\expandafter\def\csname ev@practice-transfer/4B/eligible\endcsname{3}
\expandafter\def\csname ev@practice-transfer/4B/evaluations\endcsname{3}
\expandafter\def\csname ev@practice-transfer/4B/p-max\endcsname{\ensuremath{1.7\times 10^{-16}}}
\expandafter\def\csname ev@practice-transfer/8B/base-pct-high\endcsname{15}
\expandafter\def\csname ev@practice-transfer/8B/base-pct-low\endcsname{15}
\expandafter\def\csname ev@practice-transfer/8B/cand-pct-high\endcsname{51}
\expandafter\def\csname ev@practice-transfer/8B/cand-pct-low\endcsname{45}
\expandafter\def\csname ev@practice-transfer/8B/eligible\endcsname{2}
\expandafter\def\csname ev@practice-transfer/8B/evaluations\endcsname{3}
\expandafter\def\csname ev@practice-transfer/8B/p-max\endcsname{\ensuremath{2.9\times 10^{-24}}}
\expandafter\def\csname ev@practice-transfer/8b-s1/base-counted\endcsname{893}
\expandafter\def\csname ev@practice-transfer/8b-s1/base-passes\endcsname{139}
\expandafter\def\csname ev@practice-transfer/8b-s1/base-pct\endcsname{15.57}
\expandafter\def\csname ev@practice-transfer/8b-s1/cand-counted\endcsname{912}
\expandafter\def\csname ev@practice-transfer/8b-s1/cand-passes\endcsname{463}
\expandafter\def\csname ev@practice-transfer/8b-s1/cand-pct\endcsname{50.77}
\expandafter\def\csname ev@practice-transfer/8b-s1/ci-hi-pp\endcsname{+41.53}
\expandafter\def\csname ev@practice-transfer/8b-s1/ci-lo-pp\endcsname{+30.49}
\expandafter\def\csname ev@practice-transfer/8b-s1/diff-pp\endcsname{+36.06}
\expandafter\def\csname ev@practice-transfer/8b-s1/losses\endcsname{14}
\expandafter\def\csname ev@practice-transfer/8b-s1/outcome\endcsname{excluded}
\expandafter\def\csname ev@practice-transfer/8b-s1/p-recorded\endcsname{\ensuremath{8.9\times 10^{-25}}}
\expandafter\def\csname ev@practice-transfer/8b-s1/protocol\endcsname{practice-transfer}
\expandafter\def\csname ev@practice-transfer/8b-s1/task-p\endcsname{\ensuremath{8.9\times 10^{-25}}}
\expandafter\def\csname ev@practice-transfer/8b-s1/tasks\endcsname{228}
\expandafter\def\csname ev@practice-transfer/8b-s1/ties\endcsname{85}
\expandafter\def\csname ev@practice-transfer/8b-s1/uncounted\endcsname{19}
\expandafter\def\csname ev@practice-transfer/8b-s1/uncounted-pct\endcsname{2.08}
\expandafter\def\csname ev@practice-transfer/8b-s1/unit\endcsname{task}
\expandafter\def\csname ev@practice-transfer/8b-s1/wins\endcsname{129}
\expandafter\def\csname ev@practice-transfer/8b-s2/base-counted\endcsname{903}
\expandafter\def\csname ev@practice-transfer/8b-s2/base-passes\endcsname{138}
\expandafter\def\csname ev@practice-transfer/8b-s2/base-pct\endcsname{15.28}
\expandafter\def\csname ev@practice-transfer/8b-s2/cand-counted\endcsname{908}
\expandafter\def\csname ev@practice-transfer/8b-s2/cand-passes\endcsname{461}
\expandafter\def\csname ev@practice-transfer/8b-s2/cand-pct\endcsname{50.77}
\expandafter\def\csname ev@practice-transfer/8b-s2/ci-hi-pp\endcsname{+40.99}
\expandafter\def\csname ev@practice-transfer/8b-s2/ci-lo-pp\endcsname{+30.45}
\expandafter\def\csname ev@practice-transfer/8b-s2/diff-pp\endcsname{+35.77}
\expandafter\def\csname ev@practice-transfer/8b-s2/losses\endcsname{10}
\expandafter\def\csname ev@practice-transfer/8b-s2/outcome\endcsname{improvement supported}
\expandafter\def\csname ev@practice-transfer/8b-s2/p-recorded\endcsname{\ensuremath{3.1\times 10^{-30}}}
\expandafter\def\csname ev@practice-transfer/8b-s2/protocol\endcsname{practice-transfer}
\expandafter\def\csname ev@practice-transfer/8b-s2/task-p\endcsname{\ensuremath{3.1\times 10^{-30}}}
\expandafter\def\csname ev@practice-transfer/8b-s2/tasks\endcsname{227}
\expandafter\def\csname ev@practice-transfer/8b-s2/ties\endcsname{79}
\expandafter\def\csname ev@practice-transfer/8b-s2/uncounted\endcsname{9}
\expandafter\def\csname ev@practice-transfer/8b-s2/uncounted-pct\endcsname{0.99}
\expandafter\def\csname ev@practice-transfer/8b-s2/unit\endcsname{task}
\expandafter\def\csname ev@practice-transfer/8b-s2/wins\endcsname{138}
\expandafter\def\csname ev@practice-transfer/8b-s3/base-counted\endcsname{897}
\expandafter\def\csname ev@practice-transfer/8b-s3/base-passes\endcsname{134}
\expandafter\def\csname ev@practice-transfer/8b-s3/base-pct\endcsname{14.94}
\expandafter\def\csname ev@practice-transfer/8b-s3/cand-counted\endcsname{912}
\expandafter\def\csname ev@practice-transfer/8b-s3/cand-passes\endcsname{408}
\expandafter\def\csname ev@practice-transfer/8b-s3/cand-pct\endcsname{44.74}
\expandafter\def\csname ev@practice-transfer/8b-s3/ci-hi-pp\endcsname{+35.35}
\expandafter\def\csname ev@practice-transfer/8b-s3/ci-lo-pp\endcsname{+25.42}
\expandafter\def\csname ev@practice-transfer/8b-s3/diff-pp\endcsname{+30.32}
\expandafter\def\csname ev@practice-transfer/8b-s3/losses\endcsname{14}
\expandafter\def\csname ev@practice-transfer/8b-s3/outcome\endcsname{improvement supported}
\expandafter\def\csname ev@practice-transfer/8b-s3/p-recorded\endcsname{\ensuremath{2.9\times 10^{-24}}}
\expandafter\def\csname ev@practice-transfer/8b-s3/protocol\endcsname{practice-transfer}
\expandafter\def\csname ev@practice-transfer/8b-s3/task-p\endcsname{\ensuremath{2.9\times 10^{-24}}}
\expandafter\def\csname ev@practice-transfer/8b-s3/tasks\endcsname{228}
\expandafter\def\csname ev@practice-transfer/8b-s3/ties\endcsname{87}
\expandafter\def\csname ev@practice-transfer/8b-s3/uncounted\endcsname{15}
\expandafter\def\csname ev@practice-transfer/8b-s3/uncounted-pct\endcsname{1.64}
\expandafter\def\csname ev@practice-transfer/8b-s3/unit\endcsname{task}
\expandafter\def\csname ev@practice-transfer/8b-s3/wins\endcsname{127}
\expandafter\def\csname ev@practice-transfer/c-s1/base-counted\endcsname{912}
\expandafter\def\csname ev@practice-transfer/c-s1/base-passes\endcsname{142}
\expandafter\def\csname ev@practice-transfer/c-s1/base-pct\endcsname{15.57}
\expandafter\def\csname ev@practice-transfer/c-s1/cand-counted\endcsname{912}
\expandafter\def\csname ev@practice-transfer/c-s1/cand-passes\endcsname{336}
\expandafter\def\csname ev@practice-transfer/c-s1/cand-pct\endcsname{36.84}
\expandafter\def\csname ev@practice-transfer/c-s1/ci-hi-pp\endcsname{+26.32}
\expandafter\def\csname ev@practice-transfer/c-s1/ci-lo-pp\endcsname{+16.45}
\expandafter\def\csname ev@practice-transfer/c-s1/diff-pp\endcsname{+21.27}
\expandafter\def\csname ev@practice-transfer/c-s1/losses\endcsname{17}
\expandafter\def\csname ev@practice-transfer/c-s1/outcome\endcsname{improvement supported}
\expandafter\def\csname ev@practice-transfer/c-s1/p-recorded\endcsname{\ensuremath{1.7\times 10^{-16}}}
\expandafter\def\csname ev@practice-transfer/c-s1/protocol\endcsname{practice-transfer}
\expandafter\def\csname ev@practice-transfer/c-s1/task-p\endcsname{\ensuremath{1.7\times 10^{-16}}}
\expandafter\def\csname ev@practice-transfer/c-s1/tasks\endcsname{228}
\expandafter\def\csname ev@practice-transfer/c-s1/ties\endcsname{108}
\expandafter\def\csname ev@practice-transfer/c-s1/uncounted\endcsname{0}
\expandafter\def\csname ev@practice-transfer/c-s1/uncounted-pct\endcsname{0.00}
\expandafter\def\csname ev@practice-transfer/c-s1/unit\endcsname{task}
\expandafter\def\csname ev@practice-transfer/c-s1/wins\endcsname{103}
\expandafter\def\csname ev@practice-transfer/d-s1/base-counted\endcsname{910}
\expandafter\def\csname ev@practice-transfer/d-s1/base-passes\endcsname{141}
\expandafter\def\csname ev@practice-transfer/d-s1/base-pct\endcsname{15.49}
\expandafter\def\csname ev@practice-transfer/d-s1/cand-counted\endcsname{912}
\expandafter\def\csname ev@practice-transfer/d-s1/cand-passes\endcsname{360}
\expandafter\def\csname ev@practice-transfer/d-s1/cand-pct\endcsname{39.47}
\expandafter\def\csname ev@practice-transfer/d-s1/ci-hi-pp\endcsname{+28.98}
\expandafter\def\csname ev@practice-transfer/d-s1/ci-lo-pp\endcsname{+19.10}
\expandafter\def\csname ev@practice-transfer/d-s1/diff-pp\endcsname{+24.07}
\expandafter\def\csname ev@practice-transfer/d-s1/losses\endcsname{13}
\expandafter\def\csname ev@practice-transfer/d-s1/outcome\endcsname{improvement supported}
\expandafter\def\csname ev@practice-transfer/d-s1/p-recorded\endcsname{\ensuremath{2.3\times 10^{-20}}}
\expandafter\def\csname ev@practice-transfer/d-s1/protocol\endcsname{practice-transfer}
\expandafter\def\csname ev@practice-transfer/d-s1/task-p\endcsname{\ensuremath{2.3\times 10^{-20}}}
\expandafter\def\csname ev@practice-transfer/d-s1/tasks\endcsname{228}
\expandafter\def\csname ev@practice-transfer/d-s1/ties\endcsname{106}
\expandafter\def\csname ev@practice-transfer/d-s1/uncounted\endcsname{2}
\expandafter\def\csname ev@practice-transfer/d-s1/uncounted-pct\endcsname{0.22}
\expandafter\def\csname ev@practice-transfer/d-s1/unit\endcsname{task}
\expandafter\def\csname ev@practice-transfer/d-s1/wins\endcsname{109}
\expandafter\def\csname ev@practice-transfer/g-s3/base-counted\endcsname{912}
\expandafter\def\csname ev@practice-transfer/g-s3/base-passes\endcsname{138}
\expandafter\def\csname ev@practice-transfer/g-s3/base-pct\endcsname{15.13}
\expandafter\def\csname ev@practice-transfer/g-s3/cand-counted\endcsname{912}
\expandafter\def\csname ev@practice-transfer/g-s3/cand-passes\endcsname{378}
\expandafter\def\csname ev@practice-transfer/g-s3/cand-pct\endcsname{41.45}
\expandafter\def\csname ev@practice-transfer/g-s3/ci-hi-pp\endcsname{+31.47}
\expandafter\def\csname ev@practice-transfer/g-s3/ci-lo-pp\endcsname{+21.38}
\expandafter\def\csname ev@practice-transfer/g-s3/diff-pp\endcsname{+26.32}
\expandafter\def\csname ev@practice-transfer/g-s3/losses\endcsname{16}
\expandafter\def\csname ev@practice-transfer/g-s3/outcome\endcsname{improvement supported}
\expandafter\def\csname ev@practice-transfer/g-s3/p-recorded\endcsname{\ensuremath{3.5\times 10^{-21}}}
\expandafter\def\csname ev@practice-transfer/g-s3/protocol\endcsname{practice-transfer}
\expandafter\def\csname ev@practice-transfer/g-s3/task-p\endcsname{\ensuremath{3.5\times 10^{-21}}}
\expandafter\def\csname ev@practice-transfer/g-s3/tasks\endcsname{228}
\expandafter\def\csname ev@practice-transfer/g-s3/ties\endcsname{92}
\expandafter\def\csname ev@practice-transfer/g-s3/uncounted\endcsname{0}
\expandafter\def\csname ev@practice-transfer/g-s3/uncounted-pct\endcsname{0.00}
\expandafter\def\csname ev@practice-transfer/g-s3/unit\endcsname{task}
\expandafter\def\csname ev@practice-transfer/g-s3/wins\endcsname{120}
\expandafter\def\csname ev@practice/8b-s1/all/base-counted\endcsname{893}
\expandafter\def\csname ev@practice/8b-s1/all/base-passes\endcsname{139}
\expandafter\def\csname ev@practice/8b-s1/all/base-pct\endcsname{15.57}
\expandafter\def\csname ev@practice/8b-s1/all/cand-counted\endcsname{912}
\expandafter\def\csname ev@practice/8b-s1/all/cand-passes\endcsname{463}
\expandafter\def\csname ev@practice/8b-s1/all/cand-pct\endcsname{50.77}
\expandafter\def\csname ev@practice/8b-s1/all/ci-hi-pp\endcsname{+41.53}
\expandafter\def\csname ev@practice/8b-s1/all/ci-lo-pp\endcsname{+30.49}
\expandafter\def\csname ev@practice/8b-s1/all/diff-pp\endcsname{+36.06}
\expandafter\def\csname ev@practice/8b-s1/all/losses\endcsname{14}
\expandafter\def\csname ev@practice/8b-s1/all/outcome\endcsname{excluded}
\expandafter\def\csname ev@practice/8b-s1/all/p-recorded\endcsname{\ensuremath{8.9\times 10^{-25}}}
\expandafter\def\csname ev@practice/8b-s1/all/protocol\endcsname{practice}
\expandafter\def\csname ev@practice/8b-s1/all/task-p\endcsname{\ensuremath{8.9\times 10^{-25}}}
\expandafter\def\csname ev@practice/8b-s1/all/tasks\endcsname{228}
\expandafter\def\csname ev@practice/8b-s1/all/ties\endcsname{85}
\expandafter\def\csname ev@practice/8b-s1/all/uncounted\endcsname{19}
\expandafter\def\csname ev@practice/8b-s1/all/uncounted-pct\endcsname{2.08}
\expandafter\def\csname ev@practice/8b-s1/all/unit\endcsname{task}
\expandafter\def\csname ev@practice/8b-s1/all/wins\endcsname{129}
\expandafter\def\csname ev@practice/8b-s1/composites/base-counted\endcsname{222}
\expandafter\def\csname ev@practice/8b-s1/composites/base-passes\endcsname{0}
\expandafter\def\csname ev@practice/8b-s1/composites/base-pct\endcsname{0.00}
\expandafter\def\csname ev@practice/8b-s1/composites/cand-counted\endcsname{240}
\expandafter\def\csname ev@practice/8b-s1/composites/cand-passes\endcsname{7}
\expandafter\def\csname ev@practice/8b-s1/composites/cand-pct\endcsname{2.92}
\expandafter\def\csname ev@practice/8b-s1/composites/ci-hi-pp\endcsname{+6.01}
\expandafter\def\csname ev@practice/8b-s1/composites/ci-lo-pp\endcsname{+0.44}
\expandafter\def\csname ev@practice/8b-s1/composites/diff-pp\endcsname{+2.70}
\expandafter\def\csname ev@practice/8b-s1/composites/losses\endcsname{0}
\expandafter\def\csname ev@practice/8b-s1/composites/outcome\endcsname{excluded}
\expandafter\def\csname ev@practice/8b-s1/composites/p-recorded\endcsname{0.062}
\expandafter\def\csname ev@practice/8b-s1/composites/protocol\endcsname{practice}
\expandafter\def\csname ev@practice/8b-s1/composites/task-p\endcsname{0.062}
\expandafter\def\csname ev@practice/8b-s1/composites/tasks\endcsname{60}
\expandafter\def\csname ev@practice/8b-s1/composites/ties\endcsname{56}
\expandafter\def\csname ev@practice/8b-s1/composites/uncounted\endcsname{0}
\expandafter\def\csname ev@practice/8b-s1/composites/uncounted-pct\endcsname{0.00}
\expandafter\def\csname ev@practice/8b-s1/composites/unit\endcsname{task}
\expandafter\def\csname ev@practice/8b-s1/composites/wins\endcsname{4}
\expandafter\def\csname ev@practice/8b-s1/exercises/base-counted\endcsname{671}
\expandafter\def\csname ev@practice/8b-s1/exercises/base-passes\endcsname{139}
\expandafter\def\csname ev@practice/8b-s1/exercises/base-pct\endcsname{20.72}
\expandafter\def\csname ev@practice/8b-s1/exercises/cand-counted\endcsname{672}
\expandafter\def\csname ev@practice/8b-s1/exercises/cand-passes\endcsname{456}
\expandafter\def\csname ev@practice/8b-s1/exercises/cand-pct\endcsname{67.86}
\expandafter\def\csname ev@practice/8b-s1/exercises/ci-hi-pp\endcsname{+53.65}
\expandafter\def\csname ev@practice/8b-s1/exercises/ci-lo-pp\endcsname{+40.51}
\expandafter\def\csname ev@practice/8b-s1/exercises/diff-pp\endcsname{+47.09}
\expandafter\def\csname ev@practice/8b-s1/exercises/losses\endcsname{14}
\expandafter\def\csname ev@practice/8b-s1/exercises/outcome\endcsname{excluded}
\expandafter\def\csname ev@practice/8b-s1/exercises/p-recorded\endcsname{\ensuremath{9.4\times 10^{-24}}}
\expandafter\def\csname ev@practice/8b-s1/exercises/protocol\endcsname{practice}
\expandafter\def\csname ev@practice/8b-s1/exercises/task-p\endcsname{\ensuremath{9.4\times 10^{-24}}}
\expandafter\def\csname ev@practice/8b-s1/exercises/tasks\endcsname{168}
\expandafter\def\csname ev@practice/8b-s1/exercises/ties\endcsname{29}
\expandafter\def\csname ev@practice/8b-s1/exercises/uncounted\endcsname{0}
\expandafter\def\csname ev@practice/8b-s1/exercises/uncounted-pct\endcsname{0.00}
\expandafter\def\csname ev@practice/8b-s1/exercises/unit\endcsname{task}
\expandafter\def\csname ev@practice/8b-s1/exercises/wins\endcsname{125}
\expandafter\def\csname ev@practice/8b-s2/all/base-counted\endcsname{903}
\expandafter\def\csname ev@practice/8b-s2/all/base-passes\endcsname{138}
\expandafter\def\csname ev@practice/8b-s2/all/base-pct\endcsname{15.28}
\expandafter\def\csname ev@practice/8b-s2/all/cand-counted\endcsname{908}
\expandafter\def\csname ev@practice/8b-s2/all/cand-passes\endcsname{461}
\expandafter\def\csname ev@practice/8b-s2/all/cand-pct\endcsname{50.77}
\expandafter\def\csname ev@practice/8b-s2/all/ci-hi-pp\endcsname{+40.99}
\expandafter\def\csname ev@practice/8b-s2/all/ci-lo-pp\endcsname{+30.45}
\expandafter\def\csname ev@practice/8b-s2/all/diff-pp\endcsname{+35.77}
\expandafter\def\csname ev@practice/8b-s2/all/losses\endcsname{10}
\expandafter\def\csname ev@practice/8b-s2/all/outcome\endcsname{retention supported}
\expandafter\def\csname ev@practice/8b-s2/all/p-recorded\endcsname{\ensuremath{3.1\times 10^{-30}}}
\expandafter\def\csname ev@practice/8b-s2/all/protocol\endcsname{practice}
\expandafter\def\csname ev@practice/8b-s2/all/task-p\endcsname{\ensuremath{3.1\times 10^{-30}}}
\expandafter\def\csname ev@practice/8b-s2/all/tasks\endcsname{227}
\expandafter\def\csname ev@practice/8b-s2/all/ties\endcsname{79}
\expandafter\def\csname ev@practice/8b-s2/all/uncounted\endcsname{9}
\expandafter\def\csname ev@practice/8b-s2/all/uncounted-pct\endcsname{0.99}
\expandafter\def\csname ev@practice/8b-s2/all/unit\endcsname{task}
\expandafter\def\csname ev@practice/8b-s2/all/wins\endcsname{138}
\expandafter\def\csname ev@practice/8b-s2/composites/base-counted\endcsname{231}
\expandafter\def\csname ev@practice/8b-s2/composites/base-passes\endcsname{0}
\expandafter\def\csname ev@practice/8b-s2/composites/base-pct\endcsname{0.00}
\expandafter\def\csname ev@practice/8b-s2/composites/cand-counted\endcsname{236}
\expandafter\def\csname ev@practice/8b-s2/composites/cand-passes\endcsname{5}
\expandafter\def\csname ev@practice/8b-s2/composites/cand-pct\endcsname{2.12}
\expandafter\def\csname ev@practice/8b-s2/composites/ci-hi-pp\endcsname{+4.27}
\expandafter\def\csname ev@practice/8b-s2/composites/ci-lo-pp\endcsname{+0.43}
\expandafter\def\csname ev@practice/8b-s2/composites/diff-pp\endcsname{+2.16}
\expandafter\def\csname ev@practice/8b-s2/composites/losses\endcsname{0}
\expandafter\def\csname ev@practice/8b-s2/composites/outcome\endcsname{retention supported}
\expandafter\def\csname ev@practice/8b-s2/composites/p-recorded\endcsname{0.031}
\expandafter\def\csname ev@practice/8b-s2/composites/protocol\endcsname{practice}
\expandafter\def\csname ev@practice/8b-s2/composites/task-p\endcsname{0.031}
\expandafter\def\csname ev@practice/8b-s2/composites/tasks\endcsname{59}
\expandafter\def\csname ev@practice/8b-s2/composites/ties\endcsname{54}
\expandafter\def\csname ev@practice/8b-s2/composites/uncounted\endcsname{0}
\expandafter\def\csname ev@practice/8b-s2/composites/uncounted-pct\endcsname{0.00}
\expandafter\def\csname ev@practice/8b-s2/composites/unit\endcsname{task}
\expandafter\def\csname ev@practice/8b-s2/composites/wins\endcsname{5}
\expandafter\def\csname ev@practice/8b-s2/exercises/base-counted\endcsname{672}
\expandafter\def\csname ev@practice/8b-s2/exercises/base-passes\endcsname{138}
\expandafter\def\csname ev@practice/8b-s2/exercises/base-pct\endcsname{20.54}
\expandafter\def\csname ev@practice/8b-s2/exercises/cand-counted\endcsname{672}
\expandafter\def\csname ev@practice/8b-s2/exercises/cand-passes\endcsname{456}
\expandafter\def\csname ev@practice/8b-s2/exercises/cand-pct\endcsname{67.86}
\expandafter\def\csname ev@practice/8b-s2/exercises/ci-hi-pp\endcsname{+53.57}
\expandafter\def\csname ev@practice/8b-s2/exercises/ci-lo-pp\endcsname{+40.92}
\expandafter\def\csname ev@practice/8b-s2/exercises/diff-pp\endcsname{+47.32}
\expandafter\def\csname ev@practice/8b-s2/exercises/losses\endcsname{10}
\expandafter\def\csname ev@practice/8b-s2/exercises/outcome\endcsname{retention supported}
\expandafter\def\csname ev@practice/8b-s2/exercises/p-recorded\endcsname{\ensuremath{6.9\times 10^{-29}}}
\expandafter\def\csname ev@practice/8b-s2/exercises/protocol\endcsname{practice}
\expandafter\def\csname ev@practice/8b-s2/exercises/task-p\endcsname{\ensuremath{6.9\times 10^{-29}}}
\expandafter\def\csname ev@practice/8b-s2/exercises/tasks\endcsname{168}
\expandafter\def\csname ev@practice/8b-s2/exercises/ties\endcsname{25}
\expandafter\def\csname ev@practice/8b-s2/exercises/uncounted\endcsname{0}
\expandafter\def\csname ev@practice/8b-s2/exercises/uncounted-pct\endcsname{0.00}
\expandafter\def\csname ev@practice/8b-s2/exercises/unit\endcsname{task}
\expandafter\def\csname ev@practice/8b-s2/exercises/wins\endcsname{133}
\expandafter\def\csname ev@practice/8b-s3/all/base-counted\endcsname{897}
\expandafter\def\csname ev@practice/8b-s3/all/base-passes\endcsname{134}
\expandafter\def\csname ev@practice/8b-s3/all/base-pct\endcsname{14.94}
\expandafter\def\csname ev@practice/8b-s3/all/cand-counted\endcsname{912}
\expandafter\def\csname ev@practice/8b-s3/all/cand-passes\endcsname{408}
\expandafter\def\csname ev@practice/8b-s3/all/cand-pct\endcsname{44.74}
\expandafter\def\csname ev@practice/8b-s3/all/ci-hi-pp\endcsname{+35.35}
\expandafter\def\csname ev@practice/8b-s3/all/ci-lo-pp\endcsname{+25.42}
\expandafter\def\csname ev@practice/8b-s3/all/diff-pp\endcsname{+30.32}
\expandafter\def\csname ev@practice/8b-s3/all/losses\endcsname{14}
\expandafter\def\csname ev@practice/8b-s3/all/outcome\endcsname{retention supported}
\expandafter\def\csname ev@practice/8b-s3/all/p-recorded\endcsname{\ensuremath{2.9\times 10^{-24}}}
\expandafter\def\csname ev@practice/8b-s3/all/protocol\endcsname{practice}
\expandafter\def\csname ev@practice/8b-s3/all/task-p\endcsname{\ensuremath{2.9\times 10^{-24}}}
\expandafter\def\csname ev@practice/8b-s3/all/tasks\endcsname{228}
\expandafter\def\csname ev@practice/8b-s3/all/ties\endcsname{87}
\expandafter\def\csname ev@practice/8b-s3/all/uncounted\endcsname{15}
\expandafter\def\csname ev@practice/8b-s3/all/uncounted-pct\endcsname{1.64}
\expandafter\def\csname ev@practice/8b-s3/all/unit\endcsname{task}
\expandafter\def\csname ev@practice/8b-s3/all/wins\endcsname{127}
\expandafter\def\csname ev@practice/8b-s3/composites/base-counted\endcsname{227}
\expandafter\def\csname ev@practice/8b-s3/composites/base-passes\endcsname{0}
\expandafter\def\csname ev@practice/8b-s3/composites/base-pct\endcsname{0.00}
\expandafter\def\csname ev@practice/8b-s3/composites/cand-counted\endcsname{240}
\expandafter\def\csname ev@practice/8b-s3/composites/cand-passes\endcsname{10}
\expandafter\def\csname ev@practice/8b-s3/composites/cand-pct\endcsname{4.17}
\expandafter\def\csname ev@practice/8b-s3/composites/ci-hi-pp\endcsname{+7.89}
\expandafter\def\csname ev@practice/8b-s3/composites/ci-lo-pp\endcsname{+1.36}
\expandafter\def\csname ev@practice/8b-s3/composites/diff-pp\endcsname{+4.41}
\expandafter\def\csname ev@practice/8b-s3/composites/losses\endcsname{0}
\expandafter\def\csname ev@practice/8b-s3/composites/outcome\endcsname{retention supported}
\expandafter\def\csname ev@practice/8b-s3/composites/p-recorded\endcsname{0.0078}
\expandafter\def\csname ev@practice/8b-s3/composites/protocol\endcsname{practice}
\expandafter\def\csname ev@practice/8b-s3/composites/task-p\endcsname{0.0078}
\expandafter\def\csname ev@practice/8b-s3/composites/tasks\endcsname{60}
\expandafter\def\csname ev@practice/8b-s3/composites/ties\endcsname{53}
\expandafter\def\csname ev@practice/8b-s3/composites/uncounted\endcsname{0}
\expandafter\def\csname ev@practice/8b-s3/composites/uncounted-pct\endcsname{0.00}
\expandafter\def\csname ev@practice/8b-s3/composites/unit\endcsname{task}
\expandafter\def\csname ev@practice/8b-s3/composites/wins\endcsname{7}
\expandafter\def\csname ev@practice/8b-s3/exercises/base-counted\endcsname{670}
\expandafter\def\csname ev@practice/8b-s3/exercises/base-passes\endcsname{134}
\expandafter\def\csname ev@practice/8b-s3/exercises/base-pct\endcsname{20.00}
\expandafter\def\csname ev@practice/8b-s3/exercises/cand-counted\endcsname{672}
\expandafter\def\csname ev@practice/8b-s3/exercises/cand-passes\endcsname{398}
\expandafter\def\csname ev@practice/8b-s3/exercises/cand-pct\endcsname{59.23}
\expandafter\def\csname ev@practice/8b-s3/exercises/ci-hi-pp\endcsname{+45.24}
\expandafter\def\csname ev@practice/8b-s3/exercises/ci-lo-pp\endcsname{+32.89}
\expandafter\def\csname ev@practice/8b-s3/exercises/diff-pp\endcsname{+39.10}
\expandafter\def\csname ev@practice/8b-s3/exercises/losses\endcsname{14}
\expandafter\def\csname ev@practice/8b-s3/exercises/outcome\endcsname{retention supported}
\expandafter\def\csname ev@practice/8b-s3/exercises/p-recorded\endcsname{\ensuremath{1.8\times 10^{-22}}}
\expandafter\def\csname ev@practice/8b-s3/exercises/protocol\endcsname{practice}
\expandafter\def\csname ev@practice/8b-s3/exercises/task-p\endcsname{\ensuremath{1.8\times 10^{-22}}}
\expandafter\def\csname ev@practice/8b-s3/exercises/tasks\endcsname{168}
\expandafter\def\csname ev@practice/8b-s3/exercises/ties\endcsname{34}
\expandafter\def\csname ev@practice/8b-s3/exercises/uncounted\endcsname{0}
\expandafter\def\csname ev@practice/8b-s3/exercises/uncounted-pct\endcsname{0.00}
\expandafter\def\csname ev@practice/8b-s3/exercises/unit\endcsname{task}
\expandafter\def\csname ev@practice/8b-s3/exercises/wins\endcsname{120}
\expandafter\def\csname ev@practice/c-s1/all/base-counted\endcsname{912}
\expandafter\def\csname ev@practice/c-s1/all/base-passes\endcsname{142}
\expandafter\def\csname ev@practice/c-s1/all/base-pct\endcsname{15.57}
\expandafter\def\csname ev@practice/c-s1/all/cand-counted\endcsname{912}
\expandafter\def\csname ev@practice/c-s1/all/cand-passes\endcsname{336}
\expandafter\def\csname ev@practice/c-s1/all/cand-pct\endcsname{36.84}
\expandafter\def\csname ev@practice/c-s1/all/ci-hi-pp\endcsname{+26.32}
\expandafter\def\csname ev@practice/c-s1/all/ci-lo-pp\endcsname{+16.45}
\expandafter\def\csname ev@practice/c-s1/all/diff-pp\endcsname{+21.27}
\expandafter\def\csname ev@practice/c-s1/all/losses\endcsname{17}
\expandafter\def\csname ev@practice/c-s1/all/outcome\endcsname{retention supported}
\expandafter\def\csname ev@practice/c-s1/all/p-recorded\endcsname{\ensuremath{1.7\times 10^{-16}}}
\expandafter\def\csname ev@practice/c-s1/all/protocol\endcsname{practice}
\expandafter\def\csname ev@practice/c-s1/all/task-p\endcsname{\ensuremath{1.7\times 10^{-16}}}
\expandafter\def\csname ev@practice/c-s1/all/tasks\endcsname{228}
\expandafter\def\csname ev@practice/c-s1/all/ties\endcsname{108}
\expandafter\def\csname ev@practice/c-s1/all/uncounted\endcsname{0}
\expandafter\def\csname ev@practice/c-s1/all/uncounted-pct\endcsname{0.00}
\expandafter\def\csname ev@practice/c-s1/all/unit\endcsname{task}
\expandafter\def\csname ev@practice/c-s1/all/wins\endcsname{103}
\expandafter\def\csname ev@practice/c-s1/composites/base-counted\endcsname{240}
\expandafter\def\csname ev@practice/c-s1/composites/base-passes\endcsname{0}
\expandafter\def\csname ev@practice/c-s1/composites/base-pct\endcsname{0.00}
\expandafter\def\csname ev@practice/c-s1/composites/cand-counted\endcsname{240}
\expandafter\def\csname ev@practice/c-s1/composites/cand-passes\endcsname{0}
\expandafter\def\csname ev@practice/c-s1/composites/cand-pct\endcsname{0.00}
\expandafter\def\csname ev@practice/c-s1/composites/ci-hi-pp\endcsname{+0.00}
\expandafter\def\csname ev@practice/c-s1/composites/ci-lo-pp\endcsname{+0.00}
\expandafter\def\csname ev@practice/c-s1/composites/diff-pp\endcsname{+0.00}
\expandafter\def\csname ev@practice/c-s1/composites/losses\endcsname{0}
\expandafter\def\csname ev@practice/c-s1/composites/outcome\endcsname{retention supported}
\expandafter\def\csname ev@practice/c-s1/composites/p-recorded\endcsname{1}
\expandafter\def\csname ev@practice/c-s1/composites/protocol\endcsname{practice}
\expandafter\def\csname ev@practice/c-s1/composites/task-p\endcsname{1}
\expandafter\def\csname ev@practice/c-s1/composites/tasks\endcsname{60}
\expandafter\def\csname ev@practice/c-s1/composites/ties\endcsname{60}
\expandafter\def\csname ev@practice/c-s1/composites/uncounted\endcsname{0}
\expandafter\def\csname ev@practice/c-s1/composites/uncounted-pct\endcsname{0.00}
\expandafter\def\csname ev@practice/c-s1/composites/unit\endcsname{task}
\expandafter\def\csname ev@practice/c-s1/composites/wins\endcsname{0}
\expandafter\def\csname ev@practice/c-s1/exercises/base-counted\endcsname{672}
\expandafter\def\csname ev@practice/c-s1/exercises/base-passes\endcsname{142}
\expandafter\def\csname ev@practice/c-s1/exercises/base-pct\endcsname{21.13}
\expandafter\def\csname ev@practice/c-s1/exercises/cand-counted\endcsname{672}
\expandafter\def\csname ev@practice/c-s1/exercises/cand-passes\endcsname{336}
\expandafter\def\csname ev@practice/c-s1/exercises/cand-pct\endcsname{50.00}
\expandafter\def\csname ev@practice/c-s1/exercises/ci-hi-pp\endcsname{+35.12}
\expandafter\def\csname ev@practice/c-s1/exercises/ci-lo-pp\endcsname{+22.32}
\expandafter\def\csname ev@practice/c-s1/exercises/diff-pp\endcsname{+28.87}
\expandafter\def\csname ev@practice/c-s1/exercises/losses\endcsname{17}
\expandafter\def\csname ev@practice/c-s1/exercises/outcome\endcsname{retention supported}
\expandafter\def\csname ev@practice/c-s1/exercises/p-recorded\endcsname{\ensuremath{1.7\times 10^{-16}}}
\expandafter\def\csname ev@practice/c-s1/exercises/protocol\endcsname{practice}
\expandafter\def\csname ev@practice/c-s1/exercises/task-p\endcsname{\ensuremath{1.7\times 10^{-16}}}
\expandafter\def\csname ev@practice/c-s1/exercises/tasks\endcsname{168}
\expandafter\def\csname ev@practice/c-s1/exercises/ties\endcsname{48}
\expandafter\def\csname ev@practice/c-s1/exercises/uncounted\endcsname{0}
\expandafter\def\csname ev@practice/c-s1/exercises/uncounted-pct\endcsname{0.00}
\expandafter\def\csname ev@practice/c-s1/exercises/unit\endcsname{task}
\expandafter\def\csname ev@practice/c-s1/exercises/wins\endcsname{103}
\expandafter\def\csname ev@practice/d-s1/all/base-counted\endcsname{910}
\expandafter\def\csname ev@practice/d-s1/all/base-passes\endcsname{141}
\expandafter\def\csname ev@practice/d-s1/all/base-pct\endcsname{15.49}
\expandafter\def\csname ev@practice/d-s1/all/cand-counted\endcsname{912}
\expandafter\def\csname ev@practice/d-s1/all/cand-passes\endcsname{360}
\expandafter\def\csname ev@practice/d-s1/all/cand-pct\endcsname{39.47}
\expandafter\def\csname ev@practice/d-s1/all/ci-hi-pp\endcsname{+28.98}
\expandafter\def\csname ev@practice/d-s1/all/ci-lo-pp\endcsname{+19.10}
\expandafter\def\csname ev@practice/d-s1/all/diff-pp\endcsname{+24.07}
\expandafter\def\csname ev@practice/d-s1/all/losses\endcsname{13}
\expandafter\def\csname ev@practice/d-s1/all/outcome\endcsname{retention supported}
\expandafter\def\csname ev@practice/d-s1/all/p-recorded\endcsname{\ensuremath{2.3\times 10^{-20}}}
\expandafter\def\csname ev@practice/d-s1/all/protocol\endcsname{practice}
\expandafter\def\csname ev@practice/d-s1/all/task-p\endcsname{\ensuremath{2.3\times 10^{-20}}}
\expandafter\def\csname ev@practice/d-s1/all/tasks\endcsname{228}
\expandafter\def\csname ev@practice/d-s1/all/ties\endcsname{106}
\expandafter\def\csname ev@practice/d-s1/all/uncounted\endcsname{2}
\expandafter\def\csname ev@practice/d-s1/all/uncounted-pct\endcsname{0.22}
\expandafter\def\csname ev@practice/d-s1/all/unit\endcsname{task}
\expandafter\def\csname ev@practice/d-s1/all/wins\endcsname{109}
\expandafter\def\csname ev@practice/d-s1/composites/base-counted\endcsname{238}
\expandafter\def\csname ev@practice/d-s1/composites/base-passes\endcsname{0}
\expandafter\def\csname ev@practice/d-s1/composites/base-pct\endcsname{0.00}
\expandafter\def\csname ev@practice/d-s1/composites/cand-counted\endcsname{240}
\expandafter\def\csname ev@practice/d-s1/composites/cand-passes\endcsname{1}
\expandafter\def\csname ev@practice/d-s1/composites/cand-pct\endcsname{0.42}
\expandafter\def\csname ev@practice/d-s1/composites/ci-hi-pp\endcsname{+1.27}
\expandafter\def\csname ev@practice/d-s1/composites/ci-lo-pp\endcsname{+0.00}
\expandafter\def\csname ev@practice/d-s1/composites/diff-pp\endcsname{+0.42}
\expandafter\def\csname ev@practice/d-s1/composites/losses\endcsname{0}
\expandafter\def\csname ev@practice/d-s1/composites/outcome\endcsname{retention supported}
\expandafter\def\csname ev@practice/d-s1/composites/p-recorded\endcsname{0.5}
\expandafter\def\csname ev@practice/d-s1/composites/protocol\endcsname{practice}
\expandafter\def\csname ev@practice/d-s1/composites/task-p\endcsname{0.5}
\expandafter\def\csname ev@practice/d-s1/composites/tasks\endcsname{60}
\expandafter\def\csname ev@practice/d-s1/composites/ties\endcsname{59}
\expandafter\def\csname ev@practice/d-s1/composites/uncounted\endcsname{0}
\expandafter\def\csname ev@practice/d-s1/composites/uncounted-pct\endcsname{0.00}
\expandafter\def\csname ev@practice/d-s1/composites/unit\endcsname{task}
\expandafter\def\csname ev@practice/d-s1/composites/wins\endcsname{1}
\expandafter\def\csname ev@practice/d-s1/exercises/base-counted\endcsname{672}
\expandafter\def\csname ev@practice/d-s1/exercises/base-passes\endcsname{141}
\expandafter\def\csname ev@practice/d-s1/exercises/base-pct\endcsname{20.98}
\expandafter\def\csname ev@practice/d-s1/exercises/cand-counted\endcsname{672}
\expandafter\def\csname ev@practice/d-s1/exercises/cand-passes\endcsname{359}
\expandafter\def\csname ev@practice/d-s1/exercises/cand-pct\endcsname{53.42}
\expandafter\def\csname ev@practice/d-s1/exercises/ci-hi-pp\endcsname{+38.54}
\expandafter\def\csname ev@practice/d-s1/exercises/ci-lo-pp\endcsname{+26.19}
\expandafter\def\csname ev@practice/d-s1/exercises/diff-pp\endcsname{+32.44}
\expandafter\def\csname ev@practice/d-s1/exercises/losses\endcsname{13}
\expandafter\def\csname ev@practice/d-s1/exercises/outcome\endcsname{retention supported}
\expandafter\def\csname ev@practice/d-s1/exercises/p-recorded\endcsname{\ensuremath{4.2\times 10^{-20}}}
\expandafter\def\csname ev@practice/d-s1/exercises/protocol\endcsname{practice}
\expandafter\def\csname ev@practice/d-s1/exercises/task-p\endcsname{\ensuremath{4.2\times 10^{-20}}}
\expandafter\def\csname ev@practice/d-s1/exercises/tasks\endcsname{168}
\expandafter\def\csname ev@practice/d-s1/exercises/ties\endcsname{47}
\expandafter\def\csname ev@practice/d-s1/exercises/uncounted\endcsname{0}
\expandafter\def\csname ev@practice/d-s1/exercises/uncounted-pct\endcsname{0.00}
\expandafter\def\csname ev@practice/d-s1/exercises/unit\endcsname{task}
\expandafter\def\csname ev@practice/d-s1/exercises/wins\endcsname{108}
\expandafter\def\csname ev@practice/g-s3/all/base-counted\endcsname{912}
\expandafter\def\csname ev@practice/g-s3/all/base-passes\endcsname{138}
\expandafter\def\csname ev@practice/g-s3/all/base-pct\endcsname{15.13}
\expandafter\def\csname ev@practice/g-s3/all/cand-counted\endcsname{912}
\expandafter\def\csname ev@practice/g-s3/all/cand-passes\endcsname{378}
\expandafter\def\csname ev@practice/g-s3/all/cand-pct\endcsname{41.45}
\expandafter\def\csname ev@practice/g-s3/all/ci-hi-pp\endcsname{+31.47}
\expandafter\def\csname ev@practice/g-s3/all/ci-lo-pp\endcsname{+21.38}
\expandafter\def\csname ev@practice/g-s3/all/diff-pp\endcsname{+26.32}
\expandafter\def\csname ev@practice/g-s3/all/losses\endcsname{16}
\expandafter\def\csname ev@practice/g-s3/all/outcome\endcsname{retention supported}
\expandafter\def\csname ev@practice/g-s3/all/p-recorded\endcsname{\ensuremath{3.5\times 10^{-21}}}
\expandafter\def\csname ev@practice/g-s3/all/protocol\endcsname{practice}
\expandafter\def\csname ev@practice/g-s3/all/task-p\endcsname{\ensuremath{3.5\times 10^{-21}}}
\expandafter\def\csname ev@practice/g-s3/all/tasks\endcsname{228}
\expandafter\def\csname ev@practice/g-s3/all/ties\endcsname{92}
\expandafter\def\csname ev@practice/g-s3/all/uncounted\endcsname{0}
\expandafter\def\csname ev@practice/g-s3/all/uncounted-pct\endcsname{0.00}
\expandafter\def\csname ev@practice/g-s3/all/unit\endcsname{task}
\expandafter\def\csname ev@practice/g-s3/all/wins\endcsname{120}
\expandafter\def\csname ev@practice/g-s3/composites/base-counted\endcsname{240}
\expandafter\def\csname ev@practice/g-s3/composites/base-passes\endcsname{0}
\expandafter\def\csname ev@practice/g-s3/composites/base-pct\endcsname{0.00}
\expandafter\def\csname ev@practice/g-s3/composites/cand-counted\endcsname{240}
\expandafter\def\csname ev@practice/g-s3/composites/cand-passes\endcsname{6}
\expandafter\def\csname ev@practice/g-s3/composites/cand-pct\endcsname{2.50}
\expandafter\def\csname ev@practice/g-s3/composites/ci-hi-pp\endcsname{+5.00}
\expandafter\def\csname ev@practice/g-s3/composites/ci-lo-pp\endcsname{+0.42}
\expandafter\def\csname ev@practice/g-s3/composites/diff-pp\endcsname{+2.50}
\expandafter\def\csname ev@practice/g-s3/composites/losses\endcsname{0}
\expandafter\def\csname ev@practice/g-s3/composites/outcome\endcsname{retention supported}
\expandafter\def\csname ev@practice/g-s3/composites/p-recorded\endcsname{0.031}
\expandafter\def\csname ev@practice/g-s3/composites/protocol\endcsname{practice}
\expandafter\def\csname ev@practice/g-s3/composites/task-p\endcsname{0.031}
\expandafter\def\csname ev@practice/g-s3/composites/tasks\endcsname{60}
\expandafter\def\csname ev@practice/g-s3/composites/ties\endcsname{55}
\expandafter\def\csname ev@practice/g-s3/composites/uncounted\endcsname{0}
\expandafter\def\csname ev@practice/g-s3/composites/uncounted-pct\endcsname{0.00}
\expandafter\def\csname ev@practice/g-s3/composites/unit\endcsname{task}
\expandafter\def\csname ev@practice/g-s3/composites/wins\endcsname{5}
\expandafter\def\csname ev@practice/g-s3/exercises/base-counted\endcsname{672}
\expandafter\def\csname ev@practice/g-s3/exercises/base-passes\endcsname{138}
\expandafter\def\csname ev@practice/g-s3/exercises/base-pct\endcsname{20.54}
\expandafter\def\csname ev@practice/g-s3/exercises/cand-counted\endcsname{672}
\expandafter\def\csname ev@practice/g-s3/exercises/cand-passes\endcsname{372}
\expandafter\def\csname ev@practice/g-s3/exercises/cand-pct\endcsname{55.36}
\expandafter\def\csname ev@practice/g-s3/exercises/ci-hi-pp\endcsname{+41.22}
\expandafter\def\csname ev@practice/g-s3/exercises/ci-lo-pp\endcsname{+28.42}
\expandafter\def\csname ev@practice/g-s3/exercises/diff-pp\endcsname{+34.82}
\expandafter\def\csname ev@practice/g-s3/exercises/losses\endcsname{16}
\expandafter\def\csname ev@practice/g-s3/exercises/outcome\endcsname{retention supported}
\expandafter\def\csname ev@practice/g-s3/exercises/p-recorded\endcsname{\ensuremath{5.9\times 10^{-20}}}
\expandafter\def\csname ev@practice/g-s3/exercises/protocol\endcsname{practice}
\expandafter\def\csname ev@practice/g-s3/exercises/task-p\endcsname{\ensuremath{5.9\times 10^{-20}}}
\expandafter\def\csname ev@practice/g-s3/exercises/tasks\endcsname{168}
\expandafter\def\csname ev@practice/g-s3/exercises/ties\endcsname{37}
\expandafter\def\csname ev@practice/g-s3/exercises/uncounted\endcsname{0}
\expandafter\def\csname ev@practice/g-s3/exercises/uncounted-pct\endcsname{0.00}
\expandafter\def\csname ev@practice/g-s3/exercises/unit\endcsname{task}
\expandafter\def\csname ev@practice/g-s3/exercises/wins\endcsname{115}
\expandafter\def\csname ev@projection/4B/exam-gain-pp\endcsname{+2.75}
\expandafter\def\csname ev@projection/4B/g2/heldout-gain-pp\endcsname{+1.17}
\expandafter\def\csname ev@projection/4B/g2/observed-pp\endcsname{+0.60}
\expandafter\def\csname ev@projection/4B/g2/predicted-high-pp\endcsname{+0.24}
\expandafter\def\csname ev@projection/4B/g2/predicted-low-pp\endcsname{+0.12}
\expandafter\def\csname ev@projection/4B/g2/predicted-own-pp\endcsname{+0.12}
\expandafter\def\csname ev@projection/4B/g2/seed-high-pp\endcsname{+1.50}
\expandafter\def\csname ev@projection/4B/g2/seed-low-pp\endcsname{-0.25}
\expandafter\def\csname ev@projection/4B/g2/seeds\endcsname{5}
\expandafter\def\csname ev@projection/4B/g3/heldout-gain-pp\endcsname{+5.20}
\expandafter\def\csname ev@projection/4B/g3/observed-pp\endcsname{+0.60}
\expandafter\def\csname ev@projection/4B/g3/predicted-high-pp\endcsname{+1.06}
\expandafter\def\csname ev@projection/4B/g3/predicted-low-pp\endcsname{+0.55}
\expandafter\def\csname ev@projection/4B/g3/predicted-own-pp\endcsname{+0.55}
\expandafter\def\csname ev@projection/4B/g3/seed-high-pp\endcsname{+1.25}
\expandafter\def\csname ev@projection/4B/g3/seed-low-pp\endcsname{+0.00}
\expandafter\def\csname ev@projection/4B/g3/seeds\endcsname{5}
\expandafter\def\csname ev@projection/4B/heldout-gain-pp\endcsname{+25.87}
\expandafter\def\csname ev@projection/4B/ratio\endcsname{0.11}
\expandafter\def\csname ev@projection/4B/ratio-basis\endcsname{descriptive}
\expandafter\def\csname ev@projection/8B/exam-gain-pp\endcsname{+6.25}
\expandafter\def\csname ev@projection/8B/g2/heldout-gain-pp\endcsname{+6.12}
\expandafter\def\csname ev@projection/8B/g2/observed-pp\endcsname{-0.06}
\expandafter\def\csname ev@projection/8B/g2/predicted-high-pp\endcsname{+1.25}
\expandafter\def\csname ev@projection/8B/g2/predicted-low-pp\endcsname{+0.65}
\expandafter\def\csname ev@projection/8B/g2/predicted-own-pp\endcsname{+1.13}
\expandafter\def\csname ev@projection/8B/g2/seed-high-pp\endcsname{+0.75}
\expandafter\def\csname ev@projection/8B/g2/seed-low-pp\endcsname{-0.75}
\expandafter\def\csname ev@projection/8B/g2/seeds\endcsname{4}
\expandafter\def\csname ev@projection/8B/g3/heldout-gain-pp\endcsname{+4.93}
\expandafter\def\csname ev@projection/8B/g3/observed-pp\endcsname{+0.75}
\expandafter\def\csname ev@projection/8B/g3/predicted-high-pp\endcsname{+1.01}
\expandafter\def\csname ev@projection/8B/g3/predicted-low-pp\endcsname{+0.52}
\expandafter\def\csname ev@projection/8B/g3/predicted-own-pp\endcsname{+0.91}
\expandafter\def\csname ev@projection/8B/g3/seed-high-pp\endcsname{+1.25}
\expandafter\def\csname ev@projection/8B/g3/seed-low-pp\endcsname{+0.00}
\expandafter\def\csname ev@projection/8B/g3/seeds\endcsname{3}
\expandafter\def\csname ev@projection/8B/heldout-gain-pp\endcsname{+33.93}
\expandafter\def\csname ev@projection/8B/ratio\endcsname{0.18}
\expandafter\def\csname ev@projection/8B/ratio-basis\endcsname{excluded evaluation, descriptive}
\expandafter\def\csname ev@projection/arms/b/ratio\endcsname{0.04}
\expandafter\def\csname ev@projection/arms/c/ratio\endcsname{0.02}
\expandafter\def\csname ev@projection/arms/d/ratio\endcsname{0.03}
\expandafter\def\csname ev@projection/arms/g/ratio\endcsname{0.20}
\expandafter\def\csname ev@projection/arms/ratio-max\endcsname{0.20}
\expandafter\def\csname ev@projection/arms/ratio-min\endcsname{0.02}
\expandafter\def\csname ev@projection/arms/ratios\endcsname{4}
\expandafter\def\csname ev@projection/excluded-ratio/predicted-max-pp\endcsname{1.1}
\expandafter\def\csname ev@projection/excluded-ratio/predicted-min-pp\endcsname{0.9}
\expandafter\def\csname ev@projection/heldout-gain-max-pp\endcsname{6.1}
\expandafter\def\csname ev@projection/heldout-gain-min-pp\endcsname{1.2}
\expandafter\def\csname ev@projection/interval-half-max-pp\endcsname{3.8}
\expandafter\def\csname ev@projection/interval-half-min-pp\endcsname{2.5}
\expandafter\def\csname ev@projection/observed-max-pp\endcsname{+0.75}
\expandafter\def\csname ev@projection/observed-min-pp\endcsname{-0.06}
\expandafter\def\csname ev@projection/predicted-max-pp\endcsname{1.3}
\expandafter\def\csname ev@projection/predicted-min-pp\endcsname{0.1}
\expandafter\def\csname ev@projection/seed-diff-max-pp\endcsname{1.50}
\expandafter\def\csname ev@projection/steps\endcsname{4}
\expandafter\def\csname ev@projection/steps-within-interval\endcsname{4}
\expandafter\def\csname ev@projection/yardstick/high-ratio\endcsname{0.20}
\expandafter\def\csname ev@projection/yardstick/low-ratio\endcsname{0.11}
\expandafter\def\csname ev@repeat/4B/g1/stage2/attempts\endcsname{400}
\expandafter\def\csname ev@repeat/4B/g1/stage2/diff-pp\endcsname{+0.25}
\expandafter\def\csname ev@repeat/4B/g1/stage2/first-passes\endcsname{11}
\expandafter\def\csname ev@repeat/4B/g1/stage2/second-passes\endcsname{12}
\expandafter\def\csname ev@repeat/8B/g1/stage2/attempts\endcsname{400}
\expandafter\def\csname ev@repeat/8B/g1/stage2/diff-pp\endcsname{+1.25}
\expandafter\def\csname ev@repeat/8B/g1/stage2/first-passes\endcsname{25}
\expandafter\def\csname ev@repeat/8B/g1/stage2/second-passes\endcsname{30}
\expandafter\def\csname ev@repeat/8B/g2/stage3/attempts\endcsname{400}
\expandafter\def\csname ev@repeat/8B/g2/stage3/diff-pp\endcsname{+0.50}
\expandafter\def\csname ev@repeat/8B/g2/stage3/first-passes\endcsname{30}
\expandafter\def\csname ev@repeat/8B/g2/stage3/second-passes\endcsname{32}
\expandafter\def\csname ev@repeat/count\endcsname{3}
\expandafter\def\csname ev@repeat/max-diff-pp\endcsname{1.25}
\expandafter\def\csname ev@replications/exam/seeds-level\endcsname{3}
\expandafter\def\csname ev@replications/exam/seeds-negative\endcsname{3}
\expandafter\def\csname ev@replications/exam/seeds-positive\endcsname{11}
\expandafter\def\csname ev@replications/target/seeds\endcsname{17}
\expandafter\def\csname ev@replications/target/seeds-positive\endcsname{16}
\expandafter\def\csname ev@resolution/4B/g2/detectable-pp\endcsname{2.8}
\expandafter\def\csname ev@resolution/4B/g2/discordant\endcsname{21}
\expandafter\def\csname ev@resolution/4B/g2/pairs\endcsname{400}
\expandafter\def\csname ev@resolution/4B/g3/detectable-pp\endcsname{2.8}
\expandafter\def\csname ev@resolution/4B/g3/discordant\endcsname{21}
\expandafter\def\csname ev@resolution/4B/g3/pairs\endcsname{400}
\expandafter\def\csname ev@resolution/8B/g2/detectable-pp\endcsname{3.6}
\expandafter\def\csname ev@resolution/8B/g2/discordant\endcsname{34}
\expandafter\def\csname ev@resolution/8B/g2/pairs\endcsname{400}
\expandafter\def\csname ev@resolution/8B/g3/detectable-pp\endcsname{3.7}
\expandafter\def\csname ev@resolution/8B/g3/discordant\endcsname{36}
\expandafter\def\csname ev@resolution/8B/g3/pairs\endcsname{400}
\expandafter\def\csname ev@resolution/alpha\endcsname{0.05}
\expandafter\def\csname ev@resolution/detectable-max-pp\endcsname{3.7}
\expandafter\def\csname ev@resolution/detectable-min-pp\endcsname{2.8}
\expandafter\def\csname ev@resolution/power\endcsname{80}
\expandafter\def\csname ev@resolution/tasks-factor/high-min\endcsname{7}
\expandafter\def\csname ev@resolution/tasks-needed/high-max\endcsname{14\,000}
\expandafter\def\csname ev@resolution/tasks-needed/high-min\endcsname{720}
\expandafter\def\csname ev@resolution/tasks-needed/low-max\endcsname{52\,000}
\expandafter\def\csname ev@resolution/tasks-needed/low-min\endcsname{2\,700}
\expandafter\def\csname ev@seeds/b/lowest-validation-loss\endcsname{3}
\expandafter\def\csname ev@seeds/b/most-exam-passes\endcsname{3}
\expandafter\def\csname ev@seeds/c/lowest-validation-loss\endcsname{1}
\expandafter\def\csname ev@seeds/c/most-exam-passes\endcsname{1}
\expandafter\def\csname ev@seeds/d/lowest-validation-loss\endcsname{3}
\expandafter\def\csname ev@seeds/d/most-exam-passes\endcsname{1}
\expandafter\def\csname ev@seeds/f/lowest-validation-loss\endcsname{3}
\expandafter\def\csname ev@seeds/f/most-exam-passes\endcsname{1/2/3}
\expandafter\def\csname ev@seeds/g/lowest-validation-loss\endcsname{3}
\expandafter\def\csname ev@seeds/g/most-exam-passes\endcsname{3}
\expandafter\def\csname ev@seeds/s/lowest-validation-loss\endcsname{1}
\expandafter\def\csname ev@seeds/s/most-exam-passes\endcsname{1/2/3}
\expandafter\def\csname ev@teacher/exam/attempts\endcsname{400}
\expandafter\def\csname ev@teacher/exam/first-attempt-pct\endcsname{63}
\expandafter\def\csname ev@teacher/exam/passes\endcsname{249}
\expandafter\def\csname ev@teacher/exam/pct\endcsname{62.25}
\expandafter\def\csname ev@teacher/exam/solved\endcsname{94}

%% file: sections/00-abstract.tex
\begin{abstract}
A language model does not keep the experience of a task it completes. Where an attempt can be checked by executing it,
that experience can become training data without human-written tasks or labels. We present and evaluate a framework that generates
practice tasks, verifies attempts by execution, trains a small adapter (added weights; the model itself stays fixed) on
the verified records and repeats, one \emph{generation} per repetition; every model is measured on an examination that no
stage of the pipeline reads. In predicting the output of Lua programs, with an interpreter as the only source of labels, a
4-billion-parameter (4B) model's examination accuracy rises from \ev{lua/gen4-over-base/before-pct}\% to
\ev{lua/gen4-over-base/after-pct}\% over four generations. In editing files with Vim, a larger teacher model writes tasks
from the user manual and supplies most verified solutions; trained on them, the 4B model solves hand-written examination
tasks its base never solves (up to \ev{arms/g/exam/cand-pct}\% of attempts; the teacher passes \ev{teacher/exam/pct}\%),
while a control trained on the same records with the requests exchanged between tasks solves none. On
\ev{practice/g-s3/all/tasks} easier hand-written course exercises the pass rate rises from
\ev{practice-transfer/4B/base-pct-low}--\ev{practice-transfer/4B/base-pct-high}\% to
\ev{practice-transfer/4B/cand-pct-low}--\ev{practice-transfer/4B/cand-pct-high}\% (exploratory). An
iterative protocol then ran three generations with 4B and 8B models, every release decided by a fixed rule. In a first
generation the examination gain was a tenth to a fifth of the gain on \emph{held-out generated tasks}, new tasks from the
system's own generator. Later candidates were observed, on average over training seeds,
\ev{projection/heldout-gain-min-pp} to \ev{projection/heldout-gain-max-pp} percentage points above the models they
were compared with on such tasks; in the same proportion they
would gain \ev{projection/predicted-min-pp} to \ev{projection/predicted-max-pp} points on the examination, whose
\ev{const/exams} tasks detect \ev{resolution/detectable-min-pp} points or more. No later examination gain was established,
and the data cannot tell whether transfer stopped; later acquisition and retention are undecided because outcomes
were recorded per attempt and not per task.
\end{abstract}

%% file: sections/01-introduction.tex
\section{Introduction}\label{sec:introduction}

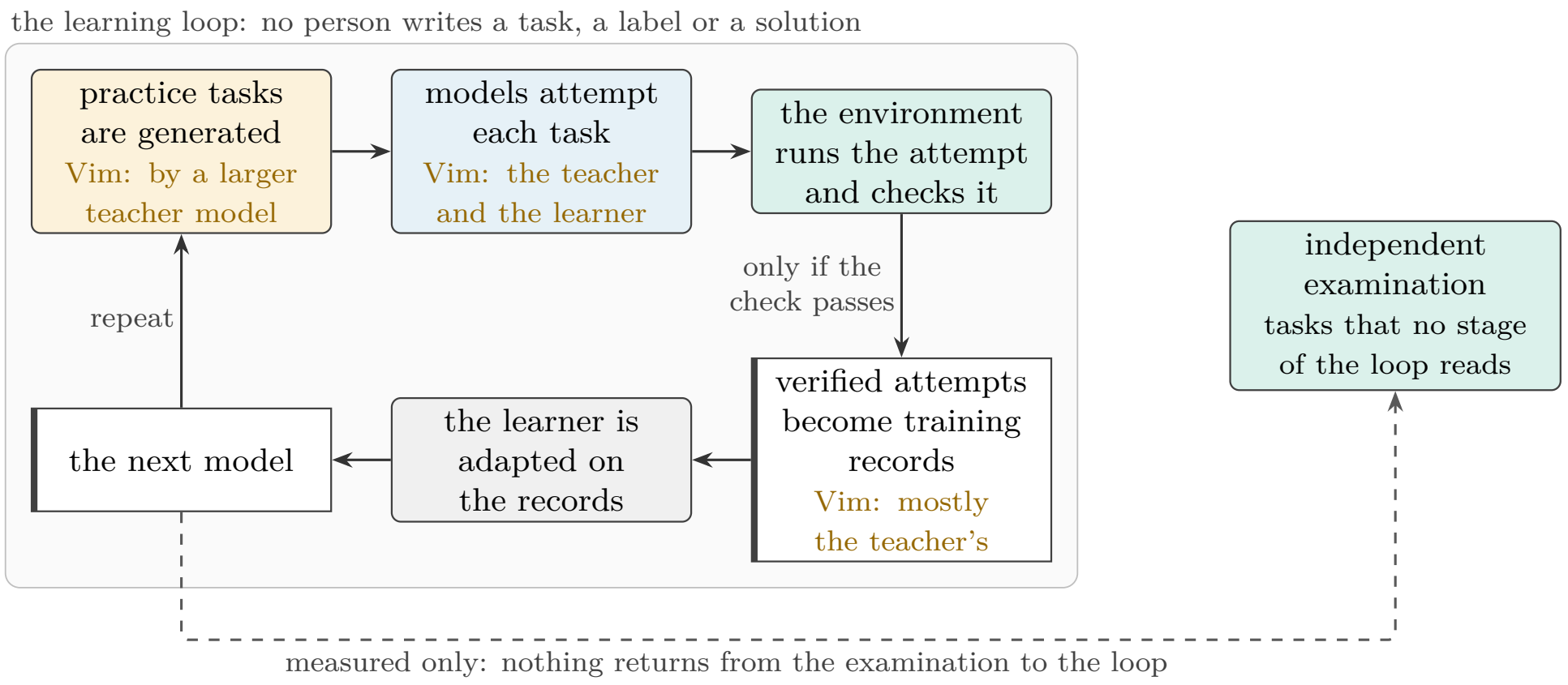
\begin{figure}[tbp]
\centering
\resizebox{0.8\linewidth}{!}{\input{figures/concept-body}}
\caption{Turning experience into competence. An executable environment decides which attempts were correct, so the verified
ones can be used as training material without a person writing a task or a solution. In one of our two environments
(Vim) a larger teacher model writes the tasks and most of the passing attempts (orange text); in the other (Lua) the
learner proposes its own. The examination stands outside the loop: it measures each model and returns nothing to
training, task generation or the choice of a model. (Conceptual diagram; no measured data.)}\label{fig:concept}
\end{figure}

A language model that has just completed a task is no better at the next similar task than it was before. What it did, what
the tools answered and whether the result was right are gone when the conversation ends. Models improve when someone prepares
a dataset and trains them on it, and preparing that dataset (writing tasks, labelling answers, checking solutions) is the
expensive part. Some environments remove that cost. A program can be run, and its output is then a fact. A file can be
edited by commands, and a check can compare the file that results with the file that was wanted. In such an
\emph{executable environment} the correctness of an attempt is decided by execution, so an attempt that passes is a worked
solution that nobody had to write. A system can then generate practice tasks, let a model attempt them, keep what the
environment verifies, train on it, and repeat (\cref{fig:concept}).

The question that decides whether such a loop is useful is whether the competence it produces extends beyond its own practice
material. Answering it needs measurements the loop cannot influence: tasks that no stage of the pipeline reads, graded
apart from it.

\paragraph{Two environments, and who supplies the solutions.} In the first environment the task is to predict what a
program prints, in Lua, a small programming language. The model proposes the programs itself and the interpreter supplies
every answer. In the second the task is to edit files with Vim, a text editor driven by typed commands. There the
environment can only say whether an edit is right, so a larger \emph{teacher} model writes the tasks from the editor's
manual and produces most of the solutions that pass, and a smaller \emph{learner} is trained on them. The first round of
the Vim loop is thus closer to distilling a larger model~\cite{hinton2015distilling} through an environment that checks
every example than to a model learning from its own experience alone. Both environments stand in for work whose result a
machine can check, such as repairing firmware until its test passes (\cref{sec:discussion}).

\paragraph{Continuous learning, and what it is not.} In the machine-learning literature, continual learning means acquiring
knowledge from experience that arrives incrementally while preserving what was learned
earlier~\cite{mccloskey1989catastrophic,kirkpatrick2017overcoming,rolnick2019experience}. It does not require that every
round improves every independent benchmark. We use \emph{continuous learning} in that sense, for a process with three
measurable parts: \emph{acquisition} (the model improves on new tasks of the kind it practised), \emph{transfer} (it
improves on an independent examination) and \emph{retention} (it keeps what earlier rounds acquired). A stronger property,
which we call \emph{generalized continuous self-improvement}, would have every successive model better on independent
tasks, with nothing lost, at falling cost. It is our long-term objective, and we do not claim it.

\paragraph{Scope and findings.} We demonstrate a functioning approach to continuous learning from executable environments,
in which language models acquire new training experience through automatically generated tasks, executable verification and
successive model adaptation, and we measure acquisition, transfer and retention separately. In short: learning from
verified practice carries over to an examination outside the loop, over the first two rounds in Lua and the first in
Vim; whether further rounds keep working we could not measure, and we show why. The process ran end to end, with every
release decided by a fixed rule. For later
generations two things limited what can be said: the runs recorded which attempts passed but not which task each
belonged to, so their planned tests over tasks cannot be computed; and the examination gain that their smaller observed
differences on generated tasks would predict is several times below what our examination can detect.

\paragraph{Research questions.} The paper is organised around five questions. The primary experiments behind them were
\emph{registered}: test and exclusions were written down and committed before the data existed, as a test is
committed before the code it checks (\cref{app:registrations}). The questions are our arrangement of those experiments,
and analyses made after the results are labelled post hoc or exploratory.
\textbf{RQ1}: can an executable environment alone supply enough supervision for a model to improve
(\cref{sec:r-pool})? \textbf{RQ2}: can automatically authored practice tasks teach skills that transfer to independent
tasks (\cref{sec:r-authoring})? \textbf{RQ3}: which sources and selections of training material produce that transfer
(\cref{sec:r-supervision})? \textbf{RQ4}: does repeating the process produce sustained improvement (\cref{sec:r-loop})?
\textbf{RQ5}: are earlier capabilities retained, and does learning become more efficient (\cref{sec:r-retention})?

\paragraph{Contributions.} (1) A working, self-contained learning framework, run end to end by three components (Splinter,
the learning controller; Sven, the agent runtime; Brain, the model server and trainer), in which executable
verification is the only source of supervision and the independent examination stays outside the loop that trains and
releases models (\cref{sec:method,app:artifacts}). (2)
Outcome-first generation of practice tasks for a tool-using agent, the request written only after a verified solution
exists (\cref{sec:m-outcome}). (3) An empirical study of transfer and of the training material that produces it, with a
shuffled-request control and registered tests (\cref{sec:r-authoring,sec:r-supervision}). (4) An analysis of iterated
learning that measures acquisition and transfer on the same candidates and quantifies how large an independent evaluation
must be to decide whether the second follows the first (\cref{sec:r-loop,sec:d-divergence}). Every empirical number in the
paper is generated from the evidence files by one pipeline, and the status of every claim is computed from that evidence
(\cref{app:claims}).

%% file: figures/concept-body.tex
\begin{tikzpicture}[fig, every node/.append style={execute at begin node={\hyphenpenalty=10000\exhyphenpenalty=10000}}]
  \def\dx{35mm}
  \def\dy{-30mm}
  \newcommand{\teacher}[1]{\\{\scriptsize\color{okOrange!65!black}#1}}
  \node[proc=okOrange!14, text width=27mm] (gen) at (0,0) {practice tasks are generated\teacher{Vim: by a larger teacher model}};
  \node[proc=okBlue!10, text width=27mm] (att) at (\dx,0) {models attempt each task\teacher{Vim: the teacher and the learner}};
  \node[proc=okGreen!14, text width=27mm] (ver) at (2*\dx,0) {the environment runs the attempt and checks it};
  \node[data, text width=27mm] (rec) at (2*\dx,\dy) {verified attempts become training records\teacher{Vim: mostly the teacher's}};
  \node[proc=black!6, text width=27mm] (train) at (\dx,\dy) {the learner is adapted on the records};
  \node[data, text width=27mm] (model) at (0,\dy) {the next model};
  \begin{scope}[on background layer]
    \node[lane=black!2, fit=(gen)(att)(ver)(rec)(train)(model), inner sep=7pt] (loop) {};
  \end{scope}
  \node[note, anchor=south west] at (loop.north west) {the learning loop: no person writes a task, a label or a solution};
  \draw[flow] (gen) -- (att); \draw[flow] (att) -- (ver); \draw[flow] (ver) -- node[note, left] {only if the\\check passes} (rec);
  \draw[flow] (rec) -- (train); \draw[flow] (train) -- (model); \draw[flow] (model) -- node[note, left] {repeat} (gen);
  \node[proc=okGreen!14, text width=30mm] (exam) at (3*\dx+13mm,-15mm) {independent examination\\{\scriptsize tasks that no
    stage of the loop reads}};
  \coordinate (below) at ($(loop.south)+(0,-5mm)$);
  \draw[measure] (model.south) |- (below) -| (exam.south);
  \node[note, anchor=north] at ($(below)+(18mm,-0.5mm)$) {measured only: nothing returns from the examination to the loop};
\end{tikzpicture}

%% file: sections/02-related.tex
\section{Related work}\label{sec:related}

Our contribution is a working, self-contained framework for learning from executable environments, together with the
measurements needed to judge it. The entire learning process (authoring practice tasks from a source document, admitting
them by execution, attempting them with an agent, selecting the verified records, training a candidate and deciding its
release) is run end to end by three components: Splinter, the learning controller; Sven, which executes agents and their
tools; and Brain, which serves the models and trains them. In the Vim experiments the process is declared as versioned
recipes (lists of stages with their inputs) that Splinter runs; the Lua cycle is a program of Splinter's repository.
What is specific to an environment (its tool and check, task admission, the authoring operators and prompts) is
supplied to the three as materials (\cref{tab:mapping}). Only the examination stands outside the
learning process, by design, so that nothing in it can see the examination's tasks. Because the process is recipes,
prompts and small operator programs that three components execute, an AI agent can read, change and extend it like any
other code. This is the property we intend
to build on towards generalized self-improvement in future work; it is not tested here. The system is incomplete,
and the following sections show where it stops, but it already carries a model from source material to examined releases
across successive generations. Its parts are not new in isolation: the training step is supervised fine-tuning on samples
that a verifier accepts, and our task authoring is a variant of backward construction~\cite{su2025learnbyinteract}. What
the paper adds is the system that joins them and the measurement around it.

\paragraph{Self-training on generated supervision.} Expert iteration alternates a slow expert with a learner that imitates
it~\cite{anthony2017thinking}. For language models, STaR fine-tunes on self-generated rationales that reach the known
answer~\cite{zelikman2022star}, ReST$^{EM}$ alternates sampling with fine-tuning on samples that a binary reward
accepts~\cite{singh2023rest}, rejection-sampling fine-tuning does the same in one round~\cite{yuan2023scaling}, and
Haluptzok et al.\ let the model write the problems as well, keeping those a Python interpreter
verifies~\cite{haluptzok2022language}. For agents, Re-ReST trains on the agent's own successful
trajectories~\cite{dou2024rerest} and SWE-Gym on trajectories that pass repository tests~\cite{pan2025swegym}. These works
evaluate mainly on the benchmark whose training split supplies the questions. Singh et al.\ also measure transfer to
held-out benchmarks and report mixed results over iterations: on a programming benchmark most of the gain came from the
first iteration and a second one regressed, there and on a held-out benchmark, while on mathematics further iterations
kept helping~\cite{singh2023rest}. We measure
gains on held-out tasks of each generation's own generator and on an outside examination for the same candidates, and
ask how large an evaluation must be to tell whether the second follows the first.

\paragraph{Writing the task after the solution.} Learn-by-interact synthesises trajectories from documentation and
summarises each into an instruction~\cite{su2025learnbyinteract}; BAGEL and
NNetNav label explored trajectories in hindsight~\cite{murty2024bagel,murty2024nnetnav}; OS-Genesis derives tasks from
observed interface changes~\cite{sun2024osgenesis}; SWE-smith injects a fault into working code and writes the issue
afterwards~\cite{yang2025swesmith}. Our outcome-first authoring (\cref{sec:m-outcome}) belongs to this family and differs
in fixing the end state before any interaction, requiring the solution to re-run, and filtering requests that give the
solution away.

\paragraph{Curricula, self-improvement and execution as supervision.} Open-ended and curriculum methods generate or order
tasks for a learner~\cite{wang2019poet,wang2023voyager,zhao2025absolutezero,bengio2009curriculum,portelas2020automatic};
self-modifying systems and meta-learning aim at systems that improve themselves or their own
learning~\cite{schmidhuber2003godel,zhang2025dgm,cai2026nosecondrun,finn2017maml}; execution has long served as a
target~\cite{zaremba2014learning,gu2024cruxeval}, as a filter~\cite{wang2018robust} and as a verifiable reward for
policy-gradient training~\cite{lambert2024tulu3,guo2025deepseekr1,yu2025dapo}. This paper claims no open-ended curriculum, no self-modification and no learning to learn. These are the next steps of
our work, to be built on the blocks shown here: verified task authoring, training on verified experience, an
independent examination, and a learning process that three components run.

\paragraph{Continual learning and contamination.} Sequential training can erase earlier
competence~\cite{mccloskey1989catastrophic}; remedies include penalising changes to important
weights~\cite{kirkpatrick2017overcoming} and replaying stored experience~\cite{rolnick2019experience}, and models trained
recursively on generated data can lose the tails of the original distribution~\cite{shumailov2024collapse}. Our Vim
protocol continues one adapter, replays earlier verified records and applies a retention rule within one environment,
which is narrower than the task-incremental benchmarks of that literature. The Lua cycle retrains a fresh adapter from
the base on the whole pool at every generation: it accumulates data, not weights, and has nothing to forget. Accuracy can fall on a fresh sample of nominally the same
distribution~\cite{recht2019imagenet} and public test items can enter pre-training data~\cite{sainz2023nlp}, a risk that
privately generated items limit~\cite{white2025livebench}: our Lua examination is private, our Vim examination public.

%% file: sections/03-problem.tex
\section{Problem formulation}\label{sec:problem}

\paragraph{Environment, task, attempt, record.} An \emph{executable environment} $E$ is a program that runs what it is
given and returns a result: here a Lua~5.1 interpreter, and the editor Vim on a set of files. A \emph{task} $x$ is
something to do in the environment, with a way to check the outcome. An \emph{attempt} is one run of a model on one task;
its \emph{trajectory} $\tau$ is the model's actions and the environment's responses in order. The verifier
$V_E(x,\tau)\in\{0,1\}$ says whether the attempt solved the task; an attempt is \emph{verified} when $V_E(x,\tau)=1$,
and a \emph{record} is the training example made from a verified attempt.

\paragraph{Models and generations.} The \emph{learner} is the model being trained and a \emph{teacher} a larger model
whose verified attempts may serve as records; \cref{tab:terms} gives the other terms. A training-step replication shows
how much one training step varies and is not a lineage.

\begin{table}[tbp]
\centering\footnotesize
\begin{tabular}{@{}>{\raggedright\arraybackslash}p{37mm}>{\raggedright\arraybackslash}p{118mm}@{}}
\toprule
Term & Meaning \\
\midrule
Base; adapter & the published model before any training of ours; a small set of additional weights trained while the
  base's own weights stay fixed. Ours are \emph{low-rank} adapters~\cite{hu2022lora}: each added weight matrix is the
  product of two thin ones, which makes an adapter cheap to train and store. A base with an adapter is what is evaluated \\
Generation; candidate $\pi_g$ & one pass of the loop of \cref{fig:concept} (generate tasks, attempt, verify, train,
  evaluate); the model that generation $g$ produces \\
Release rule; champion $\pi_{c(g)}$ & the fixed test that accepts or rejects a candidate; the most recently accepted
  model when generation $g$ is evaluated: the base at first, and not necessarily generation $g-1$ afterwards, because a
  candidate can be rejected \\
Training seed & the number that fixes the random choices of one training run (such as the order of the records); two
  seeds give two slightly different adapters from the same data \\
Lineage; training-step replication & one independent run of the whole sequence of generations from the base; one
  generation's training repeated with another training seed, from the same champion, on the same tasks and records \\
\bottomrule
\end{tabular}
\caption{Terms for models and for the repetition of the learning loop, as used throughout the paper.}\label{tab:terms}
\end{table}

\paragraph{Two kinds of unseen task.} The paper turns on the difference between two sets of tasks that a model did not
train on. Generation $g$ draws its tasks from a practice distribution $D_g$ produced by the system's own task generator. A
fixed share of them, $H_g$, is \emph{held out}: no record made from them enters training. These \emph{held-out generated tasks} are new to the model but come from the same generator as its training
material. The \emph{independent examination} $X$ is a set
of tasks that the generator did not produce and that no stage of the pipeline reads: not task generation, not training,
not the choice of a checkpoint, not the release rule. Independent means outside our pipeline, not that a base model never
saw the tasks before we received it, nor that a third party operates the examination (\cref{sec:m-examiner}).

\paragraph{What is measured.} The competence of a model $\pi$ on a set of tasks $D$ is the share of its attempts that the
environment verifies, an estimate of $A(\pi, D) = \mathbb{E}_{x\sim D,\;\tau\sim\pi(\cdot\mid x)}[V_E(x,\tau)]$: the average
verdict over tasks drawn from $D$ and attempts drawn from the model. For example, the second 8B generation passed \ev{B/8B/g2/target/cand-passes} of \ev{B/8B/g2/target/cand-counted}
attempts on its own held-out tasks and its champion \ev{B/8B/g2/target/base-passes}:
\ev{B/8B/g2/target/cand-pct}\% against \ev{B/8B/g2/target/base-pct}\%, a difference of
$\ev{B/8B/g2/target/diff-pp}$ percentage points. Three such differences are formed, each between a candidate and its
reference model measured in the same evaluation stage:
\begin{align*}
  \text{acquisition}\quad & \Delta^{\mathrm{acq}}_g = A(\pi_g, H_g) - A(\pi_{c(g)}, H_g), \\
  \text{transfer}\quad    & \Delta^{\mathrm{tr}}_g = A(\pi_g, X) - A(\pi_{c(g)}, X), \\
  \text{retention}\quad   & R_{g,j} = A(\pi_g, H_j) - A(\pi_{c(g)}, H_j) \;\ge\; -\varepsilon \quad\text{for } j<g .
\end{align*}
The retention margin $\varepsilon$ is \ev{const/retention-margin-pp} percentage points. The example above is an
observed $\Delta^{\mathrm{acq}}_2$. A positive $\Delta^{\mathrm{acq}}_g$ does not imply a positive
$\Delta^{\mathrm{tr}}_g$; how the two relate is the empirical question of the Vim experiments. Learning efficiency would need a fourth quantity, transfer per unit of compute, which these runs recorded only in
part (\cref{sec:r-efficiency}).

\paragraph{Autonomous.} A run is \emph{autonomous} when no person writes a training task, a label or a solution, selects
a checkpoint, or decides a release. People wrote the environments, the examination and the recipes, and ran the hardware.

%% file: sections/04-framework.tex
\section{Learning framework}\label{sec:method}

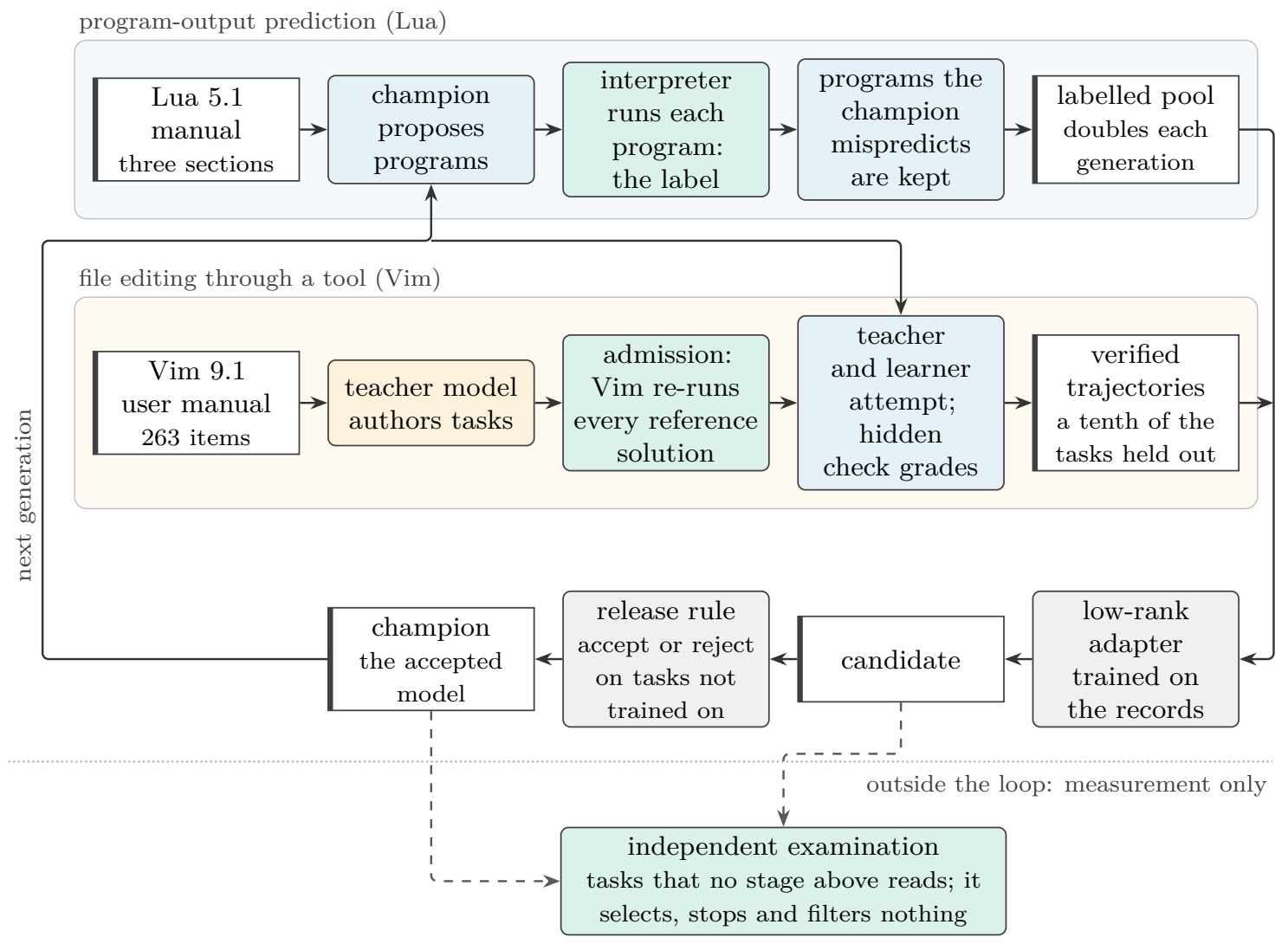
\begin{figure}[tbp]
\centering
\resizebox{0.61\linewidth}{!}{\input{figures/method-body}}
\caption{The framework as one loop over both environments. Each ends in the same steps: an adapter trained on records the
environment verified, a release rule that accepts or rejects the candidate on tasks it did not train on (in Vim the
generation's held-out tasks, in Lua development programs the learner generates for itself), and the accepted model as the
champion that proposes or attempts the next generation's tasks. The independent examination is
outside the loop (dashed): it measures candidates and champions and selects nothing. Square boxes with a heavy left edge
are stored artefacts; rounded boxes are processes, green where the environment decides.}\label{fig:method}
\end{figure}

The framework is one pattern, a loop that the two environments instantiate differently. We introduce it in the order in which a single piece of
experience moves through it (\cref{fig:example}), then as a whole (\cref{fig:method}); \cref{app:artifacts} maps its
roles to the software and pins the revisions.

\subsection{One task, one attempt, one record}\label{sec:m-teacher}

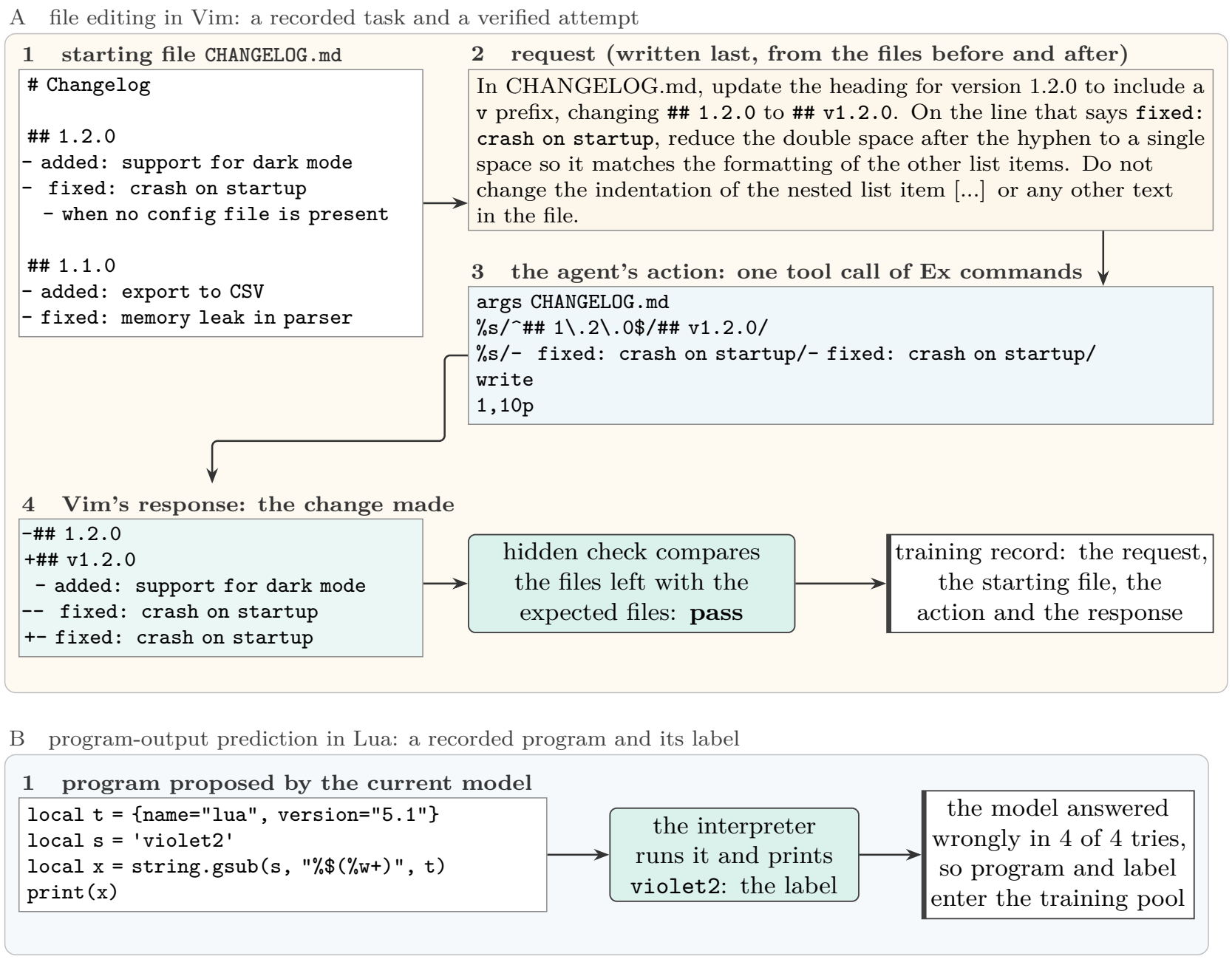
\begin{figure}[tbp]
\centering
\resizebox{0.69\linewidth}{!}{\input{figures/example-body}}
\caption{One real learning experience in each environment, from the recorded data. A: an automatically authored Vim task
(\texttt{backward-46d5246e}), the teacher model's verified attempt and what became the training record; the request is
abridged at [...], and in box~4 Vim's tool reports each changed line as removed ($-$) and added ($+$). B: a Lua program
proposed in the first generation of one lineage (\texttt{prop-f71e30f7abb31903}), the interpreter's output, and the
recorded predictions of the base model, which is the current model in a first generation. Green marks what the environment decided.}\label{fig:example}
\end{figure}

\Cref{fig:example} shows a real learning experience in each environment, copied from the recorded data. In the Vim
environment the model is an agent with a single tool: a call runs a list of Ex commands (the commands one types after a
colon in Vim) in a fresh Vim on the task's files and returns what Vim printed and what changed. A hidden check, which the
agent cannot read, compares the files left behind with the expected files. The request, the starting file, the action and
the response of a passing attempt are one training record; nothing in the one shown was written by a person. In the Lua
environment the program shown was proposed by the model itself; a pinned Lua~5.1 interpreter ran it, and because the
model had predicted its output wrongly, the program and the interpreter's output became a record.

The environments differ in one respect that matters later: the Lua interpreter supplies the correct answer whether or
not the learner found it, while Vim supplies only a verdict, so a verified trajectory must first come from the learner
itself or from a teacher.

\subsection{Generating practice}\label{sec:m-pool}

\paragraph{A pool of programs the model mispredicts.} In each Lua generation the current champion proposes Lua~5.1 programs from three sections of the language manual, and
the interpreter runs them and records their output. The champion predicts each program's output four times, sampling
freely (temperature~1.0), and a program is kept when at least one prediction is wrong. A fresh adapter is then trained
from the base on every program kept so far, each record pairing the program with the interpreter's output. The
registration fixes the size of this pool so that it doubles at every generation (\ev{compute/lua/g1/trained-records}
records over the four lineages in the first, \ev{compute/lua/g4/trained-records} in the fourth). A candidate replaces the champion only
if it passes the learner's own release rule, a sign test on development programs the learner generates for itself and
does not train on. The examination is not consulted. These programs are not a held-out share of each generation's pool,
so the Lua environment has no $H_g$ and its acquisition is not measured separately from its examination. The cycle was introduced in a companion
paper~\cite{frontier2026}, which reports earlier and shorter runs of it; the runs and the examination here are new.

\paragraph{Tasks authored from a manual.}\label{sec:m-authoring} The Vim experiments start from the text of the user manual of Vim~9.1, split into
\ev{funnel/forward/stage/sections/items} items of at most about 1500 words. The teacher model also acts as \emph{task
author}. In the first authoring mode, \emph{section by section}, the author proposes for each item tasks that need one to
six Ex commands, each with a request, starting files and a reference solution. Vim runs every reference; the author repairs,
once, a reference that Vim refuses, and in a separate reviewing pass checks each result against its request. Malformed
tasks, references that change nothing or raise an error, duplicates, and requests that quote the reference's commands are
rejected. \label{sec:m-admission}A task is then \emph{admitted} only if its reference solution, run again inside the
task's own environment, passes the task's own check, with files compared as exact bytes: a task that cannot be solved by
its own reference is never attempted, so it cannot produce a record.

\paragraph{Outcome-first authoring.}\label{sec:m-outcome} A request written before anyone knows what a correct result looks like can be ambiguous, and section-by-section tasks
tend to be as small as the section they came from.
\emph{Outcome-first authoring} reverses the order: specify the outcome, verify a solution, then write the request. (1) For
each item the author proposes the starting files and the desired end files of a job of several changes; there is no
request yet. (2) The teacher must reach each end state with the Vim tool, within four attempts. (3) The teacher's
commands must reproduce the end state when Vim runs them again, and must edit the files rather than retype them. (4) Only
then is the request written, from the files before and after, and the teacher's verified solution becomes the reference;
a request that quotes most of the target or of the solution is refused. The task of \cref{fig:example} was made this way; \cref{sec:related} relates it to the backward construction of Su et
al.~\cite{su2025learnbyinteract}.

\subsection{Training on verified attempts, and repeating}\label{sec:m-adapters}

Admitted tasks are attempted by the models whose records are wanted (the base learner eight times, the teacher four;
budgets in \cref{app:hyper}), and a tenth of the tasks is held out from every training set. An adapter is trained on the chosen records by supervised fine-tuning: the model is trained to produce the
recorded actions given the request and the environment's responses so far. There is no reward signal in the objective,
only a filter on which trajectories are imitated. The checkpoint (a saved state of the adapter) with the lowest loss on
validation records set aside from the training set is kept; no examination result selects a checkpoint, a recipe or a
mix of data, and every training seed of an experiment is reported (hyperparameters in \cref{app:hyper}).

\paragraph{Repeating the process.}\label{sec:m-loop} A generation of the Vim loop authors new outcome-first tasks, lets the champion and the teacher attempt them, trains a
candidate on verified records, evaluates candidate and champion side by side, and applies the release rule; an accepted
candidate becomes the champion. The \emph{iterative protocol} studied in \cref{sec:r-loop} fixes three choices.

\emph{Authoring rule.} Each generation's rule, written in advance, asks for larger jobs than the one before: three
to eight interacting changes built on the manual section at hand; then four to ten, combining ideas from at least two
other chapters; then five to twelve, from at least three other chapters and spanning two or more files. Both model
sizes used the same authored tasks. The evidence holds this protocol's evaluation counts but not how many tasks were
authored or records trained on per generation (\cref{app:compute}).

\emph{Training.} The first generation trains a fresh adapter from the base on the verified attempts of the teacher
and of the base learner. Later generations continue the champion's adapter at half the learning rate on the teacher's
verified attempts on the new tasks, and earlier experience is \emph{replayed}: half of the records drawn during
training come from the earlier generations' training sets and from the champion's own verified attempts on the new
tasks (\cref{app:hyper}).

\emph{Release rule.} In plain terms, a candidate is accepted if it clearly beats the champion on the new generation's
held-out tasks and is not noticeably worse on those of earlier generations. Precisely: a paired one-sided sign test at
$\alpha=\ev{const/alpha}$ with a positive difference on its own held-out tasks; no significant loss and a pass-rate
difference not below $-\ev{const/retention-margin-pp}$ percentage points on the held-out tasks of every earlier
generation; and at most
\ev{const/uncounted-bound-pct}\% of the evaluation's attempts ended without a verdict. As run, the rule's sign test
paired single attempts, not tasks (\cref{sec:d-statistics}): we report its decisions as decisions of the rule, not as
evidence of improvement. Within a protocol the examination selects, stops, filters and authors nothing; earlier
iterative runs had been examined on the same tasks before this protocol was written (\cref{app:earlier}).

\subsection{The independent examination}\label{sec:m-examiner}

The learner never controls grading. In the Lua experiments an \emph{examiner}, a separate process that holds its own seed
and secret key, draws the examination from its own program generator, excludes any item that matches a program a
candidate trained on, grades each answer against the interpreter's output, and returns aggregate verdicts that carry an
authentication code. The examiner is operated by the author of this paper,
not by a third party. Its items are of the same kind as the practice; they are drawn from \emph{program families}, each
one template of the examiner's generator. In the Vim experiments the examination is the \ev{const/exams} tasks that form
the evaluation side of a public Vim course written by hand at the author's organisation: each combines several ideas, none was authored from the manual, and
no stage that produces training data, chooses a checkpoint or releases a model reads them. The course also has
\ev{practice/g-s3/all/tasks} hand-written \emph{course exercises}, \ev{practice/g-s3/exercises/tasks} on a single idea
and \ev{practice/g-s3/composites/tasks} composite. They are not the practice tasks the system generates: no model in this paper was trained on them, and they serve as a
second, easier independent measurement.

%% file: figures/example-body.tex
\begin{tikzpicture}[fig, x=1mm, y=1mm,
    every node/.append style={execute at begin node={\hyphenpenalty=10000\exhyphenpenalty=10000}}]
  \newcommand{\q}{\textquotesingle}
  \newcommand{\bs}{\textbackslash}
  \node[listing, anchor=north west, text width=50mm] (start) at (0,0) {%
    \# Changelog\\\mbox{}\\\#\# 1.2.0\\- added: support for dark mode\\-\ \ fixed: crash on startup\\
    \ \ - when no config file is present\\\mbox{}\\\#\# 1.1.0\\- added: export to CSV\\- fixed: memory leak in parser};
  \node[caption] at (start.north west) {1\quad starting file \texttt{CHANGELOG.md}};
  \node[listing=okOrange!8, anchor=north west, text width=94mm, font=\scriptsize] (req) at (58,0) {%
    In CHANGELOG.md, update the heading for version 1.2.0 to include a \texttt{v} prefix, changing \texttt{\#\# 1.2.0} to
    \texttt{\#\# v1.2.0}. On the line that says \texttt{fixed: crash on startup}, reduce the double space after the hyphen
    to a single space so it matches the formatting of the other list items. Do not change the indentation of the nested
    list item [...] or any other text in the file.};
  \node[caption] at (req.north west) {2\quad request (written last, from the files before and after)};
  \node[listing=okBlue!6, anchor=north west, text width=94mm] (act) at (58,-28) {%
    args CHANGELOG.md\\\%s/\textasciicircum\#\# 1\bs.2\bs.0\$/\#\# v1.2.0/\\
    \%s/-\ \ fixed: crash on startup/- fixed: crash on startup/\\write\\1,10p};
  \node[caption] at (act.north west) {3\quad the agent's action: one tool call of Ex commands};
  \node[listing=okGreen!8, anchor=north west, text width=50mm] (resp) at (0,-58) {%
    -\#\# 1.2.0\\+\#\# v1.2.0\\\ - added: support for dark mode\\--\ \ fixed: crash on startup\\+- fixed: crash on startup};
  \node[caption] at (resp.north west) {4\quad Vim's response: the change made};
  \node[proc=okGreen!14, anchor=north west, text width=40mm] (check) at (58,-60) {hidden check compares the files left
    with the expected files: \textbf{pass}};
  \node[data, anchor=north west, text width=40mm] (record) at (112,-60) {training record: the request, the starting file, the action and the response};
  \draw[flow] (start.east) -- (start.east -| req.west);
  \draw[flow] (req.south -| 140,0) -- (act.north -| 140,0);
  \draw[flow] (act.west) -- ++(-3,0) |- (25,-48) -- (25,-53.5);
  \draw[flow] (resp.east |- check.west) -- (check.west);
  \draw[flow] (check) -- (record);
  \begin{scope}[on background layer]
    \node[lane=okOrange!5, fit=(start)(req)(act)(resp)(check)(record), inner xsep=5pt, inner ysep=13pt] (vim) {};
  \end{scope}
  \node[note, anchor=south west] at (vim.north west) {A\quad file editing in Vim: a recorded task and a verified attempt};
  \node[listing, anchor=north west, text width=66mm] (prog) at (0,-94) {%
    local t = \{name=\textquotedbl lua\textquotedbl, version=\textquotedbl 5.1\textquotedbl\}\\
    local s = \q violet2\q\\
    local x = string.gsub(s, \textquotedbl\%\$(\%w+)\textquotedbl, t)\\print(x)};
  \node[caption] at (prog.north west) {1\quad program proposed by the current model};
  \node[proc=okGreen!14, anchor=west, text width=30mm] (interp) at ($(prog.east)+(8,0)$) {the interpreter runs it and prints
    \texttt{violet2}: the label};
  \node[data, anchor=west, text width=33mm] (lrec) at ($(interp.east)+(8,0)$) {the model answered wrongly in 4 of 4 tries, so
    program and label enter the training pool};
  \draw[flow] (prog) -- (interp);
  \draw[flow] (interp) -- (lrec);
  \begin{scope}[on background layer]
    \node[lane=okBlue!4, fit=(prog)(interp)(lrec), inner xsep=5pt, inner ysep=13pt] (lua) {};
  \end{scope}
  \node[note, anchor=south west] at (lua.north west) {B\quad program-output prediction in Lua: a recorded program and its label};
\end{tikzpicture}

%% file: figures/method-body.tex
\begin{tikzpicture}[fig, every node/.append style={execute at begin node={\hyphenpenalty=10000\exhyphenpenalty=10000}}]
  \def\dx{27.5mm}
  \node[data] (lsrc) at (0,0) {Lua 5.1 manual\\{\scriptsize three sections}};
  \node[proc=okBlue!10] (lprop) at (\dx,0) {champion proposes programs};
  \node[proc=okGreen!14] (llab) at (2*\dx,0) {interpreter runs each program: the label};
  \node[proc=okBlue!10] (lfr) at (3*\dx,0) {programs the champion mispredicts are kept};
  \node[data] (lpool) at (4*\dx,0) {labelled pool\\{\scriptsize doubles each generation}};
  \node[data] (vsrc) at (0,-32mm) {Vim 9.1 user manual\\{\scriptsize \ev{funnel/forward/stage/sections/items} items}};
  \node[proc=okOrange!14] (vaut) at (\dx,-32mm) {teacher model authors tasks};
  \node[proc=okGreen!14] (vora) at (2*\dx,-32mm) {admission: Vim re-runs every reference solution};
  \node[proc=okBlue!10] (vatt) at (3*\dx,-32mm) {teacher and learner attempt; hidden check grades};
  \node[data] (vtraj) at (4*\dx,-32mm) {verified trajectories\\{\scriptsize a tenth of the tasks held out}};
  \node[proc=black!6] (train) at (4*\dx,-62mm) {low-rank adapter trained on the records};
  \node[data] (cand) at (3*\dx,-62mm) {candidate};
  \node[proc=black!6] (gate) at (2*\dx,-62mm) {release rule\\{\scriptsize accept or reject on tasks not trained on}};
  \node[data] (champ) at (\dx,-62mm) {champion\\{\scriptsize the accepted model}};
  \begin{scope}[on background layer]
    \node[lane=okBlue!4, fit=(lsrc)(lprop)(llab)(lfr)(lpool), inner sep=6pt] (lane1) {};
    \node[lane=okOrange!5, fit=(vsrc)(vaut)(vora)(vatt)(vtraj), inner sep=6pt] (lane2) {};
  \end{scope}
  \node[note, anchor=south west] at (lane1.north west) {program-output prediction (Lua)};
  \node[note, anchor=south west] at (lane2.north west) {file editing through a tool (Vim)};
  \draw[flow] (lsrc) -- (lprop); \draw[flow] (lprop) -- (llab); \draw[flow] (llab) -- (lfr); \draw[flow] (lfr) -- (lpool);
  \draw[flow] (vsrc) -- (vaut); \draw[flow] (vaut) -- (vora); \draw[flow] (vora) -- (vatt); \draw[flow] (vatt) -- (vtraj);
  \draw[flow] (lpool.east) -- ++(4mm,0) |- (train.east);
  \draw[flow] (vtraj.east) -- ++(4mm,0);
  \draw[flow] (train) -- (cand);
  \draw[flow] (cand) -- (gate);
  \draw[flow] (gate) -- (champ);
  \coordinate (gapy) at ($(lane1.south)+(0,-2.5mm)$);
  \draw[flow] (champ.west) -- (-18mm,-62mm) |- (gapy -| lprop) -- (lprop.south);
  \draw[flow] (gapy -| lprop) -| (vatt.north);
  \node[note, anchor=east, rotate=90] at (-20mm,-32mm) {next generation};
  \node[proc=okGreen!14, text width=50mm] (exam) at (2.5*\dx,-88mm) {independent examination\\{\scriptsize tasks that no
    stage above reads; it selects, stops and filters nothing}};
  \draw[measure] (cand.south) -- ++(0,-6mm) -| (exam.north);
  \draw[measure] (champ.south) |- (exam.west);
  \draw[black!35, densely dotted, line width=0.6pt] (-22mm,-74mm) -- (4*\dx+16mm,-74mm);
  \node[note, anchor=north east] at (4*\dx+16mm,-75mm) {outside the loop: measurement only};
\end{tikzpicture}

%% file: sections/05-design.tex
\section{Experimental design}\label{sec:design}

\label{sec:d-setup}All models are open-weight models run locally; the number in a model's name is its size in billions of parameters, so
4B and 8B are the two learners. The
learner is Qwen3-4B in both environments and also Qwen3-8B in Vim; the Vim teacher and task author is Qwen3.8-27B-FP8
(the suffix names its 8-bit weight format), about seven times larger than the smaller learner; Lua has no teacher. The learners run with their reasoning mode switched
off; the teacher ran in its default reasoning mode, and its reasoning was removed from the records. The Lua examination
has \ev{lua/gen4-over-base/items} items in \ev{lua/gen4-over-base/families} program families; the Vim examination has
\ev{const/exams} tasks, each attempted \ev{const/attempts-per-task} times.

\label{sec:d-experiments}Six experiments answer the five research questions (\cref{app:registrations,app:deviations}):
the Lua pool, four lineages of four generations each (RQ1); task authoring in two
modes on the same manual (RQ2); the training-material conditions below (RQ2, RQ3); the iterative protocol of
\cref{sec:m-loop}, three generations with each learner on the same authored tasks (RQ4, RQ5); training-step replications
of each later step of that protocol, everything unchanged but the seed (RQ4); and first-generation candidates on the
hand-written course exercises (RQ2, RQ5).

\paragraph{Training-material conditions.}\label{sec:d-arms} All conditions start from the same base Qwen3-4B, use the same
training recipe and three training seeds, measure one generation on the examination and on held-out tasks against the
base, and differ only in the records they train on. \emph{All verified}: every verified attempt of teacher and learner on
the section-by-section tasks. \emph{One per task}: at most one attempt per task, the one with the fewest actions.
\emph{Productive actions}: all verified, but an action whose tool calls all returned an error is kept as context and not
trained on. \emph{Outcome first}: the all-verified rule applied to the outcome-first tasks. \emph{Shuffled requests}, the
control: all verified, with each record's request replaced by another task's. The records, the actions and the tool
format are unchanged; only the correspondence between what is asked and what is done is destroyed, so if a model trained
this way passed examination tasks, the gain of the other conditions could be exposure to the tool's format. A sixth condition, \emph{learner only}, was added after the five had been decided: the learner's own
verified attempts on the section-by-section tasks, with no teacher record.

\subsection*{Statistics}\label{sec:d-statistics}

\paragraph{Attempts are not tasks.} A model that passes eleven of four hundred examination attempts may have solved eleven
tasks once each or three tasks several times. The first is far stronger evidence of a general gain, and a count of
passing attempts cannot tell them apart. The unit of every registered test is therefore the task (in Lua, the program
family), never the attempt.

\paragraph{Tests.} In Lua, the examiner computes each difference with a 95\% bootstrap interval over program families and
lineages (with four lineages, the lineage part of the resampling is coarse); a comparison holds when the interval
excludes zero. In the training-material experiment the
registered test is a one-sided exact sign test over examination tasks: count the tasks on which a condition passes more
of its attempts than the reference and those on which it passes fewer, and ask how probable so uneven a split is by
chance. Passes are pooled over three seeds, and Holm's correction keeps the chance of any false positive among the four
non-control comparisons at $\alpha=\ev{const/alpha}$; the claim that trajectories teach transferable skills also requires
the same condition to beat the shuffled-request control. Differences in pass rate carry 95\% intervals from a bootstrap
over tasks (\cref{app:claims}).

\paragraph{The iterative protocol recorded attempts, not tasks.} Its registered test is the same sign test over tasks. The
runs did not record which task each attempt belonged to: their evaluation stages kept the counts of passes and a sign
test over paired single attempts. The planned test therefore cannot be run for those steps, and we do not use the
recorded $p$-values. From the counts we compute the smallest and largest value the task-level test could take over every
way the attempts can fall on the tasks, and report a comparison as supported only when even the largest is below
$\alpha$; otherwise it is inconclusive and the pass rates are observations (\cref{app:claims}). Retention, where outcomes
per task exist, is tested as \emph{non-inferiority}: the candidate is no worse than the reference by more than a margin
when the 95\% interval of the difference lies wholly above $-\ev{const/retention-margin-pp}$ percentage points, the
margin the release rule fixed in advance. The release rule itself compared the observed difference with the margin,
without an interval; passing it is reported as passing the rule, not as retention.

\paragraph{Exclusions.} Attempts that end without a verdict for a reason outside the task, such as a model server that
drops a request, are never counted as failures. An evaluation in which more than \ev{const/uncounted-bound-pct}\% of the
scheduled attempts ended this way is reported and excluded from claims. The bound was applied at two scopes, as each
experiment's registration fixed it, and the difference matters: at the stricter one, used in the iterative protocol,
the outcome-first and all-verified conditions would be excluded too, and the registered transfer claim would rest on the
other two conditions (\cref{app:claims}).

\paragraph{Status words.} Four words say how sure a result is, in the text and in every table. \emph{Supported}: a test
over the claim's unit of analysis (the task, or the program family) decides it; \cref{tab:claims} says whether that
test was registered in advance. \emph{Observed}: the measured difference has the stated direction and no test decides
it; it is not established. \emph{Excluded}: the evaluation lost more attempts than the bound allows and supports
nothing. \emph{Exercised}: a mechanism of the system ran in the reported runs and produced the counted products, which
says that it works as built, not that it has an effect. A comparison that is neither supported nor excluded is
\emph{inconclusive}. Every status is computed from the evidence (\cref{app:claims}). Intervals on per-attempt pass rates
in the figures are Wilson intervals, which treat attempts as independent; they are descriptive.

%% file: sections/06-results.tex
\section{Results}\label{sec:results}

\subsection{RQ1: Can an executable environment alone supply enough supervision?}\label{sec:r-pool}

\begin{figure}[tbp]
\centering
\input{figures/lua-generations-body}
\caption{Supervision from an interpreter alone raises examination accuracy, with diminishing successive gains. Left: the
share of examination items answered correctly by the base and by each generation, for each of four lineages of Qwen3-4B
(grey circles) and their mean; all measurements are on the same examination, so the means are joined.
Right: each registered comparison with its 95\% interval from the examiner's bootstrap over program families and
lineages (unadjusted for the six comparisons, as registered; values in \cref{tab:lua}); filled: the interval excludes zero; open: it does
not.}\label{fig:lua-generations}
\end{figure}
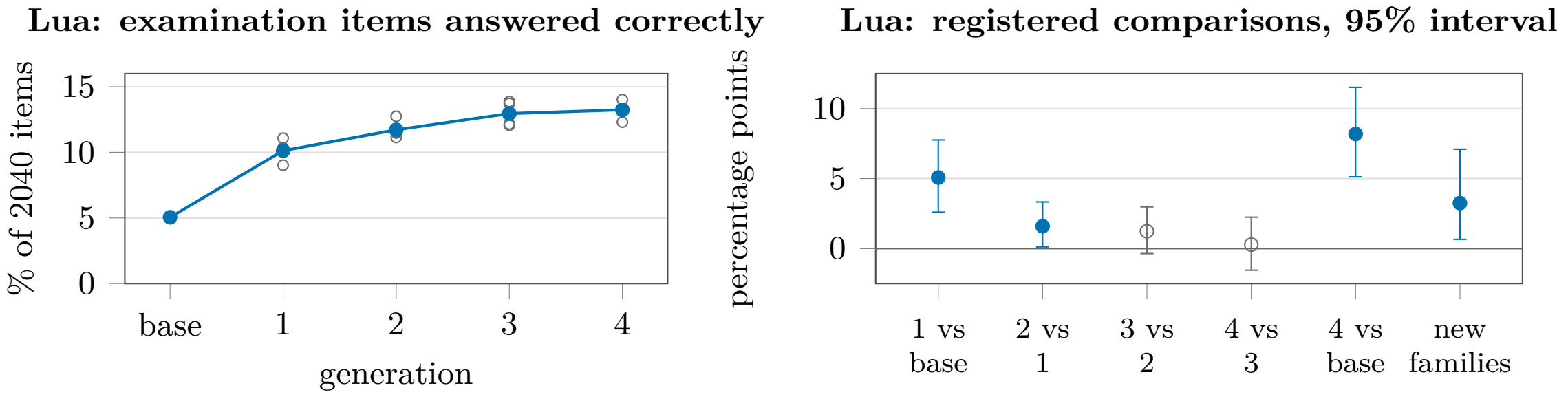

\paragraph{Test and result.} Four independent lineages of Qwen3-4B each ran four generations of the Lua cycle of
\cref{sec:m-pool}; no larger model and no person supplied anything. Every candidate was handed to the examiner;
\ev{lua/gen4-over-base/new-families/items} of its items come from \ev{lua/gen4-over-base/new-families/families} program
families that the examiner had used in no earlier examination, including those of the companion paper. The base model answers
\ev{lua/gen4-over-base/before-pct}\% of the examination correctly. After four generations the lineages answer
\ev{lua/gen4-over-base/after-pct}\% on average, a gain of $\ev{lua/gen4-over-base/diff-pp}$ percentage points (95\%
interval $\ev{lua/gen4-over-base/ci-lo-pp}$ to $\ev{lua/gen4-over-base/ci-hi-pp}$), which more than doubles the base's
accuracy. The lineages agree (\ev{lua/group/gen4/min-pct}\% to \ev{lua/group/gen4/max-pct}\%; \cref{fig:lua-generations}). The gain extends to the families never examined before, by
$\ev{lua/gen4-over-base/new-families/diff-pp}$ points from \ev{lua/gen4-over-base/new-families/before-pct}\%; on the
\ev{lua/gen4-over-base/old-families/families} families that earlier examinations had used, which the base finds
easier (\ev{lua/gen4-over-base/old-families/before-pct}\%), it is $\ev{lua/gen4-over-base/old-families/diff-pp}$
points, so most of the total is on families that were not new (\cref{tab:lua}). It is concentrated early: the first generation adds
$\ev{lua/step1/diff-pp}$ points and the second $\ev{lua/step2/diff-pp}$, both supported (each interval excludes zero;
the intervals are not adjusted for the six registered comparisons, and the second step's lower end is
$\ev{lua/step2/ci-lo-pp}$); the third and fourth add $\ev{lua/step3/diff-pp}$ and $\ev{lua/step4/diff-pp}$ points with
intervals that include zero. The registered
criterion that every generation improve on its predecessor is therefore not met; the criteria on the total gain and on the
new families are (\cref{app:registrations}).

\paragraph{Interpretation and answer.} The observed gains diminish, which is a result and not a failure of the method:
the cumulative gain is large and replicated across lineages, and each further step was observed to add less. The design cannot say why: the pool doubles at
every generation and is also made of the programs the current model gets wrong, so the amount and the choice of data
change together, and there is no control trained on as many randomly chosen programs. The learner's own release rule
accepted all four generations, including the two whose step the examination does not confirm (\cref{app:other}).
The answer to RQ1 is yes: with an interpreter as the only source of labels, a model more than doubled a low accuracy
on a privately generated examination outside the loop; successive improvement is supported for the first two generations and not
established for the next two.

\subsection{RQ2: Can automatically authored practice tasks teach transferable skills?}\label{sec:r-authoring}

\paragraph{Authoring yield.} Both authoring modes were run over the same \ev{funnel/forward/stage/sections/items} items
of the manual. Section by section, \ev{funnel/forward/oracle/tasks-solved} of \ev{funnel/forward/admission/candidates}
proposed tasks were admitted; outcome first, \ev{funnel/backward/admission/admitted} of
\ev{funnel/backward/proposals/proposed} proposed pairs of starting and end files became tasks, which are harder for both
models (\cref{fig:vim-funnel,tab:vim-attempts} in \cref{app:full}). Most generated material is rejected by automatic
checks, the last of them by execution, before any model trains on it.

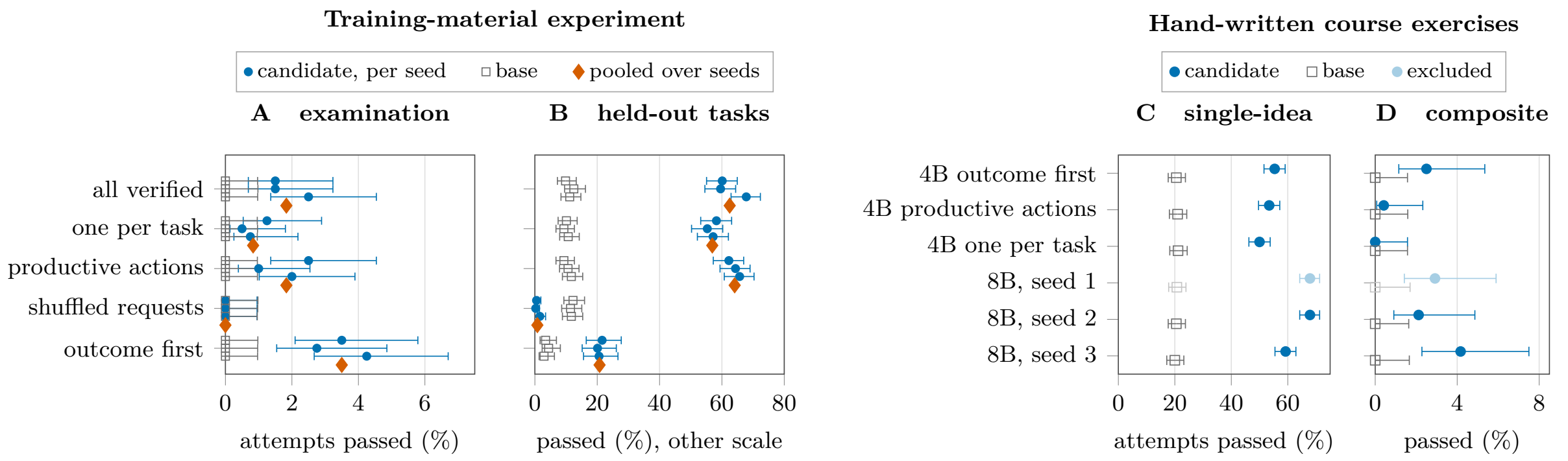
\begin{figure}[tbp]
\centering
\resizebox{\linewidth}{!}{\input{figures/exams-body}\hspace{5mm}\input{figures/practice-body}}
\caption{Training on verified records of authored tasks transfers to hand-written tasks, except when the requests are
shuffled. Left, the training-material experiment: each condition on (A) the
\ev{const/exams} examination tasks and (B) its held-out generated tasks, on different scales; a blue point is one
training seed, a grey square the base measured in the same stage, a diamond the pass rate pooled over the three seeds
(tests in \cref{tab:vim-results,tab:heldout}). Right, the hand-written course exercises, which nothing was trained on:
first-generation candidates (one training seed each) against their base on (C) the
\ev{practice/g-s3/exercises/tasks} single-idea and (D) the \ev{practice/g-s3/composites/tasks} composite exercises;
exploratory, and the faded 8B evaluation lost more attempts than the bound allows and is excluded from claims. Qwen3-4B
unless marked 8B; \ev{const/attempts-per-task} attempts per task; Wilson 95\% intervals, which treat attempts as
independent.}\label{fig:vim-seeds}\label{fig:practice}
\end{figure}

\paragraph{Transfer to the examination.} Fifteen adapters were trained from the base Qwen3-4B: the five training-material
conditions of \cref{sec:d-arms}, three training seeds each. Each adapter attempted every one of the \ev{const/exams}
examination tasks \ev{const/attempts-per-task} times, beside a base model run in the same evaluation stage. A typical
examination task asks the agent to make every underline in a document exactly as long as the heading above it. The base
passed none of its counted attempts in any stage. Every trained condition except the
control solved examination tasks, in every seed (\cref{fig:vim-seeds}). The outcome-first condition passed
\ev{arms/g/exam/cand-passes} of its \ev{arms/g/exam/cand-counted} attempts (\ev{arms/g/exam/cand-pct}\%) and solved
\ev{arms/g/exam/tasks-solved} of the \ev{const/exams} tasks at least once; the all-verified condition passed
\ev{arms/b/exam/cand-passes} and solved \ev{arms/b/exam/tasks-solved}. In the registered test each of the four non-control
conditions beats its base, with Holm's correction, and beats the shuffled-request control (\cref{tab:vim-results}): the
registered claim, that trajectories on tasks authored from the manual alone teach skills the model applies to examination
tasks, is supported.

The effect is real and small: the teacher passes \ev{teacher/exam/pct}\% of its attempts on the same examination. A
model that passes a few percent of attempts is not a usable editing agent; what the result establishes is that verified
practice moved a model from never solving such tasks to sometimes solving them. On the
held-out generated tasks the same adapters gain many times more (the all-verified condition passes
\ev{arms/b/target/cand-pct}\% of attempts where the base passes \ev{arms/b/target/base-pct}\%; \cref{tab:heldout}),
a gap that returns in \cref{sec:r-loop}.

\begin{table}[tbp]
\centering\footnotesize
\setlength{\tabcolsep}{3pt}
\begin{tabular}{@{}lrlrrll@{}}
\toprule
Training material & Records (teacher's) & Passes (per seed) & Tasks & Base & vs base, $p$ & vs control, $p$ \\
\midrule
\inputrows{generated/vim-transfer-table.tex}
\bottomrule
\end{tabular}
\caption{The training-material conditions on the \ev{const/exams} examination tasks (Qwen3-4B, three training seeds,
\ev{const/attempts-per-task} attempts per task and seed; registered analysis). Read ``Passes'' (of
\ev{arms/b/exam/cand-counted} attempts; the three seeds in parentheses) against ``Base'' (the untrained model in the same
stages, over its counted attempts). ``Records'': training records, and how many are the teacher's; ``Tasks'':
examination tasks solved at least once. The last two columns give tasks won and lost and the $p$ of a one-sided exact
sign test over tasks; all four comparisons with the base hold under Holm's correction. A comparison with the base
counts, per task, as many of a condition's attempts as the base has counted ones, hence \ev{arms/g/exam/wins} tasks won
of \ev{arms/g/exam/tasks-solved} solved for outcome first. The bases of all verified and outcome first lost
\ev{arms/b/exam/base-uncounted-pct}\% and \ev{arms/g/exam/base-uncounted-pct}\% of their attempts
(\cref{sec:d-statistics}).}\label{tab:vim-results}
\end{table}

\paragraph{Transfer to hand-written course exercises.} A base that passes nothing leaves little room to see differences.
The \ev{practice/g-s3/all/tasks} course exercises of \cref{sec:m-examiner} are easier: the base passes about
\ev{practice-transfer/4B/base-pct-low}\% of its attempts on them, and no model here was trained on them. Three
first-generation 4B candidates, one training seed each of the outcome-first, productive-actions and one-per-task
conditions, were measured on them against their base (\cref{fig:practice}). No rule fixed in advance chose these seeds: the
outcome-first one is seed~\ev{seeds/g/lowest-validation-loss}, which has the lowest validation loss and was the 4B
first generation of the earlier iterative runs (\cref{app:earlier}), and the other two are
seed~\ev{seeds/c/most-exam-passes}, with no rule recorded. Each is also the seed of its condition with the most
examination passes (\cref{tab:vim-results}), known by then, so another seed might gain less. The outcome-first
candidate passed
\ev{practice-transfer/g-s3/cand-passes} of \ev{practice-transfer/g-s3/cand-counted} attempts against the base's
\ev{practice-transfer/g-s3/base-passes}, winning \ev{practice-transfer/g-s3/wins} tasks and losing
\ev{practice-transfer/g-s3/losses} (one-sided sign test over tasks, $p=\ev{practice-transfer/g-s3/task-p}$), and the
improvement is supported for all three (\cref{tab:retention}). An 8B adapter trained on outcome-first tasks, in three
training seeds (the first generation of the earlier runs of \cref{app:earlier}; all three were measured), gains at least as much:
supported in \ev{practice-transfer/8B/eligible} of the \ev{practice-transfer/8B/evaluations} evaluations, those
within the bound on lost attempts, the third being excluded. Almost all of
the gain is on the single-idea exercises; on the composite ones every model stays near zero. This measurement is
exploratory: it was made as a check that training had not damaged what the base could already do, which the
registration of the iterative protocol names without fixing a test or a margin, and reading it as transfer is post hoc.
The label says how the analysis was chosen, not that the effect is doubtful: the test is over tasks, every eligible
evaluation agrees, and it is the largest transfer effect in the paper.

\paragraph{Answer.} Yes, under the tested conditions: one environment, and a teacher that authored the tasks and supplied
most of the records. Such tasks taught a model to solve hand-written examination tasks its base never solved and, in an
exploratory measurement, more than doubled its pass rate on hand-written course exercises.

\subsection{RQ3: Which training material produces transfer?}\label{sec:r-supervision}

\paragraph{The pairing of request and action.} The shuffled-request control was trained on exactly the records of the
all-verified condition, with every request exchanged for another task's. It passed
\ev{arms/f-vs-base/exam/cand-passes} of \ev{arms/f-vs-base/exam/cand-counted} examination attempts, in all three seeds,
and on the held-out generated tasks it did worse than the untrained base (\ev{arms/f-vs-base/target/cand-passes} passes
against \ev{arms/f-vs-base/target/base-passes}; exploratory): trained on actions that answer some other request, a model
learns to disregard what is asked. The control is destructive, not neutral. What it shows is that the
transfer of the other conditions needs records in which the action answers the request. It does not show that exposure
to the tool's format contributes nothing, because a model taught to ignore requests may lose what such exposure gave.

\paragraph{Outcome-first against section-by-section tasks.} Trained with the same selection rule, the outcome-first
condition passed \ev{arms/g-vs-b/exam/cand-passes} examination attempts against \ev{arms/g-vs-b/exam/base-passes} for the
all-verified condition, from \ev{material/g/train-records} training records against \ev{material/b/train-records}. Per
examination task it won \ev{arms/g-vs-b/exam/wins} and lost \ev{arms/g-vs-b/exam/losses} (two-sided sign test
$p=\ev{arms/g-vs-b/exam/p-two-sided}$; a registered secondary analysis, \cref{tab:secondary}). The modes are
compared as they were run: with a third of the records and a checkpoint selected at optimiser step \ev{material/g/selected-step} against
\ev{material/b/selected-step}, the outcome-first adapters had seen each record about twice as often, and their tasks
are harder, so the difference is not attributed to the authoring order.

\paragraph{Selection, and whose records.} Between leaving unproductive actions untrained and training on them no
difference was detected (interval $\ev{arms/d-vs-b/exam/ci-lo-pp}$ to $\ev{arms/d-vs-b/exam/ci-hi-pp}$ points), and
keeping one attempt per task instead of all verified attempts gave fewer examination passes on a quarter of the records
(\cref{app:other}). A sixth condition
trained on the learner's own \ev{material/s/train-records} verified records alone and passed
\ev{arms/s/exam/cand-passes} of \ev{arms/s/exam/cand-counted} counted examination attempts. This result is
\emph{excluded} from claims: two of its three evaluation runs lost more than \ev{const/uncounted-bound-pct}\% of their
attempts, its training set is a fifth the size of the all-verified one, and its held-out tasks were not evaluated. Whether
a learner's own verified attempts can produce transfer here is therefore open in both directions.

\paragraph{Answer.} Transfer needs requests paired with the actions that satisfy them; more of it came from fewer, harder,
outcome-first tasks, with the confounds named above; and all supported transfer here comes from training sets that are
mostly the teacher's.

\subsection{RQ4: Does repeating the process produce sustained improvement?}\label{sec:r-loop}

\begin{figure}[tbp]
\centering
\input{figures/iterative-body}
\caption{The first generation gains on both measures; afterwards each candidate sits beside its champion on the
examination. Each candidate (blue) beside the reference it was measured against in the same evaluation stage (grey
square: the base for generation~1, the champion afterwards). A, B: the generation's own held-out generated tasks
(\ev{B/8B/g1/target/tasks}, \ev{B/8B/g2/target/tasks} and \ev{B/8B/g3/target/tasks} tasks; a different, more demanding
set each time, so a pair is read within itself and the panels are not curves). C, D: the \ev{const/exams} examination
tasks, on a finer scale. One training run per point, \ev{const/attempts-per-task} attempts per task; Wilson 95\%
intervals, which treat attempts as independent. Faded: excluded from claims. Open circle: not accepted, so the next
candidate is compared with the same champion again. A stage's champion is the previous accepted candidate measured
anew (\cref{sec:r-projection}).}\label{fig:iterative}
\end{figure}
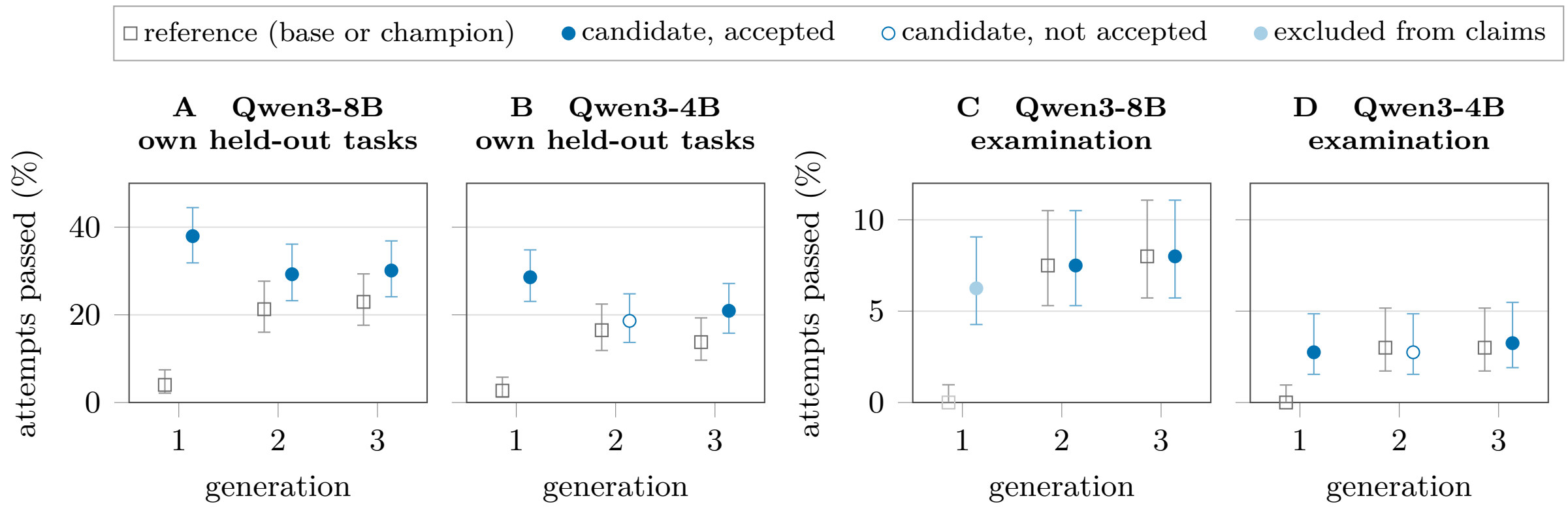

\paragraph{What was tested.} The iterative protocol of \cref{sec:m-loop} ran for three generations with Qwen3-8B and with
Qwen3-4B as the learner; no person's decision entered between generations. In every generation the candidate and its
reference model (the base, then the current champion) attempted, in the same evaluation stage, the generation's own
held-out generated tasks, every earlier generation's, and the examination tasks (\cref{fig:iterative}).

\paragraph{Held-out generated tasks.} On its own held-out tasks the first 8B generation passed
\ev{B/8B/g1/target/cand-passes} of \ev{B/8B/g1/target/cand-counted} attempts where the base passed
\ev{B/8B/g1/target/base-passes}, and the first 4B generation \ev{B/4B/g1/target/cand-passes} where its base passed
\ev{B/4B/g1/target/base-passes}. Whichever way the recorded attempts fall on the tasks, the sign test over tasks is
significant (at most $p=\ev{B/8B/g1/target/task-p-high}$ at 8B and $\ev{B/4B/g1/target/task-p-high}$ at 4B): this
improvement is supported. Later differences are smaller and undecided (counts in \cref{tab:iterative}). At 8B the release
rule accepted the second and third generations; at 4B it rejected the second and accepted the third, compared with the
same champion. In these later steps the candidates were observed higher, which is not an established gain: the test over
tasks cannot be computed for them, and the values it could take include both significance and its absence. A later
candidate's higher rate is also partly its first exposure to that generation's kind of task, which its champion never
practised.

\paragraph{The examination.} The first generations pass examination attempts where the bases pass none:
\ev{B/8B/g1/exam/cand-passes} of \ev{B/8B/g1/exam/cand-counted} at 8B and \ev{B/4B/g1/exam/cand-passes} at 4B. Neither is
a confirmed result of this protocol. The 8B comparison is excluded, because the base lost
\ev{B/8B/g1/exam/base-uncounted} of its \ev{B/8B/g1/exam/base-scheduled} attempts (\ev{B/8B/g1/exam/uncounted-pct}\%,
above the bound). The 4B comparison is inconclusive only because the tasks of its passing attempts were not recorded;
where they were, in the similar first generation of \cref{tab:vim-results}, the gain is supported. Supported first-generation transfer is therefore that of the 4B model in the training-material experiment; for 8B it
rests on the exploratory course exercises alone. After the first generation no examination gain was
established: at 8B the second and third generations pass exactly as many attempts as their champions
(\ev{B/8B/g2/exam/cand-passes} and \ev{B/8B/g3/exam/cand-passes} of \ev{B/8B/g2/exam/cand-counted}), and at 4B the
candidates pass \ev{B/4B/g2/exam/cand-passes} and \ev{B/4B/g3/exam/cand-passes} against the champion's
\ev{B/4B/g2/exam/base-passes}.

\begin{figure}[tbp]
\centering
\input{figures/resolution-body}
\caption{If transfer continued in proportion, its examination gain would be several times smaller than the smallest gain
this examination detects. For each later step of the iterative protocol: each training seed's examination difference
from the champion measurement the seeds share, and their mean; the first seed's 95\% interval, with attempts taken as
independent and the pairing ignored; the gain predicted from the
step's held-out difference by the first generation's ratio (bar: ratio \ev{projection/yardstick/low-ratio} to
\ev{projection/yardstick/high-ratio}); and the smallest gain a paired comparison of this size detects (one-sided
$\alpha=\ev{resolution/alpha}$, \ev{resolution/power}\% power). Post hoc and descriptive; values in
\cref{tab:projection}.}\label{fig:resolution}
\end{figure}
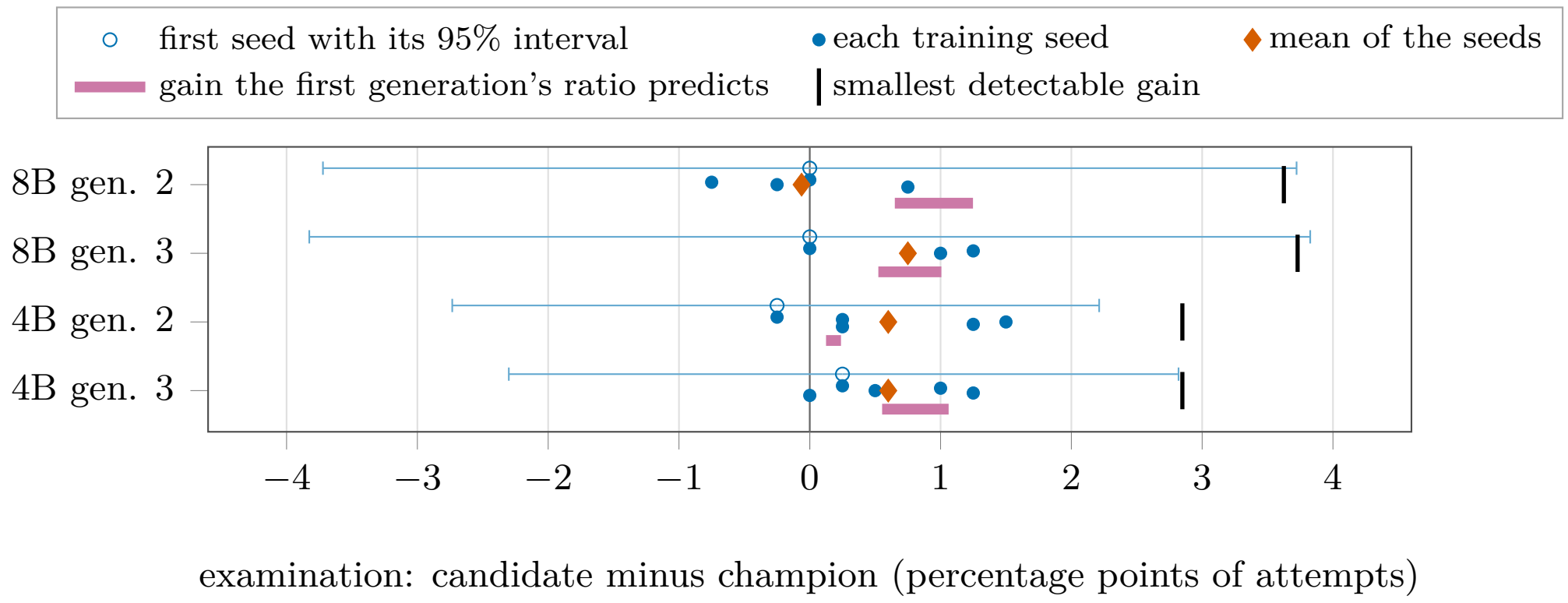

\paragraph{Training-step replications.}\label{sec:r-seeds} One training run per step could be an accident of its seed, so
each later step was repeated with further training seeds (\cref{fig:resolution}; \cref{fig:seeds,tab:replications} in
\cref{app:full}). On held-out generated tasks the candidate passed more attempts than the champion in
\ev{replications/target/seeds-positive} of the \ev{replications/target/seeds} runs. On the examination no seed differs
from its champion by more than \ev{projection/seed-diff-max-pp} percentage points: \ev{replications/exam/seeds-positive}
seeds are above it, \ev{replications/exam/seeds-negative} below and \ev{replications/exam/seeds-level} level. That
direction is untested, because the seeds of a step share their tasks and one measurement of the champion, so a low
reading of the champion would move them all together. The adapters of the iterative protocol and of its replications
were not kept, so these evaluations cannot be run again with the task of each attempt recorded.

\paragraph{What the examination can resolve.}\label{sec:r-projection} Does the lack of an established examination gain
mean that transfer stopped after the first generation? A calculation made after the results, not registered, says the
data cannot tell (\cref{fig:resolution}; \cref{tab:ratios,tab:projection} in \cref{app:full}). It has three steps.

\emph{How much examination gain went with a held-out gain.} In the first generation the examination gain was a tenth to
a fifth of the held-out gain: a ratio of \ev{projection/4B/ratio} in the 4B run of this protocol and of
\ev{projection/arms/g/ratio} in the outcome-first condition of the training-material experiment, which trained the same
kind of first generation on the same kind of task in three seeds and whose two gains are both supported. The 8B run
gives \ev{projection/8B/ratio}; its examination is excluded from claims, so it is shown as a check only, and it lies
between the other two. The ratio is not a constant of the method: the conditions trained on section-by-section tasks
give \ev{projection/arms/c/ratio} to \ev{projection/arms/b/ratio}.

\emph{What that predicts.} Averaged over seeds, the later candidates were \ev{projection/heldout-gain-min-pp} to
\ev{projection/heldout-gain-max-pp} points above their champions on held-out generated tasks. A tenth to a fifth of
that is an examination gain of \ev{projection/predicted-min-pp} to \ev{projection/predicted-max-pp} points. The
argument does not need the later held-out differences to be real gains: if they are smaller, the prediction is smaller
still.

\emph{What the examination detects.} A paired comparison of two trained models on \ev{const/exams} tasks at
\ev{const/attempts-per-task} attempts detects a difference of \ev{resolution/detectable-min-pp} to
\ev{resolution/detectable-max-pp} points (one-sided $\alpha=\ev{resolution/alpha}$, \ev{resolution/power}\% power), and
that figure flatters the examination, because it treats the attempts of one task as independent. To detect the largest
gain predicted for a step, the same comparison would need \ev{resolution/tasks-needed/high-min} to
\ev{resolution/tasks-needed/high-max} tasks, at least \ev{resolution/tasks-factor/high-min} times the hundred there are.
The measurements themselves show the noise: the same adapter, measured in two stages, passed
\ev{repeat/8B/g1/stage2/first-passes} and then \ev{repeat/8B/g1/stage2/second-passes} of
\ev{repeat/8B/g1/stage2/attempts} attempts, \ev{repeat/max-diff-pp} points apart with nothing changed but the sampling
of its attempts. The observed differences, averaged over seeds, lie between $\ev{projection/observed-min-pp}$ and
$\ev{projection/observed-max-pp}$ points; they and the predictions are inside the interval of every later comparison.
The ratio is a yardstick, not a model, since later held-out tasks are a different and harder set: the comparison shows
only that transfer continuing in proportion and transfer having stopped are both compatible with these measurements.

\paragraph{Answer.} The process repeats: three generations ran at two model sizes, and the release rule accepted five
candidates and rejected one. Acquisition is supported for the first generation; afterwards candidates were observed
higher than their champions on held-out generated tasks, by several times less, and no test decides it. Sustained improvement on the independent examination is not established, and neither is its absence. Deciding the question needs outcomes recorded per task and a larger or easier independent evaluation, not more
generations measured the same way.

\subsection{RQ5: Are earlier capabilities retained, and does learning become more efficient?}\label{sec:r-retention}

\paragraph{Retention across generations.} Each later candidate was measured against its champion on the held-out tasks of
every earlier generation (\cref{fig:retention} in \cref{app:full}). The observed differences range from
$\ev{B/8B/g3/retention-g2/diff-pp}$ to $\ev{B/4B/g3/retention-g1/diff-pp}$ percentage points, and all six comparisons satisfied
the retention condition of the release rule. Its margin is as large as the later differences on held-out tasks, so
it guards only against a large loss. Showing retention as non-inferiority needs the outcomes per task, which the runs did not
record, and the descriptive intervals cross the margin in several comparisons. We therefore report that the candidates
satisfied the specified rule and leave the retention claim inconclusive.

Training on authored tasks could also damage skills the base has. On the course exercises of \cref{sec:r-authoring},
which the base partly solves, it did not: retention is supported for the three 4B candidates and in the
\ev{practice-transfer/8B/eligible} eligible 8B evaluations, although each candidate also loses some tasks
(\cref{tab:retention}).

\paragraph{Efficiency.}\label{sec:r-efficiency} Whether learning makes further learning cheaper needs the cost of each
generation beside its gain, and the evidence records only part of that cost (\cref{app:compute}). For Lua it records
\ev{compute/lua/g1/gpu-hours}, \ev{compute/lua/g2/gpu-hours}, \ev{compute/lua/g3/gpu-hours} and
\ev{compute/lua/g4/gpu-hours} accelerator-hours (hours of one graphics card) of proposing and training for generations
one to four, the last of which added the least. For the Vim iterative protocol it records no cost.

\paragraph{Answer.} Every accepted candidate satisfied the retention rule; as a tested non-inferiority claim, retention is
supported for the course exercises only. No improvement in learning efficiency is established.

\paragraph{What was and was not established.}\label{sec:r-ledger} \Cref{tab:ledger} collects the capabilities the paper
set out to show; each status is computed from the claim registry (\cref{app:claims}).

\begin{table}[!htb]
\centering\footnotesize
\begin{tabular}{@{}>{\raggedright\arraybackslash}p{98mm}>{\raggedright\arraybackslash}p{56mm}@{}}
\toprule
Capability & Status \\
\midrule
\inputrows{generated/ledger.tex}
\bottomrule
\end{tabular}
\caption{Summary of claims (the ledger of \cref{app:claims}), in the status words of \cref{sec:d-statistics}. \emph{Exercised} says that the mechanisms
ran and produced the counted products, not that they have an effect. \emph{Observed higher} and \emph{release rule
satisfied}: the measured differences have the claimed direction, or lie within the rule's margin, and no test decides
them.}\label{tab:ledger}
\end{table}

%% file: figures/lua-generations-body.tex
\begin{tikzpicture}
\begin{groupplot}[group style={group size=2 by 1, horizontal sep=20mm}, paper, height=36mm]
\nextgroupplot[width=68mm, title={Lua: examination items answered correctly}, xmin=-0.4, xmax=4.4, ymin=0, ymax=16, xtick={0,1,2,3,4},
    xticklabels={base,1,2,3,4}, xlabel={generation}, ylabel={\% of \ev{lua/gen4-over-base/items} items}]
  \addplot[only marks, mark=o, mark size=1.4pt, okGrey] table[x=generation, y expr=100*\thisrow{share}]
    {generated/lua-generations.csv};
  \addplot[okBlue, thick, mark=*, mark size=1.6pt, discard if not={lineage}{0}]
    table[x=generation, y expr=100*\thisrow{mean}] {generated/lua-generations.csv};
\nextgroupplot[width=78mm, title={Lua: registered comparisons, 95\% interval}, xmin=0.4, xmax=6.6, ymin=-2.5, ymax=12.5, xtick={1,...,6},
    xticklabels={\strut 1 vs\\\strut base, \strut 2 vs\\\strut 1, \strut 3 vs\\\strut 2, \strut 4 vs\\\strut 3, \strut 4 vs\\\strut base, \strut new\\\strut families}, x tick label style={align=center, font=\scriptsize},
    ylabel={percentage points}, extra y ticks={0}, extra y tick style={grid style={black!60}, tick label style={opacity=0}}]
  \addplot[only marks, mark=*, mark size=1.8pt, okBlue, discard if not={holds}{1}, error bars/.cd, y dir=both, y explicit]
    table[x=x, y=diff, y error plus expr=\thisrow{hi}-\thisrow{diff}, y error minus expr=\thisrow{diff}-\thisrow{lo}]
    {generated/lua-steps.csv};
  \addplot[only marks, mark=o, mark size=1.8pt, okGrey, discard if not={holds}{0}, error bars/.cd, y dir=both, y explicit]
    table[x=x, y=diff, y error plus expr=\thisrow{hi}-\thisrow{diff}, y error minus expr=\thisrow{diff}-\thisrow{lo}]
    {generated/lua-steps.csv};
\end{groupplot}
\end{tikzpicture}

%% file: figures/exams-body.tex
\begin{tikzpicture}
\pgfplotsset{arms/.style={paper, width=49mm, height=45mm, ymin=0.35, ymax=5.85, ytick={1,2,3,4,5}, ymajorgrids=false,
    xmajorgrids, xtick pos=bottom, clip=false, title style={font=\footnotesize\bfseries, align=center}}}
\begin{groupplot}[group style={group size=2 by 1, horizontal sep=8mm}]
\nextgroupplot[arms, title={\strut A\quad examination}, xlabel={attempts passed (\%)}, xmin=0, xmax=7.5,
    yticklabels={outcome first, shuffled requests, productive actions, one per task, all verified},
    legend to name=armslegend, legend columns=3,
    legend style={/tikz/every even column/.append style={column sep=4mm}}]
  \addplot[only marks, mark=*, mark size=1.5pt, okBlue, discard if not={side}{candidate},
    error bars/.cd, x dir=both, x explicit]
    table[y expr=6-\thisrow{x}, x expr=100*\thisrow{rate}, x error plus expr=100*(\thisrow{hi}-\thisrow{rate}),
      x error minus expr=100*(\thisrow{rate}-\thisrow{lo})] {generated/vim-transfer-exam-seeds.csv};
  \addlegendentry{candidate, per seed}
  \addplot[only marks, mark=square, mark size=1.5pt, okGrey, discard if not={side}{baseline},
    error bars/.cd, x dir=both, x explicit]
    table[y expr=6-\thisrow{x}, x expr=100*\thisrow{rate}, x error plus expr=100*(\thisrow{hi}-\thisrow{rate}),
      x error minus expr=100*(\thisrow{rate}-\thisrow{lo})] {generated/vim-transfer-exam-seeds.csv};
  \addlegendentry{base}
  \addplot[only marks, mark=diamond*, mark size=2.8pt, okVermillion, discard if not={eval}{exam}]
    table[y expr=6-\thisrow{x}-0.42, x expr=100*\thisrow{rate}] {generated/vim-transfer-pooled.csv};
  \addlegendentry{pooled over seeds}
\nextgroupplot[arms, title={\strut B\quad held-out tasks}, xlabel={passed (\%), other scale}, xmin=0, xmax=80, yticklabels={}]
  \addplot[only marks, mark=*, mark size=1.5pt, okBlue, discard if not={side}{candidate},
    error bars/.cd, x dir=both, x explicit]
    table[y expr=6-\thisrow{x}, x expr=100*\thisrow{rate}, x error plus expr=100*(\thisrow{hi}-\thisrow{rate}),
      x error minus expr=100*(\thisrow{rate}-\thisrow{lo})] {generated/vim-transfer-t1-seeds.csv};
  \addplot[only marks, mark=square, mark size=1.5pt, okGrey, discard if not={side}{baseline},
    error bars/.cd, x dir=both, x explicit]
    table[y expr=6-\thisrow{x}, x expr=100*\thisrow{rate}, x error plus expr=100*(\thisrow{hi}-\thisrow{rate}),
      x error minus expr=100*(\thisrow{rate}-\thisrow{lo})] {generated/vim-transfer-t1-seeds.csv};
  \addplot[only marks, mark=diamond*, mark size=2.8pt, okVermillion, discard if not={eval}{t1}]
    table[y expr=6-\thisrow{x}-0.42, x expr=100*\thisrow{rate}] {generated/vim-transfer-pooled.csv};
\end{groupplot}
\node[anchor=south] (legend) at ($(group c1r1.north)!0.5!(group c2r1.north)+(0,7mm)$) {\pgfplotslegendfromname{armslegend}};
\node[anchor=south, font=\footnotesize\bfseries] at (legend.north) {Training-material experiment};
\end{tikzpicture}

%% file: figures/practice-body.tex
\begin{tikzpicture}
\newcommand{\exercisepoints}[5]{%
  \addplot[#4, discard if not={part}{#1}, discard if not={side}{#2}, discard if not={status}{#3},
    error bars/.cd, x dir=both, x explicit]
    table[y expr=\thisrow{y}+#5, x=pct, x error plus expr=\thisrow{hi}-\thisrow{pct},
      x error minus expr=\thisrow{pct}-\thisrow{lo}] {generated/retention.csv};
}
\newcommand{\exercises}[1]{%
  \exercisepoints{#1}{candidate}{counted}{candidate}{-0.12}
  \exercisepoints{#1}{baseline}{counted}{reference}{0.12}
  \exercisepoints{#1}{candidate}{excluded}{candidate, okBlue!35}{-0.12}
  \exercisepoints{#1}{baseline}{excluded}{reference, okGrey!40}{0.12}
}
\pgfplotsset{ret/.style={paper, height=45mm, ytick={1,2,3,4,5,6}, y dir=reverse, ymin=0.5, ymax=6.5, ymajorgrids=false,
    xmajorgrids, xmin=0, title style={font=\footnotesize\bfseries, align=center}}}
\begin{groupplot}[group style={group size=2 by 1, horizontal sep=6mm}]
\nextgroupplot[ret, width=44mm, xmax=75, title={\strut C\quad single-idea},
    xlabel={attempts passed (\%)},
    yticklabels={{4B outcome first}, {4B productive actions}, {4B one per task}, {8B, seed 1}, {8B, seed 2}, {8B, seed 3}},
    legend to name=practicelegend, legend columns=3, legend style={/tikz/every even column/.append style={column sep=3mm}}]
  \addlegendimage{candidate}\addlegendentry{candidate}
  \addlegendimage{reference}\addlegendentry{base}
  \addlegendimage{candidate, okBlue!35}\addlegendentry{excluded}
  \exercises{exercises}
\nextgroupplot[ret, width=39mm, xmax=8.5, xtick={0,4,8}, title={\strut D\quad composite},
    xlabel={passed (\%)}, yticklabels={}]
  \exercises{composites}
\end{groupplot}
\node[anchor=south] (legend) at ($(group c1r1.north west)!0.5!(group c2r1.north east)+(0,7mm)$) {\pgfplotslegendfromname{practicelegend}};
\node[anchor=south, font=\footnotesize\bfseries] at (legend.north) {Hand-written course exercises};
\end{tikzpicture}

%% file: figures/iterative-body.tex
\begin{tikzpicture}
\newcommand{\pairs}[2]{%
  \addplot[scatter, only marks, mark size=1.8pt, point meta=explicit symbolic, discard if not={size}{#1},
    discard if not={kind}{#2}, discard if not={side}{baseline},
    scatter/classes={reference={mark=square, okGrey}, excluded={mark=square, okGrey!40}},
    error bars/.cd, x dir=none, y dir=both, y explicit, error bar style={okGrey!70}]
    table[x expr=\thisrow{generation}-0.14, y=pct, meta=marker, y error plus expr=\thisrow{hi}-\thisrow{pct},
      y error minus expr=\thisrow{pct}-\thisrow{lo}] {generated/iterative-arms.csv};
  \addplot[scatter, only marks, mark size=1.8pt, point meta=explicit symbolic, discard if not={size}{#1},
    discard if not={kind}{#2}, discard if not={side}{candidate},
    scatter/classes={accepted={mark=*, okBlue}, unpromoted={mark=*, draw=okBlue, fill=white},
      excluded={mark=*, okBlue!35}},
    error bars/.cd, x dir=none, y dir=both, y explicit, error bar style={okBlue!60}]
    table[x expr=\thisrow{generation}+0.14, y=pct, meta=marker, y error plus expr=\thisrow{hi}-\thisrow{pct},
      y error minus expr=\thisrow{pct}-\thisrow{lo}] {generated/iterative-arms.csv};
}
\pgfplotsset{iter/.style={paper, width=46mm, height=38mm, xmin=0.5, xmax=3.5, xtick={1,2,3}, xmajorgrids=false,
    xlabel={generation}, title style={font=\scriptsize\bfseries, align=center}}}
\begin{groupplot}[group style={group size=4 by 1, horizontal sep=4mm}]
\nextgroupplot[iter, title={A\quad Qwen3-8B\\own held-out tasks}, ymin=0, ymax=50, ylabel={attempts passed (\%)},
    legend to name=iterativelegend, legend columns=4,
    legend style={/tikz/every even column/.append style={column sep=4mm}}]
  \addlegendimage{reference}\addlegendentry{reference (base or champion)}
  \addlegendimage{candidate}\addlegendentry{candidate, accepted}
  \addlegendimage{only marks, mark=*, mark size=1.8pt, draw=okBlue, fill=white}\addlegendentry{candidate, not accepted}
  \addlegendimage{only marks, mark=*, mark size=1.8pt, okBlue!35}\addlegendentry{excluded from claims}
  \pairs{8B}{target}
\nextgroupplot[iter, title={B\quad Qwen3-4B\\own held-out tasks}, ymin=0, ymax=50, yticklabels={}]
  \pairs{4B}{target}
\nextgroupplot[iter, title={C\quad Qwen3-8B\\examination}, ymin=0, ymax=12, xshift=11mm, ylabel={attempts passed (\%)}]
  \pairs{8B}{exam}
\nextgroupplot[iter, title={D\quad Qwen3-4B\\examination}, ymin=0, ymax=12, yticklabels={}]
  \pairs{4B}{exam}
\end{groupplot}
\node[anchor=south] at ($(group c1r1.north west)!0.5!(group c4r1.north east)+(0,11mm)$) {\pgfplotslegendfromname{iterativelegend}};
\end{tikzpicture}

%% file: figures/resolution-body.tex
\begin{tikzpicture}
\begin{axis}[paper, width=118mm, height=40mm, xmin=-4.6, xmax=4.6, ymin=0.45, ymax=4.6, y dir=reverse,
    ytick={1,2,3,4}, yticklabels from table={generated/resolution-steps.csv}{label}, ymajorgrids=false, xmajorgrids,
    xlabel={examination: candidate minus champion (percentage points of attempts)},
    extra x ticks={0}, extra x tick labels={}, extra x tick style={grid style={black!60}},
    legend to name=resolutionlegend, legend columns=3,
    legend style={/tikz/every even column/.append style={column sep=3mm}}]
  \addplot[only marks, mark=o, mark size=1.6pt, okBlue, error bars/.cd, x dir=both, x explicit,
      error bar style={okBlue!60}]
    table[y expr=\thisrow{y}-0.24, x=single, x error plus expr=\thisrow{hi}-\thisrow{single},
      x error minus expr=\thisrow{single}-\thisrow{lo}] {generated/resolution-steps.csv};
  \addlegendentry{first seed with its 95\% interval}
  \addplot[candidate, mark size=1.4pt] table[y expr=\thisrow{y}+0.035*(\thisrow{seed}-3), x=diff]
    {generated/resolution-seeds.csv};
  \addlegendentry{each training seed}
  \addplot[pooled] table[y=y, x=observed] {generated/resolution-steps.csv};
  \addlegendentry{mean of the seeds}
  \addplot[okPurple, line width=2.6pt, forget plot, quiver={u=\thisrow{predicted_hi}-\thisrow{predicted_lo}, v=0}]
    table[y expr=\thisrow{y}+0.27, x=predicted_lo] {generated/resolution-steps.csv};
  \addlegendimage{okPurple, line width=2.6pt}
  \addlegendentry{gain the first generation's ratio predicts}
  \addplot[only marks, mark=|, mark size=4.5pt, line width=1pt, black] table[y=y, x=detectable]
    {generated/resolution-steps.csv};
  \addlegendentry{smallest detectable gain}
\end{axis}
\node[anchor=south] at ($(current axis.north)+(0,1mm)$) {\pgfplotslegendfromname{resolutionlegend}};
\end{tikzpicture}

%% file: sections/07-discussion.tex
\section{Discussion}\label{sec:discussion}

\label{sec:d-divergence}\emph{Interpretation.} Held-out generated tasks measure competence on the generator's distribution, not on that of an
examination written by people, so a gain on the first is not by itself evidence of the second. In these data the two are
linked in the first generation by a small ratio, and at that ratio the examination is too coarse an instrument for every
later step (\cref{fig:resolution}). Two different limits stand behind ``not established''. For later acquisition and for retention the limit is
the record: outcomes were kept per attempt, so tests over tasks that the collected data might have decided cannot be
computed. For later transfer the limit is resolution: even with outcomes per task, an examination of this size could not
have detected the predicted effect. Neither limit was the supply of tasks, records or accepted candidates: the first was an omission of these runs, and
the second could have been computed before them.

\emph{Hypotheses, if the examination gain is in fact absent.} Four explanations are possible and the data isolate none. Later generations' tasks combine more ideas, but a task
that is harder to author is not thereby closer to what the examination demands.
Verified attempts are concentrated on tasks a model can already solve, so training on them reinforces what is already
present. The first generation's examination passes may be what imitation of the teacher transfers, exhausted in one round. And authored requests may describe the edit more closely than examination tasks do:
the request of \cref{fig:example} names the change while the examination task quoted in \cref{sec:r-authoring} states a
goal, a difference whose generality we did not measure.

\paragraph{Evaluation resolution.}\label{sec:d-resolution} Three practices follow for systems of this kind. Record the
outcome of every attempt with its task, so that the test over tasks can always be computed. Size the independent
evaluation, before the run, for the effect that the previous generation's ratio predicts and not for the first
generation's large one: here that means \ev{resolution/tasks-needed/high-min} tasks or more where the examination has
\ev{const/exams}. And keep an easier, larger tier of independent tasks on which the base is partly competent: on the course exercises a
first-generation difference was decided at $p\le\ev{practice-transfer/4B/p-max}$ with one training seed per candidate.

\paragraph{What the results mean in practice, and elsewhere.} With a capable teacher, one round of this procedure moved a small model from never solving a class of hand-written
tasks to solving a few percent of attempts, far below its teacher, and more than doubled its pass rate on easier ones;
nothing here shows that repeating the procedure adds more. The framework
needs three things from an environment: actions that can be executed, a check that decides whether the outcome is the
wanted one, and source material for authoring tasks. For debugging firmware on an emulated microcontroller, the source
material is a datasheet and a driver; the outcome-first author proposes a working firmware and a variant with an
injected fault; the teacher must repair the fault with a debugger tool; the check is the firmware's test passing on the
emulator; and the request, written last, is the bug report. The independent examination would be faults from real projects. If our results carry over, a gain on generated faults comes first, and transfer to real ones must be
measured with an instrument sized for a small effect.

%% file: sections/08-limitations.tex
\section{Limitations}\label{sec:limitations}

\paragraph{Scope and controls.} Each environment was studied under one task protocol, with two sizes of one model family,
rank-16 adapters, and one author and teacher from the same family. The teacher reasons before it acts while the learners
are trained on its actions alone, which may bound what imitation transfers. There is no matched-data control for the Lua
pool, so the gain is not attributed to the selection of mispredicted programs as opposed to the volume of labelled
programs; it was measured with reasoning switched off, and a descriptive examination of the same candidates with
reasoning on was inconclusive (\cref{app:other}). The comparison of authoring modes is confounded with passes per record and task difficulty; the learner-only
condition is smaller and its evaluation excluded; the shuffled-request control is destructive; and the iterative protocol was run as a whole, with
no ablation of its choices and no control of its own.

\paragraph{Statistical validity.} The iterative protocol and its replications recorded outcomes per attempt, not per task,
so only the first generation's gain on held-out generated tasks is decided, and every later step, retention comparison
and examination comparison of that protocol is an observation. The replications are not independent. The resolution calculation of \cref{sec:r-projection} and the reading of the
course exercises as transfer were made after the results. The bound on lost attempts was applied at two scopes (\cref{sec:d-statistics}). Supported
first-generation transfer to the examination is shown for the 4B model only.

\paragraph{The examination.} The Vim examination is \ev{const/exams} tasks from one course and is public, as are the
single-idea course exercises: a base model, the teacher or the task author may have seen them in pre-training~\cite{sainz2023nlp}. The course
comes from the same organisation as the authoring rules, and the same \ev{const/exams} tasks measured every Vim
experiment, including the earlier iterative runs whose results were known when the iterative protocol was written
(\cref{app:earlier}): the examination is outside the pipeline, not outside the design of the study. Only the Lua examination is private, and its examiner is
operated by the author.

\paragraph{Autonomy and cost.} The runs were autonomous in the sense of \cref{sec:problem} and not unattended: the 4B
iterative run was paused by the operator for scheduling and resumed from its recorded state, and one leftover sandbox
was removed by hand (\cref{app:deviations}). The cost of a generation is recorded only in part.

%% file: sections/09-reproducibility.tex
\section{Reproducibility}\label{sec:reproducibility}

Reproduction has four levels (\cref{tab:levels} in \cref{app:repro-vim}; every artefact with its access and pinned
revision in \cref{tab:artifacts}). \emph{Level 1}: every table, figure, statistic and claim status is regenerated from
the evidence files committed beside the paper's sources, on a CPU in minutes; it is exercised at every build and checks
the analysis, not the measurements. \emph{Levels 2 and 3}, evaluating the recorded adapters and training on the recorded
training sets, need private artefacts and reach RQ2 and RQ3 only. \emph{Level 4}, full replication, starts from public
material: the three code repositories, the base models, the course and the reproduction materials (task environments,
authoring prompts and operators, every recipe and the examiner; one public repository pinned by commit), with the
scripts in \nolinkurl{reproduce/}. A small end-to-end run was exercised and the materials' recipes were validated by
the controller; no full experiment was run from them. A replication is a new sample
(\cref{app:repro-vim-limits}): the recorded tasks, adapters and evidence archives are private, the Lua examination
cannot be redrawn nor its recorded verdicts verified by a reader, and the iterative runs recorded no outcomes per task
and kept no adapters, so the recorded RQ4 can be checked at Level~1 only. No independent party has rerun the procedure.

%% file: sections/10-conclusion.tex
\section{Conclusion}\label{sec:conclusion}

We set out to turn what a model did in an executable environment, and what the environment said about it, into competence
that is kept. In Lua, an interpreter alone more than doubled a model's accuracy on an independent examination over four
generations. In Vim, tasks that a larger teacher model authored from a manual, with that teacher's verified solutions,
taught a 4B model to solve hand-written examination tasks and, in an exploratory measurement, hand-written course
exercises; a control showed that without the pairing of request and action nothing transferred.

The framework is a result in its own right: three components (Splinter, Sven and Brain) carried a model from source material to examined releases
over three generations at two model sizes, every release decided by rule, as a process an AI agent can itself read and
extend; that is where work towards generalized self-improvement starts, and it is not tested here. Continuous learning has three parts that can be measured separately. Acquisition and
transfer are supported in the first generation, transfer for the 4B model; sustained acquisition, transfer and
retention across generations are not established. Later candidates were observed higher than their champions on tasks
from the system's own generator, by several times less than the first generation gained; neither those differences nor
retention can be tested, because outcomes were recorded per attempt and not per task. For transfer the limit is
resolution: in proportion to the first generation the later steps would gain \ev{projection/predicted-min-pp} to
\ev{projection/predicted-max-pp} points on the examination, and a hundred-task examination detects
\ev{resolution/detectable-min-pp} points or more. The next step is an evaluation built for small effects: outcomes
recorded per task, several times more tasks, and a tier on which the base is partly competent. Open-ended curricula,
self-modification and learning to learn are future directions that build on these blocks.

%% file: sections/90-appendix-artifacts.tex
\FloatBarrier
\section{Artefacts and builds}\label{app:artifacts}

\Cref{tab:mapping} maps the roles of the framework to the software that implements them. \Cref{tab:artifacts} gives
every dataset, model, evidence archive and code repository the paper relies on, with its access and the revision its
numbers come from (full hashes in \texttt{data/artifacts.csv}). \Cref{tab:builds} lists the source commits and the frozen
binaries of the Vim runs (full hashes in \texttt{data/builds.csv}). The three code repositories and the reproduction materials are public at
\nolinkurl{github.com/swedishembedded}; the materials are cited at commit \texttt{\MaterialsCommit}
(\nolinkurl{\MaterialsRepo}). The Brain commit \texttt{b726c045bb34}, from which the servers of the
training-material experiment were built, corresponds to the public commit \texttt{d6bd3baaff53}, given in parentheses: the
two trees differ only in the package's version number and the lock file's entries for it. A binary named
\texttt{<repository>-<commit>/<file>} is the frozen copy every run of that stage used; its sha256 is given to 16 hex
digits.

\begin{table}[!htb]
\centering\footnotesize
\begin{tabular}{@{}>{\raggedright\arraybackslash}p{70mm}>{\raggedright\arraybackslash}p{84mm}@{}}
\toprule
Role in the framework & Implementation \\
\midrule
Learning controller: runs the stages of a generation, builds training sets, applies the release rule & Splinter \\
Agent and tool execution: one attempt of a model on a task & Sven \\
Inference and optimisation: serving models, training adapters & Brain \\
Vim task environment, authoring prompts, schemas, operators and recipes; Lua examination generator, its check and the
examiner & the reproduction materials of this paper (\nolinkurl{\MaterialsName}) \\
Lua learning cycle: proposing programs, keeping the mispredicted ones, training, the learner's own release rule &
the \texttt{lineage} program of Splinter's repository (\nolinkurl{samples/campaign-lua}), at the commit pinned in
\texttt{prereg/rsi5-launch.md} \\
\bottomrule
\end{tabular}
\caption{Roles of the framework and the software that implements them.}\label{tab:mapping}
\end{table}

\begin{table}[!htb]
\centering\scriptsize
\setlength{\tabcolsep}{3pt}
\begin{tabular}{@{}>{\raggedright\arraybackslash}p{12mm}>{\raggedright\arraybackslash}p{53mm}>{\raggedright\arraybackslash}p{20mm}>{\raggedright\arraybackslash}p{13mm}>{\raggedright\arraybackslash}p{53mm}@{}}
\toprule
Kind & Repository & Revision & Access & Holds \\
\midrule
\inputrows{generated/artifacts.tex}
\bottomrule
\end{tabular}
\caption{Every artefact the paper relies on: datasets, models, evidence archives, code and the reproduction materials,
with the revision every number was taken from and its access. Private artefacts are available from the author on request.
For the code repositories the revision is that of the iterative protocol's runs; \cref{tab:builds} has the
others.}\label{tab:artifacts}
\end{table}

\begin{table}[!htb]
\centering\scriptsize
\setlength{\tabcolsep}{3pt}
\begin{tabular}{@{}>{\raggedright\arraybackslash}p{11mm}>{\raggedright\arraybackslash}p{21mm}>{\raggedright\arraybackslash}p{49mm}>{\raggedright\arraybackslash}p{24mm}>{\raggedright\arraybackslash}p{42mm}@{}}
\toprule
Repo. & Commit & Frozen binary & sha256 & Used for \\
\midrule
\inputrows{generated/builds.tex}
\bottomrule
\end{tabular}
\caption{Source commits and frozen binaries of the Vim runs. The digests are given as recorded; two Brain builds record
the same digest for their accelerator worker. The builds of the Lua runs are pinned in \texttt{prereg/rsi5-launch.md}.}
\label{tab:builds}
\end{table}

\paragraph{Where the evidence is.} Lua: the examiner's verdict over the four lineages is
\nolinkurl{data/rsi5-groups-G.json}; it carries an authentication code (HMAC-SHA256) made with the examiner's secret key, which
cannot be verified without that key; adapters, lineage records and the verdicts per candidate are in
\nolinkurl{swedishembedded/generalized-continuous-learning-evidence}. Vim: the aggregate evidence, the analysis code and
the data of every figure are in \nolinkurl{swedishembedded/source-to-transfer-evidence}; the files under
\nolinkurl{data/vim-transfer/} are byte copies of that dataset's (\texttt{make vim-data}). Outcomes per examination item
stay with the examiner's private evidence. The registered decision of the training-material experiment is recomputed by
\texttt{decide.py} and \texttt{secondary.py} of the evidence dataset from the outcomes per task; the paper's tables,
figures and macros are made by \nolinkurl{scripts/paper_data.py} from \nolinkurl{data/}.

\FloatBarrier
\section{Training settings}\label{app:hyper}

\paragraph{Attempts.} The base learner makes eight attempts at each admitted task (temperature 0.7, top-$p$ 0.8, top-$k$
20; these three settings control how freely the model samples its next word) and the teacher four, each within 60 turns
and 16384 completion tokens.

\paragraph{Training-material experiment.} Low-rank adapters of rank 16 (scaling factor 32) on Qwen3-4B; learning rate
$2\times10^{-4}$ on a cosine schedule to a tenth of the peak after a warm-up of a twentieth of the steps; up to 800 optimiser steps of
four records; context of 16384 tokens; the checkpoint with the lowest validation loss, evaluated every 40 steps, stopping
after 4 evaluations without improvement; training seeds 1, 2 and 3.

\paragraph{Iterative protocol.} The same adapter shape. The first generation trains from the base at $2\times10^{-4}$;
later generations continue the champion's adapter at $10^{-4}$, stopping after 8 evaluations without improvement. Each
generation trains for about 1.9 passes (epochs) over its records. In every generation an action whose tool calls all
failed is kept as context and not trained on. The champion makes four attempts per new task; of its verified attempts at most one per task, chosen at random, enters
the replay. The release rule is that of
\cref{sec:m-loop}.

\paragraph{Lua.} A fresh adapter is trained from the base at every generation on the whole pool; predictions for the pool
are sampled four times at temperature 1.0. The examiner counts an answer as correct when it equals the interpreter's
output after whitespace is normalised. The learner's own release rule is a one-sided sign test on development programs
it generates for itself, with a retention margin. The launch script of a lineage is
\nolinkurl{reproduce/13-lua-lineage.sh}, with the machine's paths as variables. The settings are pinned in \texttt{prereg/rsi5-launch.md}.

%% file: sections/91-appendix-claims.tex
\FloatBarrier
\section{Claim registry and unit of analysis}\label{app:claims}

Every statement of result in the abstract, the results and the conclusion corresponds to a claim in
\nolinkurl{data/claims.json}. A claim names its question, its unit of analysis, the comparisons it rests on and the rule
that combines them; its status is computed from the outcomes of those comparisons by \nolinkurl{scripts/paper_claims.py},
and the build fails when the committed registry differs from what the evidence gives. \Cref{tab:claims} lists the claims.
Its column ``Reg.'' says whether the claim's test was registered before the result as a primary or as a secondary
analysis, or not at all. Four points help in reading it. A claim resting on an excluded comparison cannot be supported;
where a claim is stated per evaluation, an excluded evaluation is dropped and named, and the others decide. A claim marked
exercised states only that a mechanism ran and produced the counts named. Each row of the ledger (\cref{tab:ledger})
names the claims behind it in the same file and takes the weakest of their statuses. The claim about the shuffled-request control
asks whether the control improves on its base; it passes nothing, as designed, so the claim is inconclusive by
construction, and the control's role is in the claim it is the reference of (\texttt{vim-transfer-arms}). And the same
question can have two statuses in two runs: the first 4B generation's examination gain is supported in the earlier
iterative runs, where outcomes per task were recorded, and inconclusive in the iterative protocol, where they were not.
Claims about the earlier iterative runs (\cref{app:earlier}) are kept for the record and are not used in the main text.
The projection of \cref{sec:r-projection} is a description computed by the same pipeline, not a claim.

{\scriptsize
\setlength{\tabcolsep}{3pt}
\begin{longtable}{@{}>{\raggedright\arraybackslash}p{39.5mm}>{\raggedright\arraybackslash}p{41mm}>{\raggedright\arraybackslash}p{31.5mm}>{\raggedright\arraybackslash}p{14mm}>{\raggedright\arraybackslash}p{19mm}@{}}
\caption{The claim registry: each claim, its unit of analysis, how its test was registered before the result, and its
computed status.}\label{tab:claims}\\
\toprule
Claim & Question & Unit of analysis & Reg. & Status \\
\midrule
\endfirsthead
\toprule
Claim & Question & Unit of analysis & Reg. & Status \\
\midrule
\endhead
\inputrows{generated/claims-table.tex}
\bottomrule
\end{longtable}
}

\paragraph{How differences are computed.} A difference with a task-bootstrap interval (the training-material experiment,
the course exercises) is computed per task over the attempts both models have a verdict for, so it can differ in the
second decimal from the difference of the two rates printed beside it. A difference of the iterative protocol and of
its replications is the difference of the two rates shown, each over its own counted attempts; the evaluation stages
also recorded a difference over the attempt slots both models have a verdict for, which the release rule read and
which the paper does not print. The seeds of one replicated step share their tasks and one measurement of the champion,
so they are not tested together. For the course exercises the retention margin is applied after the fact.

\paragraph{The two scopes of the bound on lost attempts.} The bound on lost attempts was applied at two scopes, as each experiment's registration and analysis fixed it: over
all attempts of an evaluation run, candidate and base together, in the training-material experiment, and over each
model's attempts separately in the iterative protocol. The second is stricter. At the stricter scope the bases of the
all-verified and outcome-first conditions (\ev{arms/b/exam/base-uncounted-pct}\% and
\ev{arms/g/exam/base-uncounted-pct}\% lost) would be excluded as the first 8B generation's examination of the iterative
protocol is (\ev{B/8B/g1/exam/uncounted-pct}\%). The registered transfer claim would then rest on the one-per-task and
productive-actions conditions, whose bases are within the bound at either scope, and the outcome-first numbers would be
descriptive. In all of these stages the base passed none of its counted examination attempts.

\paragraph{Unit of every recorded test.} \Cref{tab:units} gives, for each evidence file, the unit over which its recorded
wins and losses were counted, established from the software that computed them. Where the unit is the attempt slot
(attempt $k$ of a task by one model paired with attempt $k$ by the other), the paper never uses the recorded $p$-value as
a test. It reports the exact range of the task-level test that the counts allow: for example, the
\ev{B/4B/g1/exam/cand-passes} passing examination attempts of the first 4B generation may belong to as few as
\ev{B/4B/g1/exam/task-wins-low} tasks or to \ev{B/4B/g1/exam/task-wins-high}, so the sign test over tasks lies between
$p=\ev{B/4B/g1/exam/task-p-low}$ and $\ev{B/4B/g1/exam/task-p-high}$, and for the second 8B generation's held-out tasks
between \ev{B/8B/g2/target/task-p-low} and \ev{B/8B/g2/target/task-p-high}.

\begin{table}[htbp]
\centering\scriptsize
\setlength{\tabcolsep}{3pt}
\begin{tabular}{@{}>{\raggedright\arraybackslash}p{58mm}r>{\raggedright\arraybackslash}p{28mm}>{\raggedright\arraybackslash}p{33mm}>{\raggedright\arraybackslash}p{24mm}@{}}
\toprule
Evidence file & Tests & Unit as computed & Unit the file states & Task-level test \\
\midrule
\inputrows{generated/unit-audit.tex}
\bottomrule
\end{tabular}
\caption{The unit of the recorded paired tests, per evidence file. ``Recorded'': the file's test is the task-level test.
``Bounds only'': the file holds counts over attempt slots, from which the task-level test is bounded. The last kind of row
sums seeds that share tasks and a champion measurement and is not a test.}\label{tab:units}
\end{table}

%% file: sections/92-appendix-results.tex
\FloatBarrier
\section{Complete results}\label{app:full}

\paragraph{Lua.} \Cref{tab:lua} gives the registered comparisons of the Lua experiment.

\begin{table}[htbp]
\centering\small
\begin{tabular}{lrrrl}
\toprule
Comparison & After & Before & Difference [95\% interval] & Outcome \\
\midrule
\inputrows{generated/lua-steps.tex}
\bottomrule
\end{tabular}
\caption{Lua examination accuracy: the mean share of items answered correctly over four lineages of Qwen3-4B, and
differences in percentage points with 95\% intervals from the examiner's bootstrap over program families and lineages
(unadjusted for the six comparisons, as registered). The last two rows split the comparison with the base into the
families never examined before (a registered criterion) and those that earlier examinations had used (descriptive).}
\label{tab:lua}
\end{table}

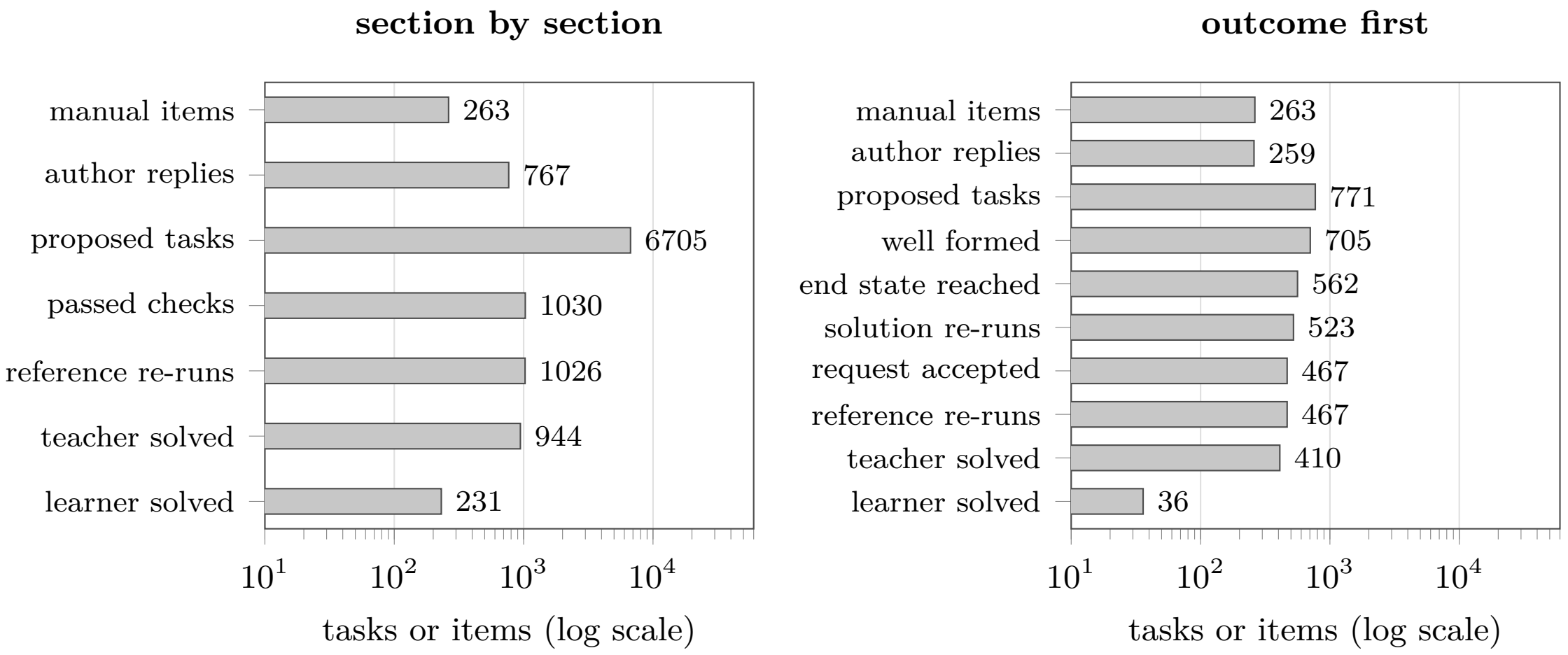
\begin{figure}[htbp]
\centering
\input{figures/funnel-body}
\caption{Automatic checks reject most generated material before training. Counts of items surviving each step of the
two authoring modes over the same manual, on a logarithmic scale; the author and teacher are Qwen3.8-27B-FP8 and the learner
Qwen3-4B. ``Author replies'': the outputs of the proposing stages, each holding several proposed tasks. ``Teacher solved'' and
``learner solved'' count admitted tasks with at least one verified attempt. The modes'
proposal units differ, so the two panels are not a comparison of efficiency.}\label{fig:vim-funnel}
\end{figure}

\paragraph{Authoring.} \Cref{fig:vim-funnel} shows how many items survive each step of the two authoring modes, and
\cref{tab:vim-attempts} the attempts of the base learner and of the teacher on the admitted tasks.

\begin{table}[htbp]
\centering\small
\begin{tabular}{llrrrr}
\toprule
Authoring mode & Model & Counted attempts & Pass rate & Tasks & Tasks solved \\
\midrule
\inputrows{generated/vim-transfer-attempts.tex}
\bottomrule
\end{tabular}
\caption{Attempts on the admitted tasks, by authoring mode and model (Qwen3-4B: the base learner, eight attempts per task;
Qwen3.8-27B-FP8: the teacher, four). A task is solved when at least one attempt was verified.}\label{tab:vim-attempts}
\end{table}

\paragraph{Iterative protocol.} \Cref{tab:iterative} gives every comparison of the iterative protocol: the candidate and
its reference in the same evaluation stage, the counts the stage recorded over attempt slots, and the range of the sign
test over tasks that those counts allow.

\begin{table}[htbp]
\centering\scriptsize
\setlength{\tabcolsep}{2pt}
\begin{tabular}{@{}>{\raggedright\arraybackslash}p{32mm}>{\raggedright\arraybackslash}p{25mm}rrrr>{\raggedright\arraybackslash}p{28mm}>{\raggedright\arraybackslash}p{22mm}@{}}
\toprule
Step & Tasks & Candidate & Reference & Diff. & Slots & Task-level $p$, range & Outcome \\
\midrule
\inputrows{generated/iterative-steps.tex}
\bottomrule
\end{tabular}
\caption{The iterative protocol, every comparison (one training run per step, \ev{const/attempts-per-task} attempts per
task). ``Candidate'' and ``Reference'': passes over counted attempts; ``Diff.'': the difference of those two rates in
percentage points (the evaluation stage also recorded a difference over the slots both models have a verdict for, which
the release rule read and which differs where attempts were lost); ``Slots'':
paired attempt slots won and lost as the evaluation stage recorded them, which are not independent; ``Task-level $p$,
range'': the smallest and largest value the one-sided sign test over tasks can take given those counts. A comparison is
supported only when the whole range is below $\alpha$ (for the examination, with Holm's correction over the generations of
a size); the 8B first-generation examination is excluded (attempts lost to infrastructure above the bound).}
\label{tab:iterative}
\end{table}

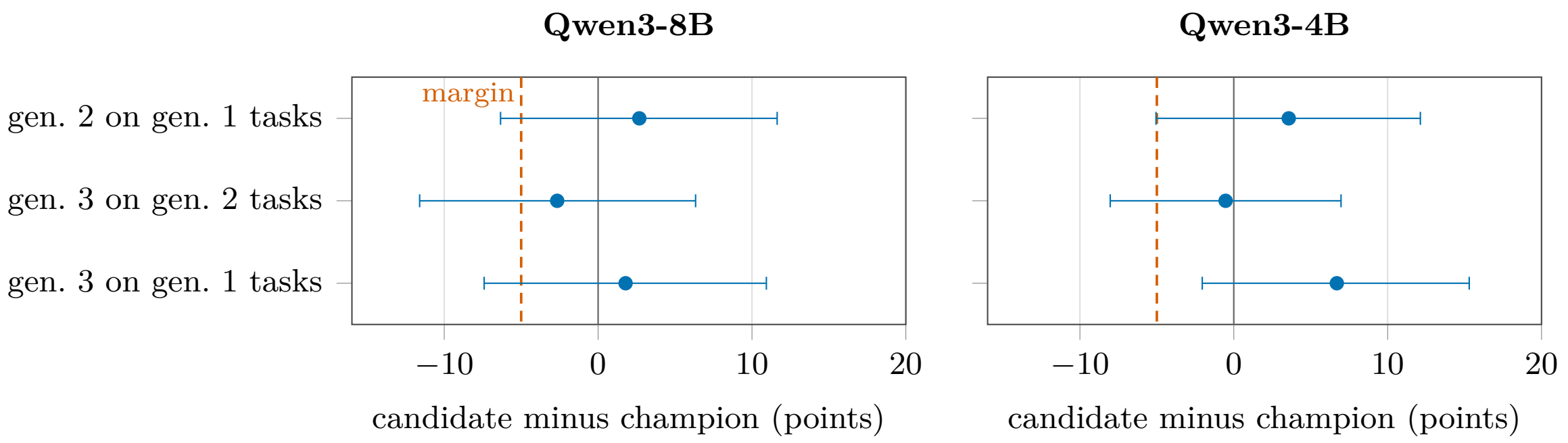
\begin{figure}[htbp]
\centering
\input{figures/retention-body}
\caption{Later candidates satisfied the retention rule on earlier generations' tasks; the data do not establish
non-inferiority. Each point is a later candidate's pass rate minus its champion's on an earlier generation's held-out
generated tasks, measured in the same stage (one training run each, \ev{const/attempts-per-task} attempts per task), with a
95\% interval that treats attempts as independent. Dashed: the retention margin of the release rule. The 4B third
generation is measured on the second generation's tasks although that generation was not accepted: the rule checks every
earlier generation's tasks.}\label{fig:retention}
\end{figure}

\paragraph{Retention across generations.} \Cref{fig:retention} shows each later candidate against its champion on every
earlier generation's held-out tasks.

\paragraph{Training-step replications.} \Cref{tab:replications} gives the passes of every training seed of every later
step beside the one champion measurement the seeds share.

\begin{table}[htbp]
\centering\scriptsize
\setlength{\tabcolsep}{2pt}
\begin{tabular}{@{}>{\raggedright\arraybackslash}p{25mm}>{\raggedright\arraybackslash}p{27mm}r>{\raggedright\arraybackslash}p{24mm}rrr>{\raggedright\arraybackslash}p{17mm}@{}}
\toprule
Step & Tasks & Seeds & Passes per seed & Champion & Above/below & Slots, summed & Outcome \\
\midrule
\inputrows{generated/replication-steps.tex}
\bottomrule
\end{tabular}
\caption{Training-step replications. The first seed of each step is the run of \cref{tab:iterative}. All seeds of a step
are compared with one champion measurement on the same tasks (``Above/below'': seeds that pass more or fewer attempts than
it), so the summed slot counts are not independent observations;
they are given as recorded and no test is made of them.}\label{tab:replications}
\end{table}

\begin{figure}[htbp]
\centering
\input{figures/seeds-body}
\caption{Across training seeds, later candidates were observed several points above their champion on held-out generated
tasks and within \ev{projection/seed-diff-max-pp} points of it, in either direction, on the examination. Each point is one training seed of one
later step: the candidate's pass rate minus that of the single champion measurement the step's seeds share, in percentage
points of attempts (A, B: the step's held-out generated tasks; C, D: the \ev{const/exams} examination tasks;
\ev{const/attempts-per-task} attempts per task). The seeds of a step have the same tasks, records and starting champion:
these are replications of one training step, not independent lineages, and no test is made over them.}
\label{fig:seeds}
\end{figure}
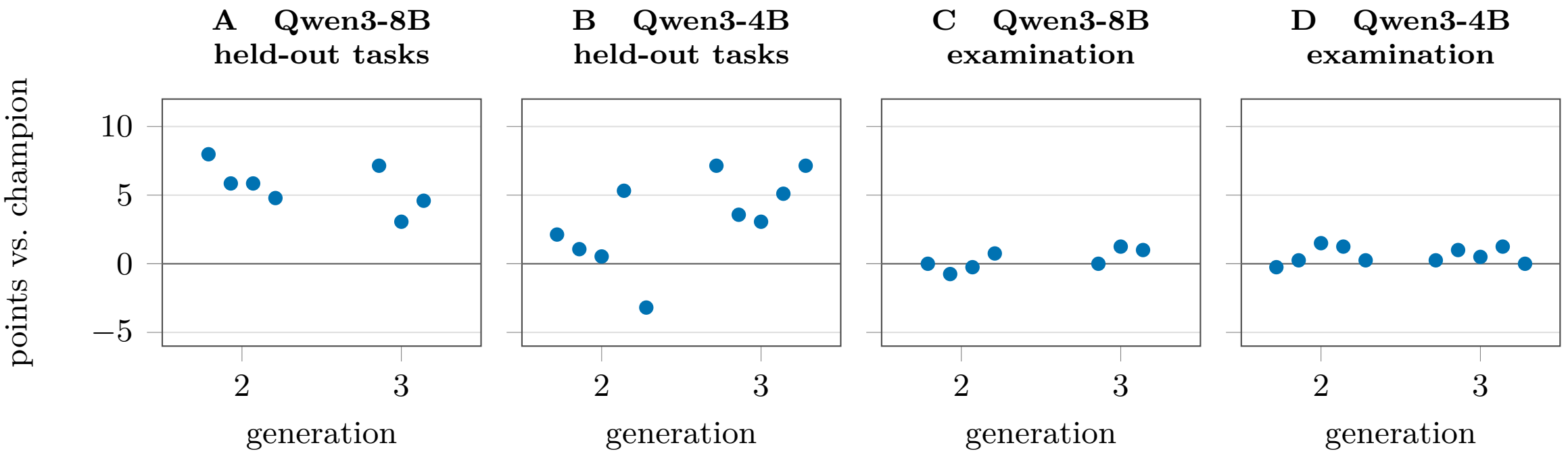

\paragraph{What the examination can resolve.} \Cref{tab:ratios,tab:projection} give the numbers behind
\cref{fig:resolution} (\cref{sec:r-projection}); the calculation was made after the results and is descriptive.

\begin{table}[!htb]
\centering\footnotesize
\begin{tabular}{@{}lrrrl@{}}
\toprule
First generation & Held-out & Examination & Ratio & Comparison \\
\midrule
\inputrows{generated/transfer-ratios.tex}
\bottomrule
\end{tabular}
\caption{The ratio of examination gain to held-out gain in every first generation measured (``Held-out'' and
``Examination'': the gain, the difference of the pass rates of candidate and base in percentage points of attempts;
``Comparison'': the status of the examination comparison). The prediction of \cref{sec:r-projection} uses
the two ratios on outcome-first tasks that do not come from an excluded evaluation, \ev{projection/yardstick/low-ratio}
and \ev{projection/yardstick/high-ratio}. \textsuperscript{a}From an evaluation excluded from claims; shown as a check
and not used.}\label{tab:ratios}
\end{table}

\begin{table}[!htb]
\centering\scriptsize
\setlength{\tabcolsep}{3pt}
\begin{tabular}{@{}lrlrllrl@{}}
\toprule
 & Held-out & \multicolumn{4}{c}{Examination difference} & & \\
\cmidrule(lr){3-6}
Step & diff. & predicted & mean & seeds, range & first seed, 95\% & Detectable & Tasks needed \\
\midrule
\inputrows{generated/transfer-projection.tex}
\bottomrule
\end{tabular}
\caption{What the first generation's ratio predicts for each later step, and what the examination detects (post hoc,
descriptive; percentage points of attempts). ``Held-out difference'' and ``mean'': averaged over the step's training
seeds, each against the champion measurement they share. ``Predicted'': the two ratios of \cref{tab:ratios} times the
held-out difference. The interval takes attempts as independent and ignores the pairing; a task-level interval may be
wider or narrower. ``Detectable'': the smallest difference a paired comparison detects at one-sided
$\alpha=\ev{resolution/alpha}$ with \ev{resolution/power}\% power, from the discordant attempt slots the first seed's
stage recorded (\ev{resolution/8B/g2/discordant} of \ev{resolution/8B/g2/pairs} in the first row). ``Tasks needed'':
tasks at \ev{const/attempts-per-task} attempts at which the higher and the lower prediction become detectable, rounded
to two digits. Both take slots as independent, so they are lower bounds.}\label{tab:projection}
\end{table}

\paragraph{Held-out generated tasks in the training-material experiment.} \Cref{tab:heldout} is the counterpart of
\cref{tab:vim-results} on the tenth of each authoring mode's tasks that no condition trained on (registered as a secondary
analysis).

\begin{table}[htbp]
\centering\scriptsize
\setlength{\tabcolsep}{2pt}
\begin{tabular}{@{}lrlrll@{}}
\toprule
Training material & Tasks & Passes (rate) & Base & vs base: won/lost, $p$ & Difference [95\%] \\
\midrule
\inputrows{generated/vim-transfer-t1.tex}
\bottomrule
\end{tabular}
\caption{The training-material conditions on held-out generated tasks (Qwen3-4B, three seeds pooled,
\ev{const/attempts-per-task} attempts per task and seed): one-sided exact sign tests over tasks and task-bootstrap 95\%
intervals of the pass-rate difference in percentage points, computed per task over the attempts both models have a
verdict for. The control's comparison is two-sided and exploratory: it is a supported regression.}\label{tab:heldout}
\end{table}

\paragraph{Comparisons between conditions.} \Cref{tab:secondary} gives the registered secondary comparisons between
conditions on the examination.

\begin{table}[!htb]
\centering\footnotesize
\begin{tabular}{lrll}
\toprule
Comparison & Tasks won/lost & $p$ & Pass-rate difference [95\% interval] \\
\midrule
\inputrows{generated/vim-transfer-secondary.tex}
\bottomrule
\end{tabular}
\caption{Comparisons between training-material conditions on the examination (secondary analyses): two-sided exact
sign tests over tasks and task-bootstrap 95\% intervals of the pass-rate difference in percentage
points.}\label{tab:secondary}
\end{table}

\paragraph{Course exercises.} \Cref{tab:retention} gives every candidate measured on the hand-written course exercises
(\cref{sec:r-authoring,sec:r-retention}).

\begin{table}[htbp]
\centering\footnotesize
\setlength{\tabcolsep}{3.5pt}
\begin{tabular}{@{}lrrll@{}}
\toprule
Candidate & Passes & Base & Tasks won/lost, $p$ & Difference [95\%] \\
\midrule
\inputrows{generated/retention.tex}
\bottomrule
\end{tabular}
\caption{The \ev{practice/g-s3/all/tasks} hand-written course exercises, each first-generation candidate against its base
(exploratory): passes over counted attempts, the one-sided exact sign test over tasks, and the task-bootstrap 95\%
interval of the pass-rate difference in percentage points, computed per task over the attempts both models have a
verdict for. The 8B candidates are three training seeds of the first
generation of the earlier iterative runs (\cref{app:earlier}). \textsuperscript{a}Excluded from claims: the base lost
more than \ev{const/uncounted-bound-pct}\% of its attempts.}\label{tab:retention}
\end{table}

\FloatBarrier
\section{Compute accounting}\label{app:compute}

\Cref{tab:compute} gives what the evidence records of the work and the accelerator time of each stage (a card is one accelerator). Accelerator time is
stated only where the evidence names the devices a stage ran on; where it does not, the row gives counts and no time.
\Cref{tab:compute-missing} lists the quantities a complete accounting would need that no evidence file holds. None of them
is estimated anywhere in the paper.

\begin{table}[htbp]
\centering\scriptsize
\setlength{\tabcolsep}{3pt}
\begin{tabular}{@{}>{\raggedright\arraybackslash}p{48mm}>{\raggedright\arraybackslash}p{28mm}>{\raggedright\arraybackslash}p{58mm}r@{}}
\toprule
Stage & Devices & Work recorded & Hours \\
\midrule
\inputrows{generated/compute-accounting.tex}
\bottomrule
\end{tabular}
\caption{Compute accounting from the recorded evidence; ``Hours'' are accelerator-hours. Lua generations: proposing and training over the four lineages as
the runs recorded them, without the examiner's evaluation. Vim training: the fifteen adapters of the training-material
experiment, three seeds per condition. Physical attempts include the repetitions of an attempt whose request was lost;
counted attempts are those with a verdict.}\label{tab:compute}
\end{table}

\begin{table}[htbp]
\centering\scriptsize
\setlength{\tabcolsep}{3pt}
\begin{tabular}{@{}>{\raggedright\arraybackslash}p{30mm}>{\raggedright\arraybackslash}p{50mm}>{\raggedright\arraybackslash}p{66mm}@{}}
\toprule
Quantity & Scope & Why it is not in the evidence \\
\midrule
\inputrows{generated/compute-missing.tex}
\bottomrule
\end{tabular}
\caption{Quantities that the evidence does not record.}\label{tab:compute-missing}
\end{table}

%% file: figures/funnel-body.tex
\begin{tikzpicture}
\begin{groupplot}[group style={group size=2 by 1, horizontal sep=30mm}, paper, width=62mm, height=58mm,
    xmode=log, xmin=10, xmax=60000, ytick=data, y tick label style={font=\scriptsize}, xlabel={tasks or items (log scale)},
    every axis plot/.append style={xbar, /pgf/bar width=2.4mm}, xmajorgrids, ymajorgrids=false,
    nodes near coords, nodes near coords style={font=\scriptsize, anchor=west}, nodes near coords align={horizontal},
    point meta=explicit symbolic, enlarge y limits=0.07]
\nextgroupplot[title={\strut section by section}, yticklabels from table={generated/vim-transfer-funnel-forward.csv}{step}]
  \addplot[draw=black!70, fill=black!22] table[x=count, y=y, meta=count] {generated/vim-transfer-funnel-forward.csv};
\nextgroupplot[title={\strut outcome first}, yticklabels from table={generated/vim-transfer-funnel-backward.csv}{step}]
  \addplot[draw=black!70, fill=black!22] table[x=count, y=y, meta=count] {generated/vim-transfer-funnel-backward.csv};
\end{groupplot}
\end{tikzpicture}

%% file: figures/retention-body.tex
\begin{tikzpicture}
\begin{groupplot}[group style={group size=2 by 1, horizontal sep=8mm}, paper, width=70mm, height=40mm, y dir=reverse,
    ymin=0.5, ymax=3.5, ytick={1,2,3}, ymajorgrids=false, xmajorgrids, xmin=-16, xmax=20,
    xlabel={candidate minus champion (points)}]
\nextgroupplot[title={Qwen3-8B}, yticklabels={{gen.~2 on gen.~1 tasks}, {gen.~3 on gen.~2 tasks}, {gen.~3 on gen.~1 tasks}}]
  \draw[black!60] (axis cs:0,0.5) -- (axis cs:0,3.5);
  \draw[okVermillion, densely dashed, line width=0.7pt] (axis cs:-\ev{const/retention-margin-pp},0.5) --
    (axis cs:-\ev{const/retention-margin-pp},3.5);
  \node[note, anchor=north east, text=okVermillion] at (axis cs:-\ev{const/retention-margin-pp},0.5) {margin};
  \addplot[candidate, discard if not={size}{8B}, error bars/.cd, x dir=both, x explicit]
    table[y=y, x=diff, x error plus expr=\thisrow{hi}-\thisrow{diff}, x error minus expr=\thisrow{diff}-\thisrow{lo}]
    {generated/retention-steps.csv};
\nextgroupplot[title={Qwen3-4B}, yticklabels={}]
  \draw[black!60] (axis cs:0,0.5) -- (axis cs:0,3.5);
  \draw[okVermillion, densely dashed, line width=0.7pt] (axis cs:-\ev{const/retention-margin-pp},0.5) --
    (axis cs:-\ev{const/retention-margin-pp},3.5);
  \addplot[candidate, discard if not={size}{4B}, error bars/.cd, x dir=both, x explicit]
    table[y=y, x=diff, x error plus expr=\thisrow{hi}-\thisrow{diff}, x error minus expr=\thisrow{diff}-\thisrow{lo}]
    {generated/retention-steps.csv};
\end{groupplot}
\end{tikzpicture}

%% file: figures/seeds-body.tex
\begin{tikzpicture}
\newcommand{\seedpoints}[2]{%
  \addplot[candidate, discard if not={size}{#1}, discard if not={kind}{#2}] table[x=x, y=diff]
    {generated/replication-seeds.csv};
}
\begin{groupplot}[group style={group size=4 by 1, horizontal sep=4mm}, paper, width=47mm, height=40mm,
    xmin=1.5, xmax=3.5, xtick={2,3}, xmajorgrids=false, ymin=-6, ymax=12, xlabel={generation},
    extra y ticks={0}, extra y tick style={grid style={black!60}, tick label style={opacity=0}},
    title style={font=\scriptsize\bfseries, align=center}]
\nextgroupplot[title={A\quad Qwen3-8B\\held-out tasks}, ylabel={points vs.\ champion}]
  \seedpoints{8B}{target}
\nextgroupplot[title={B\quad Qwen3-4B\\held-out tasks}, yticklabels={}]
  \seedpoints{4B}{target}
\nextgroupplot[title={C\quad Qwen3-8B\\examination}, yticklabels={}]
  \seedpoints{8B}{exam}
\nextgroupplot[title={D\quad Qwen3-4B\\examination}, yticklabels={}]
  \seedpoints{4B}{exam}
\end{groupplot}
\end{tikzpicture}

%% file: sections/93-appendix-records.tex
\FloatBarrier
\section{Registrations and internal names}\label{app:registrations}

\begin{sloppypar}\hbadness=4000
\paragraph{Lua.} The Lua experiment is registered as RSI5 (\texttt{prereg/rsi5.md}, launch record
\texttt{prereg/rsi5-launch.md}). Its examination is the examiner's exam~G. Its three criteria are evaluated by
\texttt{scripts/rsi5\_criteria.py} from the examiner's verdict: G1, every generation better than its predecessor
(\ev{lua/criterion/G1}); G2, the fourth generation at least five points above the base (\ev{lua/criterion/G2}); G3, a gain
on the families never examined before (\ev{lua/criterion/G3}).
\end{sloppypar}

\paragraph{Vim.} The Vim experiments are registered in the roadmap file
\nolinkurl{.agents/roadmap/source-to-transfer.md} of the learning controller's repository: the training-material
experiment at commit \texttt{c319ff4c3}, before any condition trained, with the outcome-first condition added at
\texttt{66277f180}, also before any condition trained; the iterative protocol per model size as the primary estimand at
\texttt{7ddad366e}, before any second-generation examination result; and the training-step replications before any result
of a replication. The registration letters its conditions, and the evidence files keep the letters
(\texttt{arm-<letter>-s<seed>}): A the base, B all verified, C one per task, D productive actions, F the shuffled-request
control, G outcome first, S learner only. The held-out generated tasks are its measurement T1 and the examination T2 and T4;
\texttt{unseen} in a file name also means the held-out generated tasks.
In the evidence files the iterative protocol of the main text is the \texttt{rerun} record of each model size, and the
earlier iterative runs of \cref{app:earlier} are the \texttt{loop} record. The evidence files, the claim identifiers and
the reproduction materials also keep earlier names: \texttt{exam} for an examination task, \texttt{practice} for the
course exercises, \texttt{student} for the learner, \texttt{arm} for a condition, \texttt{forward} or
\texttt{transfer.yaml} for section-by-section authoring and \texttt{backward} for outcome-first authoring.

\paragraph{The long-term objective as an operational definition.} After the first Vim generation had been examined, and
before any verdict against it, the following definition was written down so that claims would be scoped by a fixed
statement. A loop \emph{improves itself continuously} when, run from a recipe with no human-written training tasks, no human
labels and no human intervention between generations, it shows across at least three successive generations: (a) each
generation beats its predecessor on examinations that no generation trained on, under a registered paired test; (b) no
generation regresses on skills earlier generations had, by a registered rule; (c) the gains hold across seeds and beat a
shuffled-request control; (d) the loop restarts end to end from the recipe and frozen binaries with no manual step. A part
that was not measured is unmet. This is the long-term objective of \cref{sec:introduction}, not the criterion of success of
this paper. Scored against it: part (a) holds for the first two Lua generations and is not established for any Vim step
after the first; part (b) was met as a release rule and not established as a test; part (c) holds for the first Vim
generation of the training-material experiment; part (d) was not measured. The definition is not met.

\FloatBarrier
\section{Deviations}\label{app:deviations}

\paragraph{Lua.} Four deviations, all made before any candidate was examined, are recorded beside the registration: the
planning target of the learner's own release rule, the proposer's state moved to memory, one lineage's last generation
moved to another accelerator, and the fourth generation's proposal budget raised. None changes a proposal, a training record
or an acceptance decision.

\paragraph{Vim.} In the order they were made, one line each.
\begin{enumerate}
\item The outcome-first condition was added, and the comparison family grew to four, before any condition trained.
\item The source is the text of the manual as Vim ships it, not a scan of its printed form as first planned.
\item The authoring run was restarted once after a hardware failure. Its authoring, review, task and admission stages were
  reused and every attempt was collected again; no attempt of the first run is used.
\item The model servers ran without speculative decoding; the sampled distribution is unchanged.
\item The operational definition of \cref{app:registrations} was written after the first generation was examined and before
  any verdict against it; it adds no registered comparison.
\item The learner-only condition was added after the registered family had been decided, and is decided outside it. Two of
  its three evaluation stages lost more attempts to infrastructure than the bound allows and were not repeated; its held-out
  tasks were not evaluated. The result is excluded from claims.
\item The iterative protocol per model size became the primary estimand before any second-generation examination
  result and after the first-generation results of the training-material experiment.
\item The registration named a second teacher model from the second generation on; it was not used.
\item Two generations were first run under an earlier iterative protocol (\cref{app:earlier}). The iterative protocol of the
  main text (\cref{sec:m-loop}) then replaced it, and three generations were run under it from the authoring stage. The
  examination and the registered tests are the same in both, and the earlier protocol's examination results were known
  when the present one was written.
\item In the iterative protocol, the 8B first-generation examination left \ev{B/8B/g1/exam/base-uncounted} of the base's
  \ev{B/8B/g1/exam/base-scheduled} attempts uncounted, above the bound. In each, the base model repeated itself inside a
  tool call until the agent ended the attempt; the attempt was recorded as a refused request, retried and left uncounted.
  One counting rule was kept for all generations, and the comparison is excluded as registered.
\item The evaluation stages of the iterative protocol and of its replications recorded their paired tests over attempt
  slots, where the registration specifies tasks; the paper reports the bounds of the task-level test
  (\cref{sec:d-statistics,app:claims}).
\item The 4B run of the iterative protocol was paused during its first generation and resumed from its recorded state
  after the 8B run.
\item The routing of requests across accelerators was changed during the runs. It changes throughput, not what is measured.
\item One interrupted stage left an attempt's sandbox behind, which was removed by hand: a manual step.
\item Additional teacher attempts were made on tasks the teacher had not solved in the second and third generations, for the
  published dataset only; nothing was trained on them.
\item Registered analyses of the iterative protocol that are not reported, because they need outcomes per task: the trend
  of examination passes over generations (a permutation over examination tasks) and the step test summed over the two
  model sizes. An earlier registration named a trend test over attempts (Cochran--Armitage) and each champion against the
  base; the first was replaced by the permutation test before the protocol's runs, and later champions were measured
  against their predecessor only. The third generation was measured against the first only where the first was still the
  champion (4B).
\item The registration of the replications names a test on wins and losses summed over the seeds of a step. The sums are
  given in \cref{tab:replications} and are not used as a test, because the seeds share their tasks and one measurement of
  the champion.
\item The measurement on hand-written course exercises is named in the registration of the iterative protocol as a
  retention check against the base, without a test or a margin. It was made for first-generation candidates only, three
  of the six from the training-material experiment (seed~\ev{seeds/g/lowest-validation-loss} of outcome first, and
  seed~\ev{seeds/c/most-exam-passes} of productive actions and of one per task; no rule for the choice was registered,
  and each is the seed of its condition with the most examination passes); the margin and the bound on lost attempts were applied afterwards, and its
  reading as transfer (\cref{sec:r-authoring}) is post hoc. The paper labels it exploratory.
\item The projection of examination gains from the first generation's ratio and the calculation of what the examination
  detects (\cref{sec:r-projection}) are not registered.
\end{enumerate}

\FloatBarrier
\section{Earlier iterative runs}\label{app:earlier}

This section records, for completeness, the results of the earlier iterative protocol. They are not part of the paper's
argument, except that the three 8B adapters of their first generation are the 8B candidates measured on the course
exercises in \cref{sec:r-authoring}, and they are not a first part of the sequence shown in \cref{sec:r-loop}. Under the earlier protocol each generation
trained a fresh adapter from the base on the verified attempts of teacher and learner together, with earlier generations'
records as replay, under a fixed number of training steps.

\emph{First generation.} The 4B first generation is seed~\ev{seeds/g/lowest-validation-loss} of the outcome-first
condition of \cref{sec:r-authoring}: the seed with the lowest validation loss, as the registration names it, and also
the one with the most examination passes, which were known when it was registered. It
passed \ev{A/4B/g1/exam/cand-passes} of \ev{A/4B/g1/exam/cand-counted} examination attempts against its base's
\ev{A/4B/g1/exam/base-passes} of \ev{A/4B/g1/exam/base-counted}, winning \ev{A/4B/g1/exam/wins} tasks and losing
\ev{A/4B/g1/exam/losses} (one-sided sign test over tasks, $p=\ev{A/4B/g1/exam/task-p}$): \ev{A/4B/g1/exam/outcome}. The 8B
first generation passed \ev{A/8B/g1/exam/cand-passes}, \ev{A/8B/g1.seed1/exam/cand-passes} and
\ev{A/8B/g1.seed2/exam/cand-passes} of \ev{A/8B/g1/exam/cand-counted} attempts in three training seeds against bases that
passed none. In all three the base lost more attempts to infrastructure than the bound allows
(\ev{A/8B/g1/exam/uncounted}, \ev{A/8B/g1.seed1/exam/uncounted} and \ev{A/8B/g1.seed2/exam/uncounted} of
\ev{A/8B/g1/exam/base-scheduled}), so these comparisons are excluded.

\emph{Second generation.} Against its first-generation champion in the same stage, the 4B second generation passed
\ev{A/4B/g2/exam/cand-passes} examination attempts against \ev{A/4B/g2/exam/base-passes}
(\ev{A/4B/g2/exam/wins} tasks won, \ev{A/4B/g2/exam/losses} lost, $p=\ev{A/4B/g2/exam/task-p}$), and the 8B second
generation \ev{A/8B/g2/exam/cand-passes} against \ev{A/8B/g2/exam/base-passes} (\ev{A/8B/g2/exam/wins} won,
\ev{A/8B/g2/exam/losses} lost, $p=\ev{A/8B/g2/exam/task-p}$): both \ev{A/8B/g2/exam/outcome}. On the first generation's
held-out tasks the second generations passed \ev{A/4B/g2/retention-g1/cand-passes} against
\ev{A/4B/g2/retention-g1/base-passes} at 4B and \ev{A/8B/g2/retention-g1/cand-passes} against
\ev{A/8B/g2/retention-g1/base-passes} at 8B; against a margin of \ev{const/retention-margin-pp} points, applied post hoc,
retention is supported. Neither second generation was accepted by that protocol's release rule.

\FloatBarrier
\section{Other measurements}\label{app:other}

\paragraph{Lua.} The learner's own release rule, on its development programs, accepted every generation, including the
third and fourth, whose steps on the examination are inconclusive (\cref{tab:lua}): a loop's own rule kept accepting
candidates whose gain the examination did not confirm. A
descriptive examination of the same candidates with the reasoning mode switched on
(\nolinkurl{data/rsi5-groups-G-thinking.json}) was inconclusive.

\paragraph{Vim.} Between leaving unproductive actions untrained and training on them no difference was detected
(interval $\ev{arms/d-vs-b/exam/ci-lo-pp}$ to $\ev{arms/d-vs-b/exam/ci-hi-pp}$ points), and keeping one solution per task
passed fewer examination attempts than keeping all verified solutions, from fewer records (the interval of the difference
excludes zero; two-sided sign test $p=\ev{arms/c-vs-b/exam/p-two-sided}$; \cref{app:full}). On the held-out generated tasks the
shuffled-request control passed \ev{arms/f-vs-base/target/cand-passes} of \ev{arms/f-vs-base/target/cand-counted} attempts
against its base's \ev{arms/f-vs-base/target/base-passes} (\ev{arms/f-vs-base/target/wins} tasks won,
\ev{arms/f-vs-base/target/losses} lost; exploratory, outside the registration).

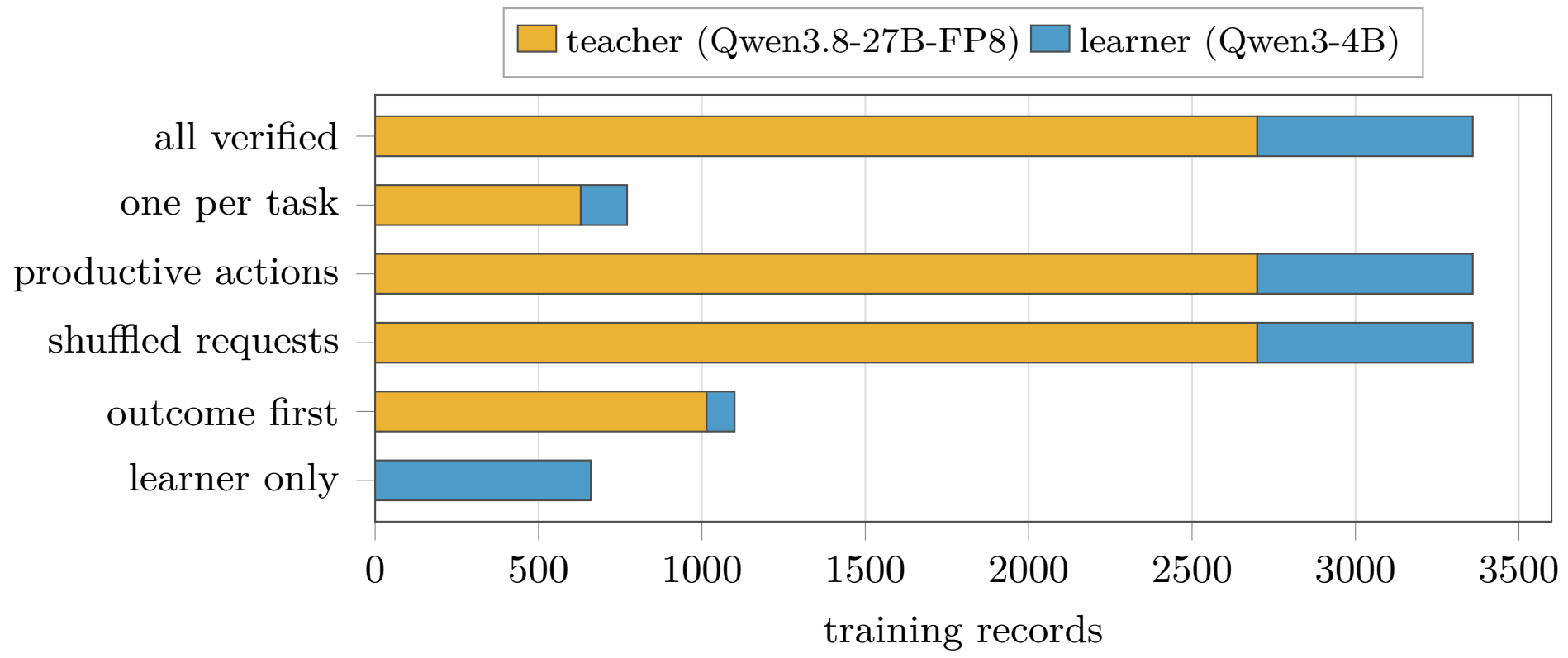
\begin{figure}[htbp]
\centering
\input{figures/records-body}
\caption{Most training records come from the teacher. Training records of each condition of the training-material
experiment by the model that produced them (teacher: Qwen3.8-27B-FP8; learner: Qwen3-4B). The learner-only condition's
evaluation is excluded from claims (\cref{sec:r-supervision}).}\label{fig:records}
\end{figure}

\paragraph{Who produced the training records.} \Cref{fig:records} shows, for each condition of the training-material
experiment, how many of its training records came from the teacher and how many from the learner.

\begin{figure}[htbp]
\centering
\input{figures/claims-body}
\caption{A model's declaration that it is done is not evidence that it is. Of the attempts that ended with the model
declaring the task complete, the share the environment's check rejected, for the base (grey squares) and for the first
generation of the iterative protocol (blue), on that generation's held-out generated tasks and on the examination; Wilson
95\% intervals over such attempts, which treat attempts as independent. Descriptive: the 8B examination evaluation is
the one excluded from claims for its lost base attempts (\cref{sec:r-loop}). The vertical axis starts at 50\%.}\label{fig:claims}
\end{figure}
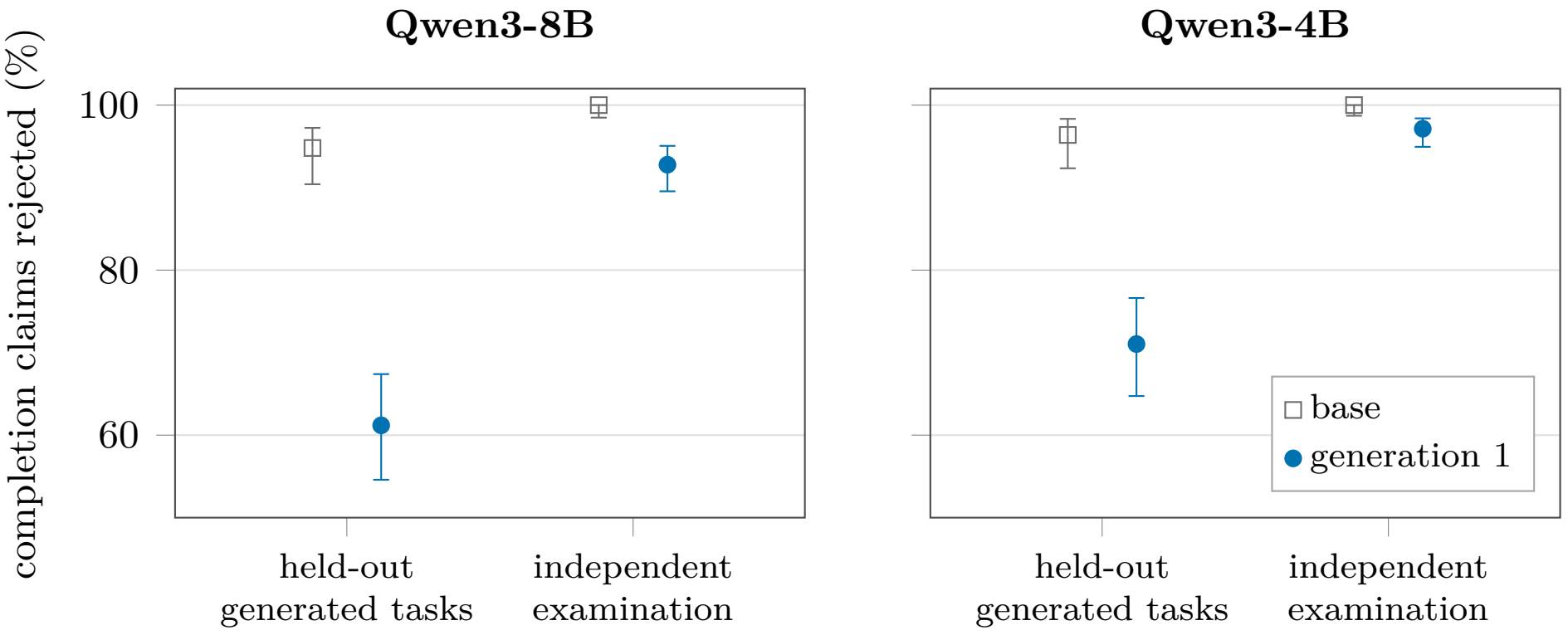

\paragraph{What a model's own report is worth.} One behavioural measurement is recorded for the first generation of the
iterative protocol (\cref{fig:claims}). An attempt normally ends with the model declaring the task done. For the base
models almost every such declaration was rejected by the environment's check. After one generation the share rejected fell
markedly on held-out generated tasks and only slightly on the examination. The trained models are still wrong most of the
times they say they are finished; the environment, not the model's report, is what makes a record trustworthy.

%% file: figures/records-body.tex
\begin{tikzpicture}
\begin{axis}[paper, width=110mm, height=50mm, ytick={1,...,6}, y dir=reverse, ymin=0.4, ymax=6.6, ymajorgrids=false,
    xmajorgrids, xbar stacked, /pgf/bar width=3.2mm, xmin=0, xmax=3600,
    yticklabels from table={generated/vim-records.csv}{arm}, xlabel={training records},
    legend style={at={(0.5,1.04)}, anchor=south}, legend columns=2]
  \addplot[draw=black!70, fill=okOrange!80] table[x=teacher, y=y] {generated/vim-records.csv};
  \addlegendentry{teacher (Qwen3.8-27B-FP8)}
  \addplot[draw=black!70, fill=okBlue!70] table[x=student, y=y] {generated/vim-records.csv};
  \addlegendentry{learner (Qwen3-4B)}
\end{axis}
\end{tikzpicture}

%% file: figures/claims-body.tex
\begin{tikzpicture}
\newcommand{\claimpoints}[1]{%
  \addplot[reference, discard if not={size}{#1}, discard if not={side}{baseline}, error bars/.cd, y dir=both, y explicit]
    table[x expr=\thisrow{y}-0.12, y=pct, y error plus expr=\thisrow{hi}-\thisrow{pct},
      y error minus expr=\thisrow{pct}-\thisrow{lo}] {generated/completion-claims.csv};
  \addplot[candidate, discard if not={size}{#1}, discard if not={side}{candidate}, error bars/.cd, y dir=both, y explicit]
    table[x expr=\thisrow{y}+0.12, y=pct, y error plus expr=\thisrow{hi}-\thisrow{pct},
      y error minus expr=\thisrow{pct}-\thisrow{lo}] {generated/completion-claims.csv};
}
\begin{groupplot}[group style={group size=2 by 1, horizontal sep=10mm}, paper, width=66mm, height=50mm, xmin=0.4,
    xmax=2.6, ymin=50, ymax=102, xtick={1,2}, xticklabels={held-out\\generated tasks, independent\\examination},
    x tick label style={font=\scriptsize, align=center}, xmajorgrids=false]
\nextgroupplot[title={Qwen3-8B}, ylabel={completion claims rejected (\%)}]
  \claimpoints{8b}
\nextgroupplot[title={Qwen3-4B}, yticklabels={}, legend style={at={(0.96,0.06)}, anchor=south east}]
  \claimpoints{4b}
  \legend{base, generation 1}
\end{groupplot}
\end{tikzpicture}

%% file: appendix-reproduction.tex

\FloatBarrier
\section{Reproduction procedures}\label{app:repro-vim}

\begin{sloppypar}
The commands that repeat the experiments from the listed artefacts are scripts in the directory
\nolinkurl{reproduce/} beside the paper's sources (\nolinkurl{papers/generalized-continuous-learning/reproduce/} of the
learning controller's repository). Scripts \texttt{01} to \texttt{10} repeat the Vim training-material experiment
(\cref{sec:r-authoring,sec:r-supervision}) from the recorded tasks; \texttt{11} and \texttt{12} take the reproduction
materials and run the iterative protocol from their recipes; \texttt{13} runs one lineage of the Lua experiment. Each
script states at its top what it does and needs; \nolinkurl{reproduce/env.sh} holds the pinned revisions and every
path. The scripts keep the names of the registration and of the evidence files (\cref{app:registrations}). They only
configure and call the learning controller, which orchestrates every stage.
\end{sloppypar}

\Cref{tab:levels} places this procedure among the four levels of reproduction of \cref{sec:reproducibility}.

\begin{table}[htbp]
\centering\footnotesize
\setlength{\tabcolsep}{3pt}
\begin{tabular}{@{}>{\raggedright\arraybackslash}p{23mm}>{\raggedright\arraybackslash}p{44mm}>{\raggedright\arraybackslash}p{43mm}>{\raggedright\arraybackslash}p{43mm}@{}}
\toprule
Level & Needs & Reaches & Exercised \\
\midrule
1. The publication: every table, figure, statistic and claim status from the committed evidence &
  the public source repository; Python~3 and a \LaTeX{} distribution on a CPU; minutes &
  every research question, as arithmetic on recorded aggregates: it checks the analysis, not the measurements &
  yes, at every build; the build fails on a broken invariant of the evidence or a claim registry that differs from what
  the evidence gives \\
2. Evaluation: examine the recorded adapters again &
  the adapters (private); the Vim course (public); one accelerator that holds a 4B model &
  RQ2 and RQ3 only. The adapters of RQ4 were not retained; the Lua adapters are private and the Lua examination stays
  with the examiner &
  no: no script here evaluates a recorded adapter \\
3. Training: train again on the recorded training sets &
  the exact training sets (private; a public copy of the trajectories, without held-out tasks, exists) and the authored
  tasks (private); one accelerator; a recorded training took \ev{compute/arms/c/train-minutes-per-seed} to
  \ev{compute/arms/b/train-minutes-per-seed} minutes &
  RQ2 and RQ3 &
  in part: four training steps on one recorded training set, on which the base's validation loss equalled the one the
  recorded run logged \\
4. Full replication: author tasks and run every stage &
  the reproduction materials (public; \nolinkurl{\MaterialsName}, commit \texttt{\MaterialsCommitShort}), the Vim
  manual and a model that can author; for the Lua cycle the controller's repository and a Lua~5.1.5 interpreter; the
  recorded Vim run used four nodes of four accelerators; training its fifteen adapters took
  \ev{compute/arms/total/train-gpu-hours} accelerator-hours, and the time of its attempts and evaluations was not
  recorded &
  every research question, as a new sample: authored tasks, attempts and examination items are drawn again
  (\cref{app:repro-vim-limits}) &
  build, data, task and recipe steps and a small end-to-end run of the training-material experiment: yes; the materials'
  recipes: validated, not run; a full experiment, the iterative protocol or a Lua lineage from these scripts: no \\
\bottomrule
\end{tabular}
\caption{The four levels of reproduction: what each needs, which research questions it reaches, and what was exercised
for this paper. Hardware and time are given where they are known.}\label{tab:levels}
\end{table}

\subsection{The steps, what each needs, and what was run}\label{app:repro-vim-scope}

\begin{center}\footnotesize
\begin{tabular}{@{}>{\raggedright\arraybackslash}p{36mm}>{\raggedright\arraybackslash}p{46mm}>{\raggedright\arraybackslash}p{31mm}>{\raggedright\arraybackslash}p{34mm}@{}}
\toprule
Script & What it does & Needs & Run for this paper \\
\midrule
\texttt{01-build.sh}, \texttt{02-build-brain.sh} & clone the three code repositories at pinned commits and build them &
  public sources; a Rust toolchain; for Brain, CUDA & as pasted commands \\
\texttt{03-data.sh} & fetch the five data repositories at pinned revisions & read access to three private repositories &
  as pasted commands, and again as this script \\
\texttt{04-tasks.sh} & lay out the task repository and check every task file against the digests the recorded runs pinned
  & the task and evidence datasets (private) & as pasted commands, and again as this script \\
\texttt{05-recipes.sh}, \texttt{gen-recipe.sh} & write the evaluation plans and generate and validate the 19 recipes of the training-material experiment (two
  authoring modes, their two training sets, fifteen adapters) &
  the built controller & as pasted commands, and again as these scripts \\
\texttt{06-serve.sh} & pull the two models and start the model servers & accelerators & only its two model downloads \\
\texttt{07-smoke.sh} & a few tasks through every stage, in minutes & all of the above & as pasted commands \\
\texttt{08-full.sh} & the experiment: both authoring modes' attempts, five training sets, fifteen adapters and their
  evaluations & all of the above & no; its recipes were validated \\
\texttt{09-train-published.sh} & train one condition on its recorded training set & the evidence dataset (private) &
  in the smoke's four-step form \\
\texttt{10-analysis.sh} & the registered decision, the secondary comparisons, and the paper's tables and figures &
  runs to analyse; a \LaTeX{} distribution & as pasted commands, over the smoke's runs \\
\texttt{11-materials.sh} & clone the reproduction materials at the pinned commit, place the manual and the course, and
  validate every recipe that needs no earlier run & the built controller; the course; the Vim manual as Markdown &
  as this script, from a local copy of the materials at the pinned commit: 26 of its 28 recipes valid \\
\texttt{12-iterative.sh} & the iterative protocol: the materials' three recipes, one generation each, in order &
  \texttt{11}; model servers with the author and the learner & no; tested with a stand-in for the materials' run
  script \\
\texttt{13-lua-lineage.sh} & one Lua lineage: four generations of the learning cycle with the registered settings &
  the \texttt{lineage} program built from the controller's repository; a Lua~5.1.5 interpreter; three pages of the Lua
  manual; one accelerator & no; tested with a stand-in for the program, its options checked against the program's
  source \\
\bottomrule
\end{tabular}
\end{center}

\begin{sloppypar}
``As pasted commands'' means that the commands now in the script were run, in order, in one shell session on one node of
the cluster that ran the experiment, before they were moved into these files. After the move every script was checked for
syntax, \nolinkurl{gen-recipe.sh} was confirmed to write byte-identical recipes to the generator that had been run, and
the three steps marked ``again as this script'' were run from the files on a machine without accelerators: the data
step fetched the five repositories in a few minutes, the task step found every pinned file identical, and the recipe step
validated all 19 recipes, with a controller built from a later commit than the pinned one. The other scripts were not run
again in their present form. They differ from the pasted commands in that each script stops at its first error, archives
are unpacked without restoring file owners, the number of cards is a parameter of \texttt{06} and the number of
concurrent attempts a parameter of the recipes, and \texttt{10} takes the paper's revision as a parameter.
\end{sloppypar}

\subsection{Prerequisites and hardware}\label{app:repro-vim-prereq}

\begin{sloppypar}
\textbf{Software.} Linux (the run: Ubuntu 26.04 on aarch64, in a container on a cluster node; x86-64 was not tried) with
unprivileged user namespaces or a Docker daemon (the sandbox of each attempt; \texttt{unshare -Ur true} must succeed); git
with git-lfs; a Rust toolchain (the run: rustc 1.100.0); Python~3.12 or later with \texttt{venv}; \texttt{zstd}; the
\texttt{hf} command-line client, logged in with read access to the private \texttt{swedishembedded} repositories; for the
model servers the CUDA toolkit~13 and an NVIDIA driver; network access to github.com, huggingface.co, the Debian snapshot
archive and the Docker registry (each task image is built from its pinned \texttt{debian:trixie-slim} digest and a fixed
snapshot of the archive). Vim is not needed on the host: every Vim the experiment runs is the task image's (Debian's
9.1.1230).
\end{sloppypar}

\begin{sloppypar}
\textbf{Hardware, disk and time.} The recorded run used four nodes of four NVIDIA GH200 accelerators (96~GB of device
memory each, 288 Arm cores per node): one model server per card, each node's attempts and evaluations through a router
over its cards, each adapter's training on one card. The model weights take 7.5~GiB (Qwen3-4B) and 28.8~GiB
(Qwen3.8-27B-FP8) on disk and the five data repositories about 4.3~GiB (measured when the data step was run again). A machine with fewer or smaller cards was not
tried, so no minimum is known. Training the fifteen adapters took \ev{compute/arms/total/train-gpu-hours}
accelerator-hours in the recorded run (\cref{tab:compute}); the time of its attempts and evaluations was not recorded, so
the time of the full experiment cannot be stated.
\end{sloppypar}

\subsection{Revisions}\label{app:repro-vim-build}

\begin{sloppypar}
\nolinkurl{reproduce/env.sh} pins the revisions the procedure was tested with, which are not in every case those of
\cref{tab:artifacts,tab:builds}. The controller is built at a later commit than the controllers of the recorded runs
(\cref{tab:builds}), one that holds every option the conditions used. Brain is built at the public commit that
corresponds to the recorded servers' (\cref{app:artifacts}), and the agent at the commit every attempt ran. The course and
the trajectory dataset are at the revisions of \cref{tab:artifacts}; the authored tasks, the evidence and the adapters
are at earlier revisions of the same repositories, to which the revisions of \cref{tab:artifacts} add the tasks of the
later generations, the records of the later experiments and further adapters.
\end{sloppypar}

\subsection{What to compare}\label{app:repro-vim-full}

\begin{sloppypar}
\label{app:repro-vim-verify}\texttt{04-tasks.sh} prints, for each of the three task trees, how many files are identical to
the ones the recorded runs pinned; every pinned file must be, and the section-by-section tasks whose reference did not
re-run at admission (\cref{sec:m-admission}) are named as not published. The examination tasks are the evaluation side of
the public course dataset, with each task's instruction, starting files, checker and reference answer, so a third party
grades its own candidates exactly as the experiment did; the script lays the run's copies of the environment's three files
over the published ones, after which every examination file is byte-identical to the one the run pinned.
\Cref{tab:repro-vim-outputs} lists what the full procedure produces beside what the recorded run measured. The attempts of
a repetition are new samples, and two trainings of the same records with the same seed on the same server logged the same
losses to four decimals and left adapters of different digests, so a repetition's counts are a new sample of the same
measurement, not a copy.
\end{sloppypar}

\begin{table}[htbp]\centering\footnotesize
\caption{What the full procedure produces and what the recorded run measured, with the evidence file that holds it.}
\label{tab:repro-vim-outputs}
\begin{tabular}{>{\raggedright\arraybackslash}p{34mm}>{\raggedright\arraybackslash}p{112mm}}
\toprule
Product & Recorded run \\
\midrule
section-by-section material & admission \ev{funnel/forward/oracle/tasks-solved} of \ev{funnel/forward/oracle/tasks}
tasks; learner \ev{funnel/forward/student-attempts/counted} counted attempts,
\ev{funnel/forward/student-attempts/pass-pct}\% passed, \ev{funnel/forward/student-attempts/tasks-solved} tasks solved;
teacher \ev{funnel/forward/teacher-attempts/counted}, \ev{funnel/forward/teacher-attempts/pass-pct}\%,
\ev{funnel/forward/teacher-attempts/tasks-solved} (\texttt{material/funnel.json}) \\
outcome-first material & admission \ev{funnel/backward/oracle/tasks-solved} of \ev{funnel/backward/oracle/tasks};
learner \ev{funnel/backward/student-attempts/counted}, \ev{funnel/backward/student-attempts/pass-pct}\%,
\ev{funnel/backward/student-attempts/tasks-solved} solved; teacher \ev{funnel/backward/teacher-attempts/counted},
\ev{funnel/backward/teacher-attempts/pass-pct}\%, \ev{funnel/backward/teacher-attempts/tasks-solved} \\
training sets & B and D \ev{material/b/train-records} training and \ev{material/b/validation-records} validation records of
\ev{material/b/tasks} tasks, C \ev{material/c/train-records} training records, F as B with the requests shuffled, G
\ev{material/g/train-records} training records of \ev{material/g/tasks} tasks
(\texttt{arm-datasets/*/composition.json}) \\
training & up to 800 steps, the checkpoint of lowest validation loss of every 40
(\texttt{evaluation/aggregates.json}); accelerator time in \cref{tab:compute} \\
examination & passes of \ev{arms/b/exam/cand-counted} attempts per condition: B \ev{arms/b/exam/cand-passes}, C
\ev{arms/c/exam/cand-passes}, D \ev{arms/d/exam/cand-passes}, F \ev{arms/f-vs-base/exam/cand-passes}, G
\ev{arms/g/exam/cand-passes}; the base none (\texttt{decision/decision-eval-exam.json}) \\
held-out generated tasks & B \ev{arms/b/target/cand-passes} of \ev{arms/b/target/cand-counted}, C
\ev{arms/c/target/cand-passes}, D \ev{arms/d/target/cand-passes}, F \ev{arms/f-vs-base/target/cand-passes}, G
\ev{arms/g/target/cand-passes} of \ev{arms/g/target/cand-counted} (\texttt{decision/decision-eval-unseen.json}) \\
\bottomrule
\end{tabular}
\end{table}

\subsection{The reproduction materials: what they hold and what they do not reproduce}\label{app:repro-vim-limits}

\begin{sloppypar}
The reproduction materials are public at \nolinkurl{\MaterialsRepo}, cited at commit \texttt{\MaterialsCommit}, under
the GPL-3.0-only licence. They hold the Vim task environment (the editing tool, the task image, admission); the seven
authoring operators with their prompts and output schemas; the recipes of both authoring modes, of the base and teacher
examinations, of every training-material condition and seed, of the course-exercise evaluations and of the three
generations of the iterative protocol (\nolinkurl{recipes/vim-manual/loop/}); the evaluation plans; the Lua prediction
environment with its examination generator and check; and the examiner. A recipe is run by the learning controller
(\texttt{make recipe/run} of the materials); the materials orchestrate nothing themselves.
\end{sloppypar}

\begin{sloppypar}
What they do not reproduce, stated plainly. (1) Tasks generated anew are named differently in their paths and metadata,
so they are not byte-identical to the tasks of the recorded runs, and the stage digests of a repetition differ from the
recorded ones. (2) The course-exercise evaluations run on the \ev{practice/g-s3/exercises/tasks} published single-idea
exercises, not on the \ev{practice/g-s3/all/tasks} of the recorded runs, because the
\ev{practice/g-s3/composites/tasks} composite exercises are not published. (3) The Lua interpreter of the recorded runs is
pinned by its sha256 and is not distributed; the examination generator accepts only that binary, and the materials'
tests can be pointed at another build of Lua~5.1.5. (4) The recorded Lua examination was drawn from a seed only the
examiner holds: it is not published and cannot be redrawn, and a repetition draws its own. (5) The recorded Lua verdicts
carry an HMAC made with the examiner's secret key, which only the key's holder can check, so they are not publicly
verifiable; verdicts made with the materials are signed with a key pair instead. (6) The materials hold the three
iterative recipes for the 4B learner and none for the 8B learner; the adapters of the iterative runs were not retained,
and the download revision of the 8B base model was not recorded (its checkpoint digest is). (7) The Lua learning cycle is not in the materials: it
is the \texttt{lineage} program of the controller's repository (\cref{tab:mapping}), and
\nolinkurl{reproduce/13-lua-lineage.sh} is the recorded launch script with the machine's paths as variables.
\end{sloppypar}

\begin{sloppypar}
What was run from them for this paper. Every recipe that needs no earlier run (26 of the 28) was validated by the
controller with the course in place; validation checks a recipe and everything it names, and that the inputs exist, not
their content. The second and third iterative recipes name the earlier generation's finished run as their champion and
were not validated. Not run end to end from the materials: drawing a Lua examination with the pinned interpreter,
examining candidates against a model server, and the attempt and training stages of any recipe. The Lua lineage script
was not run or built from the public repository for this paper. The recorded run's outcomes per examination task are
private; its registered decision is checked through the evidence dataset's \texttt{decision/} and \texttt{evaluation/}
files, and a repetition makes its own. No script evaluates the recorded adapters.
\end{sloppypar}